\pdfoutput=1

\RequirePackage[svgnames]{xcolor}

\documentclass[11pt]{article}

\usepackage{ACL2023}

\usepackage{times}
\usepackage{latexsym}

\usepackage[utf8]{inputenc}

\usepackage{microtype}

\usepackage{mflogo}

\usepackage[T1]{fontenc}

\usepackage{inconsolata}
\usepackage{algorithm}
\usepackage{algpseudocode}
\usepackage{amsmath,amssymb}
\usepackage{soul}
\usepackage{tikz}

\tikzset{rndblock/.style={rounded corners,rectangle,minimum size=0mm,scale=0.8,draw,outer sep=0pt}}

\newcommand{\tframed}[2][]{\tikz[baseline=(h.base)]\node[rndblock,#1] (h) {#2};}

\usetikzlibrary{shapes.geometric}
\usepackage{framed}
\usepackage{enumitem}
\setlist{leftmargin=1mm}
\newlist{RQ}{enumerate}{1}
\setlist[RQ]{label=\textbf{RQ\,\arabic*},ref={RQ\,\arabic*}}
\usepackage{comment}
\usepackage{natbib}
\usepackage{multibib}
\makeatletter
\usepackage{booktabs}
\usepackage[inkscapeformat=png]{svg}
\usepackage{graphicx}
\usepackage{caption}
\usepackage{subcaption}
\usepackage{tabularx}
\usepackage{multicol,multirow}
\usepackage{soul}
\usepackage{float}
\usepackage{enumitem}
\usepackage{pifont}
\usepackage{arydshln}
\usepackage{lipsum}

\definecolor{gre}{RGB}{101, 191, 127}

\definecolor{gree}{RGB}{7, 135, 44}

\usepackage[many]{tcolorbox}

\newtcolorbox{defin}{colback=VioletRed!5!White,enhanced,title=Contributions,
	attach boxed title to top left={xshift=-4mm},boxrule=0pt,after skip=1cm,before skip=1cm,right skip=0cm,breakable,fonttitle=\bfseries,toprule=0pt,bottomrule=0pt,rightrule=0pt,leftrule=3pt,arc=0mm,skin=enhancedlast jigsaw,sharp corners,colframe=VioletRed!55!black,colbacktitle=VioletRed!55!black,boxed title style={
		frame code={ 
			\fill[VioletRed!25!black](frame.south west)--(frame.north west)--(frame.north east)--([xshift=3mm]frame.east)--(frame.south east)--cycle;
			\draw[line width=1mm,VioletRed!25!black]([xshift=2mm]frame.north east)--([xshift=5mm]frame.east)--([xshift=2mm]frame.south east);
			\draw[line width=1mm,VioletRed!25!black]([xshift=5mm]frame.north east)--([xshift=8mm]frame.east)--([xshift=5mm]frame.south east);
			\fill[VioletRed!25!black](frame.south west)--+(4mm,-2mm)--+(4mm,2mm)--cycle;
		}
	}
}

\usetikzlibrary{shapes.geometric, arrows}
\usetikzlibrary{decorations.markings}

\usepackage{fancybox}
\usepackage{xcolor}
\usepackage{hyperref}
 \definecolor{darkblue}{rgb}{0, 0, 0.5}
  \hypersetup{colorlinks=true, citecolor=darkblue, linkcolor=darkblue, urlcolor=darkblue}

\definecolor{vgreen}{HTML}{60A917}
\definecolor{vred}{HTML}{CE3A29}
\definecolor{vpurple}{HTML}{FF00FF}

\usepackage{xstring}
\usepackage{longtable}

\usepackage{tabularray}

\DefTblrTemplate{firsthead,middlehead,lasthead}{default}{
}
\DefTblrTemplate{firstfoot}{default}{
  \UseTblrTemplate{contfoot}{default}
  \UseTblrTemplate{caption}{default}
}
\DefTblrTemplate{middlefoot}{default}{
  \UseTblrTemplate{contfoot}{default}
  \UseTblrTemplate{capcont}{default}
}
\DefTblrTemplate{lastfoot}{default}{
  \UseTblrTemplate{note}{default}
  \UseTblrTemplate{remark}{default}
  \UseTblrTemplate{capcont}{default}
}

\newcolumntype{P}[1]{>{\centering\arraybackslash}p{#1}}

\usepackage{color}
\tcbuselibrary{skins}

\usepackage[export]{adjustbox} 

\usepackage{setspace}
\usepackage[capitalise,nameinlink]{cleveref}

\crefname{section}{Sec.}{Sec.}

\makeatletter
\newcommand{\DrawLine}{%
  \begin{tikzpicture}
  \path[use as bounding box] (0,0) -- (\linewidth,0);
  \draw[color=blue!75!black,dashed,dash phase=.5pt]
        (0-\kvtcb@leftlower-\kvtcb@boxsep,0)--
        (\linewidth+\kvtcb@rightlower+\kvtcb@boxsep,0);
  \end{tikzpicture}%
  }
\makeatother

\usepackage[draft,textsize=footnotesize,textwidth=15mm]{todonotes}

\newcommand\acb[1]{\textcolor{blue}{#1}}

\usepackage[draft,textsize=footnotesize,textwidth=15mm]{todonotes}

\usepackage[draft,textsize=footnotesize,textwidth=15mm]{todonotes}

\newcommand\vril[1]{\todo[author=VR,color=green!20,inline,caption={}]{#1}}

\newcommand*{\affaddr}[1]{#1}
\newcommand*{\affmark}[1][*]{\textsuperscript{#1}}
\newcommand*{\email}[1]{\texttt{#1}}

\author{
Vipula Rawte\affmark[1]\thanks{\,\,\,Corresponding author.}, Prachi Priya\affmark[2], S.M Towhidul Islam Tonmoy\affmark[3], S M Mehedi Zaman\affmark[3], \\ \bf Aman Chadha\affmark[4,5]\thanks{\,\,\,Work does not relate to position at Amazon.}, \bf Amit Sheth\affmark[1], Amitava Das\affmark[1]  \\
\affaddr{\affmark[1]AI Institute, University of South Carolina, USA}\\
\affaddr{\affmark[2]Indian Institute of Technology, Kharagpur}\\
\affaddr{\affmark[3]Islamic University of Technology}\\
\affmark[4]Stanford University, USA, 
\affmark[5]Amazon AI, USA \\
\email{vrawte@mailbox.sc.edu}
}

\title{\includegraphics[height=0.16\textwidth]{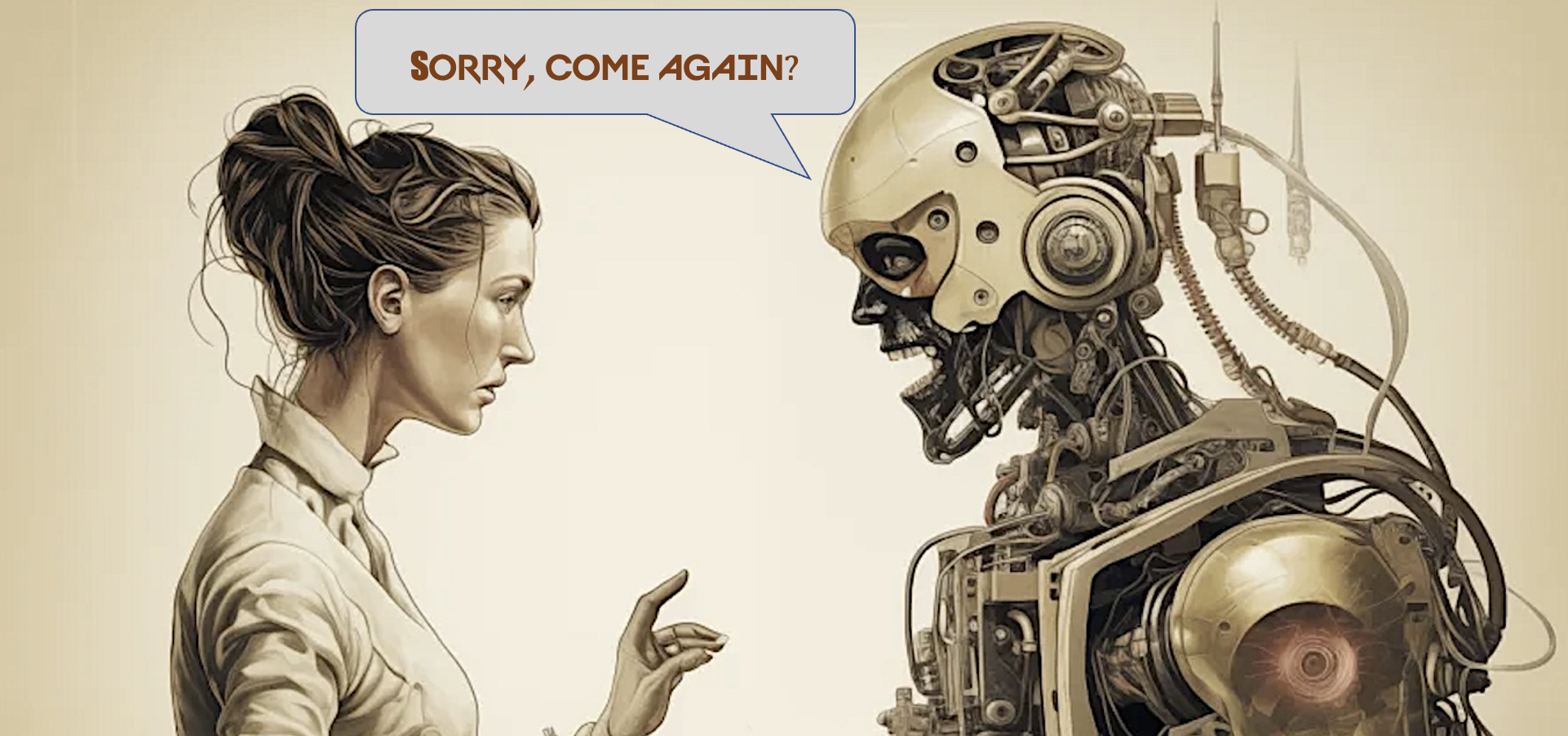}\\
``\emph{Sorry, Come Again?}'' Prompting
 -- Enhancing Comprehension and Diminishing Hallucination with \tframed[line width=0.5bp,fill=vred]{\textcolor{white}{\texttt{\textbf{[PAUSE]}}}}-injected Optimal Paraphrasing}

\let\oldtwocolumn\twocolumn
\renewcommand\twocolumn[1][]{%
    \oldtwocolumn[{#1}{
    \begin{center}
           \vspace{26mm}
           \includegraphics[width=\textwidth]{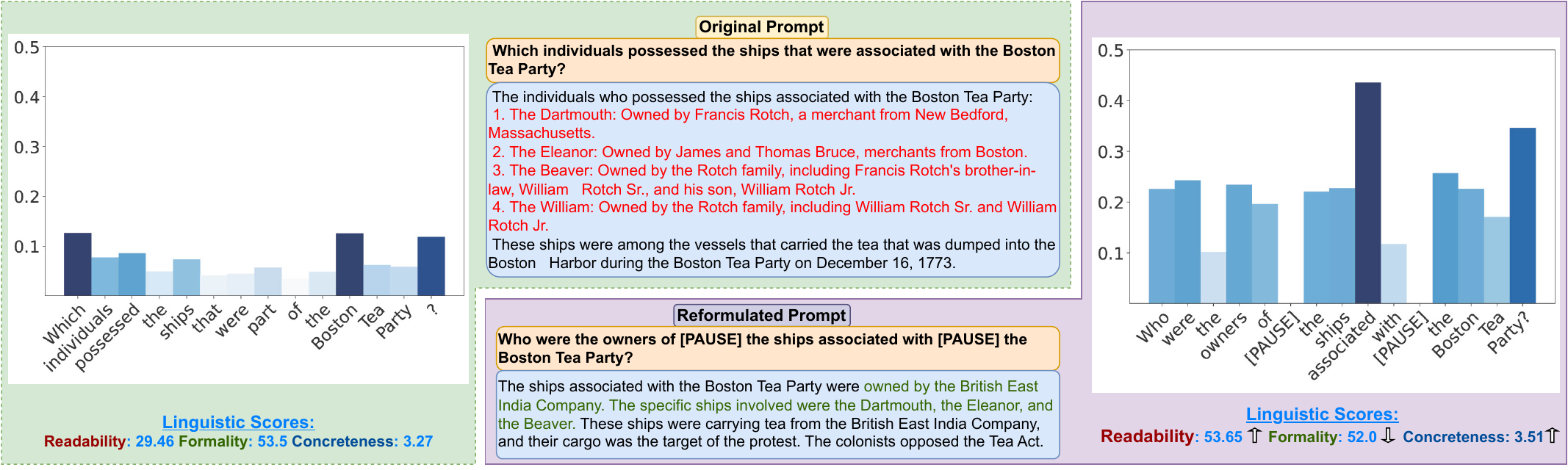}
           \vspace{-6.5mm}
           \captionof{figure}{An example demonstrating how a ``\emph{rephrased prompt}'' presented to a particular LLM can aid in avoiding  hallucination. Here, the hallucinated text is highlighted in \textcolor{red}{red}. Post reformulation, the newly generated response incorporates the factually correct (dehallucinated) text, highlighted in \textcolor{teal}{green}.}
           \label{fig:sorry_come_again_problem_definition}\vspace{2mm}
        \end{center}
    }]
}

\begin{document}
\maketitle

\begin{abstract} 
Hallucination has emerged as the most vulnerable aspect of contemporary Large Language Models (LLMs). In this paper, we introduce the \emph{Sorry, Come Again (SCA)} prompting, aimed to avoid LLM hallucinations by enhancing comprehension through: (i) optimal paraphrasing and (ii) injecting \tframed[line width=0.5bp,fill=vred]{\textcolor{white}{\texttt{\textbf{[PAUSE]}}}} tokens to delay LLM generation. First, we provide an in-depth analysis of linguistic nuances: \emph{formality, readability,} and \emph{concreteness} of prompts for 21 LLMs, and elucidate how these nuances contribute to hallucinated generation. Prompts with lower readability, formality, or concreteness pose comprehension challenges for LLMs, similar to those faced by humans. In such scenarios, an LLM tends to speculate and generate content based on its imagination (associative memory) to fill these information gaps. Although these speculations may occasionally align with factual information, their accuracy is not assured, often resulting in hallucination. Recent studies reveal that an LLM often tends to neglect the middle sections of extended prompts, a phenomenon termed as \emph{lost in the middle}.
We find that while a specific paraphrase may suit one LLM, the same paraphrased version may elicit a different response from another LLM. Therefore, we propose an optimal paraphrasing technique aimed at identifying the most comprehensible paraphrase of a given prompt, evaluated using Integrated Gradient (and its variations) to guarantee that all words are accurately processed by the LLM. Furthermore, during the reading of lengthy sentences, humans often pause at various points to better comprehend the meaning read thus far. These pauses are not only dictated by delimiters but also by semantic meaning. We have fine-tuned an LLM with injected \tframed[line width=0.5bp,fill=vred]{\textcolor{white}{\texttt{\textbf{[PAUSE]}}}} tokens, allowing the LLM to pause while reading lengthier prompts. The introduction of \tframed[line width=0.5bp,fill=vred]{\textcolor{white}{\texttt{\textbf{[PAUSE]}}}} injection has brought several key contributions: (i) determining the optimal position to inject \tframed[line width=0.5bp,fill=vred]{\textcolor{white}{\texttt{\textbf{[PAUSE]}}}}, (ii) determining the number of \tframed[line width=0.5bp,fill=vred]{\textcolor{white}{\texttt{\textbf{[PAUSE]}}}} tokens to be inserted, and (iii) introducing reverse proxy tuning to fine-tune the LLM for \tframed[line width=0.5bp,fill=vred]{\textcolor{white}{\texttt{\textbf{[PAUSE]}}}} insertion. SCA's demo is publicly available\footnote{\url{https://huggingface.co/spaces/aisafe/SCA}}.



\end{abstract}

\vspace{-11mm}
\begin{defin}
\begin{itemize}
[labelindent=-0.6em,labelsep=0.1cm,leftmargin=*]
\setlength\itemsep{0em}
\begin{spacing}{0.5}
\item[$\blacktriangleright$] 
{\footnotesize 
{\fontfamily{phv}\fontsize{8}{9}
\selectfont
Investigating the impact of three different linguistic features \acb{(formality, readability, and concreteness)} of prompts on hallucination for 21 LLMs (cf. \cref{sec:linguistic}).}
}

\item[$\blacktriangleright$] 
{\footnotesize 
{\textls[-15]{\fontfamily{phv}\fontsize{8}{9}\selectfont
Presenting SCA an optimal paraphrasing prompting framework aimed at identifying the most comprehensible paraphrase of the same prompt (cf. \cref{sec:sca}).}}
}

\item[$\blacktriangleright$] 
{\footnotesize 
{\fontfamily{phv}\fontsize{8}{9}\selectfont
\tframed[line width=0.5bp,fill=vred]{\textcolor{white}{\texttt{\textbf{[PAUSE]}}}} injection methods to delay LLM generation and aid comprehension (cf. \cref{sec:pause}) and a novel reverse proxy-tuning for it (cf. \cref{sec:rev-pt}).} 
}

\item[$\blacktriangleright$] 
{\footnotesize 
{\fontfamily{phv}\fontsize{8}{9}\selectfont
Lastly, presenting \emph{\bfseries\fontfamily{pcr}\selectfont{ACTIVATOR}}, an end-to-end framework crafted to avoid hallucination by enhancing LLMs' reading comprehension (cf. \cref{sec:activator}).}
}

\vspace{-6mm}
\end{spacing}
\end{itemize}

\end{defin}

\begin{figure*}[!ht]
\begin{tcolorbox}
[boxsep=0pt,left=2.5pt,right=5pt,top=5pt,bottom=5pt,colback=Teal!10!white,colframe=Teal!45!black]
\footnotesize
\centering
    \ul{\textbf{Original Prompt:}} Which individuals possessed the ships that were associated with the Boston Tea Party?
\end{tcolorbox}    
\end{figure*}

\begin{figure*}[!ht]
\vspace{-6mm}
        \centering
        \begin{subfigure}[b]{0.28\textwidth}
            \centering
            \includegraphics[width=\textwidth, height=3cm]{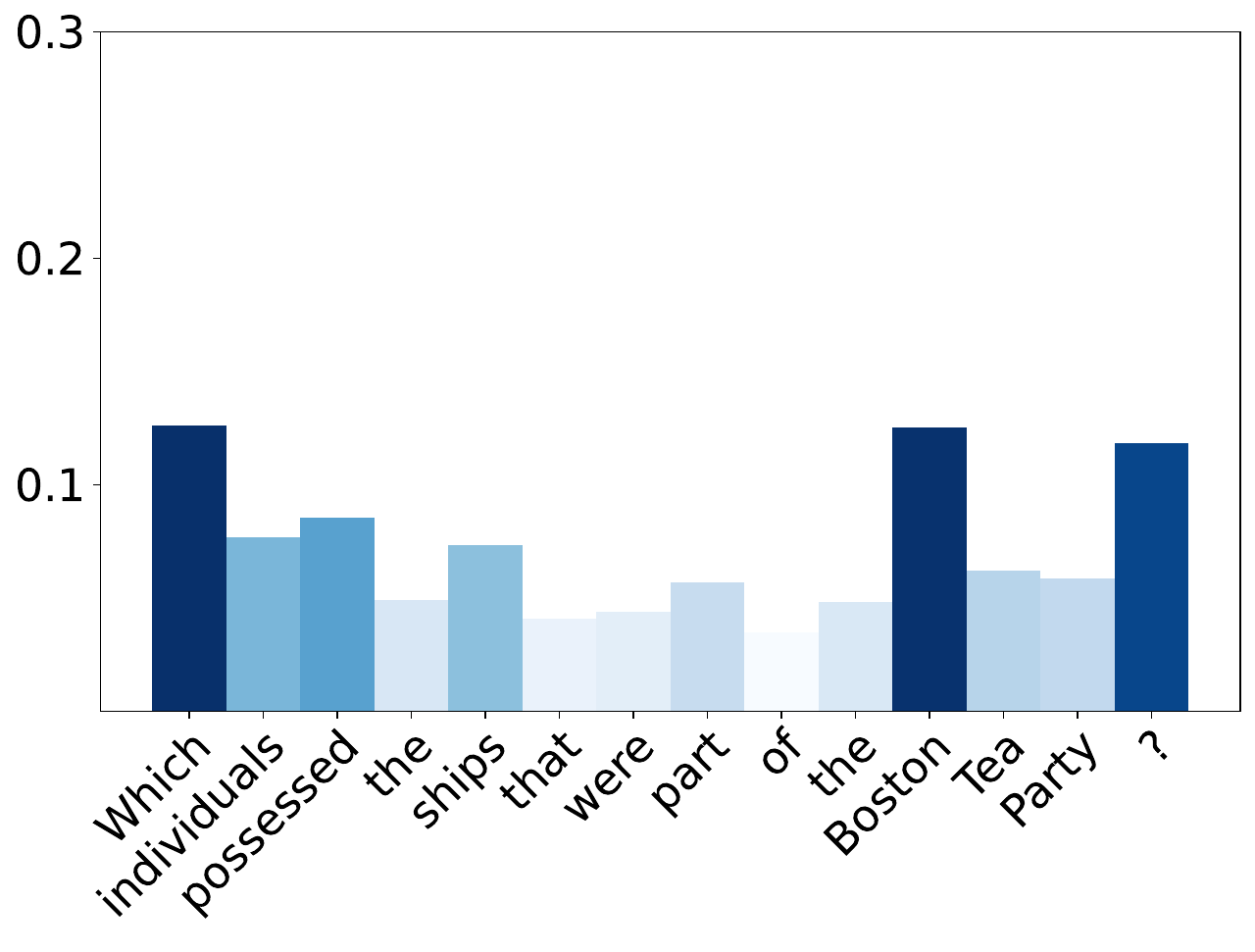}
            \caption[]%
            {{\small Falcon}}    
            \label{fig:net14}
        \end{subfigure}
        \hfill
        \begin{subfigure}[b]{0.28\textwidth}  
            \centering 
            \includegraphics[width=\textwidth, height=3cm]{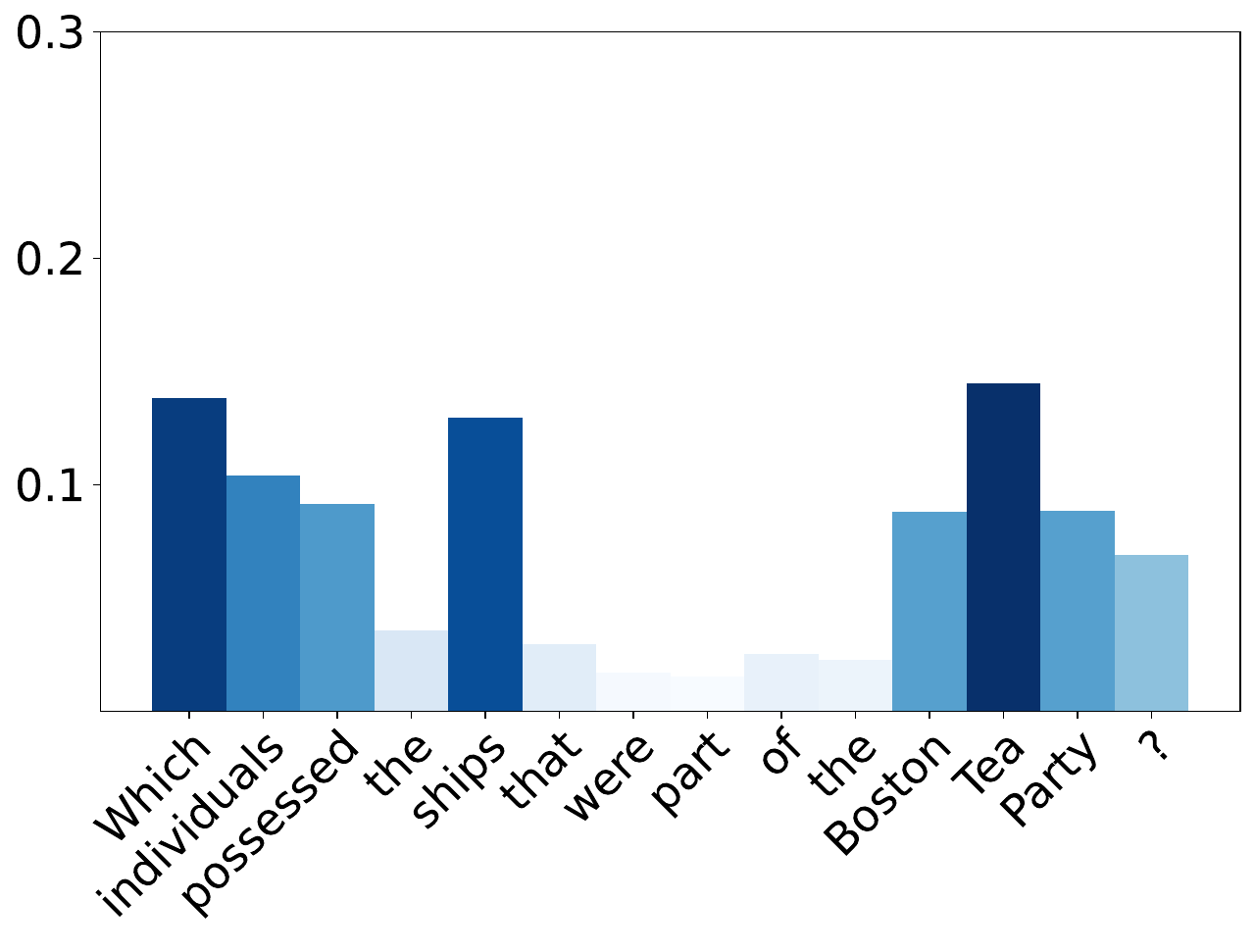}
            \caption[]%
            {{\small BLOOM}}    
            \label{fig:net24}
        \end{subfigure}
        \hfill
        \begin{subfigure}[b]{0.28\textwidth}   
            \centering 
            \includegraphics[width=\textwidth, height=3cm]{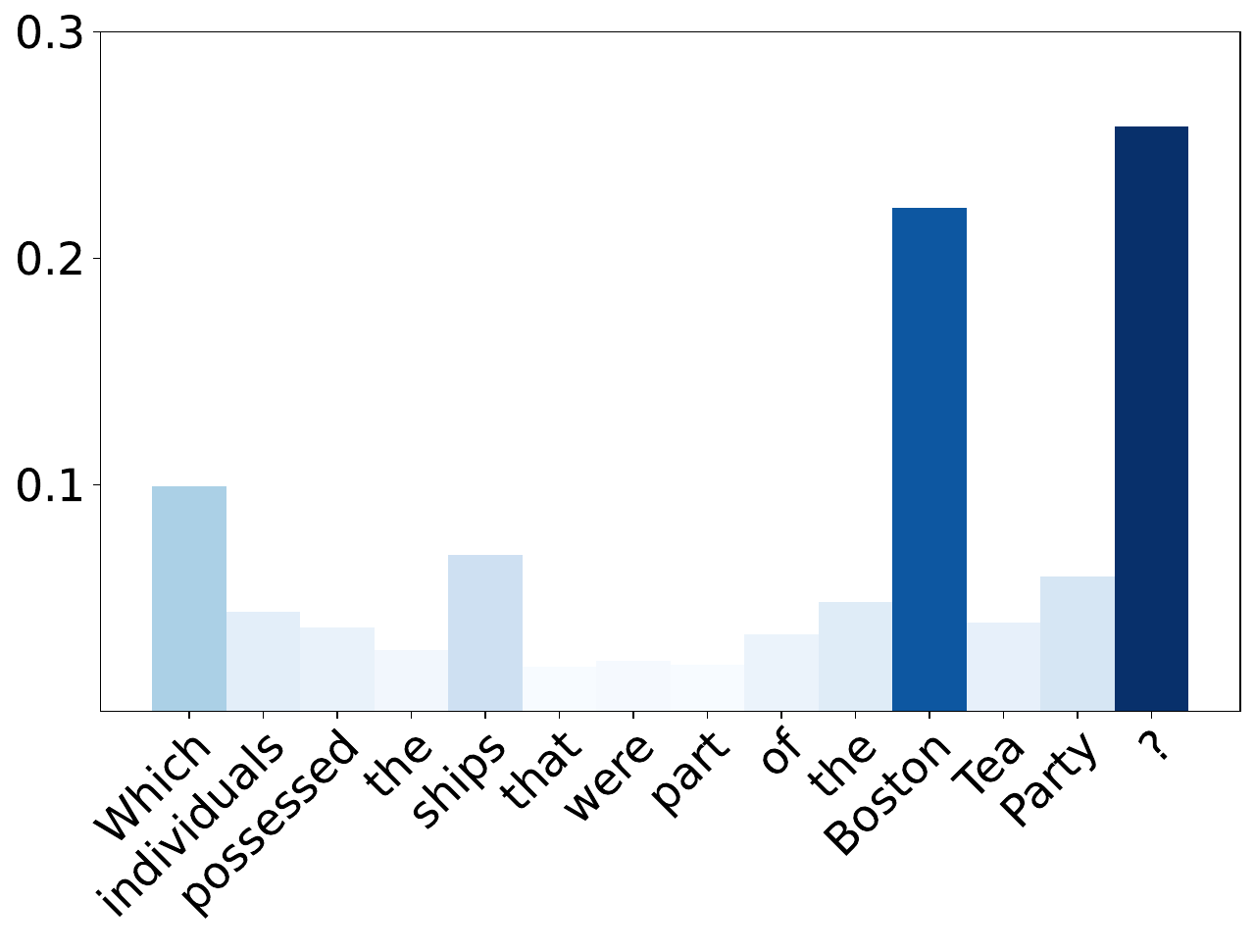}
            \caption[]%
            {{\small Dolly}}    
            \label{fig:net34}
        \end{subfigure}
        \vspace{-4mm}
        \caption[]%
            {{\small The same prompt is read by different LLMs differently.}} 
        \label{fig:sameprompt_different_llm}
\end{figure*}

\begin{figure*}[!ht]
\vspace{-1mm}
        \centering
        \begin{subfigure}[b]{0.28\textwidth}
            \begin{tcolorbox}
            [boxsep=0pt,left=2.5pt,right=5pt,top=5pt,bottom=5pt,colback=Teal!10!white,colframe=Teal!45!black]
            \footnotesize
            \centering
                Who were the proprietors of the ships associated with the Boston Tea Party?
            \end{tcolorbox}    
            \end{subfigure}
            \hfill
            \begin{subfigure}[b]{0.28\textwidth}
            \begin{tcolorbox}
            [boxsep=0pt,left=2.5pt,right=5pt,top=5pt,bottom=5pt,colback=Teal!10!white,colframe=Teal!45!black]
            \footnotesize
            \centering
                Which individuals possessed the ships that were associated with the Boston Tea Party?
            \end{tcolorbox}    
            \end{subfigure}
            \hfill
            \begin{subfigure}[b]{0.28\textwidth}
            \begin{tcolorbox}
            [boxsep=0pt,left=2.5pt,right=5pt,top=5pt,bottom=5pt,colback=Teal!10!white,colframe=Teal!45!black]
            \footnotesize
            \centering
                Who owned which ships were a part of the Boston Tea Party?
            \end{tcolorbox}    
            \end{subfigure}
            \hfill
            \vskip\baselineskip
        \vspace{-4mm}
        \begin{subfigure}[b]{0.28\textwidth}
            \centering
            \includegraphics[width=\textwidth, height=3cm]{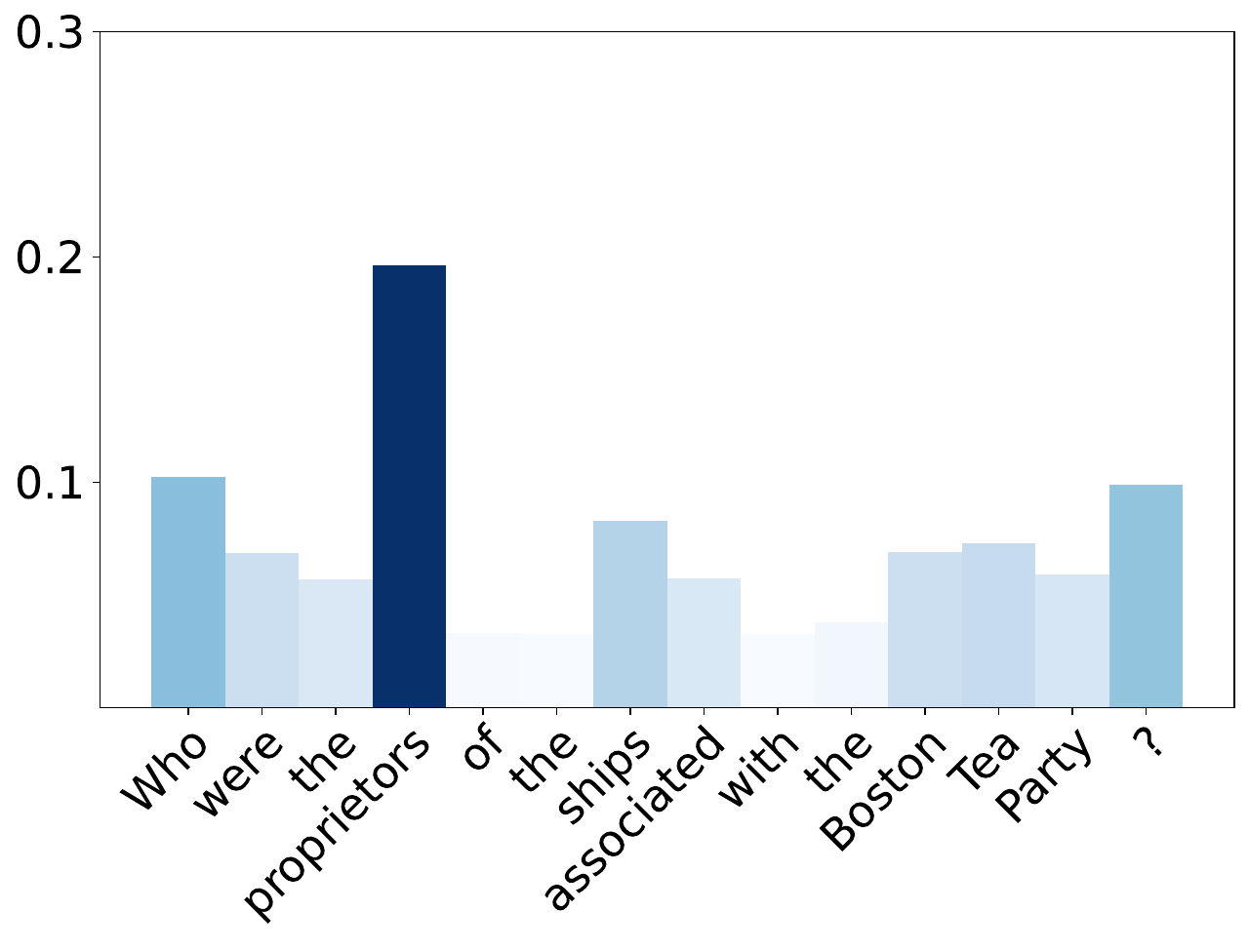}
            \caption[]%
            {{\small Optimal Prompt for Falcon}}    
            \label{fig:net14}
        \end{subfigure}
        \hfill
        \begin{subfigure}[b]{0.28\textwidth}  
            \centering 
            \includegraphics[width=\textwidth, height=3cm]{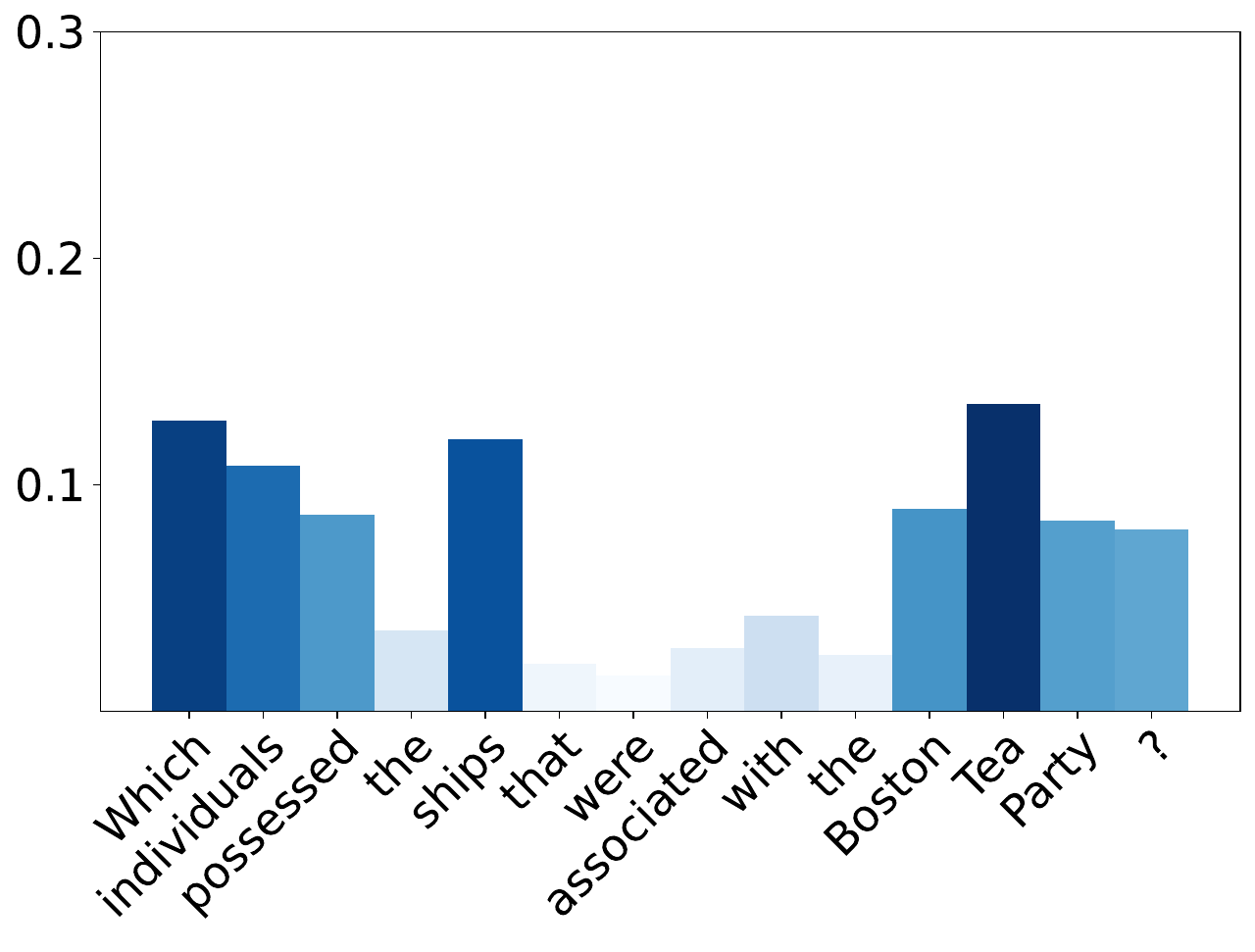}
            \caption[]%
            {{\small Optimal Prompt for BLOOM}}    
            \label{fig:net24}
        \end{subfigure}
        \hfill
        \begin{subfigure}[b]{0.28\textwidth}   
            \centering 
            \includegraphics[width=\textwidth, height=3cm]{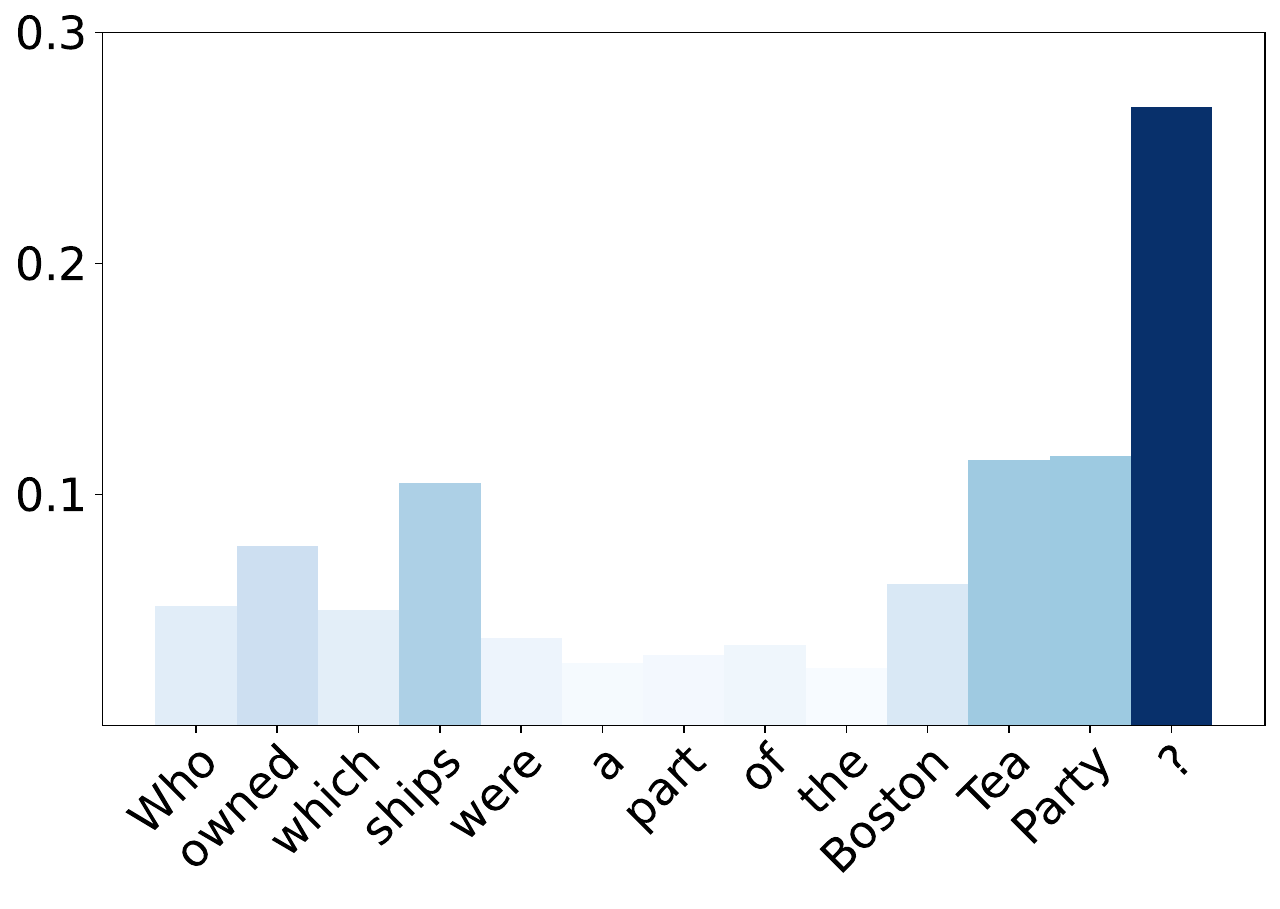}
            \caption[]%
            {{\small Optimal Prompt for Dolly}}    
            \label{fig:net34}
        \end{subfigure}
        \vspace{-2mm}
        \caption[]%
            {{\small Paraphrased versions of the aforementioned prompt with a focus on suitability for different LLMs.}} 
        \label{fig:different_paraphrase_different_llm}
\end{figure*}


\section{``Sorry, Come Again?'' -- LLM Does Not Comprehend It All in a Given Prompt} 
\label{sec:sca}

With the advent of LLMs, \emph{Prompt Engineering} has emerged as a new technical profession \cite{pe1,pe2,pe3}. While the fundamental concept revolves around framing questions or commands effectively to elicit the desired response, mastering this skill delves into several intricacies. These include (a) understanding the LLM's proficiencies (based on the tasks it was trained to accomplish), (b) trial and error-based experimentation, (c) balancing precision and flexibility, and (d) considering bias and ethical considerations, among many other nuances. Therefore, achieving an \emph{optimal prompt} is a rather daunting task. \cite{Sclar2023QuantifyingLM} has highlighted the high sensitivity of LLMs to subtle changes in prompt formatting, giving accuracy ranges from 4\%-88\% for a given task with LLaMA-2-70B and 47\%-85\% with GPT-3.5 \cite{liu2023lost} has demonstrated that LLMs struggle to read and comprehend longer prompts. Instead, they tend to focus on words at the beginning and end, often neglecting those in between. They call this phenomenon `\emph{lost in the middle}'. In \cref{fig:sorry_come_again_problem_definition}, the prompt provided on the left-hand side is not effectively read by the LLM, resulting in a hallucinated generation. However, a paraphrased version of the same prompt, incorporating \tframed[line width=0.5bp,fill=vred]{\textcolor{white}{\texttt{\textbf{[PAUSE]}}}} tokens, is read and comprehended well by the same LLM, thereby eliminating hallucinations. Continuing along the same line, the identical prompt may be read and comprehended differently by different LLMs, as depicted in \cref{fig:sameprompt_different_llm}. Among the myriad potential paraphrases of a given prompt, a specific one may emerge as optimal for a particular LLM to effectively read and comprehend, as illustrated in \cref{fig:paraphrase_same_llm}. Conversely, a different paraphrase version may be more suitable for other LLMs (see \cref{fig:different_paraphrase_different_llm}).

The premise of this work posits that improved comprehension can lead to reduced hallucination. ``\emph{Sorry, Come Again?}'' (SCA henceforth) is a common expression in human communication, indicating difficulty in understanding the previous statement. In response, the speaker typically rephrases their utterance for better clarity. LLMs cannot seek clarification or ask follow-up questions for better understanding. 
This study presents SCA, an innovative approach in optimal prompt engineering aimed at finding the clearest prompt for a given LLM, leading to a decrease in hallucination occurrences.
\section{Dissecting an LLM's Comprehension}
\label{sec:comprehension}

Due to the blackbox nature of deep neural networks, deducing the internal process of how an LLM comprehends an input prompt presents a significant challenge. Integrated Gradients \cite{pmlr-v70-sundararajan17a} serve as the cornerstone among explainability methods, calculating the gradient of the model's prediction output with respect to its input features.
Following the approach outlined by \cite{liu2023lost}, we investigate which input words are effectively comprehended by LLMs, which forms our working hypothesis of comprehension. However, this hypothesis could be subject to further scrutiny, and we engage in self-criticism. We have utilized the following SoTA explainability methods such as Discretized Integrated Gradients (DIG) \cite{sanyal-ren-2021-discretized}, and Sequential Integrated Gradients (SIG) \cite{enguehard-2023-sequential} in this study. Developing new methods for explainability is an evolving area of research, and we have yet to determine the best-performing method among IG, DIG, and SIG. Therefore, in our study, we utilize all of them and calculate an average score obtained from them at the word level.

\section{Linguistic Nuances of Prompts} \label{sec:linguistic}
Numerous practitioners advocate that proficient prompt engineering could serve as an effective method to mitigate hallucination \cite{mit1,mit2,mit3,mit4,mit5}. However, such assertions require empirical testing conducted with scientific rigor. To the best of our knowledge, there is scarce research (except one \cite{rawte2023exploring}) on the linguistic properties of prompts and their resultant impact on hallucination in generated content. In this study, we delve into an examination of three pivotal linguistic features: \emph{readability \cite{flesch1948new}, formality \cite{heylighen1999formality}, and concreteness \cite{paivio2013dual}} of a prompt, and their consequential effects on hallucination.

\paragraph{Readability (R)} assesses the ease with which a text can be read and comprehended, taking into account factors such as complexity, familiarity, legibility, and typography. The widely recognized measure of readability is the Flesch Reading Ease Score (FRES) \cite{flesch1948new}, which provides a numerical representation of a text's readability. It is computed based on sentence length and word complexity using the formula: $\text{FRES} = 206.835-1.015\cdot(\text {total words}/\text {total sentences}) - 84.6\cdot(\text {total syllables}/\text {total words})$. For instance, a simple sentence yields a high score, while a complex one results in a lower score, reflecting the ease or difficulty of comprehension, as shown below. 

\vspace{-2.5mm}

\begin{tcolorbox}
[boxsep=0pt,left=2.5pt,right=5pt,top=2pt,bottom=2pt,colback=Wheat!55!white,colframe=Wheat!45!black]
\scriptsize
\textbf{Easily readable \colorbox{cyan!30}{FRES score = 75.5}} \\
\ul{\textbf{Sentence:}} The sun rises in the east every morning.
\end{tcolorbox}

\vspace{-4.5mm}

\begin{tcolorbox}
[boxsep=0pt,left=2.5pt,right=5pt,top=2pt,bottom=2pt,colback=Wheat!55!white,colframe=Wheat!45!black]
\scriptsize
\textbf{Challenging readability \colorbox{cyan!30}{FRES score = 11.45}}  \\
\ul{\textbf{Sentence:}} The intricacies of quantum mechanics, as expounded upon by renowned physicists, continue to baffle even the most astute scholars.
\end{tcolorbox}

\paragraph{Formality (F)} in language is characterized by detachment, accuracy, rigidity, and heaviness; an informal style is more
flexible, direct, implicit, and involved, but less informative. 

\vspace{-2.5mm}

\begin{tcolorbox}
[boxsep=0pt,left=2.5pt,right=5pt,top=2pt,bottom=2pt,colback=Wheat!55!white,colframe=Wheat!45!black]
\scriptsize
\textbf{Informal sentence \colorbox{cyan!30}{Formality score = 54.5}}  \\
The big thing in the corner dates from the 18th century.
\end{tcolorbox}

\vspace{-4.5mm}

\begin{tcolorbox}
[boxsep=0pt,left=2.5pt,right=2pt,top=2pt,bottom=5pt,colback=Wheat!55!white,colframe=Wheat!45!black]
\scriptsize
\textbf{Formal sentence \colorbox{cyan!30}{Formality score = 62}} \\
In the right corner, next to the entrance, stands a 2 meter high wooden cupboard with gold inlays, that dates from the 18th century.
\end{tcolorbox}

The widely accepted method for measuring formality, proposed by \cite{heylighen1999formality}, is calculated as follows: $\text{Formality} = (\text{freq}_{noun} + \text{freq}_{adjective} + \text{freq}_{preposition} + \text{freq}_{article} - \text{freq}_{pronoun} - \text{freq}_{verb} - \text{freq}_{adverb} - \text{freq}_{interjection} + 100)/2$, where $\text{freq}_{part\,\,of\,\,speech}$ represents the frequency of the respective part of speech.


\paragraph{Concreteness (C)} measures how well a word represents a tangible concept, with concrete words being easier to process than abstract ones \cite{paivio2013dual}. The degree of concreteness is rated on a 5-point scale (1-5) from \ul{abstract} to \ul{concrete}. Concrete words relate to tangible, sensory experiences, while abstract words involve concepts not directly sensed. Concreteness ratings for over 39,000 English words are available in \cite{brysbaert2014concreteness}. In this work, to compute the concreteness of a sentence with $n$ words, an average of concreteness ratings is calculated using the formula: $\sum_{i=1}^{n}\text{concreteness rating}_{i}/n$.

\vspace{-0.5mm}

\begin{tcolorbox}
[boxsep=0pt,left=2.5pt,right=5pt,top=2pt,bottom=2pt,colback=Wheat!55!white,colframe=Wheat!45!black]
\scriptsize
    \textbf{Examples of \ul{\textit{concrete}} words} \\
    Apple\colorbox{cyan!30}{5}, Dog\colorbox{cyan!30}{4}, Chair\colorbox{cyan!30}{4}, Book\colorbox{cyan!30}{5}, Water\colorbox{cyan!30}{5}, Car\colorbox{cyan!30}{5}
\end{tcolorbox}

\vspace{-4.5mm}

\begin{tcolorbox}
[boxsep=0pt,left=2.5pt,right=2pt,top=2pt,bottom=5pt,colback=Wheat!55!white,colframe=Wheat!45!black]
\scriptsize
   \textbf{Examples of \ul{\textit{abstract}} words}  \\
   Justice\colorbox{cyan!30}{1}, Love\colorbox{cyan!30}{1}, Happiness\colorbox{cyan!30}{1}, Courage\colorbox{cyan!30}{1}, Wisdom\colorbox{cyan!30}{1}
\end{tcolorbox}

\vspace{-0.5mm}

\begin{figure*}[!ht]
        \centering
        \begin{subfigure}[b]{0.24\textwidth}
            \centering
            \includegraphics[width=\textwidth, height=2.5cm]{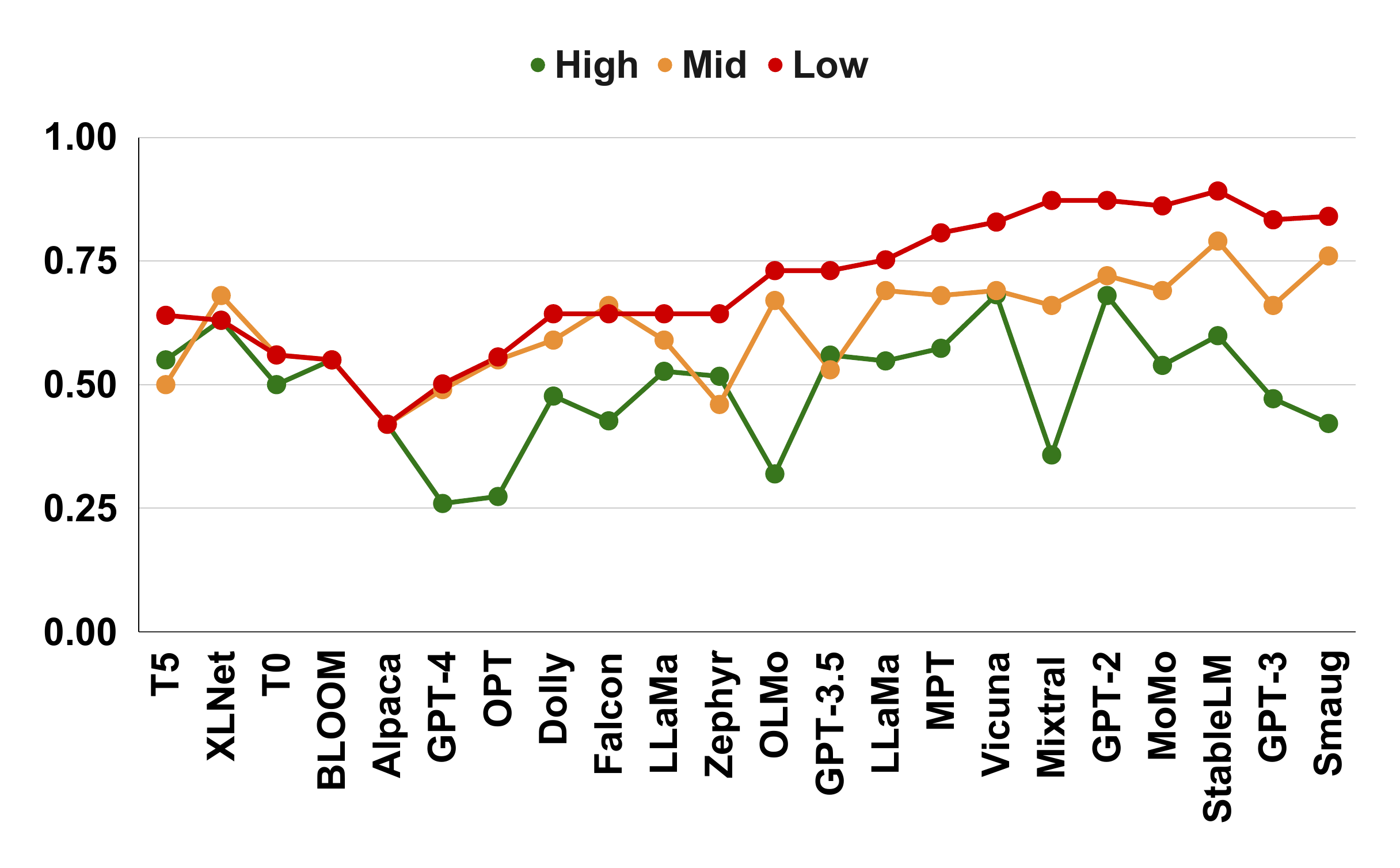}
            \caption[]%
            {{\small Person}}    
            \label{fig:mean and std of net14}
        \end{subfigure}
        \hfill
        \begin{subfigure}[b]{0.24\textwidth}  
            \centering 
            \includegraphics[width=\textwidth, height=2.5cm]{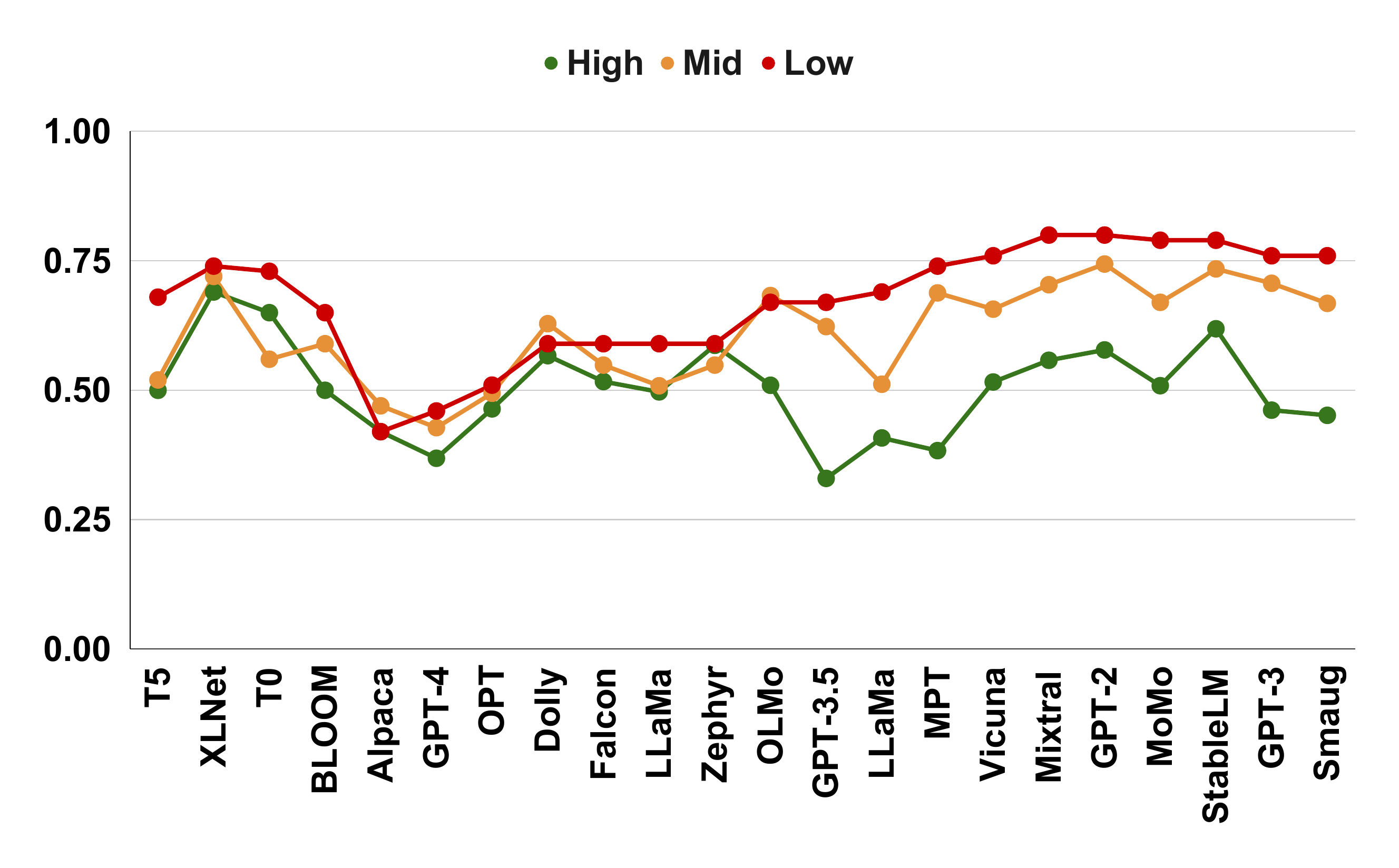}
            \caption[]%
            {{\small Location}}    
            \label{fig:mean and std of net24}
        \end{subfigure}
        \hfill
        \begin{subfigure}[b]{0.24\textwidth}   
            \centering 
            \includegraphics[width=\textwidth, height=2.5cm]{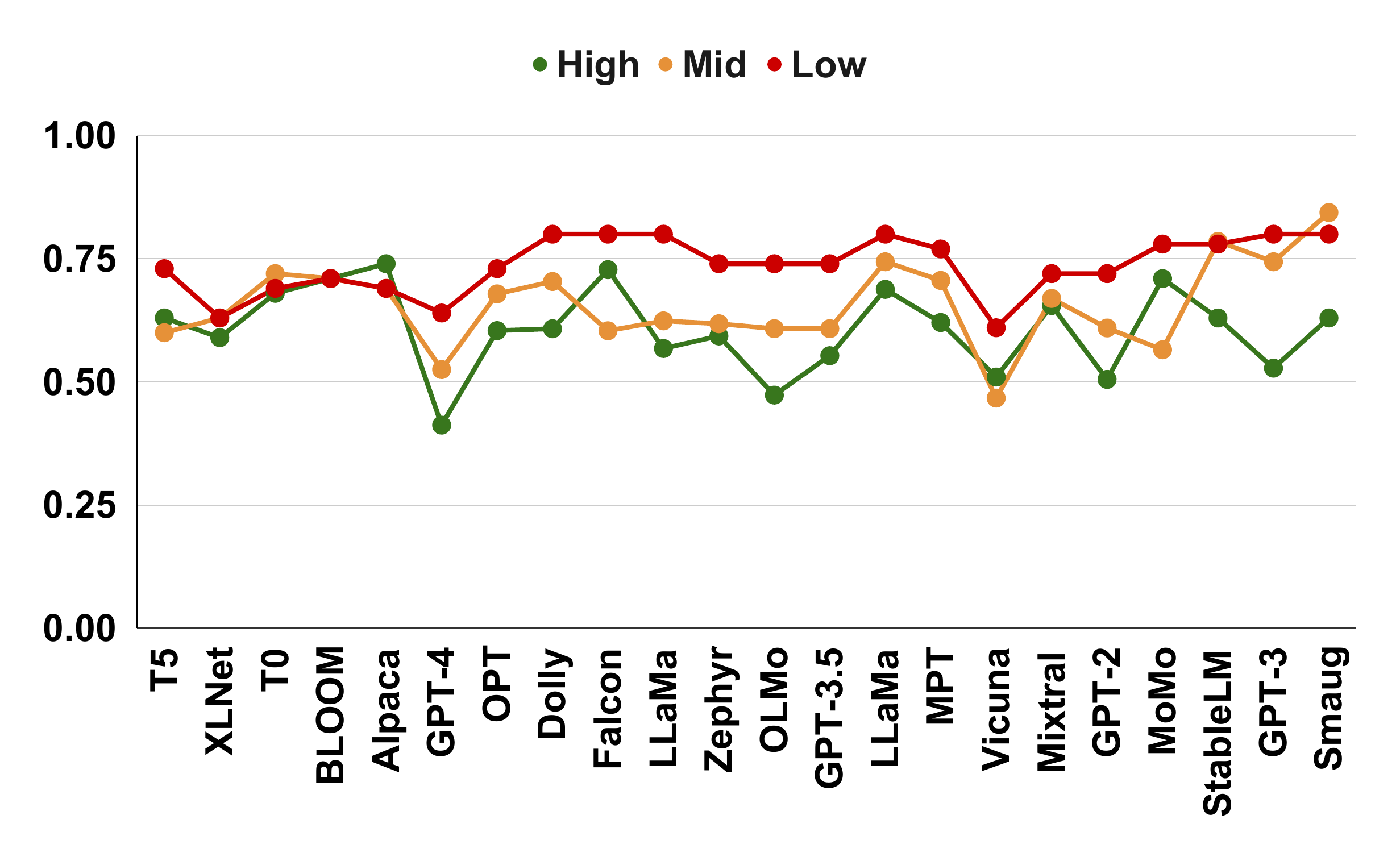}
            \caption[]%
            {{\small Number}}    
            \label{fig:mean and std of net34}
        \end{subfigure}
        \hfill
        \begin{subfigure}[b]{0.24\textwidth}   
            \centering 
            \includegraphics[width=\textwidth, height=2.5cm]{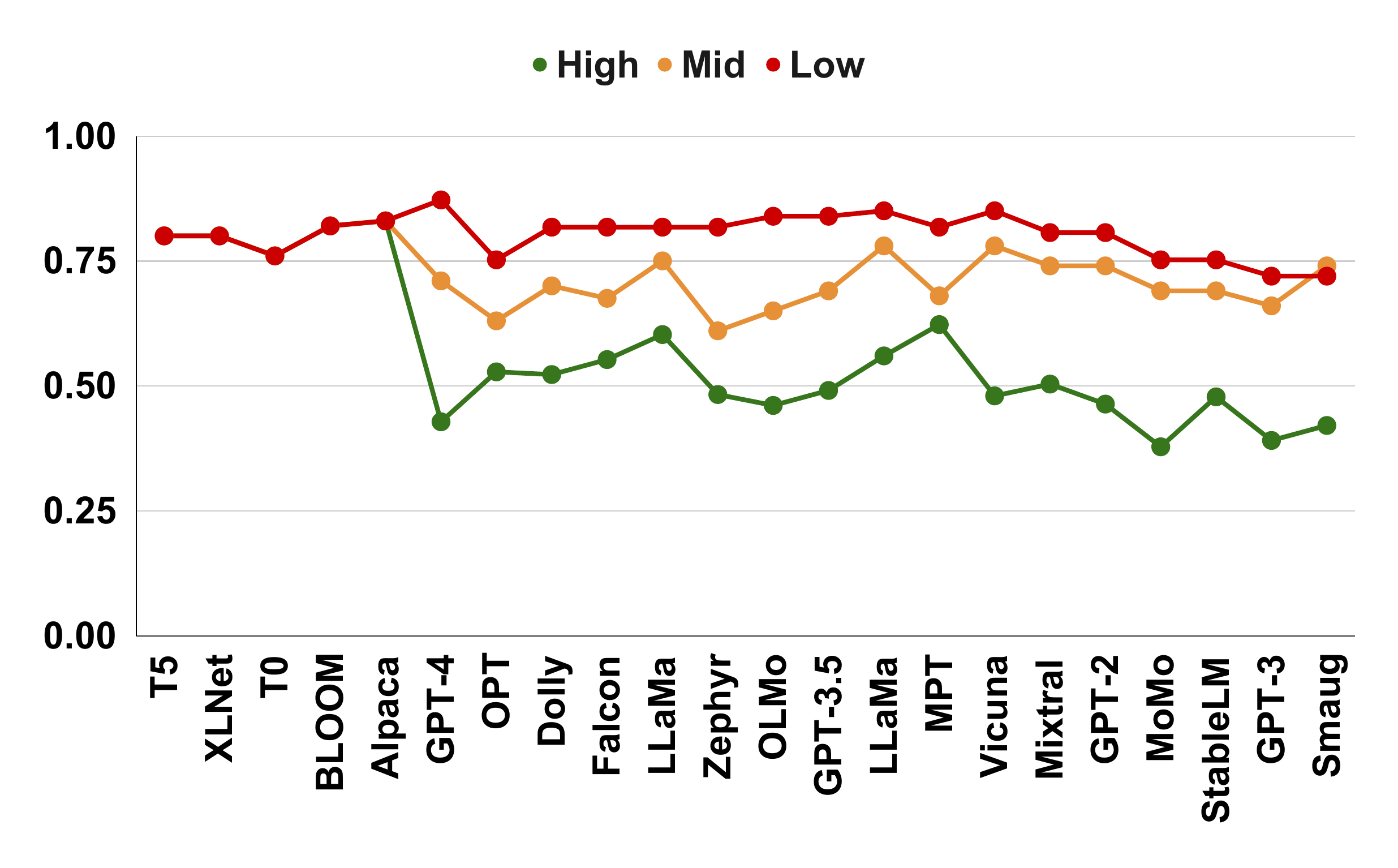}
            \caption[]%
            {{\small Time}}    
            \label{fig:mean and std of net44}
        \end{subfigure}
        \hfill
        \begin{subfigure}[b]{\textwidth}   
            \centering 
            \input{concrete}
        \end{subfigure}
        \caption[]%
            {{\small Percentage of hallucination for four different categories of hallucination for three levels of concreteness.}}   
        \label{fig:Concreteness}
\end{figure*}

We analyze the impact of linguistic characteristics on LLM hallucination by establishing specific score ranges (see \cref{tab:range}) and provide a detailed examination in \cref{fig:Readability,fig:Formality,fig:Concreteness}.

\begin{table}[!htbp]
\centering
\resizebox{0.8\columnwidth}{!}{
    \begin{tabular}{ccccc}  \toprule
    \bf Range $\rightarrow$ & \bf Low &   \bf Mid  &  \bf High &  \bf Std. dev. \\ \bf Linguistic Aspect $\downarrow$ & & & \\ \midrule
    \bf Readability   &  0-13.68 & 13.69-52.42 & 52.42-100  &  19.37    \\ 
    \bf Formality     &  0-45.65 & 45.66-70 & 70.051-100  &  12.1  \\ 
    \bf Concreteness  &  1-3.03 & 3.03-3.47 & 3.47-5  & 0.22	\\ \bottomrule
    \end{tabular}
    }
    \caption{Range(s) for prompt's three linguistic aspects.}
    \vspace{-6mm}
    \label{tab:range}
\end{table}

\section{Types of Hallucination}  \label{sec:hal-types}

The phenomenon of generating factually incorrect or imaginary responses by LLMs is commonly called \emph{hallucination} \cite{augenstein2023factuality,xu2024hallucination,wang2024factuality}. Recent studies \cite{ladhak2023pre,varshney2023stitch} have categorized various types of hallucinations. \cite{rawte-etal-2023-troubling} defined two fundamental types of hallucination: when an LLM hallucinates despite being given a factually correct prompt, it is termed as a \emph{factual mirage}, whereas when an LLM hallucinates given a factually incorrect prompt, it is termed as a \emph{silver lining}. This study confines its investigation solely to the phenomenon of factual mirage hallucination. In this study, we adopt a simplified approach by utilizing the \textit{four} distinct categories of hallucination proposed 
Additionally, we furnish descriptions and examples for each category, emphasizing the hallucinated text in \textcolor{red}{red}.

\vspace{-3mm}

\paragraph{1. Person (P):} This occurs when an LLM invents a fictional personality without any tangible proof.

\vspace{-3mm}

\begin{tcolorbox}
[boxsep=0pt,left=5pt,right=5pt,top=5pt,bottom=5pt,colback=GreenYellow!10!white,colframe=GreenYellow!35!black]
\scriptsize
\textbf{Original:} The three people who were killed in the shooting at Michigan State University were all students, the police said on Tuesday morning.

\vspace{-2mm}
\DrawLine

\textbf{\scriptsize AI-generated:} The three students who died were identified as \textcolor{red}{17 y.o. Diva Davis, 20 y.o. Thomas McDevitt and 19 y.o. Jordan Eubanks}.

\vspace{-2mm}
\DrawLine

\textbf{\scriptsize Fact:} Three students — Alexandria Verner of Clawson; Brian Fraser of Grosse Pointe; and Arielle Anderson of Grosse Pointe - lost their lives.

\end{tcolorbox}

\paragraph{2. Location (L):} This issue arises when LLMs produce an inaccurate location linked to an event.

\vspace{-2.5mm}

\begin{tcolorbox}
[boxsep=0pt,left=5pt,right=5pt,top=5pt,bottom=5pt,colback=GreenYellow!10!white,colframe=GreenYellow!35!black]
\scriptsize
\textbf{Original:} A wooden boat carrying 130 migrants broke apart against rocks near a beach town in southern Italy.

\vspace{-2mm}
\DrawLine

\textbf{\scriptsize AI-generated:} ...it ran aground at dawn on Sunday near the \textcolor{red}{beach town of Punta Imperatore, in the province of Salerno, in Campania}.

\vspace{-2mm}
\DrawLine

\textbf{\scriptsize Fact:} Many of the bodies were reported to have washed up on a tourist beach near Steccato di Cutro...

\end{tcolorbox}
  
\paragraph{3. Number (N):} This happens when an LLM produces imaginary numbers (such as age, etc.).

\vspace{-2.5mm}

\begin{tcolorbox}
[boxsep=0pt,left=5pt,right=5pt,top=5pt,bottom=5pt,colback=GreenYellow!10!white,colframe=GreenYellow!35!black]
\scriptsize
\textbf{Original:} In 1944, when the Nazis killed 643 people in a French village, Robert Hebras was one of a handful who lived to tell the story.

\vspace{-2mm}
\DrawLine

\textbf{\scriptsize AI-generated:} Robert Hebras was one of \textcolor{red}{seven} men who managed to escape the massacre.

\vspace{-2mm}
\DrawLine

\textbf{\scriptsize Fact:} Only six wounded survived, hidden under corpses.

\end{tcolorbox}
  
\paragraph{4. Time (T):} This issue involves LLMs generating text about events from various timelines.

\vspace{-2.5mm}

\begin{tcolorbox}
[boxsep=0pt,left=5pt,right=5pt,top=5pt,bottom=5pt,colback=GreenYellow!10!white,colframe=GreenYellow!35!black]
\scriptsize
\textbf{Original:} After a Chinese spy balloon was shot down this month, the U.S. has brought down at least three UFOs... 

  \vspace{-2mm}
  \DrawLine

\textbf{\scriptsize AI-generated:} \textcolor{red}{April 3,  2020}: U.S. military shot down a Chinese spy balloon.

\vspace{-2mm}
\DrawLine

\textbf{\scriptsize Fact:} Feb. 4 2023: A U.S. fighter plane shoots down the balloon.
\end{tcolorbox}

\section{Selection of LLMs} \label{sec:llm}
We have selected 21 contemporary LLMs that have consistently demonstrated outstanding performance across a wide spectrum of NLP tasks, per the Open LLM Leaderboard \cite{open-llm-leaderboard}. These models include: (i) GPT-4 \cite{openai2023gpt4}, 
(ii) GPT-3.5 \cite{ChatGPT}, 
(iii) LLama2 \cite{Touvron2023Llama2O}, 
(iv) GPT-2 \cite{radford2019language}, 
(v) MPT \cite{wang2023multitask}, 
(vi) OPT \cite{zhang2022opt}, 
(vii) LLaMA \cite{meta2023introducing}, 
(viii) BLOOM \cite{scao2022bloom}, 
(ix) Alpaca \cite{alpaca}, 
(x) Vicuna \cite{vicuna2023}, 
(xi) Dolly \cite{dolly}, 
(xii) StableLM \cite{liu2023your}, 
(xiii) XLNet \cite{yang2019xlnet}, 
(xiv) T5 \cite{raffel2020exploring}, 
(xv) T0 \cite{DBLP:conf/iclr/DeleuKFKBLB22},
(xvi) Falcon \cite{almazrouei2023falcon},
(xvii) Zephyr \cite{tunstall2023zephyr},
(xviii) Mixtral \cite{jiang2024mixtral},
(xix) OLMo \cite{groeneveld2024olmo},
(xx) MoMo \cite{chada2023momo},
(xxi) Smaug \cite{Smaug}.
\section{Dataset}  \label{sec:data}
\vspace{-1mm}
To construct the $\mathcal{SCA-}90\mathcal{K}$ dataset, we utilized NYTimes tweets \cite{nyt} primary sources of data as prompts. We selected 21 LLMs, based on the criteria delineated in \cref{sec:llm}, and used them to generate a total of 52,500 text passages, with each LLM producing 2,500 text prose entries. We follow a similar approach to \cite{rawte-etal-2023-troubling} for annotating our data. More details are in \cref{appendix:dataset}. \cref{tab:data-stats} provides detailed dataset statistics.

\vspace{-2mm}
\begin{table}[!htbp]
\centering
\scriptsize
\resizebox{0.55\columnwidth}{!}{
    \begin{tabular}{cc}  \toprule
    \bf Hallucination Category  & \bf \# Sentences   \\ \midrule
    \bf Person   &  9,570  \\ 
    \bf Location     &  32,190  \\ 
    \bf Number  &  11,745 	\\ 
    \bf Time  &  36,105 	\\  \midrule
    \bf Total  &  89,610  \\   \bottomrule
    \end{tabular}
    }
    \vspace{-1.5mm}
    \caption{Statistics of $\mathcal{SCA-}90\mathcal{K}$.}
    \label{tab:data-stats}
\vspace{-4mm}
\end{table}

\vspace{-2mm}
\section{\textls[0]{Can Paraphrasing Help in Better Comprehension?}}
\label{sec:para}
\vspace{-2mm}
\textls[-10]{As discussed, it is apparent that enhanced prompt comprehension correlates with reduced hallucination. Therefore, it is necessary to determine the optimal comprehensible prompt. This premise has led to our experiments with paraphrasing, in which we generate up to 5 paraphrases for a given prompt.}
\vspace{-2mm}

\subsection{Automatic Paraphrasing}
\vspace{-1mm}
When choosing automatic paraphrasing, there are many other factors to consider for e.g., a model may only be able to generate a limited number of paraphrase variations compared to others, but others can be more correct and/or consistent. As such, we consider three major dimensions in our evaluation: \textit{(i) \textbf{coverage}: a number of considerable generations, (ii) \textbf{correctness}: correctness in those generations, and (iii) \textbf{diversity}: linguistic diversity in those generations}. 

\vspace{-2mm}
\begin{table}[H]
\centering
\resizebox{0.6\columnwidth}{!}{%
\begin{tabular}{@{}lcccc@{}}
\toprule
\textbf{Model}           &  \textbf{Coverage}  & \textbf{Correctness} & \textbf{Diversity} \\ \midrule
\textbf{Pegasus}          &   32.46   &    94.38\%       &     3.76      \\
\textbf{T5}               &  30.26       &      83.84\%       &    3.17       \\
\textbf{GPT-3} &   35.51   &   88.16\%      &     7.72      \\
\bottomrule
\end{tabular}%
}
\vspace{-1.5mm}
\caption{\textls[-8]{Experimental results of automatic paraphrasing models based on three factors: \textit{(i) coverage, (ii) correctness, and (iii) diversity}. GPT-3 (\texttt{text-davinci-003}) is the most performant considering all three aspects.}}
\label{tab:my-table}
\end{table}
\vspace{-4mm}

We conducted experiments with three models: (a) Pegasus \cite{zhang2020pegasus}, (b) T5-Large \cite{raffel2020exploring}, and (c) GPT-3 (\texttt{text-davinci-003}) \cite{brown2020language}. Based on empirical observations, we concluded that GPT-3 outperformed all the other models. 
To offer transparency around our experimental process, we  detail coverage, correctness, and diversity, along with the experimental paraphrasing setup, in \ref{sec:appendix-Paraphrasing}.

\subsection{Choosing a Prompt's Optimal Paraphrase}
Suppose the top-performing paraphraser generates the following \ul{\emph{five}} rephrasings for the prompt \emph{``Which individuals possessed the ships that were part of the Boston Tea Party?''}. The objective is to acquire the most comprehensible paraphrase tailored to a specific LLM.

\begin{figure*}[!ht]
        \centering
        \begin{subfigure}[b]{0.41\textwidth}
            \centering
            \begin{tcolorbox}
[boxsep=0pt,left=2.5pt,right=5pt,top=2pt,bottom=5pt,colback=orange!15!white,colframe=orange!45!black]
\footnotesize
\tframed[line width=0.5bp,fill=Navy!65]{\textcolor{white}{\textbf{Original Prompt}}} \tframed[line width=0.5bp,fill=DarkCyan!100]{\textcolor{white}{\textbf{Which individuals possessed the}}} \\ 
\tframed[line width=0.5bp,fill=DarkCyan!100]{\textcolor{white} {\textbf{ ships that were part of the Boston Tea Party?}}} 

\vspace{-1mm}
  \DrawLine

\tframed[line width=0.5bp,fill=DarkViolet!80]{\textcolor{white}{\textbf{Para \S 1}}} \small Who were the owners of the ships associated with the Boston Tea Party? 

\tframed[line width=0.5bp,fill=Orange!80]{\textcolor{white}{\textbf{Para \S 2}}} \small Which individuals possessed the ships that were associated with the Boston Tea Party? 

\tframed[line width=0.5bp,fill=Green!80]{\textcolor{white}{\textbf{Para \S 3}}} \small Who owned which ships were a part of Boston Tea Party? 

\tframed[line width=0.5bp,fill=DarkCyan!60]{\textcolor{white}{\textbf{Para \S 4}}} \footnotesize{What was the identity of those who owned the ships that were associated with the Boston Tea Party?}

\tframed[line width=0.5bp,fill=LimeGreen!100]{\textcolor{white}{\textbf{Para \S 5}}} \small Can you identify the shipkeepers of Boston Tea Party?
\vspace{-1mm}

\end{tcolorbox}
\vspace{-2mm}
            \caption[]%
            {{\small Five paraphrases generated for the original prompt using the T5 paraphrasing model.}}    
            \label{fig:net14}
        \end{subfigure}
        \hfill
        \begin{subfigure}[b]{0.58\textwidth}  
            \centering 
            \includegraphics[width=\textwidth, height=5.5cm]{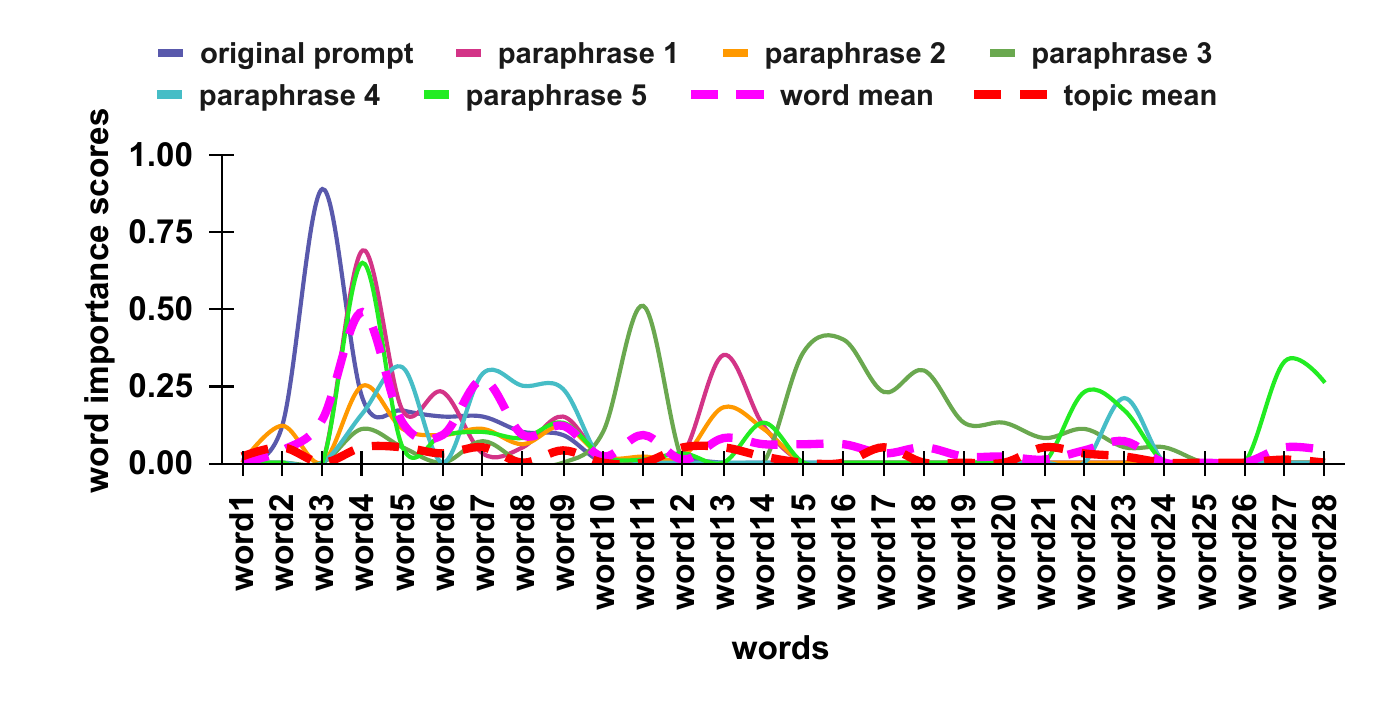}
            \caption[]%
            {{\small Word importance scores distribution for the original prompt and its five paraphrases. The \textcolor{vpurple}{purple} dashed line shows the mean of the IGs while the \textcolor{red}{red} dashed line shows the topic mean.}}    
            \label{fig:net24}
        \end{subfigure}
        \vspace{-6.6mm}
        \caption[]%
            {{\small (a) Paraphrased versions for a given prompt; (b) Per-word importance score distribution for each paraphrase.}} 
        \label{fig:pause-ig}
\end{figure*}

\begin{algorithm*}[!ht]
\caption{Finding the optimal paraphrased prompt}\label{alg:para}
\begin{algorithmic}[1]
\footnotesize
\State Find out the topics for the original prompt 

\For{$i$ in {1...5}}
    \State a: Compute the IG, DIG, and SIG and b: an \textbf{average gradient} = $\frac{IG + DIG + SIG}{3}$ for $paraphrased\_prompt_i$
    \State Compute the mean of all the gradients across various tokens
    \State Find out the topics for $paraphrased\_prompt_i$ 
    \State Calculate the \colorbox{LightSalmon!75}{distance} of the mean prompt from the $paraphrased\_prompt_i$
    \State Calculate the \colorbox{SkyBlue!85}{topic similarity} between the original prompt and the $paraphrased\_prompt_i$
\EndFor

\State Calculate a weighted average \text{\textbf{Comprehension Score}} = $(w_1 \times \text{distance} + w_2 \times \text{topic similarity})$ where, $w_1$ and $w_2$ are equal weights.

\State Select the $paraphrased\_prompt_i$ with the highest weighted average as the \textbf{optimal} $paraphrased\_prompt$  

\end{algorithmic}
\end{algorithm*}

LLM comprehension is determined by two factors: (i) whether all the words in a given prompt are well-read, indicated by having an IG score above a threshold and (ii) whether all the topic words are well-read by the LLM. The overall approach is illustrated in \cref{alg:para}. This process employs a two-step method, as described below. Further details are available in \cref{app:opt-para}.
\paragraph{Distance} We compute integrated gradients for paraphrased prompts, calculate their mean and measure the distance of each paraphrased prompt from the mean using cosine similarity.
\textls[-10]{\paragraph{Topic Modeling} To address potential oversights in hidden word patterns, we include topic modeling using LDA \cite{blei2003latent}. This involves identifying topics for both the original prompt and paraphrases. Topic similarity scores are then employed to determine the most similar topics between a paraphrase and the original prompt.
The final selection is determined by calculating distance and topic similarity for these two steps and then computing a weighted average. \emph{Having spent much of my career studying various combination methods, it has been somewhat frustrating to consistently find that the simple average performs so well empirically.} \cite{clemen2008comment}. The optimal paraphrase is chosen based on the highest average score. It is crucial to highlight that the original prompt itself may be the optimal prompt.}
\vspace{-1mm}

\begin{figure*}[!ht]
    \centering
    \includegraphics[width=\textwidth]{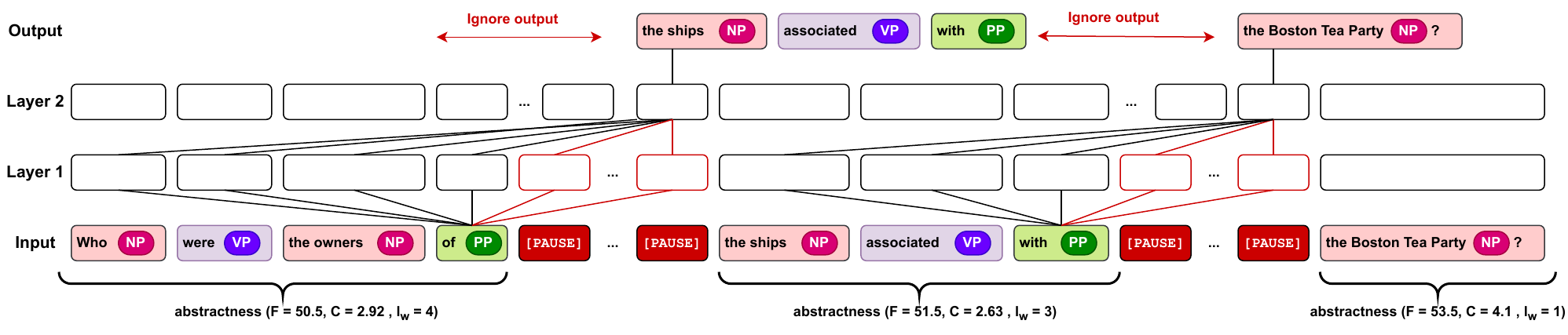}
    \vspace{-6mm}
    \caption{We use conjunct \tframed[line width=0.5bp,fill=vgreen]{\textcolor{white}{\texttt{\textbf{PP}}}} to split the long prompt. We use standard POS tagging \cite{akbik2018coling}. 
    Two \tframed[line width=0.5bp,fill=vred]{\textcolor{white}{\texttt{\textbf{[PAUSE]}}}} tokens are appended after \tframed[line width=0.5bp,fill=vgreen]{\textcolor{white}{\texttt{\textbf{PP}}}} based on the concreteness score of the chunk before the \tframed[line width=0.5bp,fill=vred]{\textcolor{white}{\texttt{\textbf{[PAUSE]}}}} tokens. Hence, it ignores, meaning it \emph{breathes} for the next two tokens, as shown by \textcolor{red}{Ignore output}.}
    \label{fig:pause}
    \vspace{-3mm}
\end{figure*}

\section{LLMs Need to Breathe While Reading!}  
\label{sec:pause}

\vspace{-1mm}
The `\emph{lost in the middle}' phenomenon, as introduced by \cite{liu2023lost}, illustrates that a substantial amount of information contained in the middle section of lengthy input prompts is overlooked during the comprehension process by LLMs. 
Recently, the introduction of \tframed[line width=0.5bp,fill=vred]{\textcolor{white}{\texttt{\textbf{[PAUSE]}}}} tokens demonstrated improvements in reasoning tasks \cite{goyal2023think}. Based on these findings and the `lost in the middle' phenomenon, we propose that inserting \tframed[line width=0.5bp,fill=vred]{\textcolor{white}{\texttt{\textbf{[PAUSE]}}}} tokens may enhance LLM comprehension of longer prompts, potentially minimizing hallucination. Empirical results support this hypothesis.

\begin{tcolorbox}[
left=3pt,right=2pt,colback=SteelBlue!5!white,colframe=SteelBlue!75!black,colbacktitle=SteelBlue!90,
  title=\footnotesize \fontfamily{qbk} \selectfont \textbf{Our Contributions related to \tframed[line width=0.5bp,fill=vred]{\textcolor{white}{\texttt{\textbf{[PAUSE]}}}} tokens} ]
  
\vspace{-2mm}
\begin{itemize}
[labelindent=0em,labelsep=0.1cm,leftmargin=*]
\setlength\itemsep{0em}
\begin{spacing}{0.65}

\item[\ding{64}] 
{\footnotesize 
{\fontfamily{phv}\fontsize{8}{9}
\selectfont
\textbf{Where to inject \tframed[line width=0.5bp,fill=vred]{\textcolor{white}{\texttt{\textbf{[PAUSE]}}}} token(s)?} We propose clause boundary aka injecting \tframed[line width=0.5bp,fill=vred]{\textcolor{white}{\texttt{\textbf{[PAUSE]}}}} after conjunction. 
}
}
\vspace{-1mm}

\item[\ding{64}] 
{\footnotesize 
{\fontfamily{phv}\fontsize{8}{9}
\selectfont
\textbf{How many \tframed[line width=0.5bp,fill=vred]{\textcolor{white}{\texttt{\textbf{[PAUSE]}}}} token(s)?} We propose a content-based method for \tframed[line width=0.5bp,fill=vred]{\textcolor{white}{\texttt{\textbf{[PAUSE]}}}} injection. 
}
}
\vspace{-1mm}
\item[\ding{64}] 
{\footnotesize
{\fontfamily{phv}\fontsize{8}{9}
\selectfont
\textbf{Best fine-tuning method(s)?} We introduce a novel finetuning paradigm named reverse proxy tuning. 
}
}

\vspace{-5.5mm}
\end{spacing}
\end{itemize}
\end{tcolorbox}
\vspace{-2mm}

\subsection{Where to Inject \tframed[line width=0.5bp,fill=vred]{\textcolor{white}{\texttt{\textbf{[PAUSE]}}}} Tokens?} \label{sec:where-pause}

In their work, \cite{goyal2023think} suggested an overall insertion of 10\% \tframed[line width=0.5bp,fill=vred]{\textcolor{white}{\texttt{\textbf{[PAUSE]}}}} tokens; however, they did not provide specific guidelines or methods for determining the optimal positions for inserting \tframed[line width=0.5bp,fill=vred]{\textcolor{white}{\texttt{\textbf{[PAUSE]}}}}.
We posit that the most effective location for injecting the \tframed[line width=0.5bp,fill=vred]{\textcolor{white}{\texttt{\textbf{[PAUSE]}}}} token should be at clause boundaries. However, identifying these boundaries comes with its own set of challenges. As a simple approach, we have opted to insert the \tframed[line width=0.5bp,fill=vred]{\textcolor{white}{\texttt{\textbf{[PAUSE]}}}} token after conjunctions, illustrated in \cref{fig:pause}.

\subsection{How Many \tframed[line width=0.5bp,fill=vred]{\textcolor{white}{\texttt{\textbf{[PAUSE]}}}} Tokens?} \label{sec:how-pause}

The study by \cite{goyal2023think} did not definitively assert the ideal quantity of \tframed[line width=0.5bp,fill=vred]{\textcolor{white}{\texttt{\textbf{[PAUSE]}}}} tokens. Their experimentation ranged from 2 to 50 tokens, with a general conclusion that around 10 tokens were optimal, though this determination varied depending on the specific task. In contrast, we propose a content-based approach. 

Our assessment of their impact on LLM comprehension revealed that readability provides a weaker signal compared to formality and concreteness. We define a combined measure called \emph{\ul{abstractness}}: $abs = \frac{\delta_1*F+\delta_1*C}{l_w}$, where $\delta_1$, and $\delta_2$ are coefficients. $F$ is the formality measure, $C$ is the concreteness measure, and $l_w$ is the length of text in terms of the words. Additionally, we divided abstractness into three ranges—high, mid, and low—based on the overall distribution, mean, and standard deviations. Our method involves utilizing the abstractness score of the text preceding a \tframed[line width=0.5bp,fill=vred]{\textcolor{white}{\texttt{\textbf{[PAUSE]}}}} token to determine the appropriate number of tokens required. Higher abstractness scores suggest a lower (2) necessity to pause, whereas lower scores indicate a greater need for the language model to pause for comprehension, necessitating more (10) tokens. For the mid range abstractness we decide to insert five \tframed[line width=0.5bp,fill=vred]{\textcolor{white}{\texttt{\textbf{[PAUSE]}}}} tokens. 
The mechanism for inserting  \tframed[line width=0.5bp,fill=vred]{\textcolor{white}{\texttt{\textbf{[PAUSE]}}}} has been illustrated in \cref{fig:pause}. Please refer to \textcolor{black}{Appendix \ref{appendix-pause} for more details.}

\begin{figure}[!ht]
    \centering
    \includegraphics[width=0.45\textwidth]{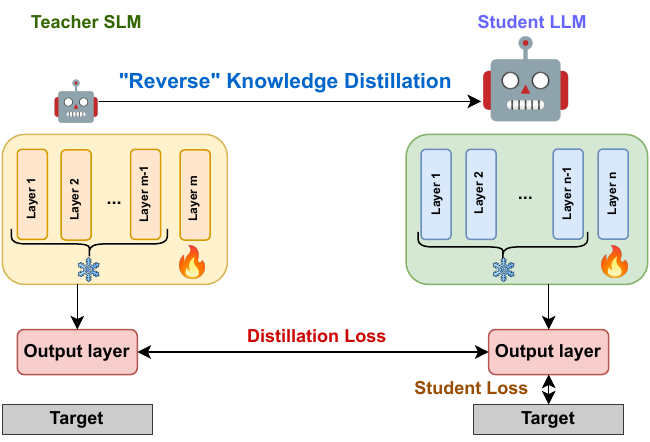}
    \caption{\textls[-10]{\textbf{Reverse Proxy-Tuning:} SLM is used to fine-tune LLM. First, SLM is fine-tuned on SQuAD where all the hidden layers except the last one are frozen. This fine-tuned SLM is further used to distill knowledge to the LLM, where all hidden layers of the LLM except the last one are frozen.}} 
    \label{fig:sca-rev-pt}
\end{figure}

\begin{figure}[!ht]
        \centering
        \begin{subfigure}[b]{0.49\linewidth}
            \centering
            \includegraphics[width=\textwidth, height=2.3cm]{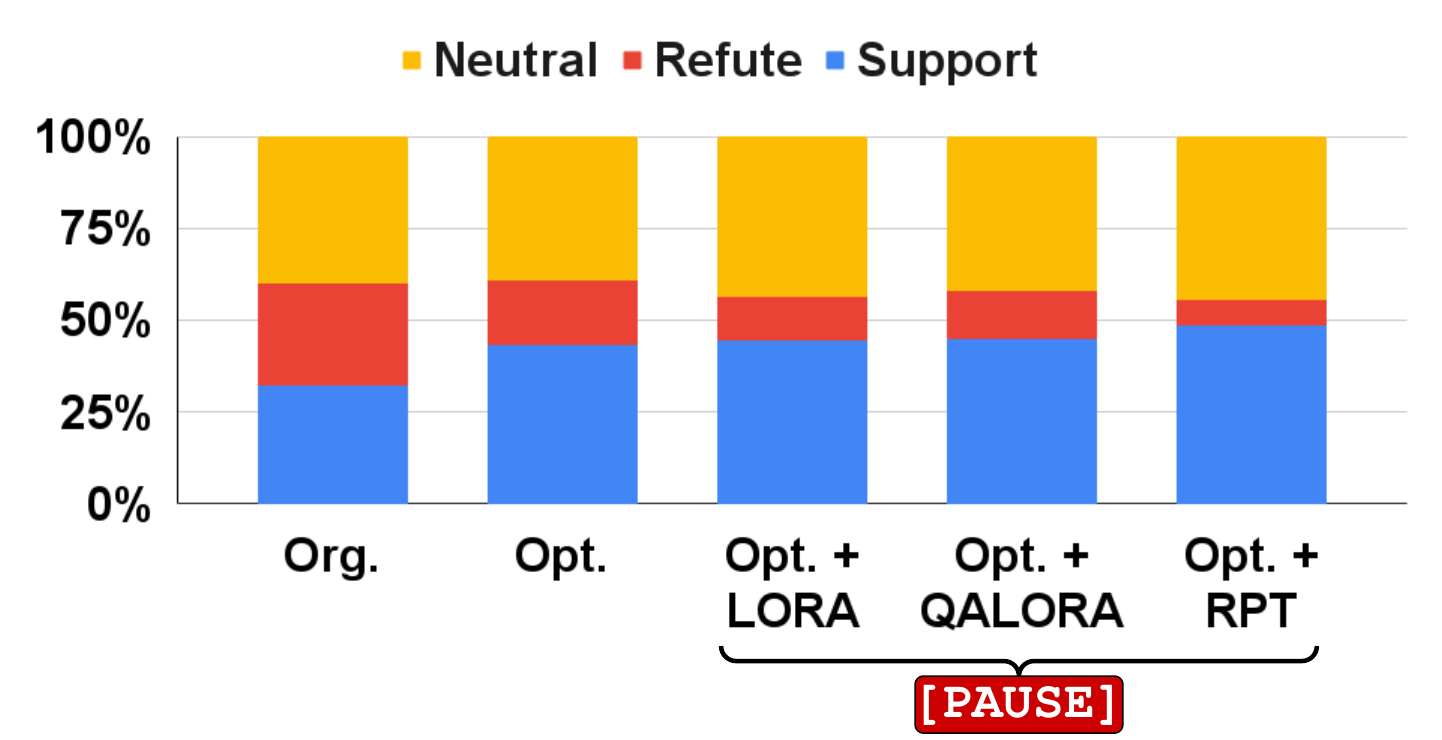}
            \caption[]%
            {{\small Person}}    
            \label{fig:net14}
        \end{subfigure}
        \hfill
        \begin{subfigure}[b]{0.49\linewidth}  
            \centering 
            \includegraphics[width=\textwidth, height=2.3cm]{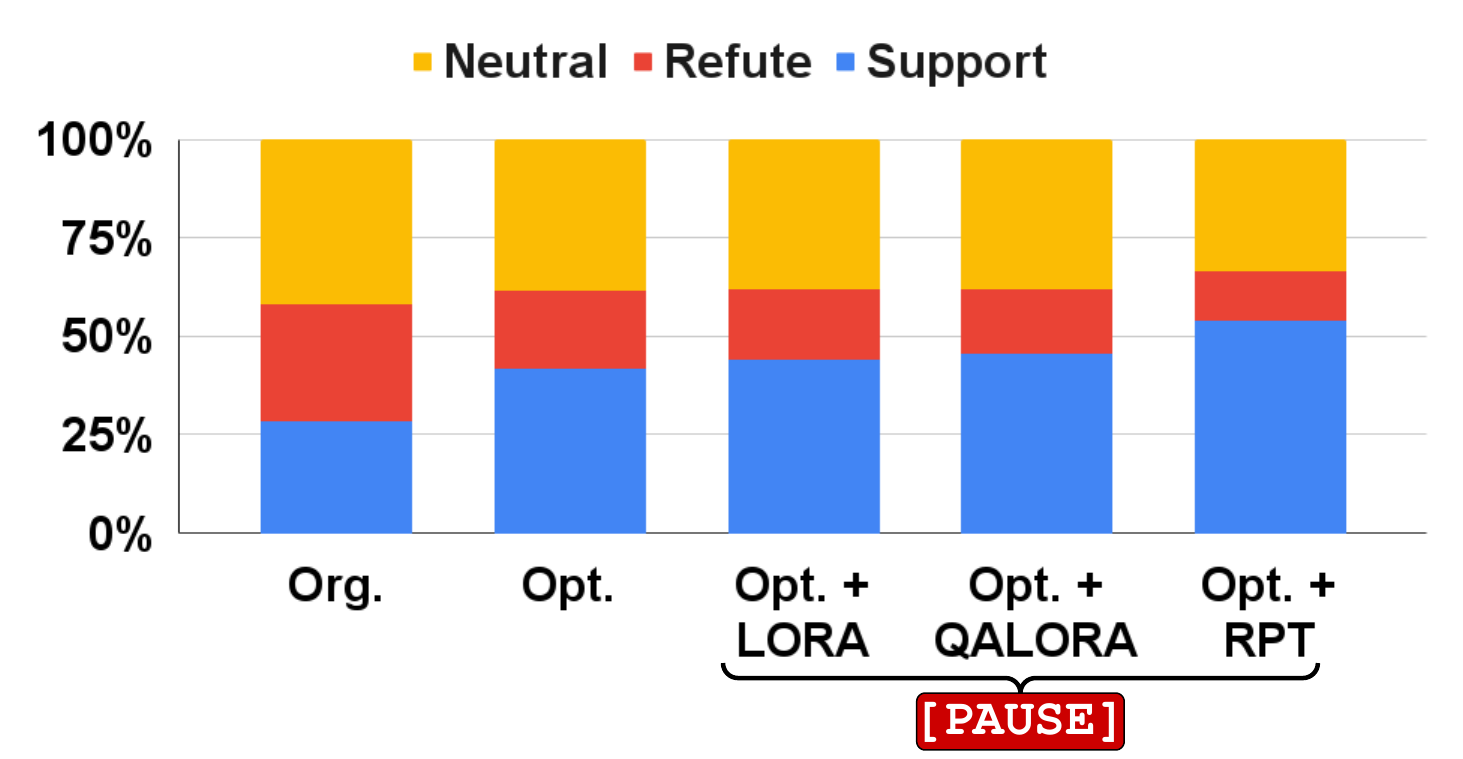}
            \caption[]%
            {{\small Location}}    
            \label{fig:net24}
        \end{subfigure}
        \hfill
        \vskip\baselineskip
        \vspace{-9mm}
        \begin{subfigure}[b]{0.49\linewidth}   
            \centering 
            \includegraphics[width=\textwidth, height=2.3cm]{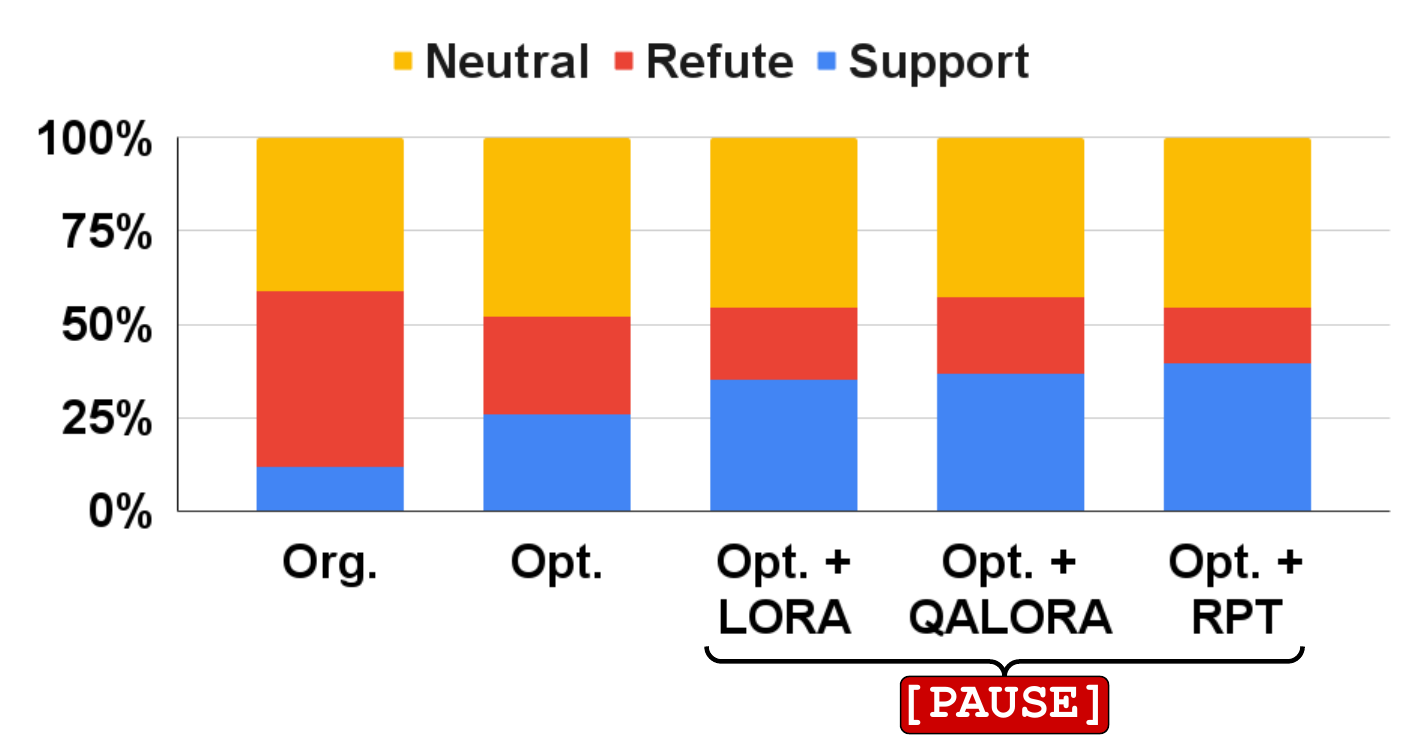}
            \caption[]%
            {{\small Numeric}}    
            \label{fig:net34}
        \end{subfigure}
        \hfill
        \begin{subfigure}[b]{0.49\linewidth}   
            \centering 
            \includegraphics[width=\textwidth, height=2.3cm]{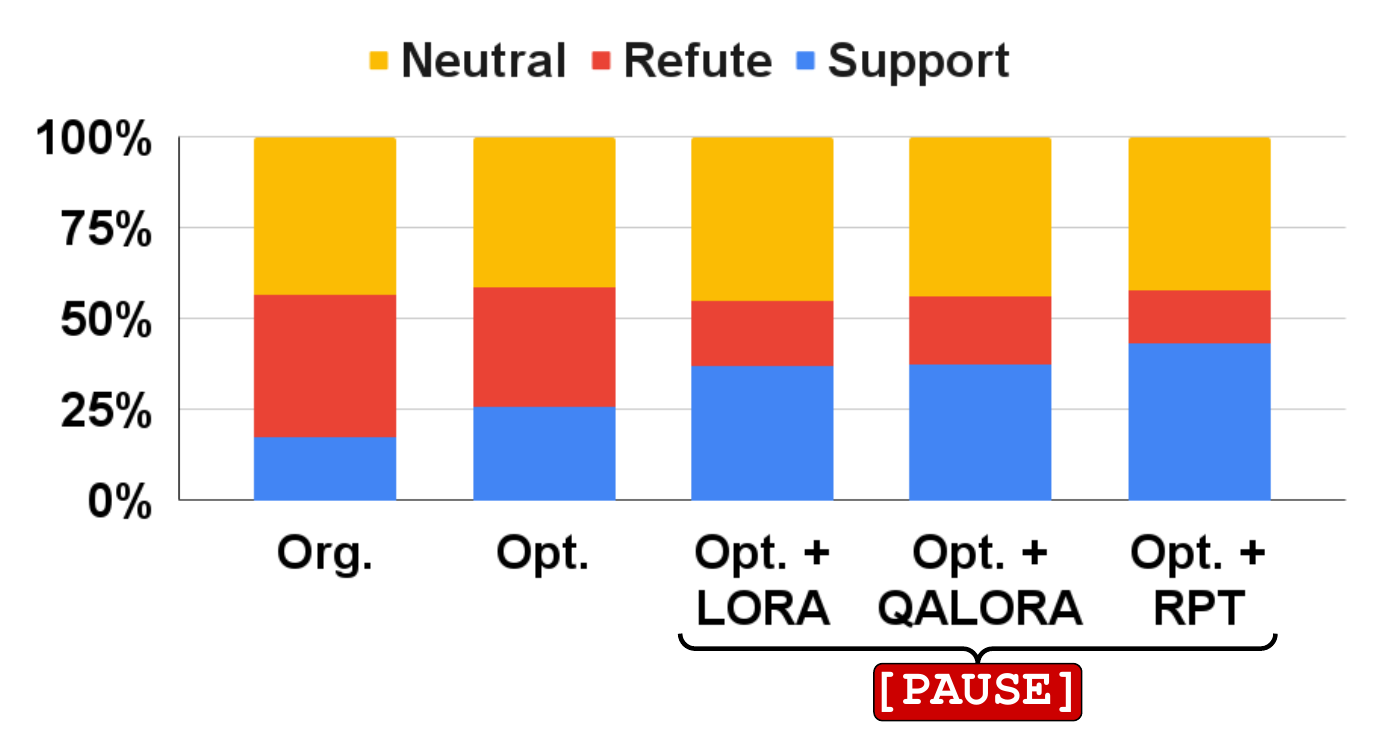}
            \caption[]%
            {{\small Time}}    
            \label{fig:net44}
        \end{subfigure}
        \caption[]%
            {{\textls[-10]{Empirical results for reverse proxy tuning using optimal prompt and \tframed[line width=0.5bp,fill=vred]{\textcolor{white}{\texttt{\textbf{[PAUSE]}}}} token for \emph{four} different hallucination categories. . \textbf{Org.}: Original Prompt and \textbf{Opt.}: Optimal Paraphrase + LDA topics. These results indicate an overall average for all the 21 LLMs.}}}
        \label{fig:nets}
\end{figure}

\begin{figure*}[!htb]
    \centering
    \includegraphics[width=0.98\textwidth, height=5.5cm]{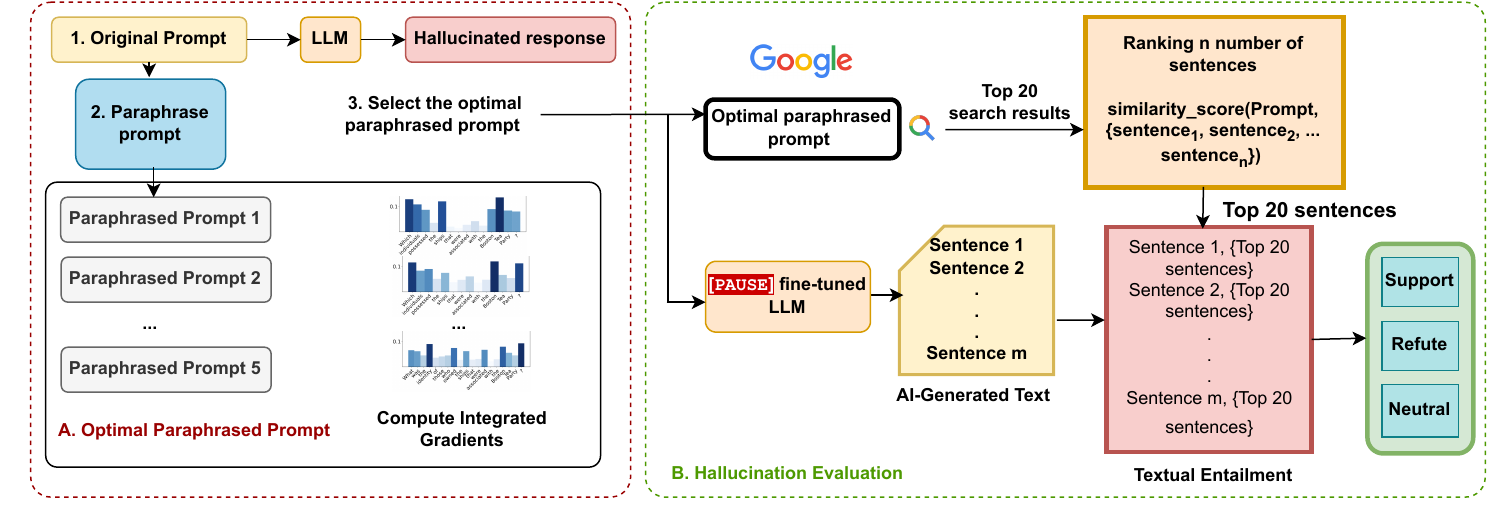}    
    \vspace{-1mm}
\caption{\emph{\bfseries\fontfamily{pcr}\selectfont{ACTIVATOR}} is a two-part end-to-end pipeline: \textbf{1. Optimal Paraphrased Prompt selection:} Using the \cref{alg:para}, an optimal prompt is selected by computing the average IG. \textbf{2. Hallucination Evaluation: } With the chosen optimal prompt, textual entailment is done to verify whether the AI-generated response is correct.}
    \label{fig:lingo}
    \vspace{-3mm}
\end{figure*}
\vspace{-1mm}

\subsection{Reverse Proxy-Tuning}  \label{sec:rev-pt}

\cite{goyal2023think} did not extensively explore a range of state-of-the-art (SoTA) fine-tuning techniques, such as LoRA, QALoRA, or ReLoRA, particularly regarding the injection of \tframed[line width=0.5bp,fill=vred]{\textcolor{white}{\texttt{\textbf{[PAUSE]}}}} tokens. These techniques fall into three broad categories: \textbf{1. Prompt Modifications:} Examples include Soft Prompt Tuning, Soft Prompt vs. Prompting, Prefix Tuning, and Hard Prompt Tuning. \textbf{2. Adapter Methods:} Such as LLaMA-Adapters. \textbf{3. Reparameterization:} Including Low Rank Adaptation (LoRA) \cite{hu2021lora}, Quantized Low-Rank Adaptation (QLoRA) \cite{dettmers2023qlora}, Quantization-Aware Low-Rank Adaptation (QALoRA) \cite{xu2023qa}, and Refined Low-Rank Adaptation (ReLoRA) \cite{lialin2023stack}.

Although the above-mentioned fine-tuning methods are much more efficient for fine-tuning LLMs, they are still computationally expensive for our purpose -- single modification to the prompt -- adding  \tframed[line width=0.5bp,fill=vred]{\textcolor{white}{\texttt{\textbf{[PAUSE]}}}} token(s). So in this work, we use the small language model (SLM) to fine-tune the larger language model. We adopt this idea from Knowledge Distillation (KD) \cite{44873, gu2023knowledge,hsieh-etal-2023-distilling}. The core concept in KD is distilling the knowledge from a larger model (Teacher) to a smaller model (Student). In this process, the Student not only learns from the expected labels but also from the Teacher. During this distillation, all the layers are updated using a loss function. However, changing weights for all layers is also computationally expensive. Therefore, in our case, we only choose the last output layer for fine-tuning and freeze all the layers. Additionally, we use an SLM to fine-tune the LLM which is reverse KD. We were further inspired by \cite{liu2024tuning}, which presents \emph{Proxy Tuning}. Here, we introduce a novel approach called \textbf{Reverse Proxy-Tuning} (RPT), depicted in \cref{fig:sca-rev-pt}, where the SLM serves as a proxy model. This method is computationally efficient as it involves updating only the last layer and utilizing an SLM to fine-tune an LLM. Experimental results are illustrated in \cref{fig:nets}.

 

\vspace{-1mm}
\subsection{Dataset and Experimental Setup for \tframed[line width=0.5bp,fill=vred]{\textcolor{white}{\texttt{\textbf{[PAUSE]}}}} finetuning} \label{sec:pause-ft}

\textls[-10]{For all our fine-tuning experiments, we use the CommonsenseQA dataset \cite{talmor-etal-2019-commonsenseqa}. We have implemented two baselines: QLoRA \cite{dettmers2023qlora} and QALoRA \cite{xu2023qa}. The proposed novel reverse proxy-tuning yielded better performance than these two baselines. Further details regarding the experimental setup, including hyperparameters and other specifics, can be found in the \cref{tab:hyperp} in the Appendix}.
\section{Does Better Comprehension Guarantee Lesser Hallucination?}
\vspace{-1mm}

\textls[-12]{This question is likely to captivate the reader's attention significantly. Enhancing comprehension and mitigating hallucinations in LLMs may initially appear as two distinct considerations. The subsequent query that naturally arises is how we discern hallucinations after furnishing an optimal prompt to the LLM. We have chosen the entailment approach to empirically evaluate whether overall support scores improve following the implementation of SCA. Support scores signify factual entailment.

While there's no assurance that the most comprehensible prompt will completely eliminate hallucinations, the results depicted in \cref{fig:nets} provide empirical evidence of improvement in overall entailment support scores across all the hallucination classes. Additional details on entailment-based fact verification are provided in \cref{sec:ent-label}}.

\begin{tcolorbox}[
left=3pt,right=2pt,colback=RosyBrown!5!white,colframe=RosyBrown!75!black,colbacktitle=RosyBrown,
  title=\footnotesize \fontfamily{qbk} \selectfont \textbf{Takeaways related to Reverse Proxy Tuning} ]
  
\vspace{-2mm}
\begin{itemize}
[labelindent=0.1em,labelsep=0.1cm,leftmargin=*]
\setlength\itemsep{0em}
\begin{spacing}{0.85}

\item[\ding{237}] 
{\footnotesize 
{\fontfamily{phv}\fontsize{8}{9}
\selectfont
Optimal paraphrase + LDA yields better results for both Number and Time categories.
}
}
\vspace{-1mm}
\item[\ding{237}] 
{\footnotesize 
{\fontfamily{phv}\fontsize{8}{9}
\selectfont
We see marginal betterment for the Person and Location categories with Lora and QALora and a significant boost for the Number and Time categories.
}
}
\vspace{-1mm}
\item[\ding{237}] 
{\footnotesize 
{\fontfamily{phv}\fontsize{8}{9}
\selectfont
Among all other fine-tuning techniques, reverse proxy-tuning performs the best across all four categories.
}
}

\vspace{-6mm}
\end{spacing}
\end{itemize}
\end{tcolorbox}
\vspace{-2mm}





\vspace{-1mm}
\section{\emph{\fontfamily{pcr}\selectfont{ACTIVATOR}} - A Reprompter} \label{sec:activator}
\vspace{-1mm}

We propose the \emph{\bfseries\fontfamily{pcr}\selectfont{ACTIVATOR}} pipeline to automatically rephrase and evaluate the prompt as shown in \cref{fig:lingo}. Activator is an end-to-end pipeline that accepts a prompt as input and outputs an entailment score. This process involves pre-processing the input prompt to add \tframed[line width=0.5bp,fill=vred]{\textcolor{white}{\texttt{\textbf{[PAUSE]}}}} tokens, paraphrasing the input prompts to identify the most optimal prompt which maximizes comprehension by minimizing distance to the mean prompt and maximizing topic similarity based on the original prompt based on a mean of the integrated gradients score. This optimal prompt undergoes sentence-level entailment based on a web lookup to yield final entailment scores. 


\vspace{-1mm}
\section{Conclusion} \label{sec:conclusion}
\vspace{-1mm}
In this preliminary research study, we begin by categorizing the primary types of hallucinations present in LLMs. Subsequently, we compile our dataset by utilizing New York Times news tweets, aligning with these established categories. Language intricacies assume a crucial role in the comprehension of language. Therefore, we delve into the examination of three significant linguistic dimensions: readability, formality, and concreteness, and their potential influence on the occurrence of hallucinations in LLMs.

\newpage
\newpage

\section{Discussion and Limitations} \label{sec:limitaions}
\vspace{-2mm}
\textbf{Discussion:} On June 14th, 2023, the European Parliament successfully passed its version of the EU AI Act \cite{euaiproposal}. Following this, many other countries began discussing their stance on the evolving realm of Generative AI. A primary agenda of policymaking is to protect citizens from political, digital, and physical security risks posed by Generative AI. While safeguarding against misuse is crucial, one of the biggest concerns among policymakers is the occurrence of unwanted errors by systems, such as hallucination (source: https://cetas.turing.ac.uk/publications/rapid-rise-generative-ai).

\textbf{Limitations:} In this paper, we present 
several key findings: (i) LLM comprehension, (ii) paraphrasing can improve LLM comprehension, (iii) optimal paraphrasing, (iv) \tframed[line width=0.5bp,fill=vred]{\textcolor{white}{\texttt{\textbf{[PAUSE]}}}} injection, and (v) finally empirically show that the overall hallucination is reducing due to better LLM comprehension. We believe the following aspects require critical attention in future endeavours.

\textbf{Limitation 1: The three linguistic properties are NOT independent.} Certainly, these factors are not mutually exclusive. Our assessment of their impact on LLM comprehension revealed that readability provides a weaker signal compared to formality and concreteness. As a result, we have chosen to prioritize concreteness as the actionable feature.

\textbf{Limitation 2: Which explainability method is the best?} Integrated Gradient (IG) has long served as a fundamental principle governing explainability methods in deep neural networks. Despite recent advancements such as DIG and SIG, which have shown improved performance in various contexts, we were uncertain about their effectiveness for our specific use case of hallucination detection. Therefore, we opted for a more cautious approach and decided to average the results obtained from all three methods. A suitable explainability method for hallucination could be a nice future direction to explore.


\textbf{Limitation 4: Is fine-tuning the ONLY method?} One could argue that instead of fine-tuning, we could have explored techniques like In-Context Learning (ICL), Zero-Shot, and Few-Shot learning for [PAUSE] insertion. Some team members believe that ICL might yield competitive results compared to fine-tuning. However, due to time constraints, we were unable to conduct these experiments. Nevertheless, we acknowledge that exploring these techniques could be a valuable direction for future research.


\section{Ethical Considerations}
Through our experiments, we have uncovered the susceptibility of LLMs to hallucination. 
While emphasizing the vulnerabilities of LLMs, our goal is to underscore their current limitations. However, it's crucial to address the potential misuse of our findings by malicious entities who might exploit AI-generated text for nefarious purposes, such as designing new adversarial attacks or creating fake news that is indistinguishable from human-written content. We strongly discourage such misuse and strongly advise against it.

\bibliography{custom}
\bibliographystyle{acl_natbib}

\newpage
\newpage
\onecolumn

    
\section*{Frequently Asked Questions (FAQs)}\label{sec:FAQs}

\begin{itemize}[leftmargin=*,nolistsep]

     \item[\ding{93}] {\fontfamily{lmss} \selectfont \textbf{Why do you select those 21 large language models?}}
     \begin{description}
     \item[\ding{224}] We want to select several language models with varying parameter sizes for our experiments - ranging from large to small. Hence, the above chosen models consist of both large models like GPT-3, LLaMa and smaller ones like T5 and T0.
     \end{description}

     \item[\ding{93}] {\fontfamily{lmss} \selectfont \textbf{Why only three linguistic properties are selected for this study?}}
     \begin{description}
     \item[\ding{224}] As far as we know, formality, readability, and concreteness appear to be the most obvious criteria for assessing LLM comprehension.
     \end{description}
\vspace{2mm}
     \item[\ding{93}] {\fontfamily{lmss} \selectfont \textbf{What is the purpose of calculating integrated gradients? Why not simply use attention scores?}}
     \begin{description}
     \item[\ding{224}] Integrated Gradient provides an explanatory score at the word level, indicating how the LLM interprets each word and generates output. In contrast, attention scores only reveal the encoding side of processing.
     \end{description}
\vspace{2mm}
     \item[\ding{93}] {\fontfamily{lmss} \selectfont \textbf{Why do you only generate five paraphrases?}}
     \begin{description}
     \item[\ding{224}] We conducted a study to assess the limit of how many ways a single sentence could be paraphrased. Our findings suggest that there is indeed a limit, as generating too many paraphrases can disrupt diversity. Through experimentation, we have observed that five paraphrases is the optimal number.
     \end{description}
\vspace{2mm}
     \item[\ding{93}] {\fontfamily{lmss} \selectfont \textbf{What are the broad implications of the \emph{\fontfamily{pcr}\selectfont{ACTIVATOR}} framework for hallucination mitigation?}}
     \vspace{-4mm}
     \begin{description}
     \item[\ding{224}] The primary aim of \texttt{ACTIVATOR} is automation. End users might lack proper training and understanding of linguistic properties like formality, readability, or concreteness. Additionally, the functioning of LLMs is often a black box for end users. \texttt{ACTIVATOR} serves to assist end users in obtaining the best non-hallucinated output from LLMs.

     \end{description}

\end{itemize}    
\newpage
\appendix

\section{Appendix}
\label{sec:appendix}

This section provides supplementary material in the form of additional examples, implementation details, etc. to bolster the reader's understanding of the concepts presented in this work.

\section{Linguistic Nuances}
Linguistic nuances refer to subtle variations in language that convey additional meaning or context beyond the literal interpretation. \textbf{Readability} pertains to how easily text can be understood, often influenced by sentence structure and vocabulary. \textbf{Formality} involves the level of politeness or professionalism in language, ranging from casual to formal expressions. \textbf{Concreteness} relates to the degree of specificity and tangible details in language, with concrete language being more explicit and tangible than abstract language. These nuances contribute to the overall tone, clarity, and effectiveness of communication.

\begin{figure*}[!ht]
        \centering
        \begin{subfigure}[b]{0.45\textwidth}
            \centering
            \includegraphics[width=\textwidth]{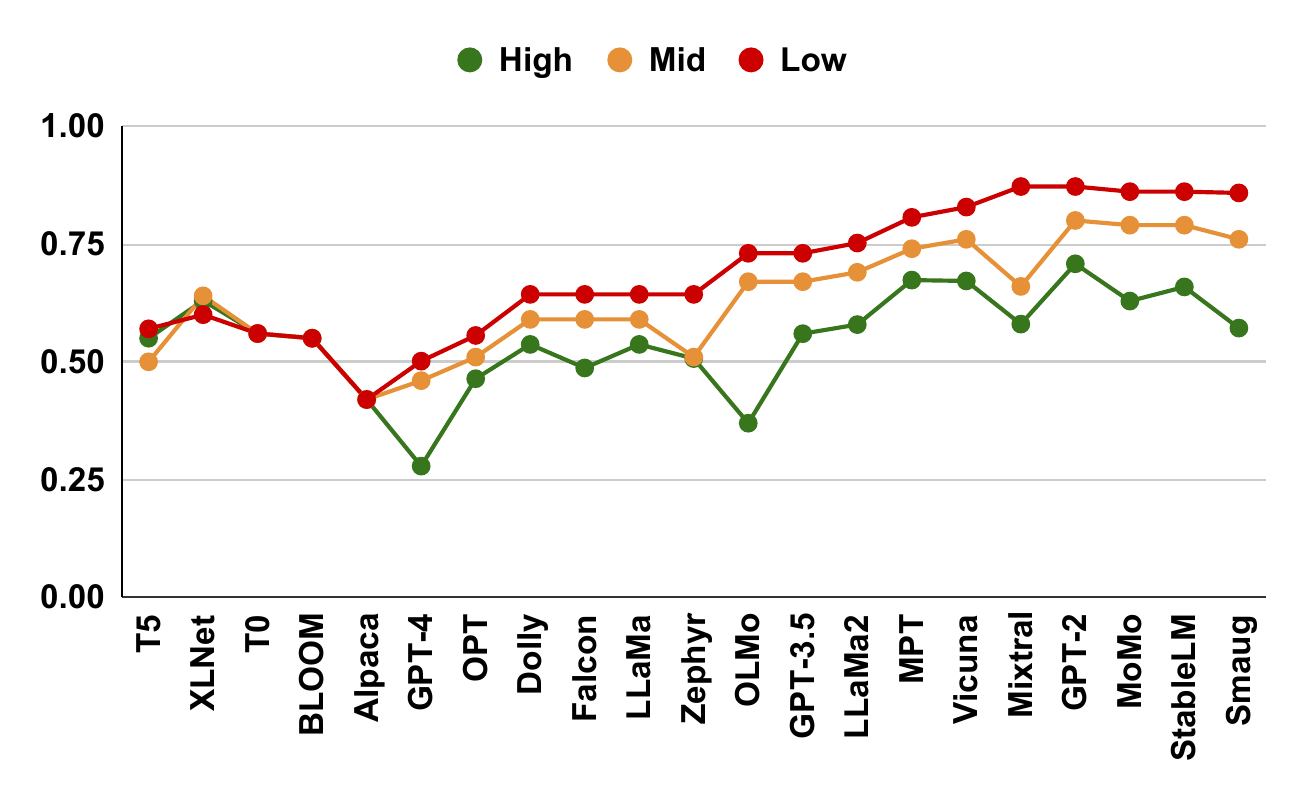}
            \caption[]%
            {{\small Person}}    
            \label{fig:mean and std of net14}
        \end{subfigure}
        \hfill
        \begin{subfigure}[b]{0.45\textwidth}  
            \centering 
            \includegraphics[width=\textwidth]{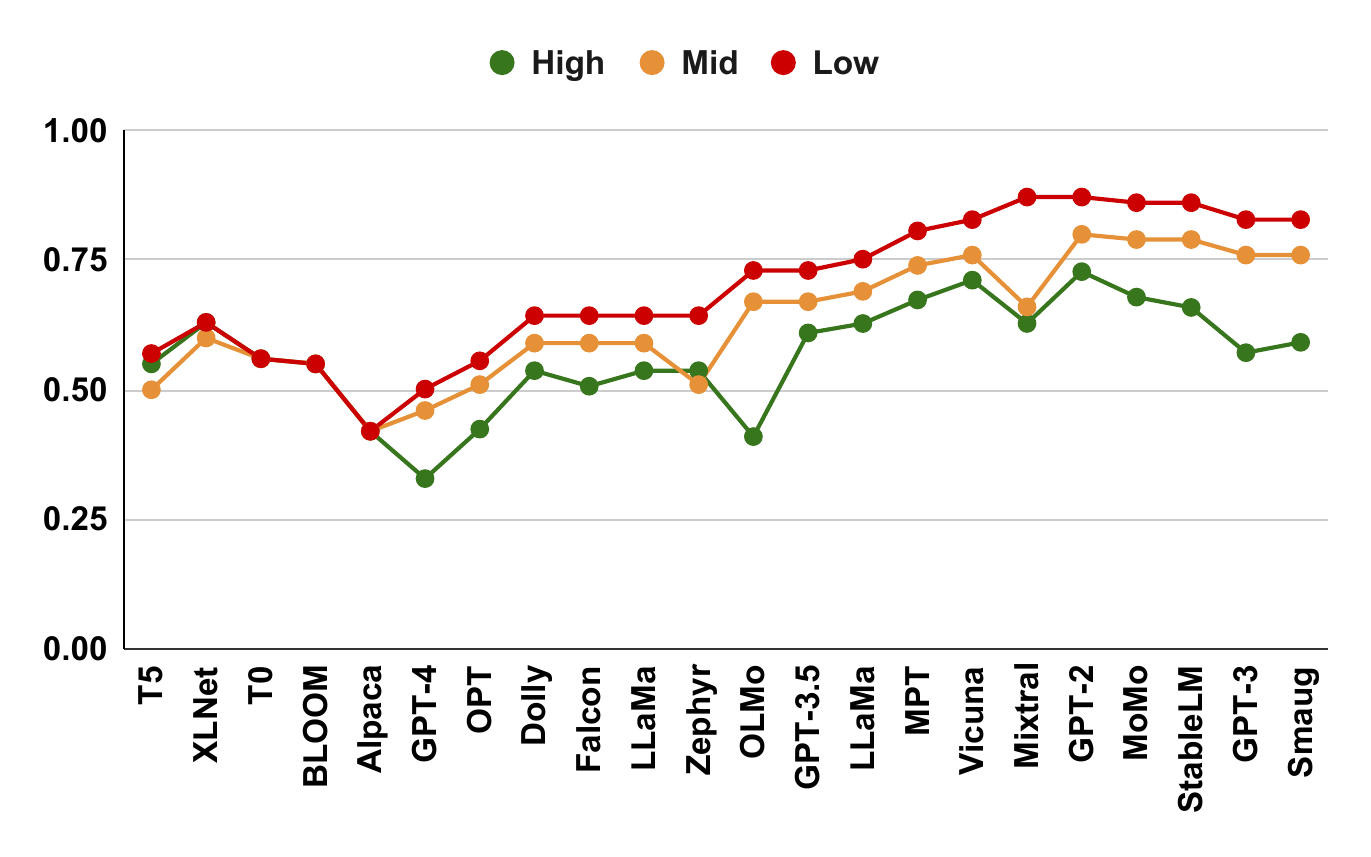}
            \caption[]%
            {{\small Location}}    
            \label{fig:mean and std of net24}
        \end{subfigure}
        \hfill
        \vskip\baselineskip
        \begin{subfigure}[b]{0.45\textwidth}   
            \centering 
            \includegraphics[width=\textwidth]{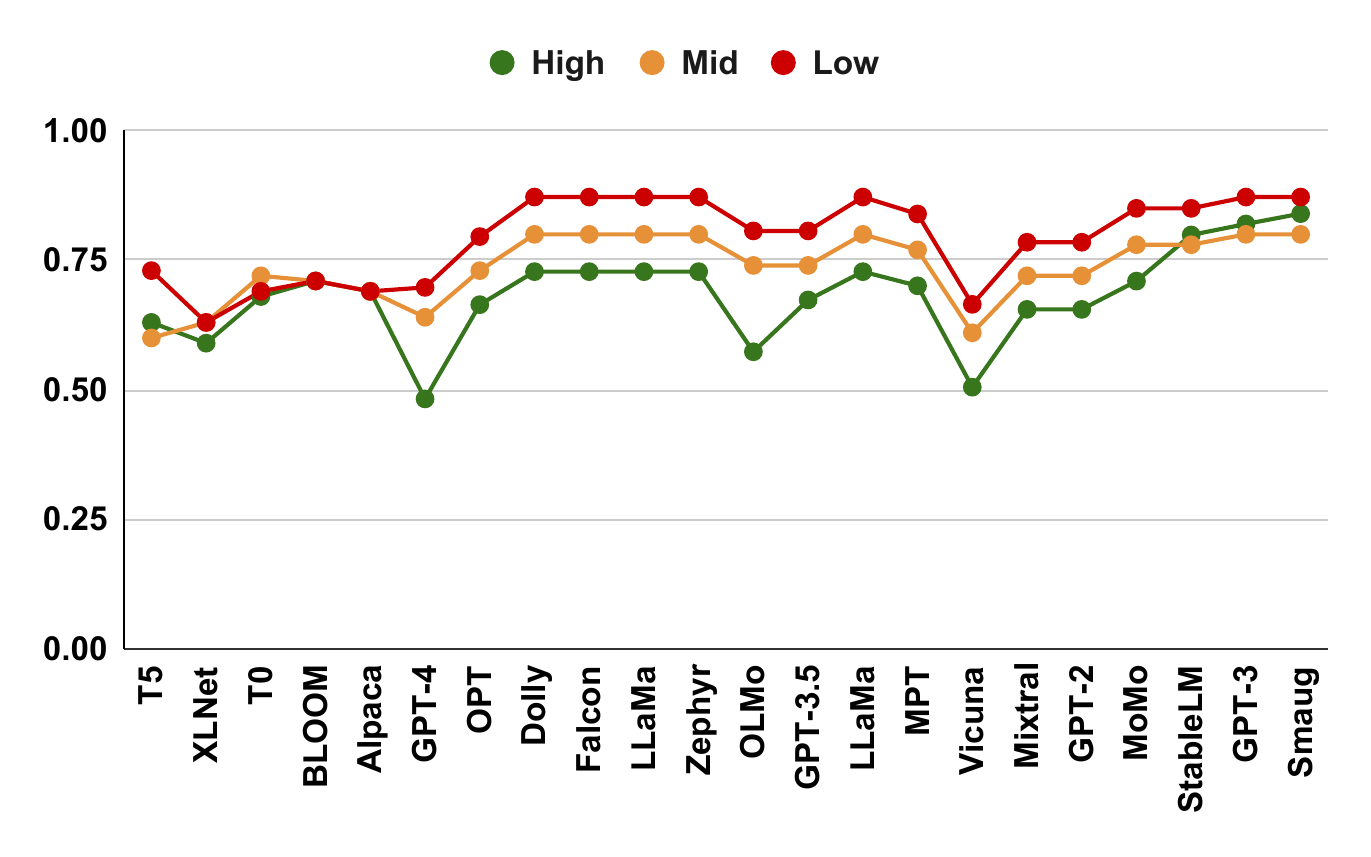}
            \caption[]%
            {{\small Number}}    
            \label{fig:mean and std of net34}
        \end{subfigure}
        \hfill
        \begin{subfigure}[b]{0.45\textwidth}   
            \centering 
            \includegraphics[width=\textwidth]{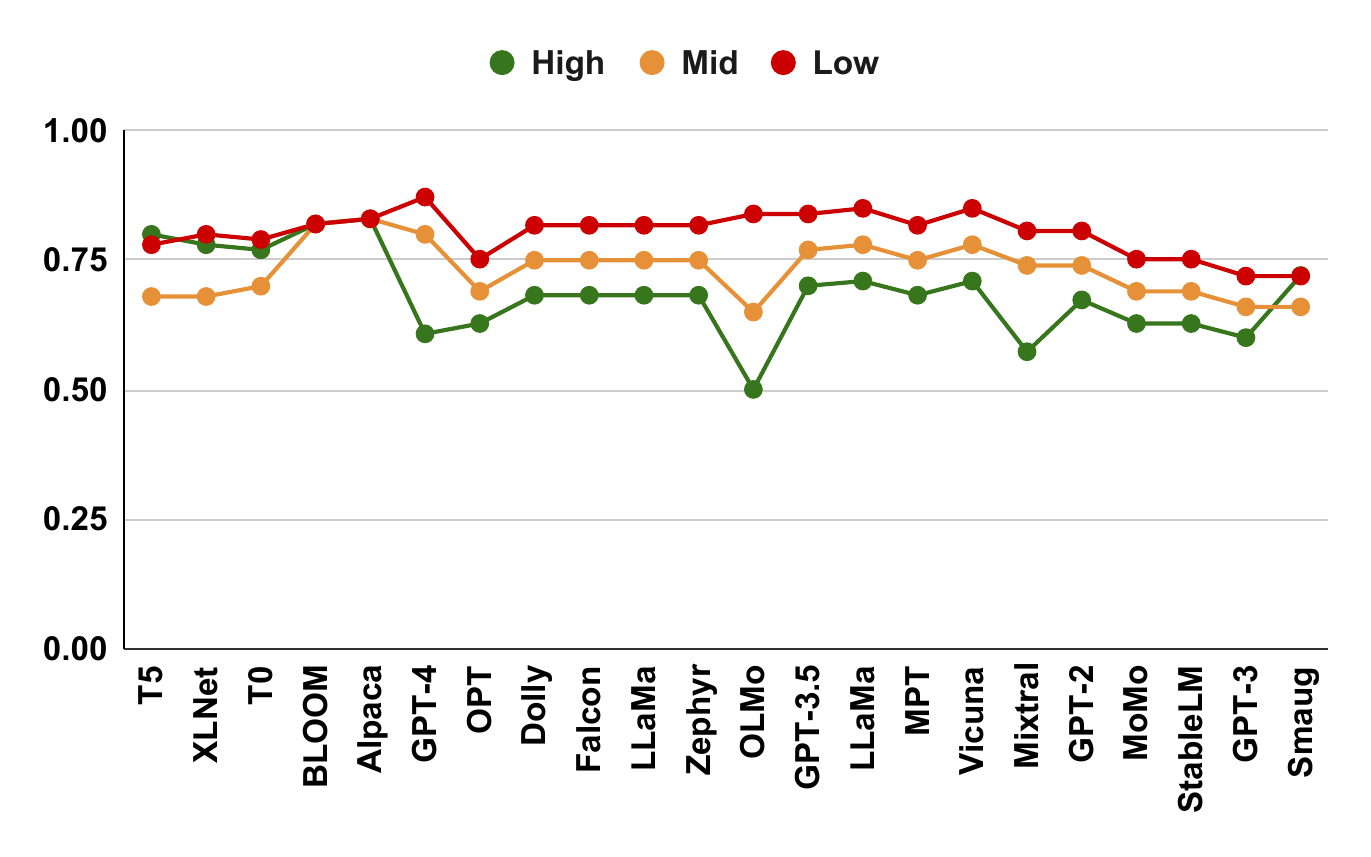}
            \caption[]%
            {{\small Time}}    
            \label{fig:mean and std of net44}
        \end{subfigure}
        \hfill
        \begin{subfigure}[b]{\textwidth}   
            \centering 
            \input{read}
        \end{subfigure}
        \caption[]%
            {{\small Readability}}   
        \label{fig:Readability}
\end{figure*}

\begin{figure*}[!ht]
        \centering
        \begin{subfigure}[b]{0.48\textwidth}
            \centering
            \includegraphics[width=\textwidth]{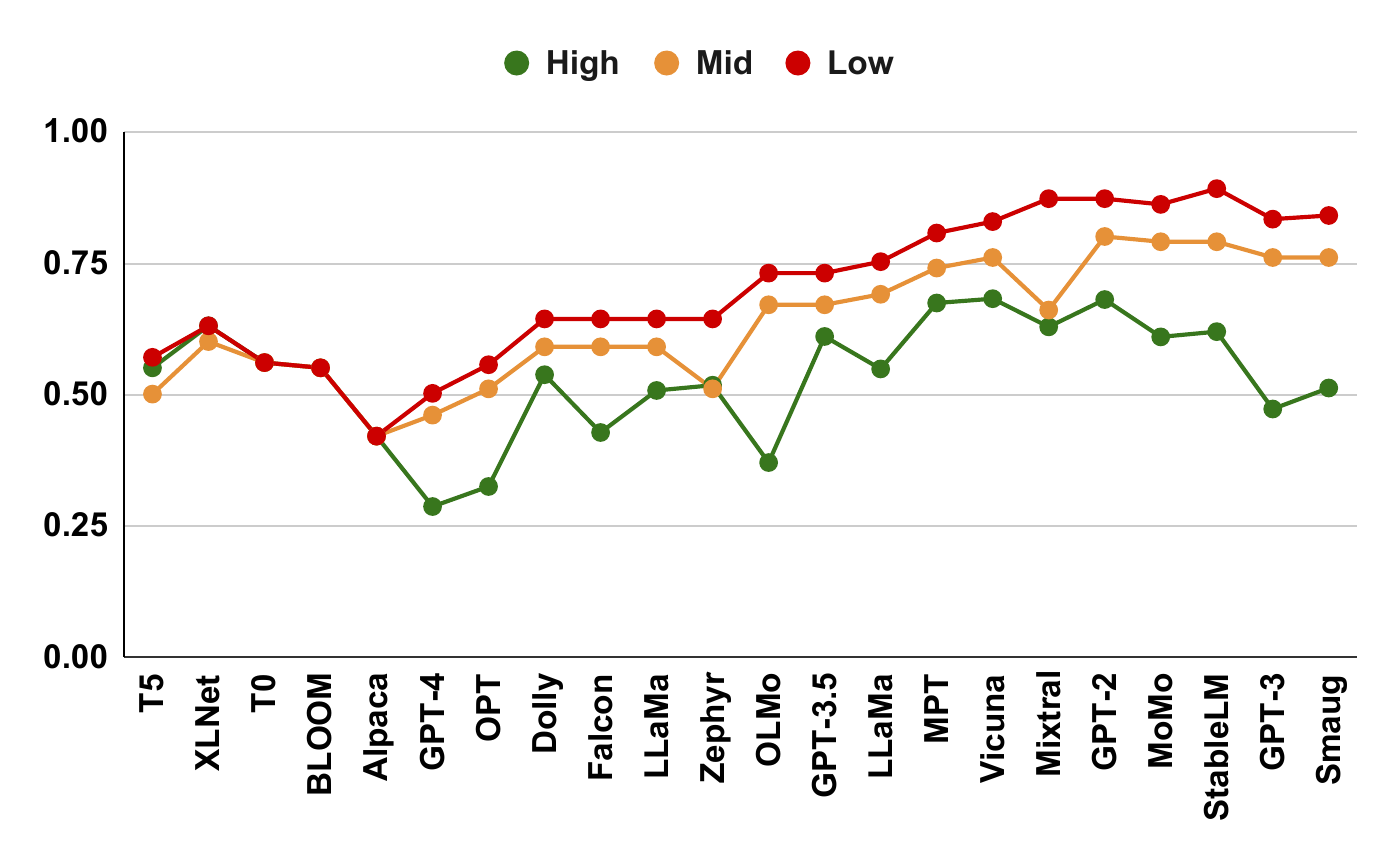}
            \caption[]%
            {{\small Person}}    
            \label{fig:mean and std of net14}
        \end{subfigure}
        \hfill
        \begin{subfigure}[b]{0.48\textwidth}  
            \centering 
            \includegraphics[width=\textwidth]{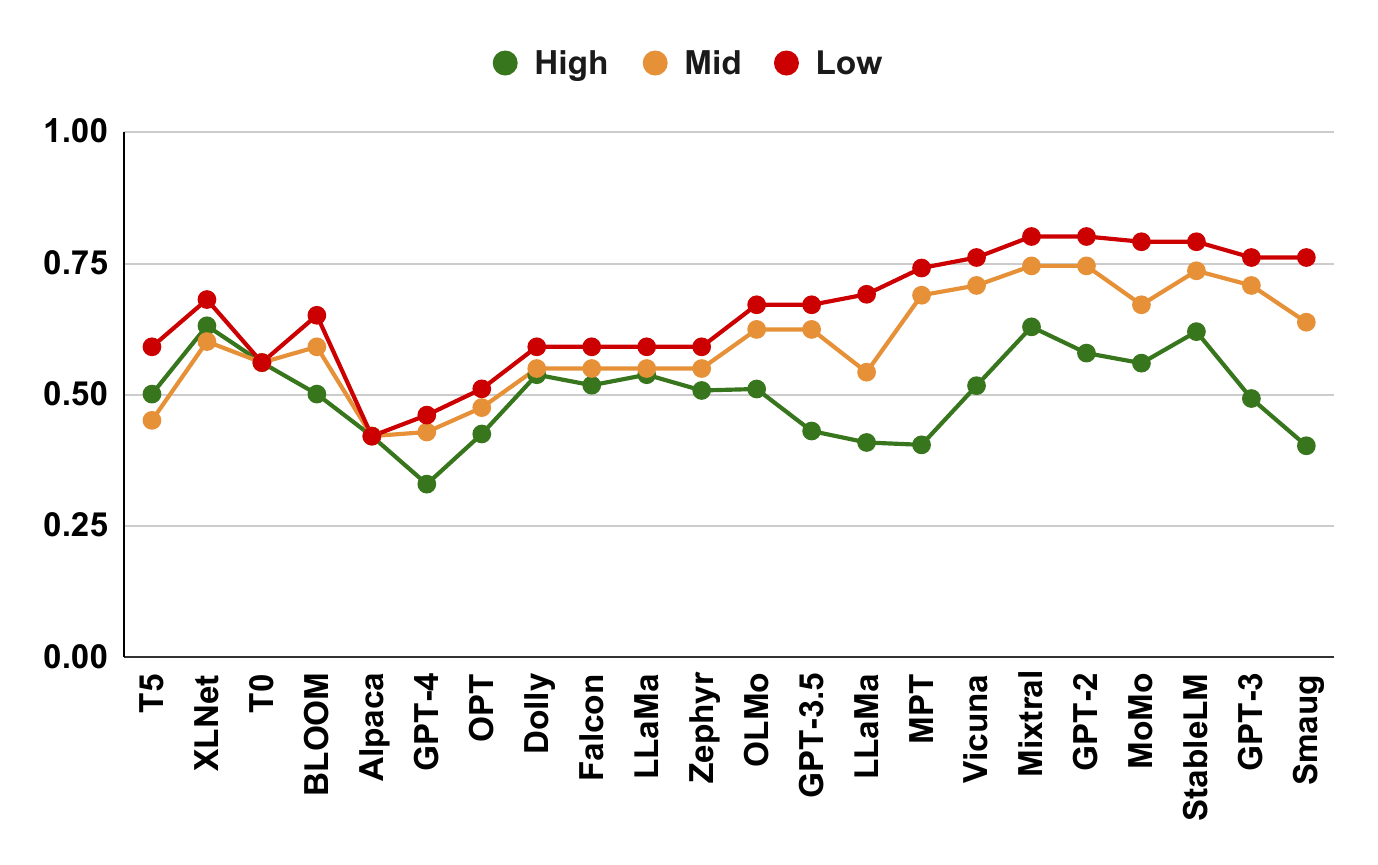}
            \caption[]%
            {{\small Location}}    
            \label{fig:mean and std of net24}
        \end{subfigure}
        \hfill
        \vskip\baselineskip
        \begin{subfigure}[b]{0.48\textwidth}   
            \centering 
            \includegraphics[width=\textwidth]{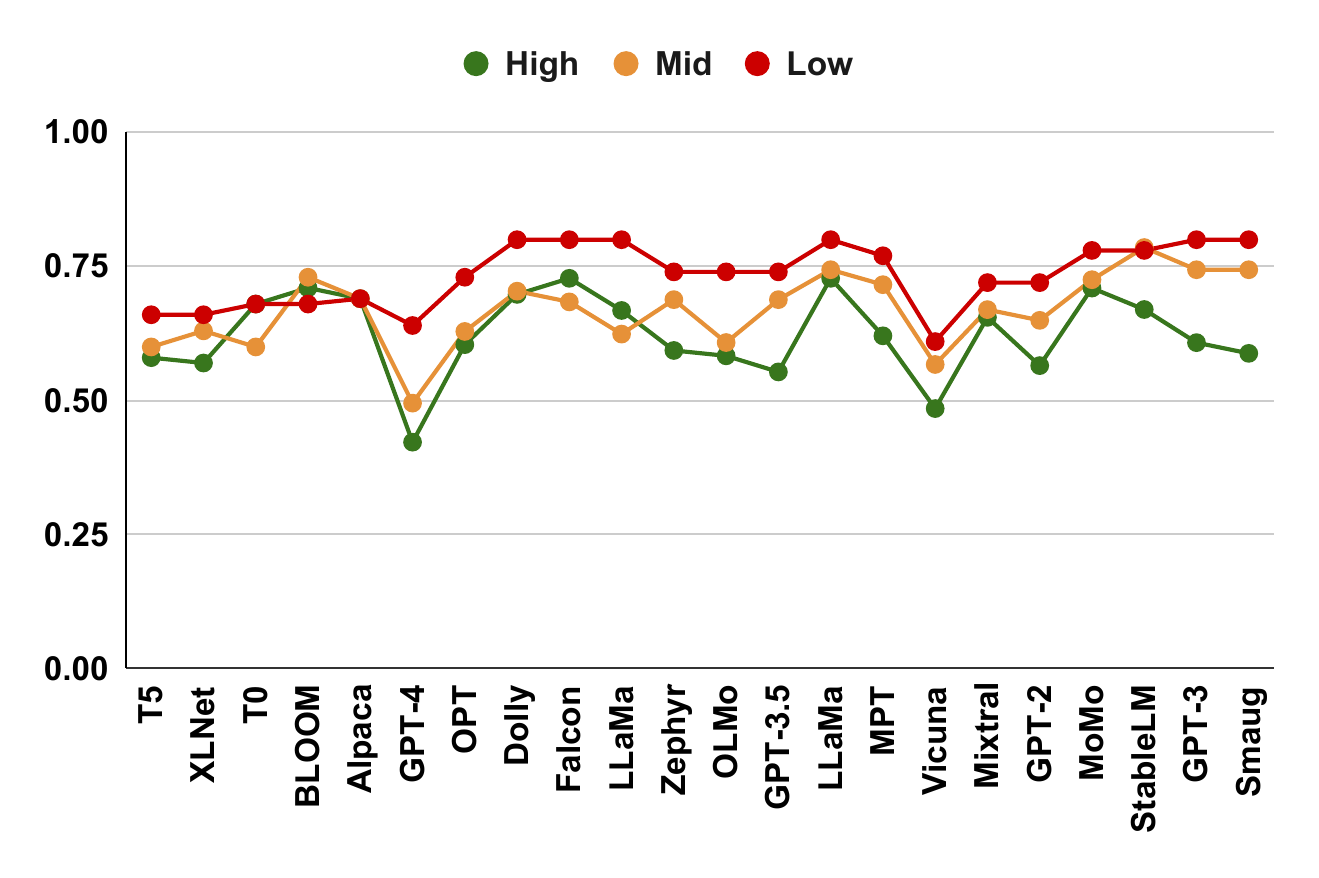}
            \caption[]%
            {{\small Number}}    
            \label{fig:mean and std of net34}
        \end{subfigure}
        \hfill
        \begin{subfigure}[b]{0.48\textwidth}   
            \centering 
            \includegraphics[width=\textwidth]{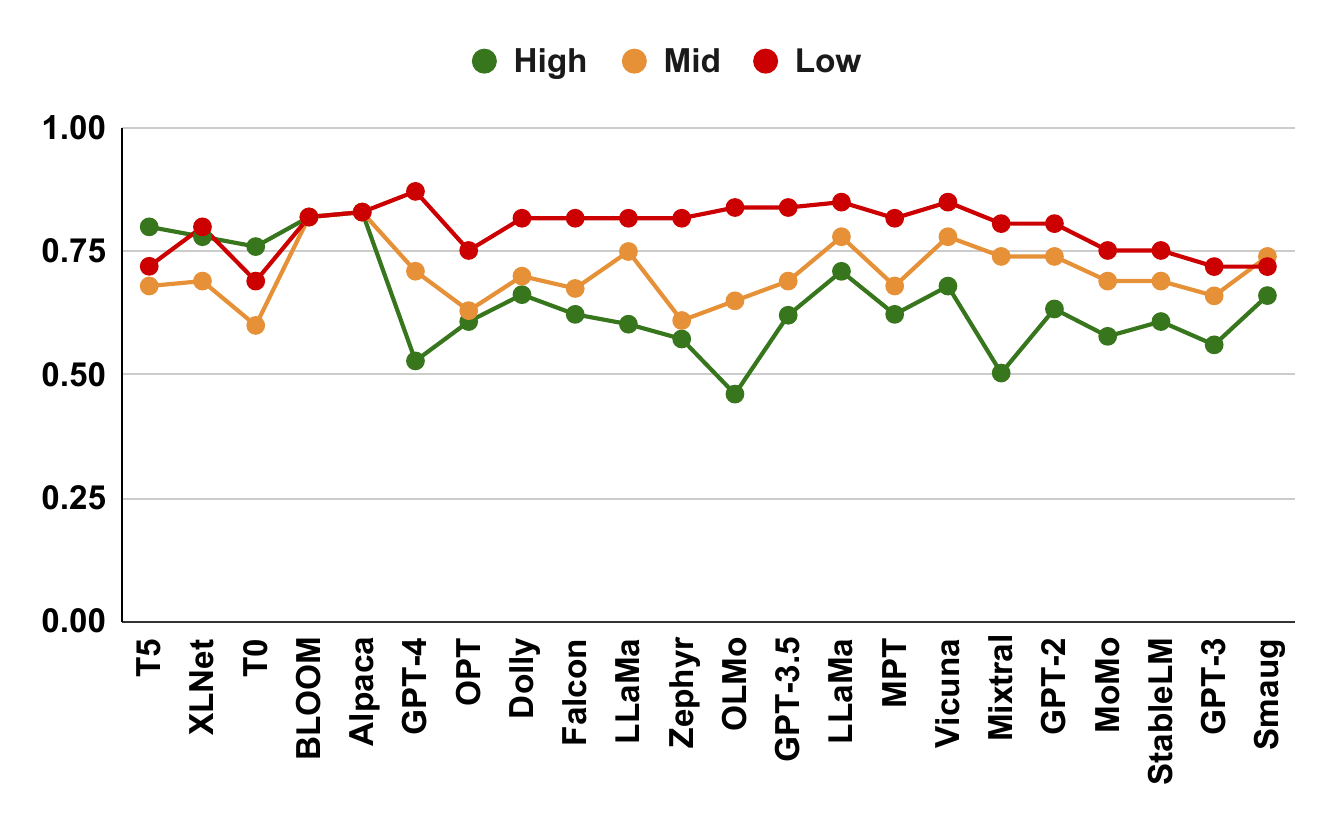}
            \caption[]%
            {{\small Time}}    
            \label{fig:mean and std of net44}
        \end{subfigure}
        \hfill
        \begin{subfigure}[b]{\textwidth}   
            \centering 
            \input{formality}
        \end{subfigure}
        \caption[]%
            {{\small Formality}}   
        \label{fig:Formality}
\end{figure*}

\newpage

\section{Dataset Annotation}  \label{appendix:dataset}

Crowdsourcing platforms are widely acknowledged for their efficiency and cost-effectiveness in annotation tasks. However, it is crucial to acknowledge that they may introduce inaccuracies or noise in annotations. To address this, we conducted an in-house annotation process involving 1,000 samples before employing crowdsourcing services. This internal process involved prompts and generated text snippets from five different LLMs, serving the dual purpose of formulating comprehensive annotation guidelines and creating a tailored annotation interface. The internal annotation aimed to ensure the quality and reliability of annotations before transitioning to crowdsourcing. We follow the similar annoatation guidelines as \cite{rawte-etal-2023-troubling} to generate the $\mathcal{SCA-}90\mathcal{K}$ dataset.

\section{Paraphrasing}\label{sec:appendix-Paraphrasing}
Paraphrasing entails the process of rephrasing or altering the wording of a text while preserving its initial meaning. This practice aims to present the content differently to improve clarity, prevent plagiarism, and tailor the language for a particular audience or purpose. Successful paraphrasing demands a thorough grasp of the source material, involving the reorganization of sentences, alteration of word selections, and retention of core ideas without replicating the exact wording from the original text. Following are the three characteristics of paraphrasing methods.

\textbf{Coverage:} Our goal is to create up to $5$ paraphrases for each claim. After generating the claims, we use the Minimum Edit Distance (MED) \cite{wagner1974string} measure (in words) for comparison. If the MED exceeds $\pm2$ for any paraphrase candidate (e.g., $c-p_1^c$) with the original claim, we include it; otherwise, we discard it. The evaluation is based on determining which model produces the highest number of meaningful paraphrases under this criterion.

\textbf{Correctness:} Following the initial filtration, we conducted pairwise entailment, retaining only paraphrase candidates endorsed as entailed by \cite{liu2019roberta} (Roberta Large), the state-of-the-art model trained on SNLI \cite{bowman2015large}.

\begin{figure}[!ht]
    \centering
    \includegraphics[width=0.5\textwidth, height=5.5cm]{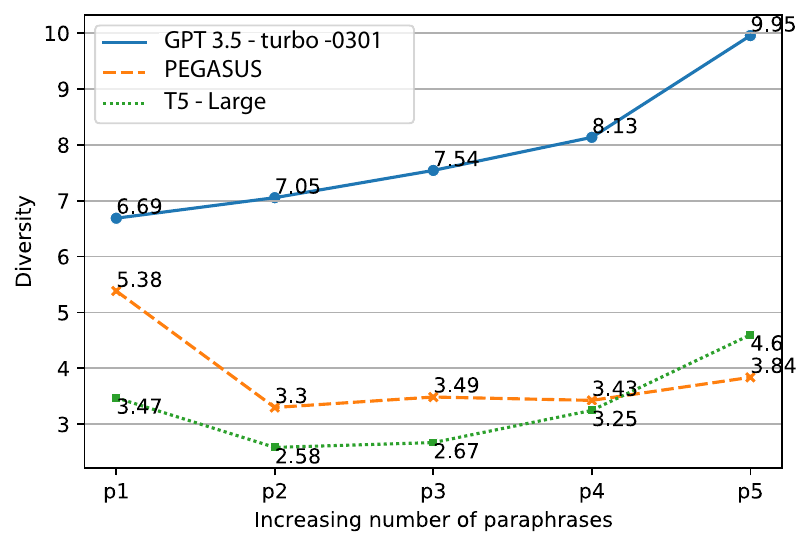}
    \caption{This figure shows the various parameters for generating paraphrases.}
    \label{fig:para-model}
\end{figure}

\textbf{Diversity:} Our focus was on selecting a model capable of producing linguistically diverse paraphrases. To assess this, we examined dissimilarities among generated paraphrase claims. For instance, we calculated dissimilarity scores for pairs like $c-p_n^c$, $p_1^c-p_n^c$, $p_2^c-p_n^c$, and so on, using the inverse of the BLEU score \cite{papineni2002bleu}. This process was repeated for all paraphrases, and the average dissimilarity score was computed. Our experiments revealed that \texttt{gpt-3.5-turbo-0301} performed the best in terms of linguistic diversity, as shown in the table. Furthermore, \texttt{gpt-3.5-turbo-0301} excelled in maximizing linguistic variations, as indicated in the diversity vs. models plot in \cref{fig:para-model}.

\section{Selecting the optimal paraphrase} \label{app:opt-para}

\subsection{Cosine Similarity}
Cosine similarity is a metric used to measure the similarity between two vectors, often in the context of high-dimensional spaces. It calculates the cosine of the angle between the two vectors, providing a numerical value that indicates how closely related the vectors are.

In the context of natural language processing, cosine similarity is often employed to assess the similarity between two documents represented as vectors in a high-dimensional space, where each dimension corresponds to a term or word. The cosine similarity ranges from -1 (completely dissimilar) to 1 (completely similar), with 0 indicating orthogonality (no similarity).

The formula for cosine similarity between two vectors A and B is given in \cref{eq:cos}.

\begin{equation}
\text{Cosine Similarity}(\mathbf{A}, \mathbf{B}) = \frac{\mathbf{A} \cdot \mathbf{B}}{\|\mathbf{A}\| \|\mathbf{B}\|}
\label{eq:cos}
\end{equation}

\subsection{Topic Modeling}
Topic modeling is a statistical technique that aims to identify topics present in a collection of text documents. The goal is to uncover the hidden thematic structure within the text data. One common algorithm used for topic modeling is Latent Dirichlet Allocation (LDA).

In the process of topic modeling, each document in the corpus is considered as a mixture of various topics, and each topic is represented as a distribution of words. The algorithm analyzes the co-occurrence patterns of words across documents to identify these latent topics. It helps in understanding the main themes or subjects present in a large collection of textual data without the need for manual annotation.

Topic modeling has applications in various NLP tasks, including document categorization, information retrieval, and content recommendation. It enables researchers and practitioners to gain insights into the underlying themes and structures within large textual datasets, making it a valuable tool for text analysis and understanding.

\subsubsection{Topic Similarity}

To overcome the issue of lengthy prompts, \cite{goyal2023think} introduce, the idea of inserting \texttt{[PAUSE]} tokens. However, it is not clear where these tokens can be added. Since they follow a rather random approach, in this work, we use a more deterministic approach.

\newpage

\section{Experimental Details}  \label{sec:exp-details}

For different fine-tuning techniques, the list of hypermaters is provided in \cref{tab:hyperp}.

\begin{table}[!h]
\small
\centering
\begin{tabular}{lc}
\toprule
\textbf{Parameter} & \textbf{Value}\\
\midrule
FC1 size  &  768 \\
FC2 size &  600 \\
Number of epochs & 5 \\
Learning rate  & 1E-03 \\
Optimizer  & AdamW  \\
Dropout probability &  0.1 \\
Batch size  &  1 \\
\bottomrule
\end{tabular}
\caption{\label{tab:hyperp}
Hyperparameters for different fine-tuning techniques.
}
\end{table}

\section{Factuality based entailment} \label{sec:ent-label}
In this approach, the prompt is submitted to the Google Search API to retrieve the top 20 relevant search results. From these 20 results, we assess a total of $n$ sentences for their pertinence to the prompt using a similarity metric. The top 20 sentences most akin to the prompt are chosen. For each of the $m$ sentences in the AI-generated text and the selected top 20 sentences, we utilize a textual entailment model to individually evaluate their credibility. Based on the entailment scores, we classify the AI-generated text into three categories: (i) \textit{support}, (ii) \textit{refute}, and (iii) \textit{not enough information}.

\section{Results after Adding \tframed[line width=0.5bp,fill=vred]{\textcolor{white}{\texttt{\textbf{[PAUSE]}}}} tokens} \label{appendix-pause}
In the \cref{tab:results} below, we show the experimental results for adding \tframed[line width=0.5bp,fill=vred]{\textcolor{white}{\texttt{\textbf{[PAUSE]}}}}.

\begin{table*}[!h]
\centering
\resizebox{\textwidth}{!}{%
\begin{tabular}{lrrrrrrrrrrrr}  \toprule
\multicolumn{1}{l|}{\multirow{2}{*}{\textbf{Fine-tuning technique}}} & \multicolumn{3}{l|}{\textbf{Person}} & \multicolumn{3}{l|}{\textbf{Location}} & \multicolumn{3}{l|}{\textbf{Numeric}} & \multicolumn{3}{l}{\textbf{Time}} \\
\multicolumn{1}{l|}{} & \multicolumn{1}{l|}{\textbf{Support}} & \multicolumn{1}{l|}{\textbf{Refute}} & \multicolumn{1}{l|}{\textbf{Neutral}} & \multicolumn{1}{l|}{\textbf{Support}} & \multicolumn{1}{l|}{\textbf{Refute}} & \multicolumn{1}{l|}{\textbf{Neutral}} & \multicolumn{1}{l|}{\textbf{Support}} & \multicolumn{1}{l|}{\textbf{Refute}} & \multicolumn{1}{l|}{\textbf{Neutral}} & \multicolumn{1}{l|}{\textbf{Support}} & \multicolumn{1}{l|}{\textbf{Refute}} & \multicolumn{1}{l}{\textbf{Neutral}} \\ \midrule
\multicolumn{1}{l|}{\textbf{Original Prompt}} & \multicolumn{1}{r|}{0.63} & \multicolumn{1}{r|}{0.54} & \multicolumn{1}{r|}{0.78} & \multicolumn{1}{r|}{0.52} & \multicolumn{1}{r|}{0.55} & \multicolumn{1}{r|}{0.77} & \multicolumn{1}{r|}{0.22} & \multicolumn{1}{r|}{0.89} & \multicolumn{1}{r|}{0.77} & \multicolumn{1}{r|}{0.29} & \multicolumn{1}{r|}{0.65} & 0.72 \\
\multicolumn{1}{l|}{\textbf{Optimal Paraphrase + LDA topics}} & \multicolumn{1}{r|}{0.65} & \multicolumn{1}{r|}{0.26} & \multicolumn{1}{r|}{0.59} & \multicolumn{1}{r|}{0.59} & \multicolumn{1}{r|}{0.28} & \multicolumn{1}{r|}{0.54} & \multicolumn{1}{r|}{0.36} & \multicolumn{1}{r|}{0.36} & \multicolumn{1}{r|}{0.66} & \multicolumn{1}{r|}{0.44} & \multicolumn{1}{r|}{0.56} & 0.7 \\  \hdashline 
\multicolumn{13}{c}{\textbf{\tframed[line width=0.5bp,fill=vred]{\textcolor{white}{\texttt{\textbf{[PAUSE]}}}} Injection}} \\ \hdashline 
\multicolumn{1}{l|}{\textbf{Optimal Paraphrase + LDA topics + w/ {[}PAUSE{]} token LoRA}} & \multicolumn{1}{r|}{0.7} & \multicolumn{1}{r|}{0.19} & \multicolumn{1}{r|}{0.69} & \multicolumn{1}{r|}{0.61} & \multicolumn{1}{r|}{0.25} & \multicolumn{1}{r|}{0.53} & \multicolumn{1}{r|}{0.53} & \multicolumn{1}{r|}{0.29} & \multicolumn{1}{r|}{0.69} & \multicolumn{1}{r|}{0.59} & \multicolumn{1}{r|}{0.29} & 0.72 \\
\multicolumn{1}{l|}{\textbf{Optimal Paraphrase + LDA topics + w/ {[}PAUSE{]} token QALoRA}} & \multicolumn{1}{r|}{0.72} & \multicolumn{1}{r|}{0.21} & \multicolumn{1}{r|}{0.67} & \multicolumn{1}{r|}{0.62} & \multicolumn{1}{r|}{0.22} & \multicolumn{1}{r|}{0.52} & \multicolumn{1}{r|}{0.58} & \multicolumn{1}{r|}{0.32} & \multicolumn{1}{r|}{0.67} & \multicolumn{1}{r|}{0.62} & \multicolumn{1}{r|}{0.31} & 0.73 \\
\multicolumn{1}{l|}{\textbf{Optimal Paraphrase + LDA topics + w/ {[}PAUSE{]} token Reverse Proxy Tuning}} & \multicolumn{1}{r|}{0.86} & \multicolumn{1}{r|}{0.12} & \multicolumn{1}{r|}{0.79} & \multicolumn{1}{r|}{0.77} & \multicolumn{1}{r|}{0.18} & \multicolumn{1}{r|}{0.48} & \multicolumn{1}{r|}{0.69} & \multicolumn{1}{r|}{0.26} & \multicolumn{1}{r|}{0.79} & \multicolumn{1}{r|}{0.68} & \multicolumn{1}{r|}{0.23} & 0.66  \\ \bottomrule
\end{tabular}%
}
\caption{Empirical results for Reverse Proxy Tuning with  \tframed[line width=0.5bp,fill=vred]{\textcolor{white}{\texttt{\textbf{[PAUSE]}}}} tokens.}
\label{tab:results}
\end{table*}

\newpage
\section{Selecting the optimal paraphrased prompt}
\begin{figure*}[!ht]
\centering
\includegraphics[width=\textwidth]{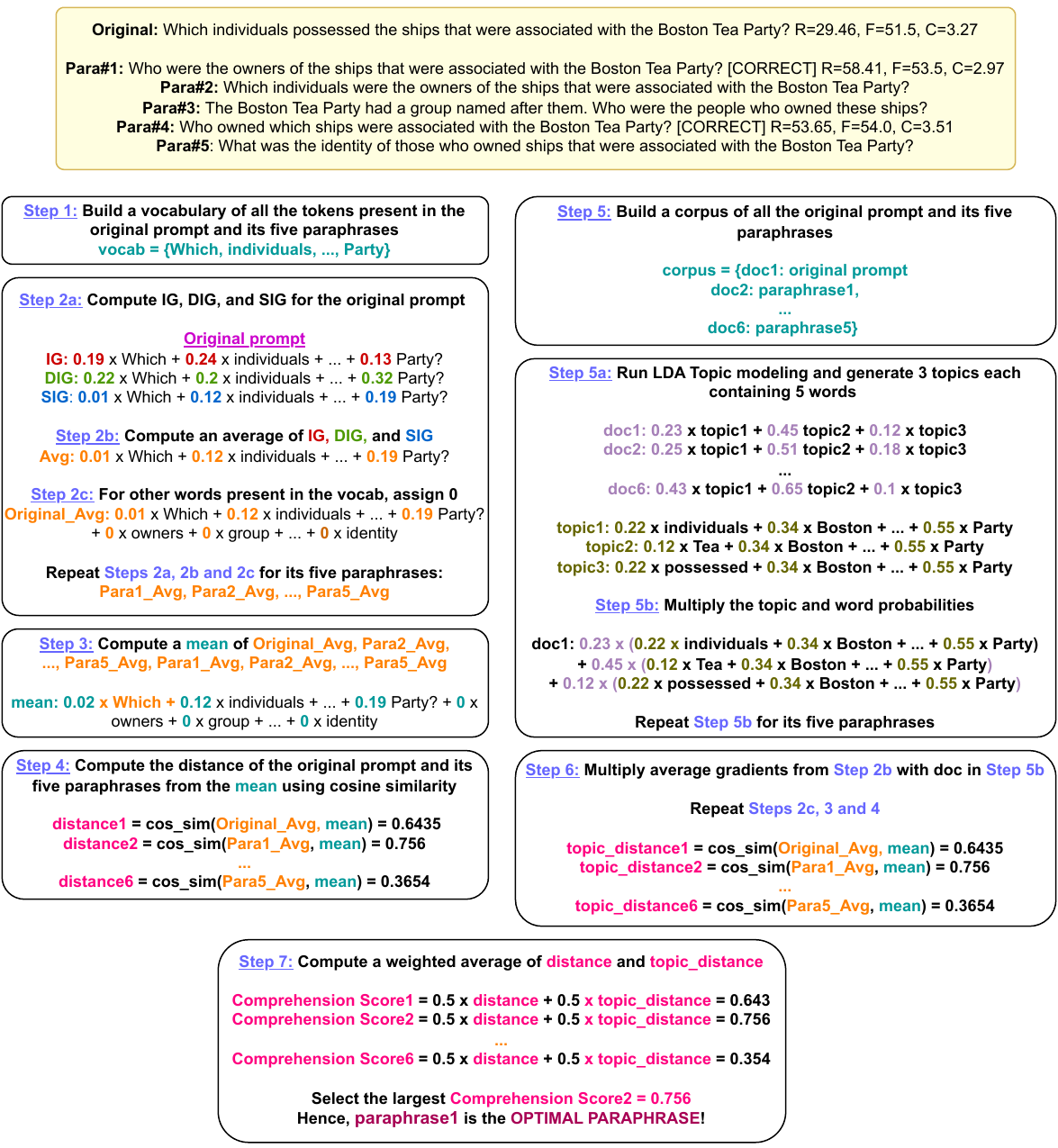}
\caption{A walkthrough of our optimal paraphrase selection process.}
\label{fig:para}
\end{figure*}

The detailed explanation of our algorithm to identify the optimal paraphrased prompt is provided in the illustration in \cref{fig:para}.

\section{Before and after adding \tframed[line width=0.5bp,fill=vred]{\textcolor{white}{\texttt{\textbf{[PAUSE]}}}} token}

\begin{figure*}[!ht]
        \centering
        \begin{subfigure}[b]{0.45\textwidth}
            \centering
            \includegraphics[width=\textwidth,height=3cm]{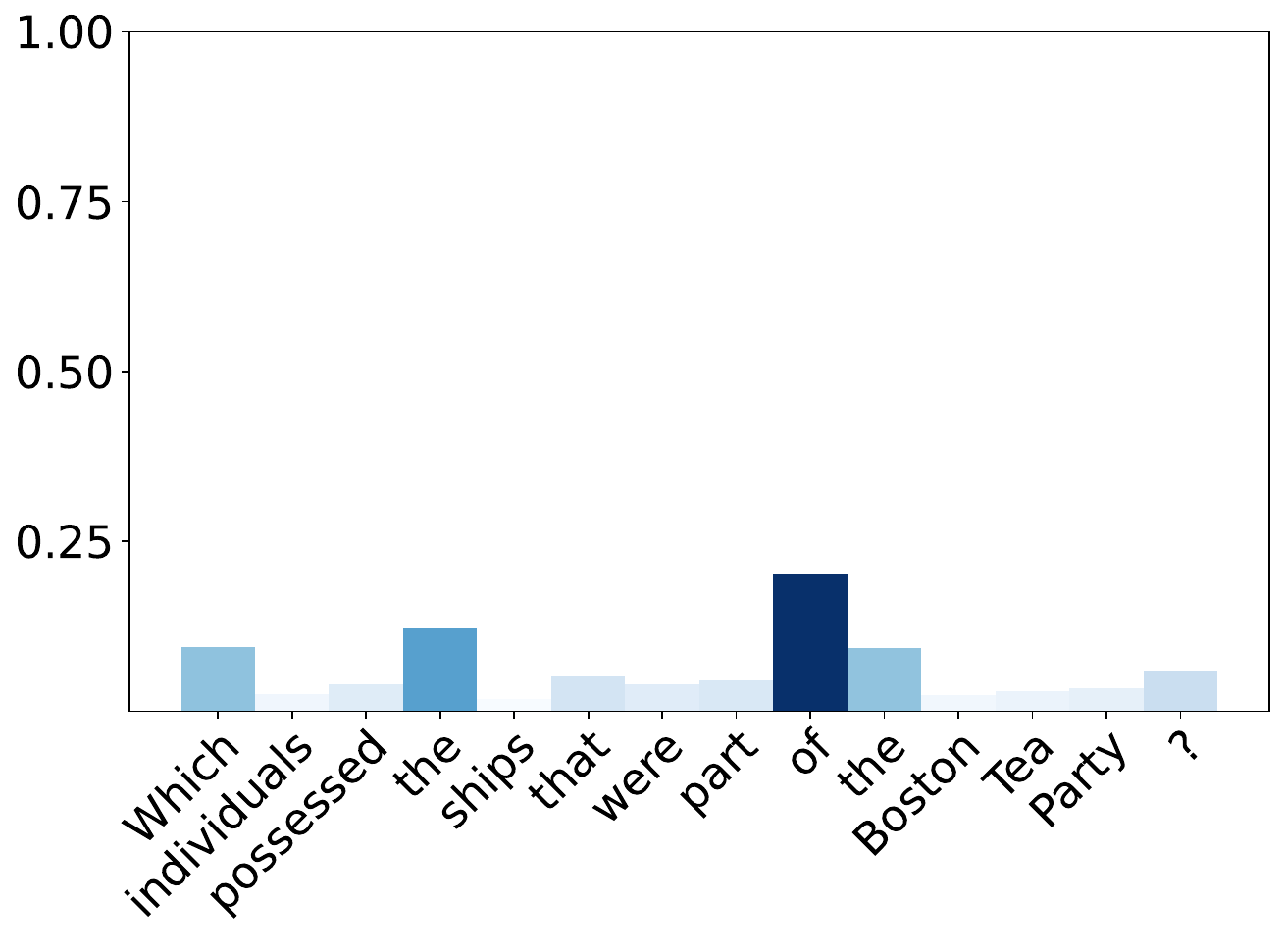}
            \caption[]%
            {{\small Before adding \tframed[line width=0.5bp,fill=vred]{\textcolor{white}{\texttt{\textbf{[PAUSE]}}}} tokens} to original prompt.}
            \label{fig:mean and std of net14}
        \end{subfigure}
        \hfill
        \begin{subfigure}[b]{0.45\textwidth}  
            \centering 
            \includegraphics[width=\textwidth,height=3cm]{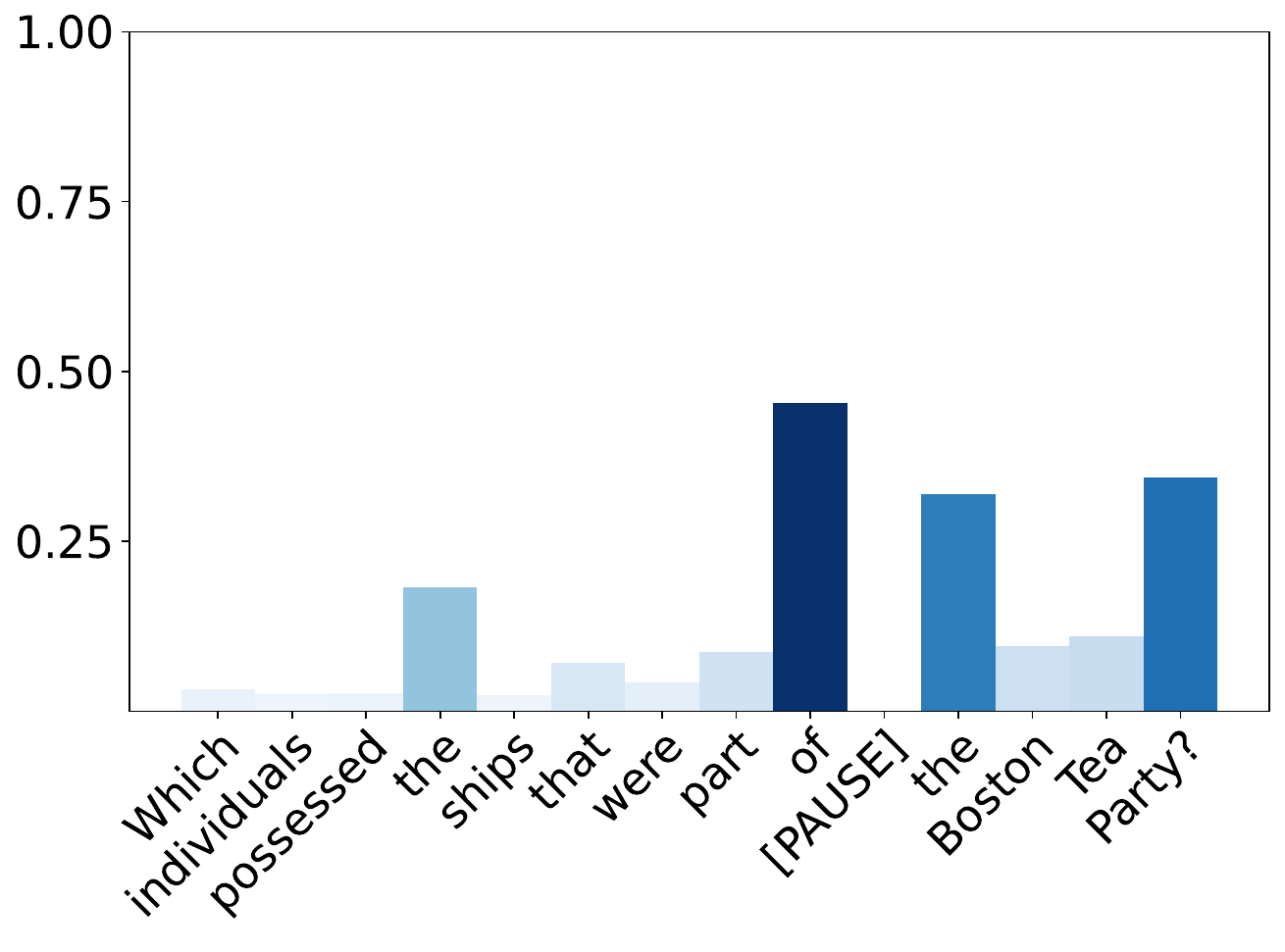}
            \caption[]%
            {{\small After adding \tframed[line width=0.5bp,fill=vred]{\textcolor{white}{\texttt{\textbf{[PAUSE]}}}} tokens} to original prompt.}    
            \label{fig:mean and std of net24}
        \end{subfigure}
        \hfill
        \vskip\baselineskip
        \begin{subfigure}[b]{0.45\textwidth}   
            \centering 
            \includegraphics[width=\textwidth,height=3cm]{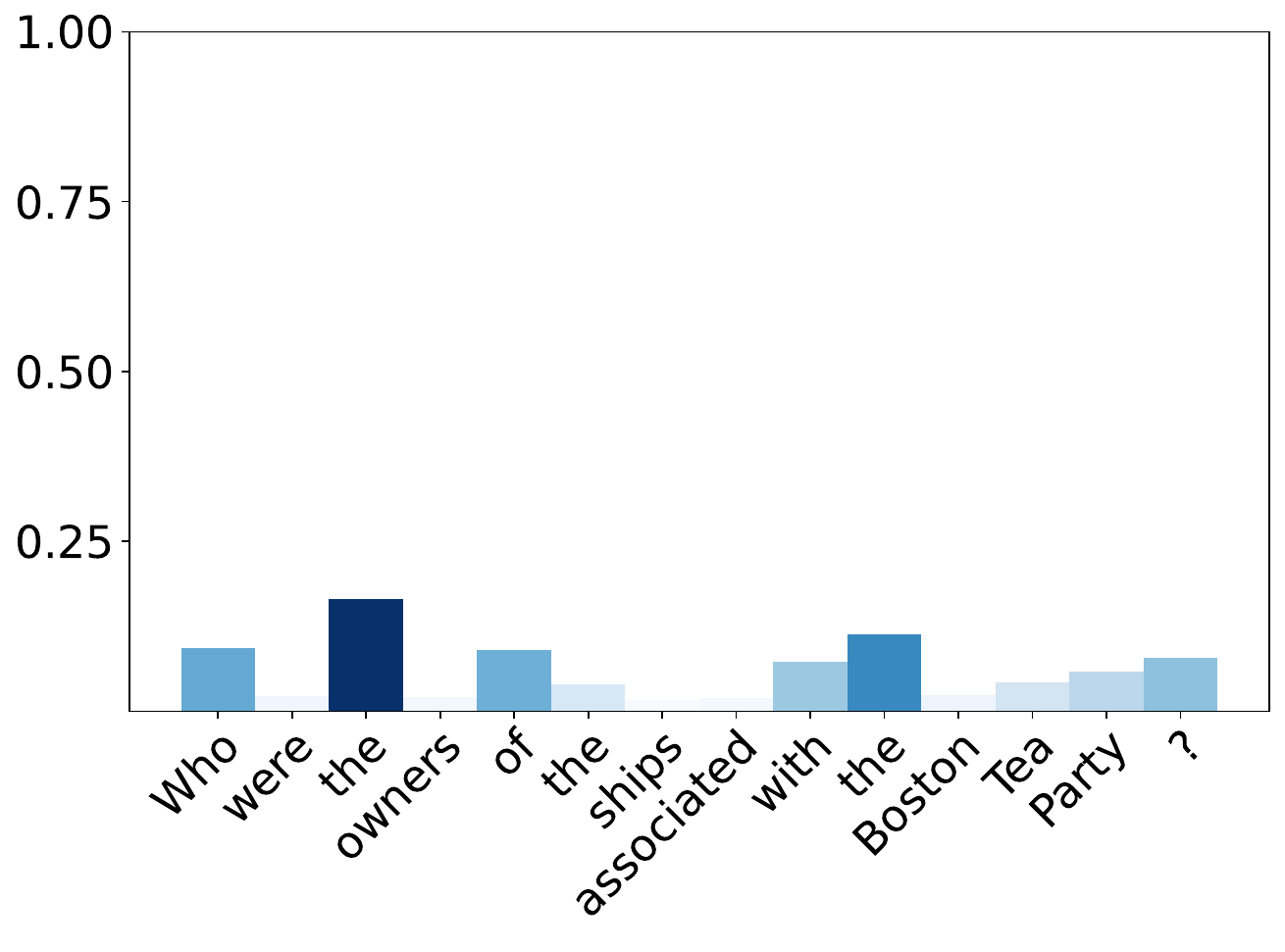}
            \caption[]%
            {{\small Before adding \tframed[line width=0.5bp,fill=vred]{\textcolor{white}{\texttt{\textbf{[PAUSE]}}}} tokens} to paraphrase 1.}    
            \label{fig:mean and std of net34}
        \end{subfigure}
        \hfill
        \begin{subfigure}[b]{0.45\textwidth}   
            \centering 
            \includegraphics[width=\textwidth,height=3cm]{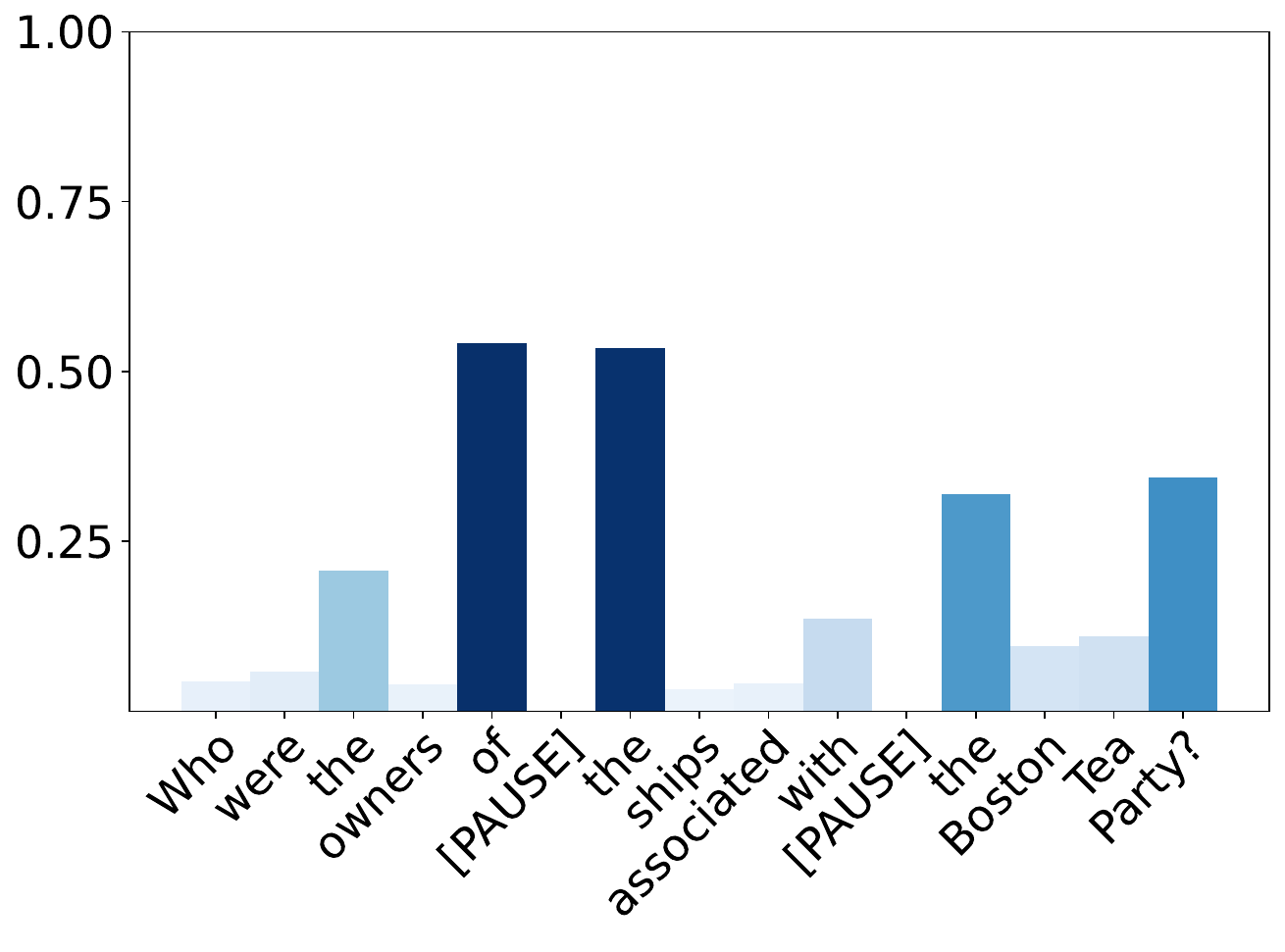}
            \caption[]%
            {{\small After adding \tframed[line width=0.5bp,fill=vred]{\textcolor{white}{\texttt{\textbf{[PAUSE]}}}} tokens} to paraphrase 1.}    
            \label{fig:mean and std of net44}
        \end{subfigure}
        \hfill
        \vskip\baselineskip
        \begin{subfigure}[b]{0.45\textwidth}   
            \centering 
            \includegraphics[width=\textwidth,height=3cm]{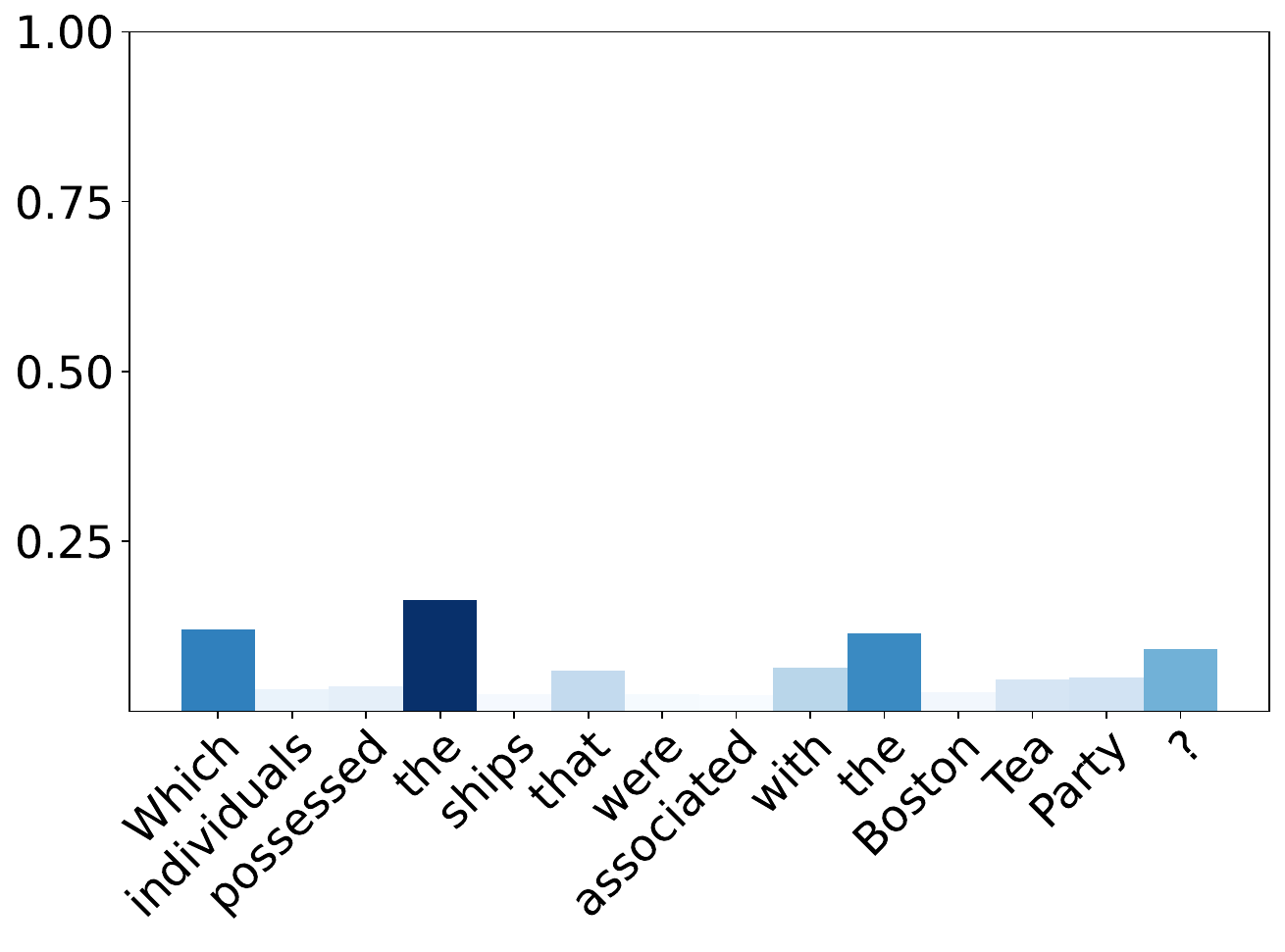}
            \caption[]%
            {{\small Before adding \tframed[line width=0.5bp,fill=vred]{\textcolor{white}{\texttt{\textbf{[PAUSE]}}}} tokens} to paraphrase 2.}
            \label{fig:mean and std of net34}
        \end{subfigure}
        \hfill
        \begin{subfigure}[b]{0.45\textwidth}   
            \centering 
            \includegraphics[width=\textwidth,height=3cm]{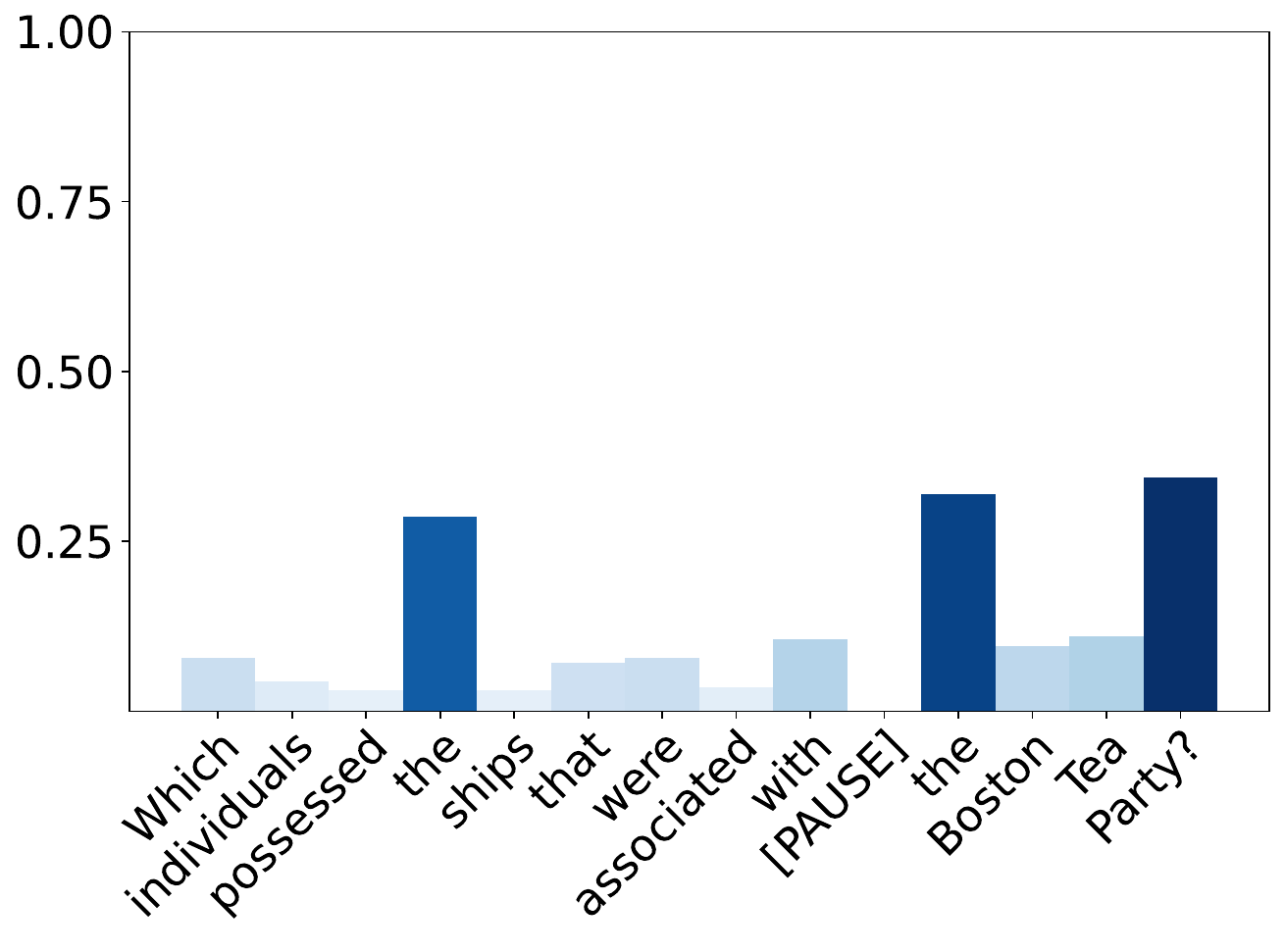}
            \caption[]%
            {{\small After adding \tframed[line width=0.5bp,fill=vred]{\textcolor{white}{\texttt{\textbf{[PAUSE]}}}} tokens} to paraphrase 2.} 
            \label{fig:mean and std of net44}
        \end{subfigure}
        \hfill
        \vskip\baselineskip
        \begin{subfigure}[b]{0.45\textwidth}   
            \centering 
            \includegraphics[width=\textwidth,height=3cm]{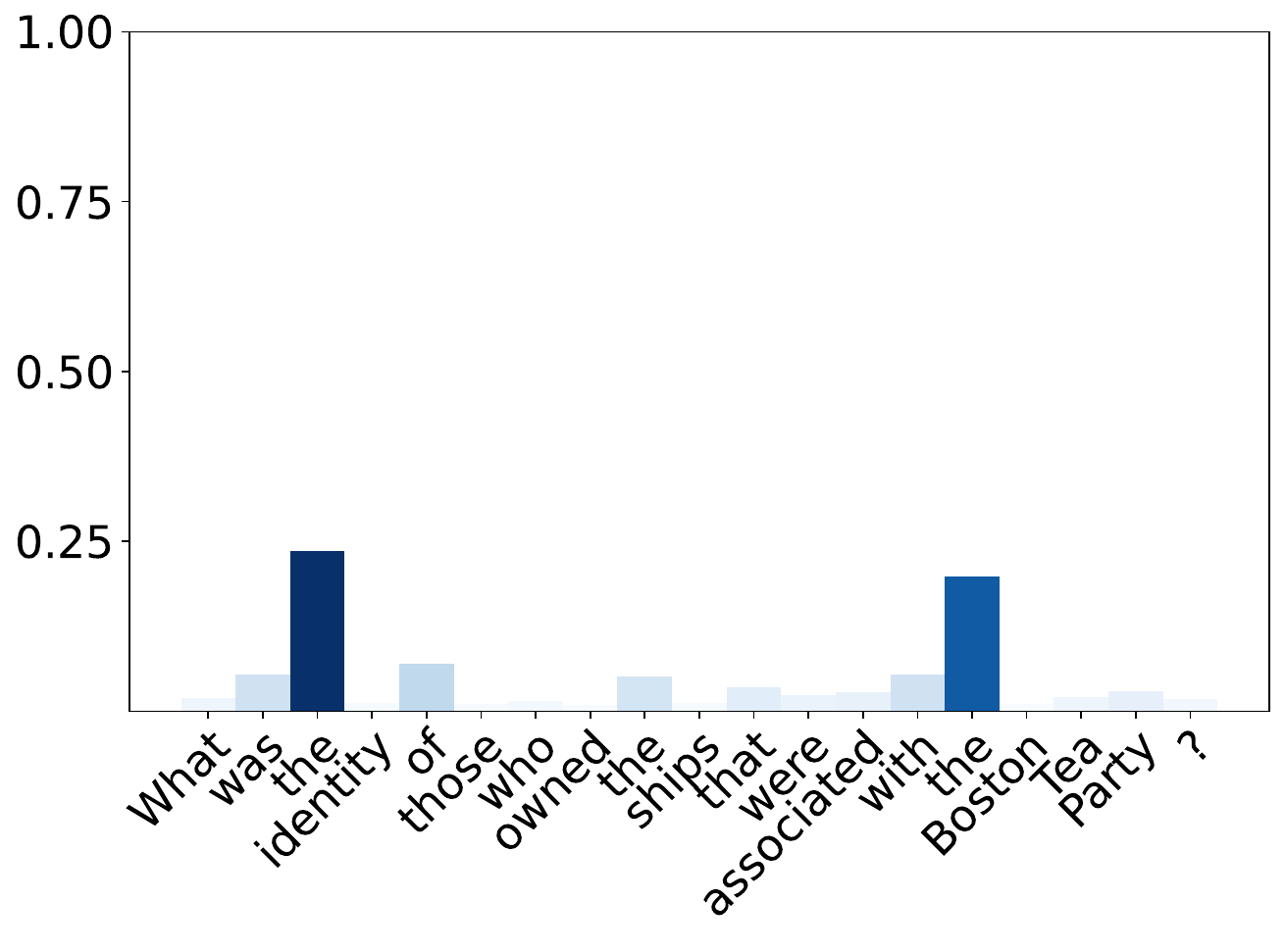}
            \caption[]%
            {{\small Before adding \tframed[line width=0.5bp,fill=vred]{\textcolor{white}{\texttt{\textbf{[PAUSE]}}}} tokens} to paraphrase 3.}
            \label{fig:mean and std of net44}
        \end{subfigure}
        \hfill
        \begin{subfigure}[b]{0.45\textwidth}   
            \centering 
            \includegraphics[width=\textwidth,height=3cm]{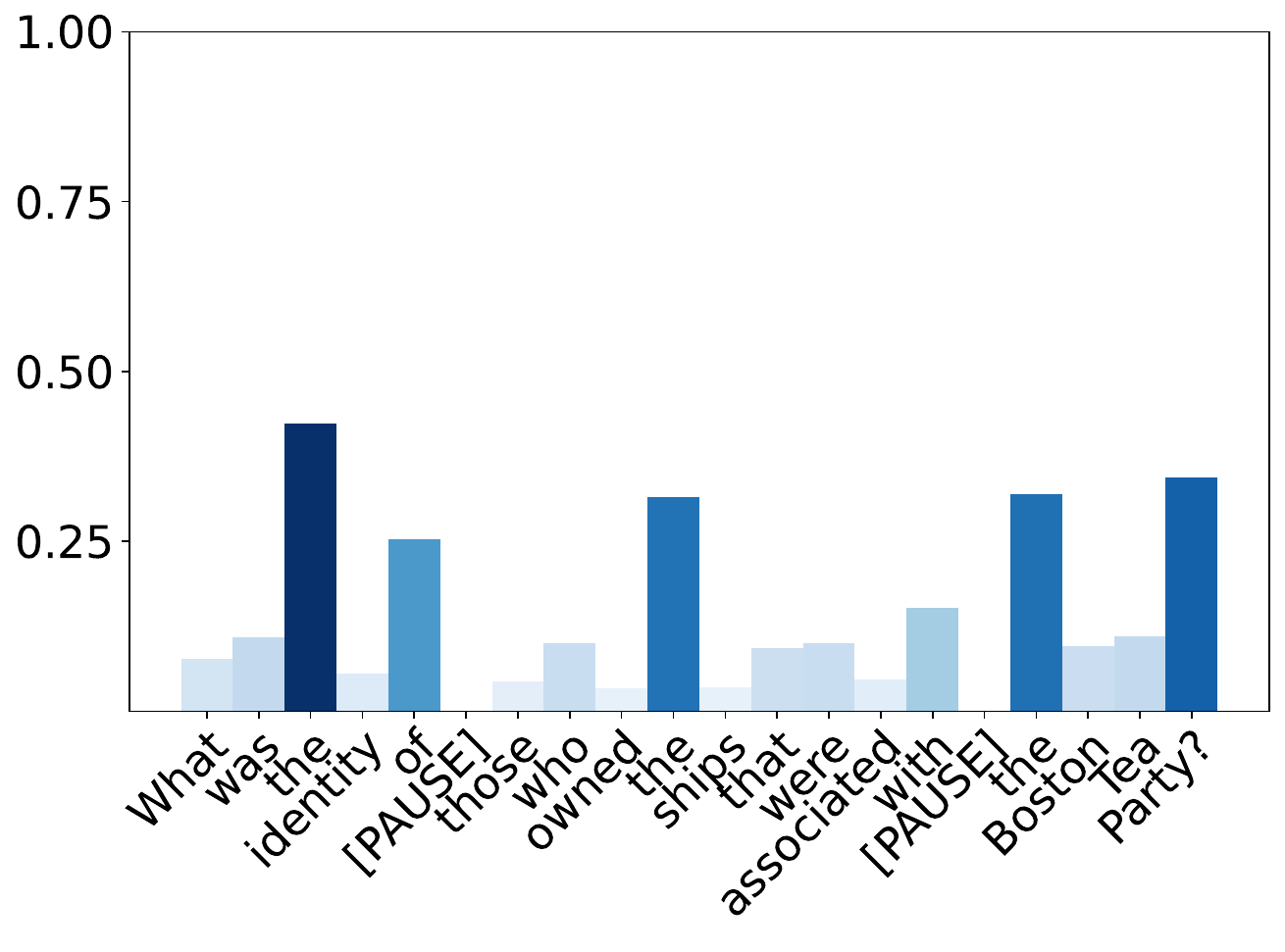}
            \caption[]%
            {{\small After adding \tframed[line width=0.5bp,fill=vred]{\textcolor{white}{\texttt{\textbf{[PAUSE]}}}} tokens} to paraphrase 3.}    
            \label{fig:mean and std of net44}
        \end{subfigure}
        \hfill
        \vskip\baselineskip
        \begin{subfigure}[b]{0.45\textwidth}   
            \centering 
            \includegraphics[width=\textwidth,height=3cm]{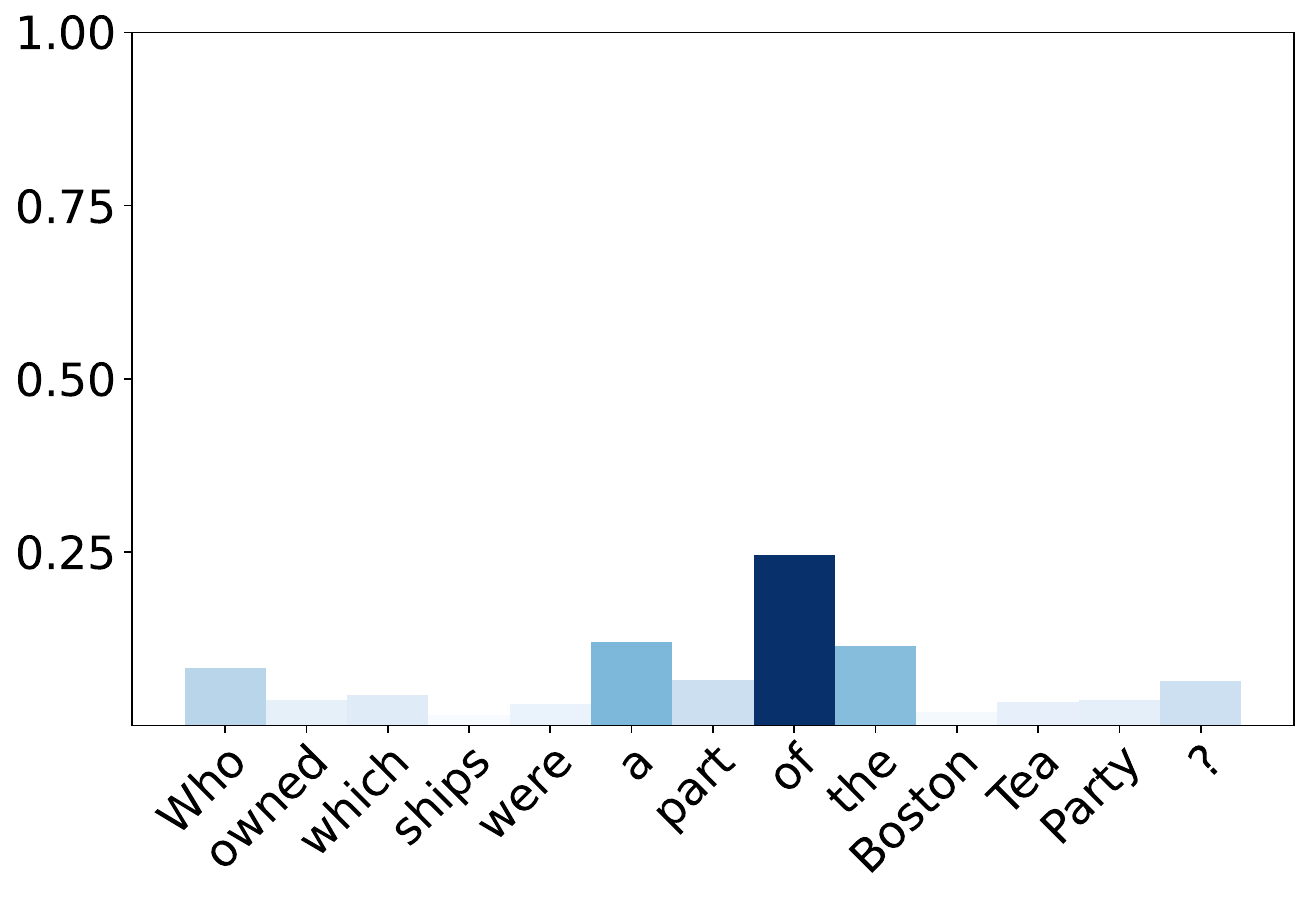}
            \caption[]%
            {{\small Before adding \tframed[line width=0.5bp,fill=vred]{\textcolor{white}{\texttt{\textbf{[PAUSE]}}}} tokens} to paraphrase 4.}    
            \label{fig:mean and std of net44}
        \end{subfigure}
        \hfill
        \begin{subfigure}[b]{0.45\textwidth}   
            \centering 
            \includegraphics[width=\textwidth,height=3cm]{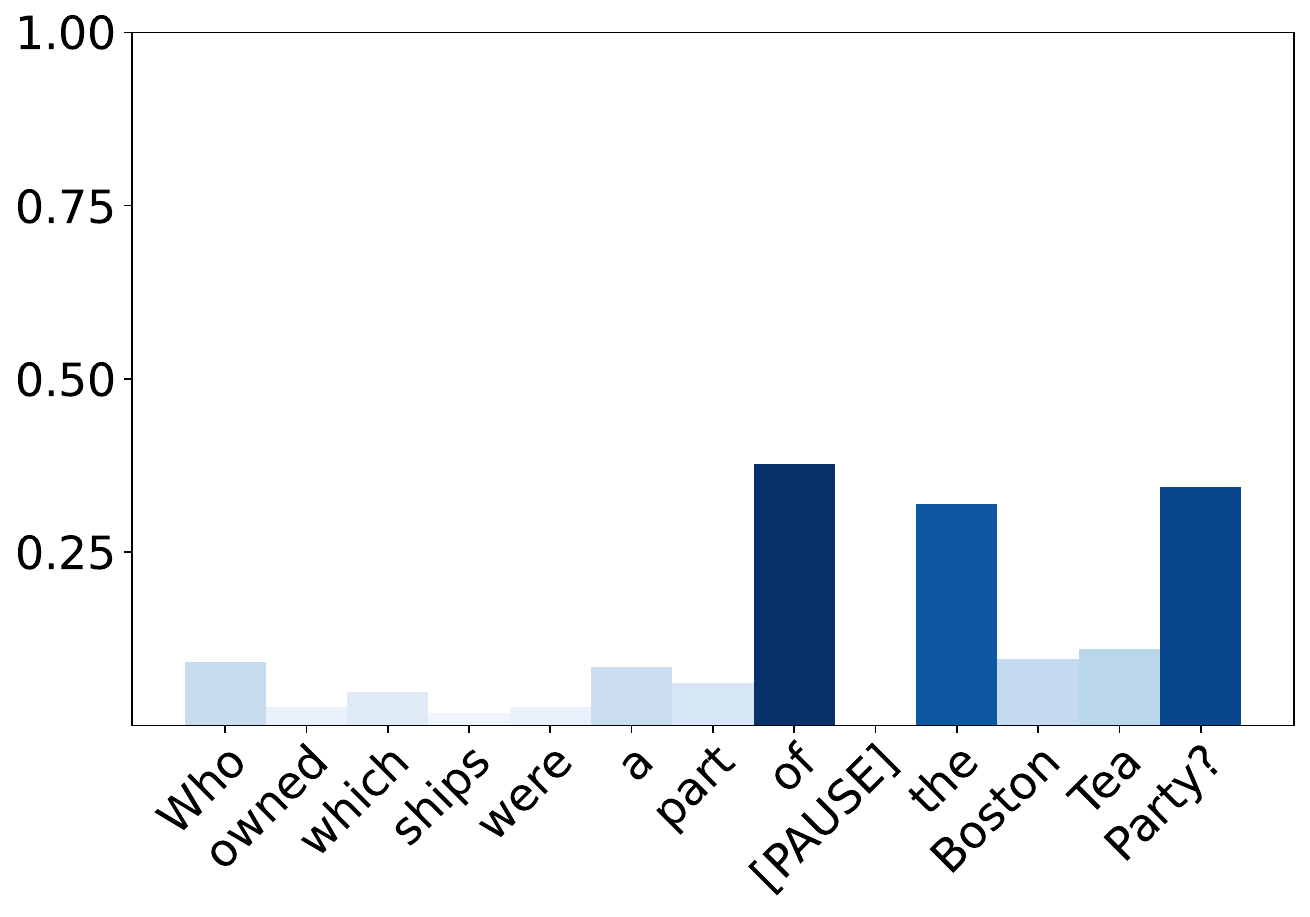}
            \caption[]%
            {{\small After adding \tframed[line width=0.5bp,fill=vred]{\textcolor{white}{\texttt{\textbf{[PAUSE]}}}} tokens} to paraphrase 4.}    
            \label{fig:mean and std of net44}
        \end{subfigure}
        \hfill
        \vskip\baselineskip
        \begin{subfigure}[b]{0.45\textwidth}   
            \centering 
            \includegraphics[width=\textwidth,height=3cm]{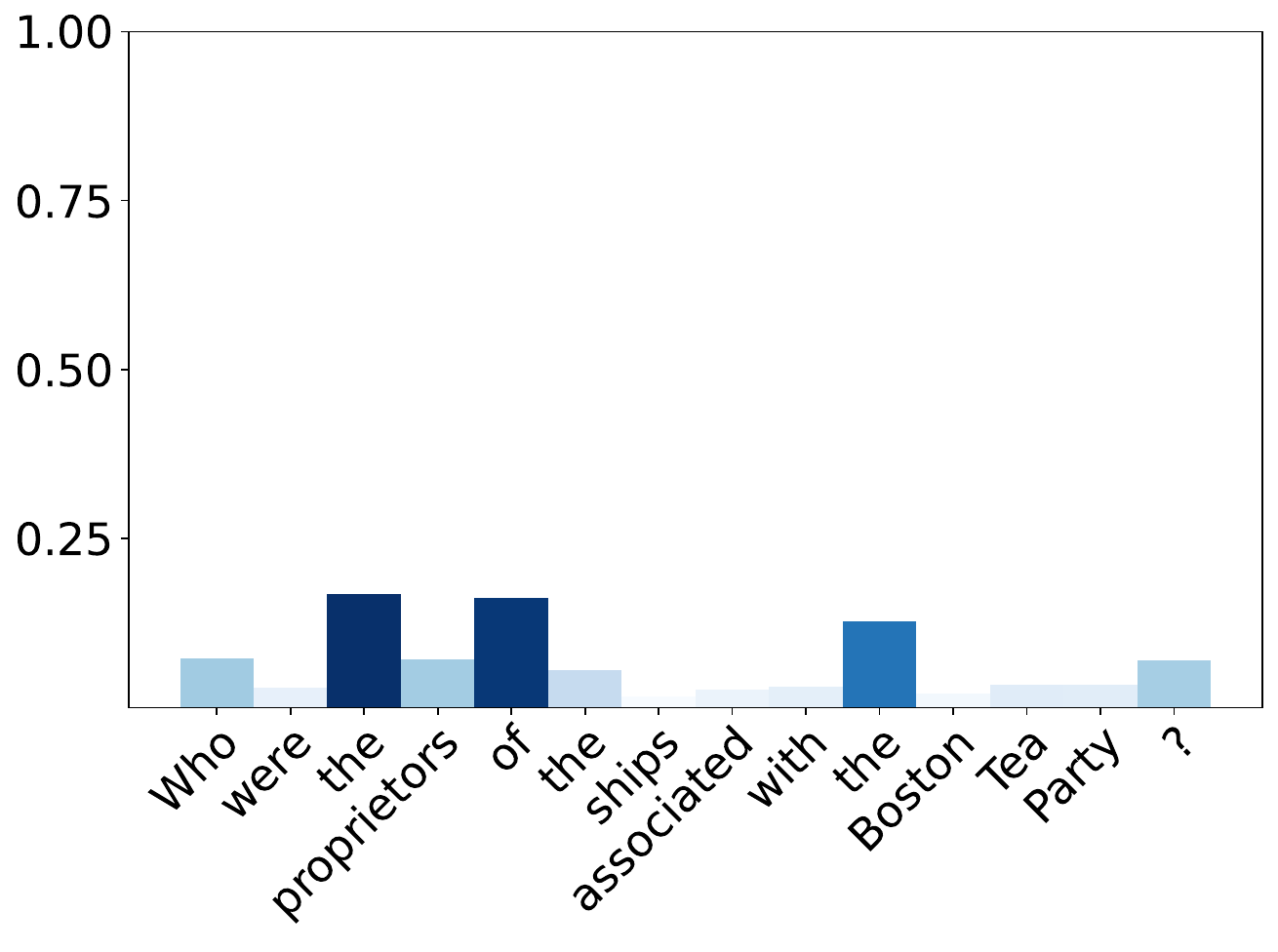}
            \caption[]%
            {{\small Before adding \tframed[line width=0.5bp,fill=vred]{\textcolor{white}{\texttt{\textbf{[PAUSE]}}}} tokens} to paraphrase 5.}    
            \label{fig:mean and std of net44}
        \end{subfigure}
        \hfill
        \begin{subfigure}[b]{0.45\textwidth}   
            \centering 
            \includegraphics[width=\textwidth,height=3cm]{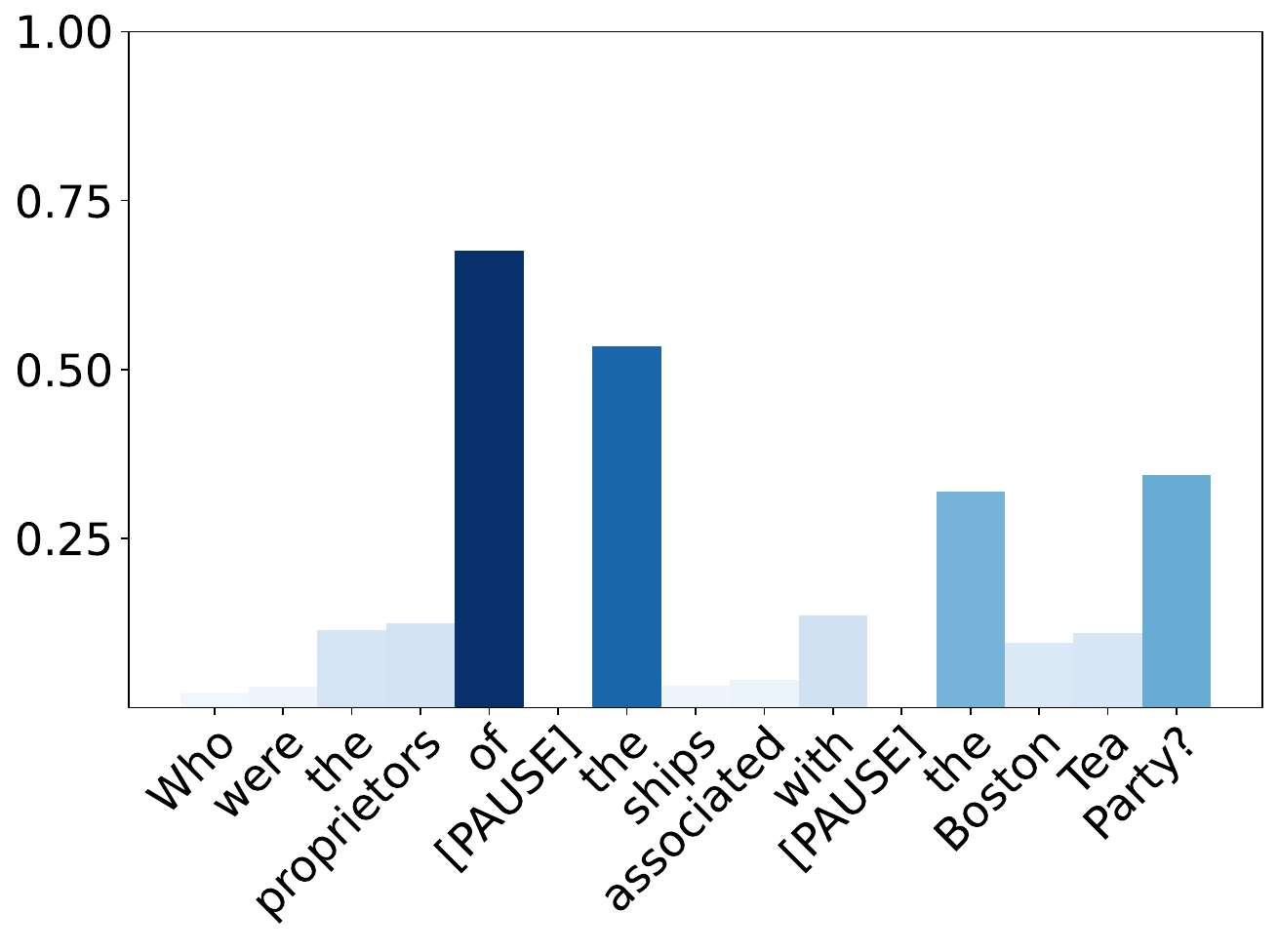}
            \caption[]%
            {{\small After adding \tframed[line width=0.5bp,fill=vred]{\textcolor{white}{\texttt{\textbf{[PAUSE]}}}} tokens} to paraphrase 5.}    
            \label{fig:mean and std of net44}
        \end{subfigure}
        \caption[]%
            {{\small The phrase \textbf{Boston Tea} gets more importance score after adding \tframed[line width=0.5bp,fill=vred]{\textcolor{white}{\texttt{\textbf{[PAUSE]}}}} token for alpaca.}}   
        \label{fig:alpaca}
\end{figure*}

\begin{figure*}[!ht]
        \centering
        \begin{subfigure}[b]{0.45\textwidth}
            \centering
            \includegraphics[width=\textwidth,height=3cm]{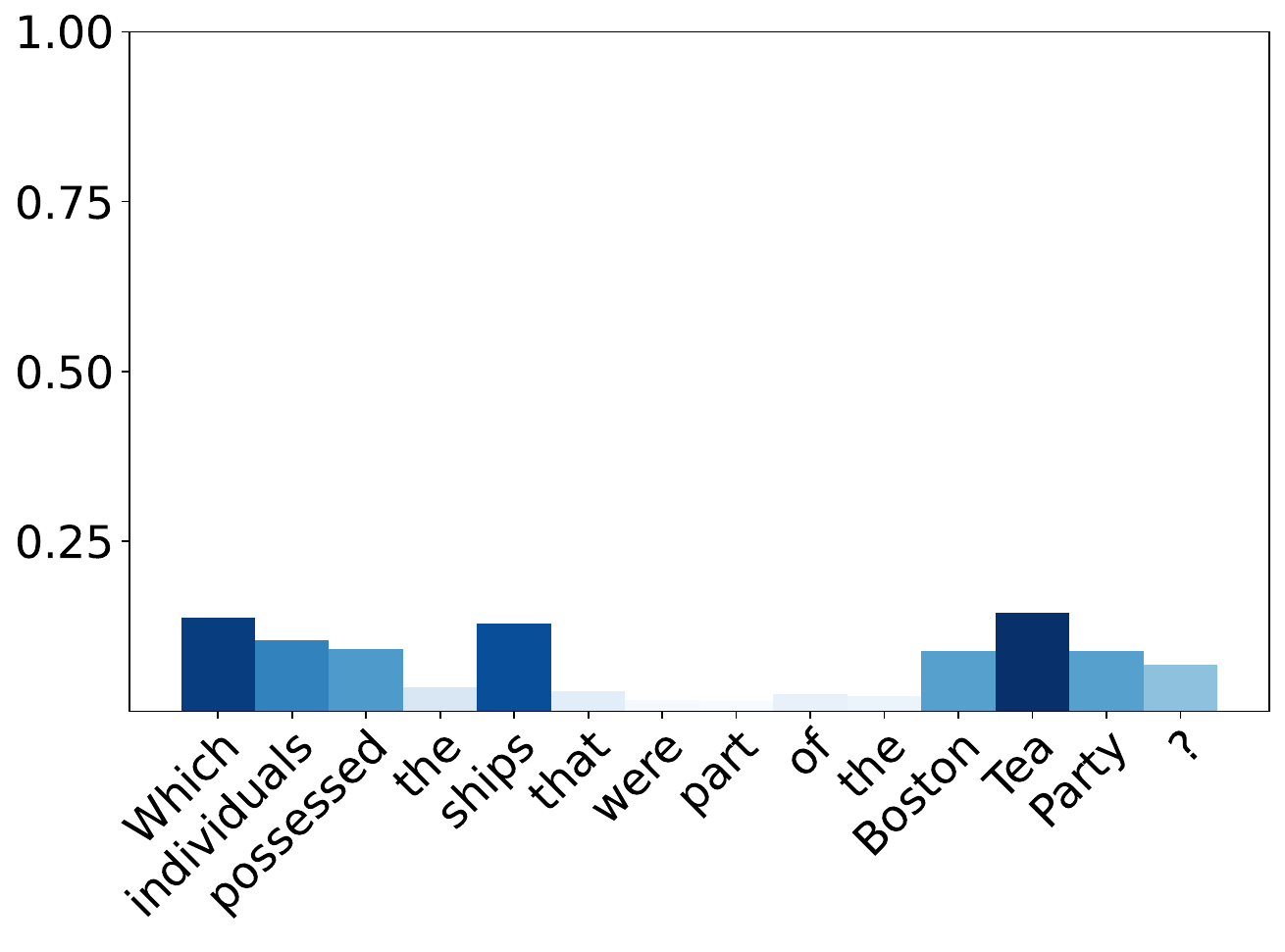}
            \caption[]%
            {{\small Before adding \tframed[line width=0.5bp,fill=vred]{\textcolor{white}{\texttt{\textbf{[PAUSE]}}}} tokens} to original prompt.}
            \label{fig:mean and std of net14}
        \end{subfigure}
        \hfill
        \begin{subfigure}[b]{0.45\textwidth}  
            \centering 
            \includegraphics[width=\textwidth,height=3cm]{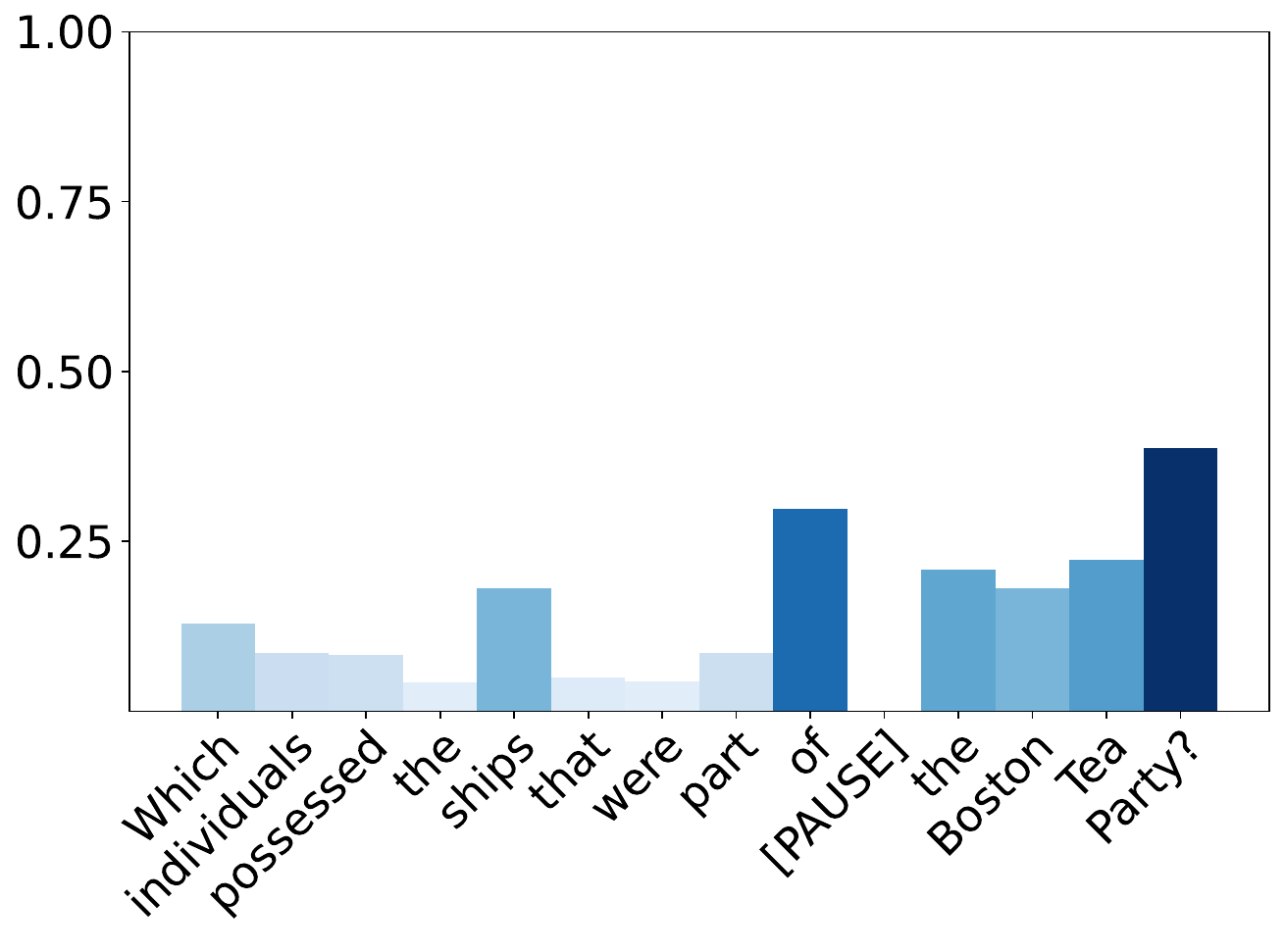}
            \caption[]%
            {{\small After adding \tframed[line width=0.5bp,fill=vred]{\textcolor{white}{\texttt{\textbf{[PAUSE]}}}} tokens} to original prompt.}    
            \label{fig:mean and std of net24}
        \end{subfigure}
        \hfill
        \vskip\baselineskip
        \begin{subfigure}[b]{0.45\textwidth}   
            \centering 
            \includegraphics[width=\textwidth,height=3cm]{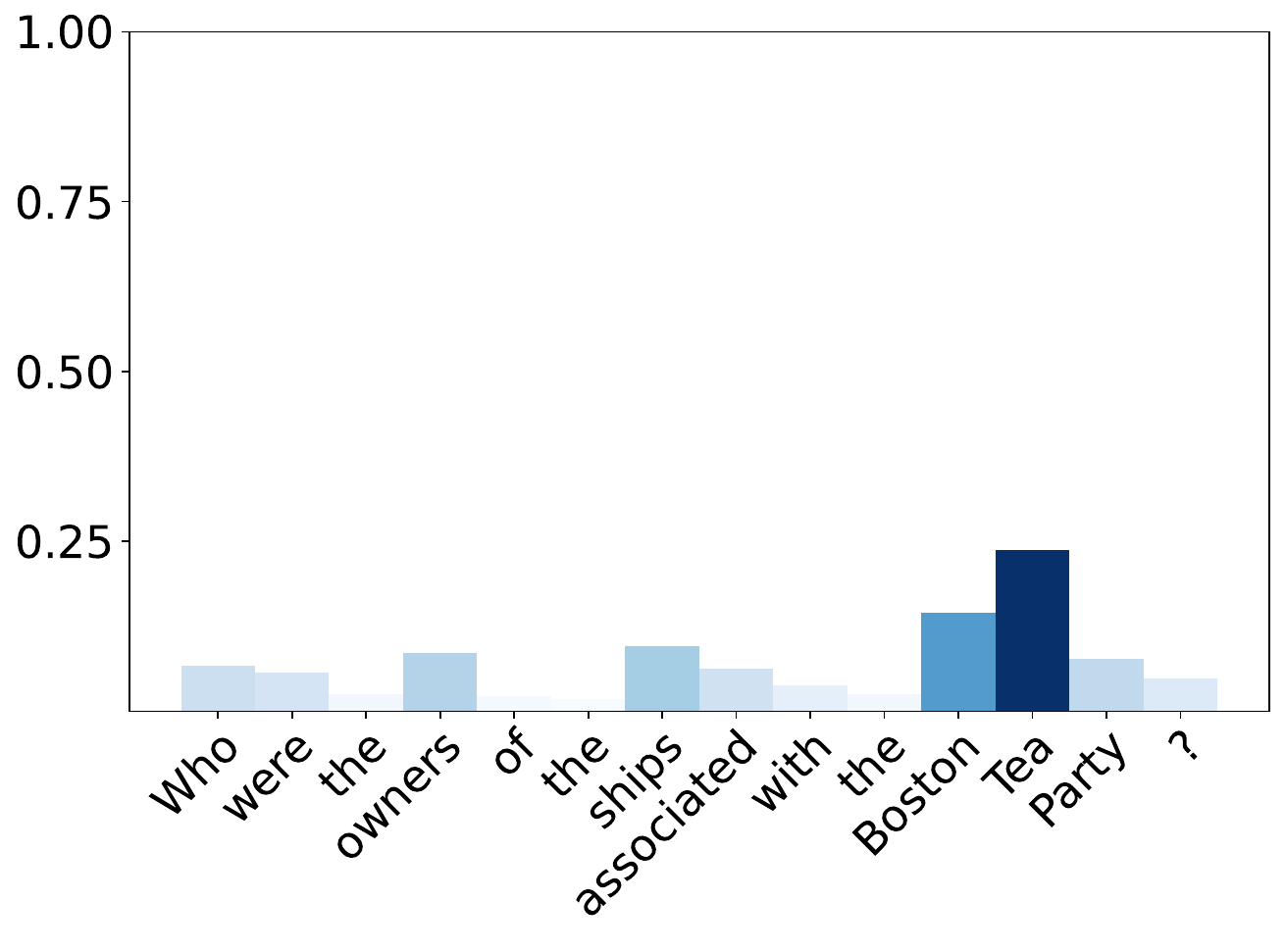}
            \caption[]%
            {{\small Before adding \tframed[line width=0.5bp,fill=vred]{\textcolor{white}{\texttt{\textbf{[PAUSE]}}}} tokens} to paraphrase 1.}    
            \label{fig:mean and std of net34}
        \end{subfigure}
        \hfill
        \begin{subfigure}[b]{0.45\textwidth}   
            \centering 
            \includegraphics[width=\textwidth,height=3cm]{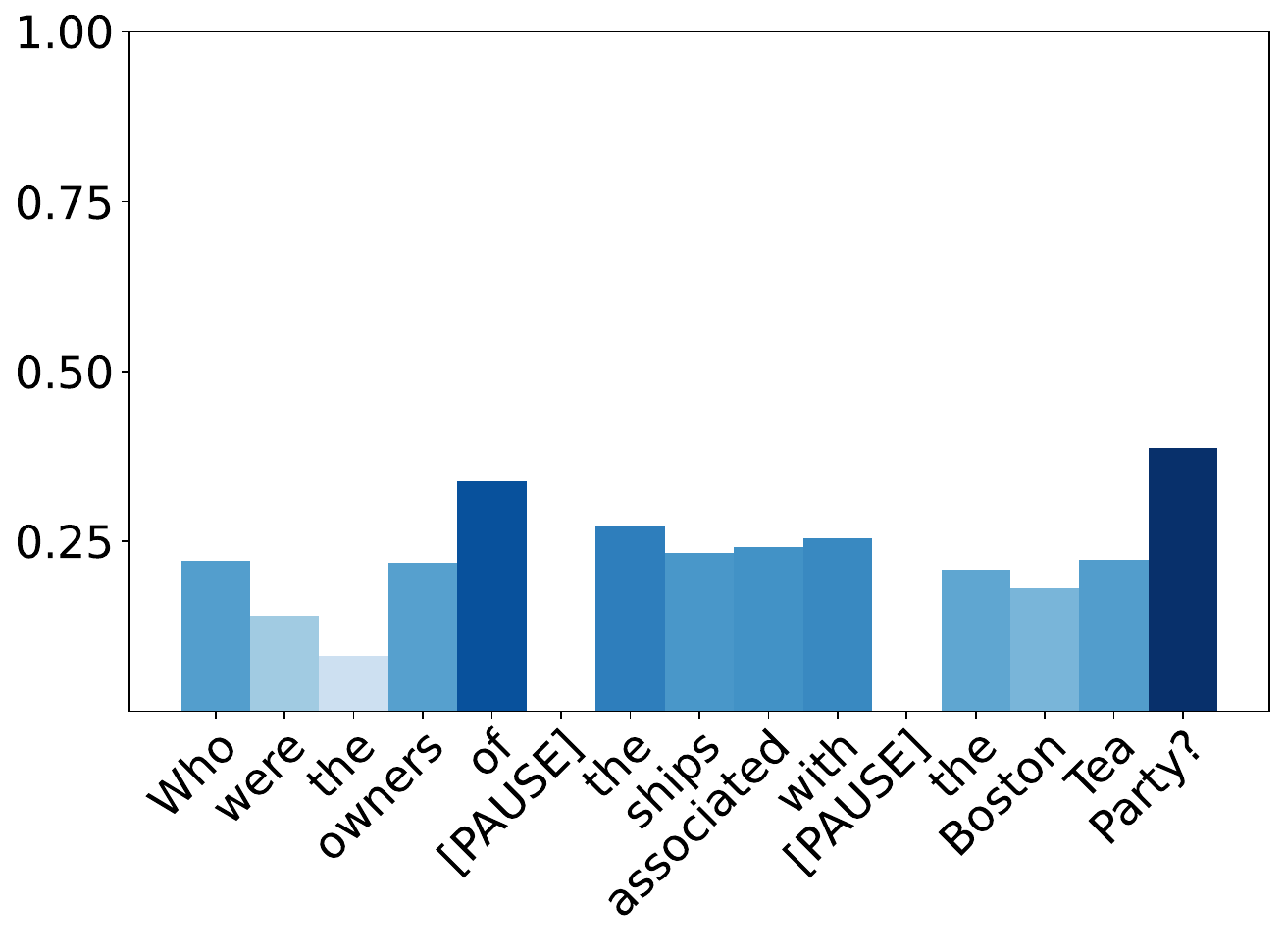}
            \caption[]%
            {{\small After adding \tframed[line width=0.5bp,fill=vred]{\textcolor{white}{\texttt{\textbf{[PAUSE]}}}} tokens} to paraphrase 1.}    
            \label{fig:mean and std of net44}
        \end{subfigure}
        \hfill
        \vskip\baselineskip
        \begin{subfigure}[b]{0.45\textwidth}   
            \centering 
            \includegraphics[width=\textwidth,height=3cm]{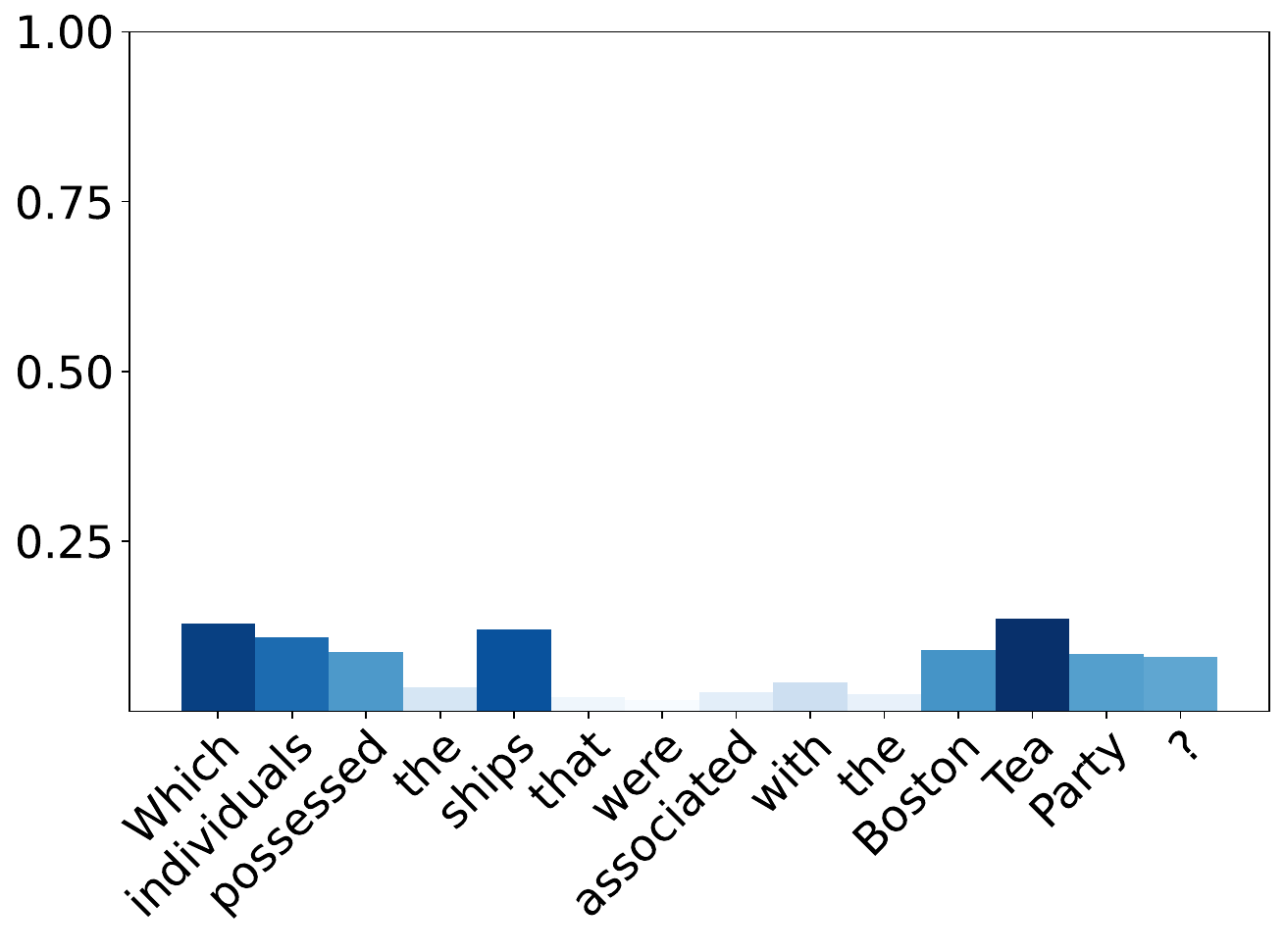}
            \caption[]%
            {{\small Before adding \tframed[line width=0.5bp,fill=vred]{\textcolor{white}{\texttt{\textbf{[PAUSE]}}}} tokens} to paraphrase 2.}
            \label{fig:mean and std of net34}
        \end{subfigure}
        \hfill
        \begin{subfigure}[b]{0.45\textwidth}   
            \centering 
            \includegraphics[width=\textwidth,height=3cm]{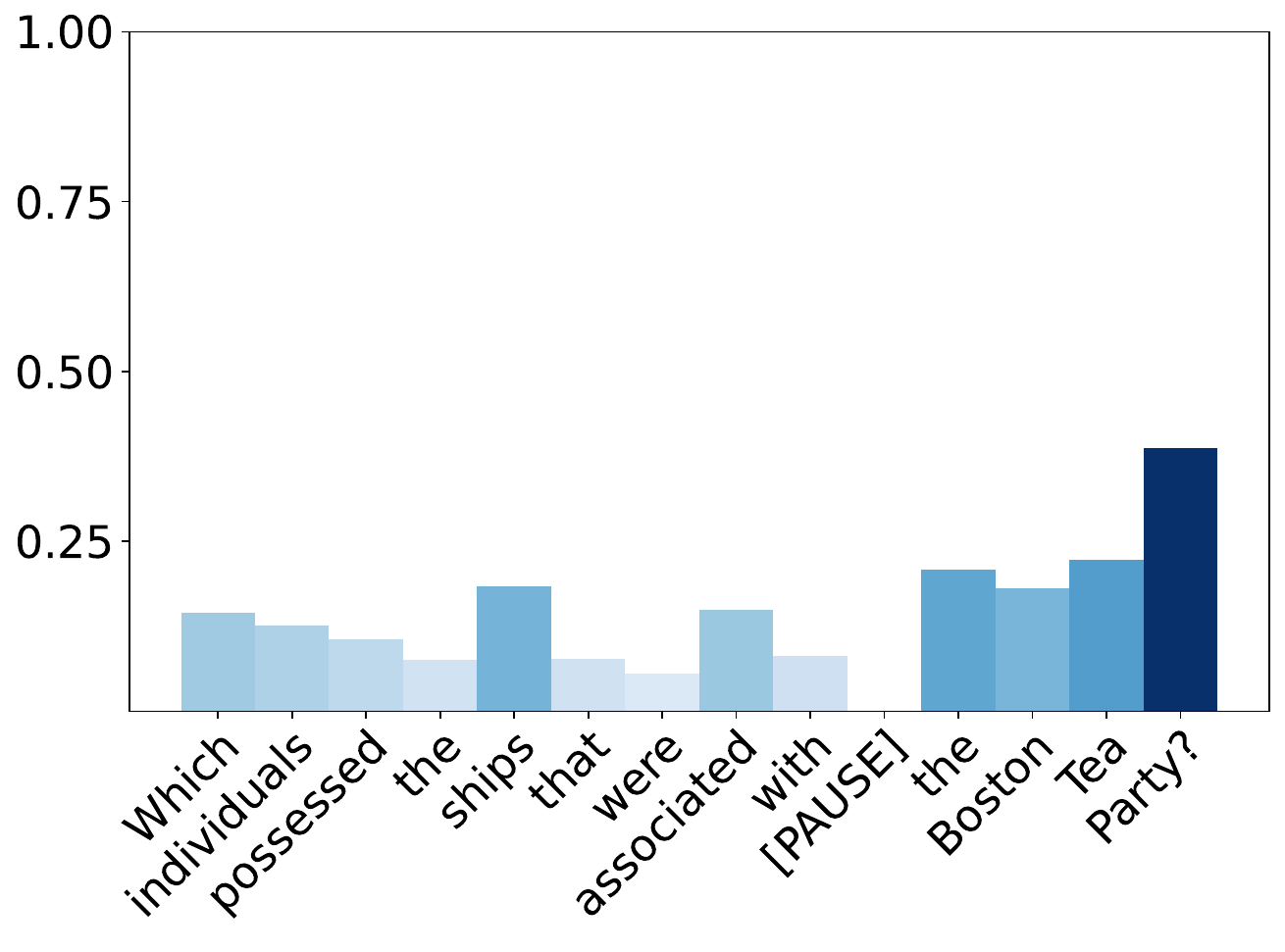}
            \caption[]%
            {{\small After adding \tframed[line width=0.5bp,fill=vred]{\textcolor{white}{\texttt{\textbf{[PAUSE]}}}} tokens} to paraphrase 2.} 
            \label{fig:mean and std of net44}
        \end{subfigure}
        \hfill
        \vskip\baselineskip
        \begin{subfigure}[b]{0.45\textwidth}   
            \centering 
            \includegraphics[width=\textwidth,height=3cm]{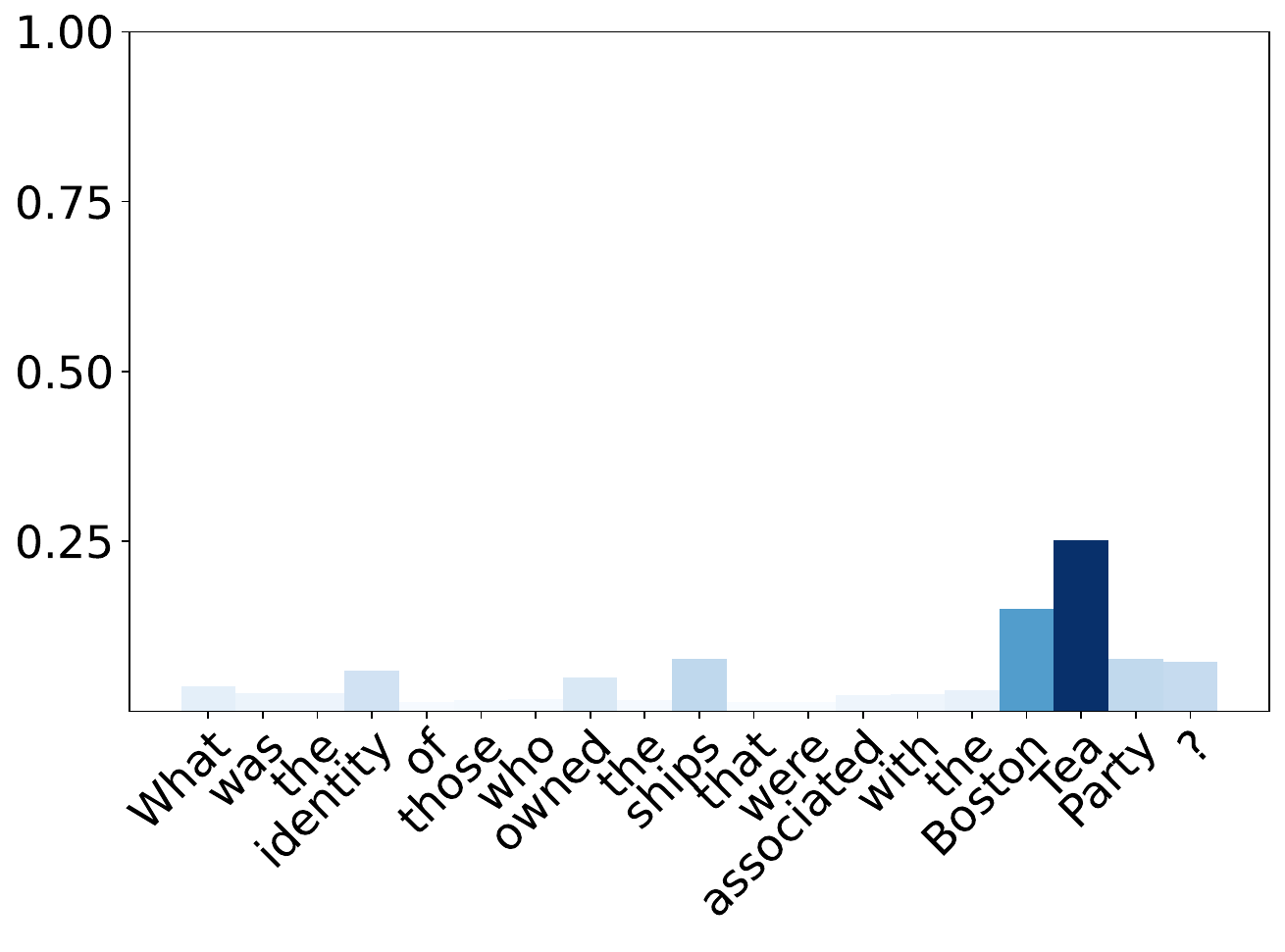}
            \caption[]%
            {{\small Before adding \tframed[line width=0.5bp,fill=vred]{\textcolor{white}{\texttt{\textbf{[PAUSE]}}}} tokens} to paraphrase 3.}
            \label{fig:mean and std of net44}
        \end{subfigure}
        \hfill
        \begin{subfigure}[b]{0.45\textwidth}   
            \centering 
            \includegraphics[width=\textwidth,height=3cm]{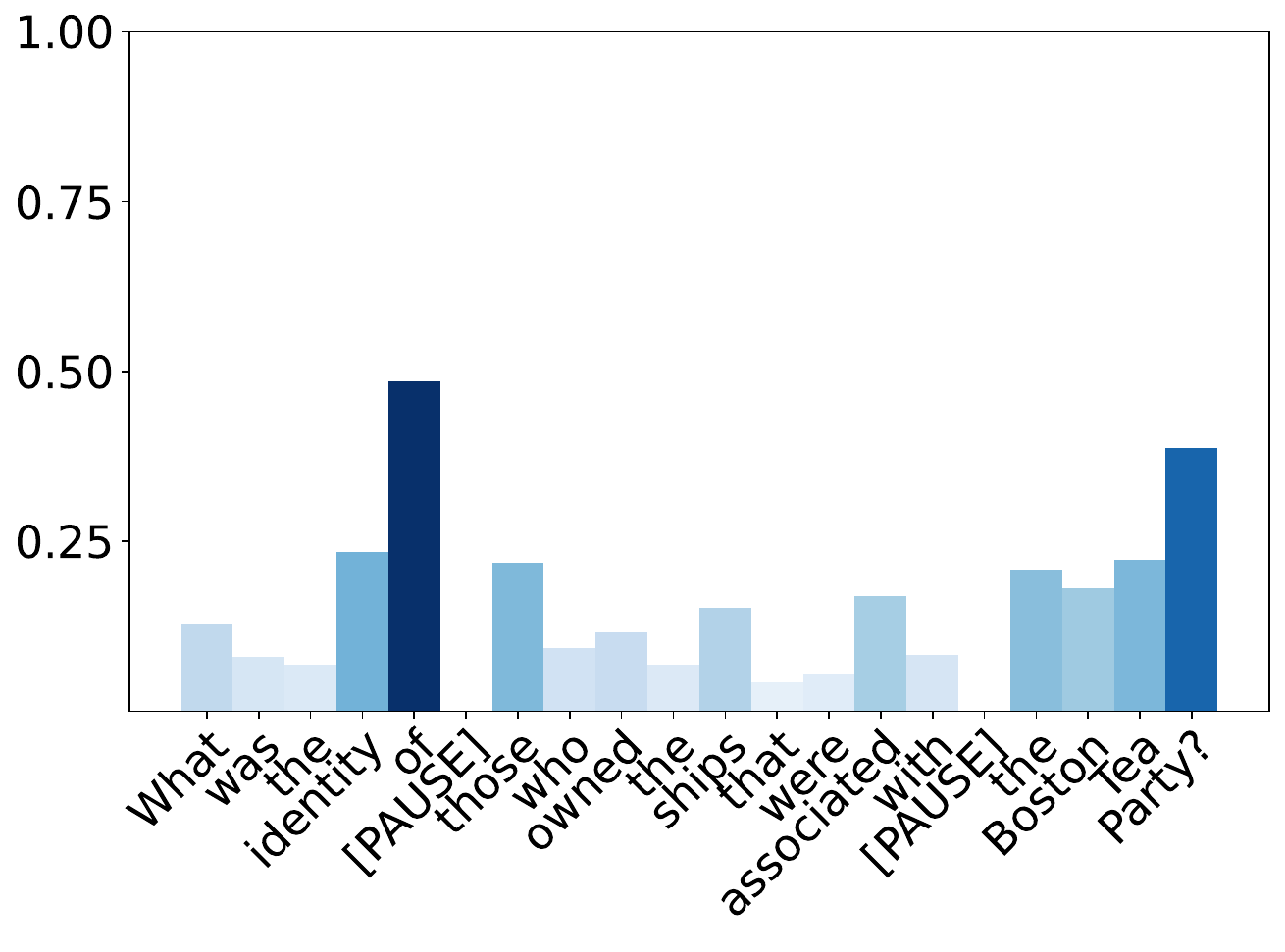}
            \caption[]%
            {{\small After adding \tframed[line width=0.5bp,fill=vred]{\textcolor{white}{\texttt{\textbf{[PAUSE]}}}} tokens} to paraphrase 3.}    
            \label{fig:mean and std of net44}
        \end{subfigure}
        \hfill
        \vskip\baselineskip
        \begin{subfigure}[b]{0.45\textwidth}   
            \centering 
            \includegraphics[width=\textwidth,height=3cm]{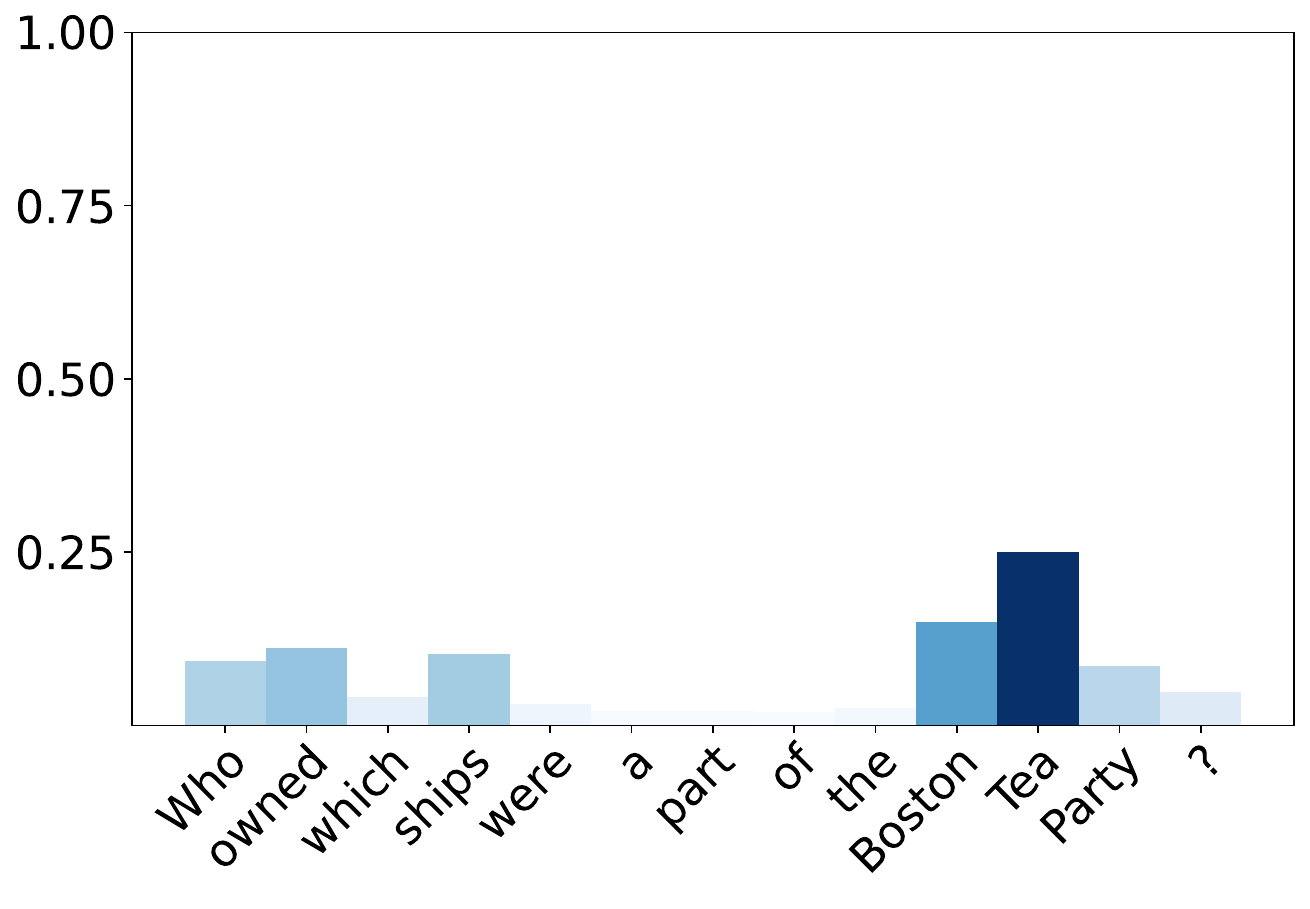}
            \caption[]%
            {{\small Before adding \tframed[line width=0.5bp,fill=vred]{\textcolor{white}{\texttt{\textbf{[PAUSE]}}}} tokens} to paraphrase 4.}    
            \label{fig:mean and std of net44}
        \end{subfigure}
        \hfill
        \begin{subfigure}[b]{0.45\textwidth}   
            \centering 
            \includegraphics[width=\textwidth,height=3cm]{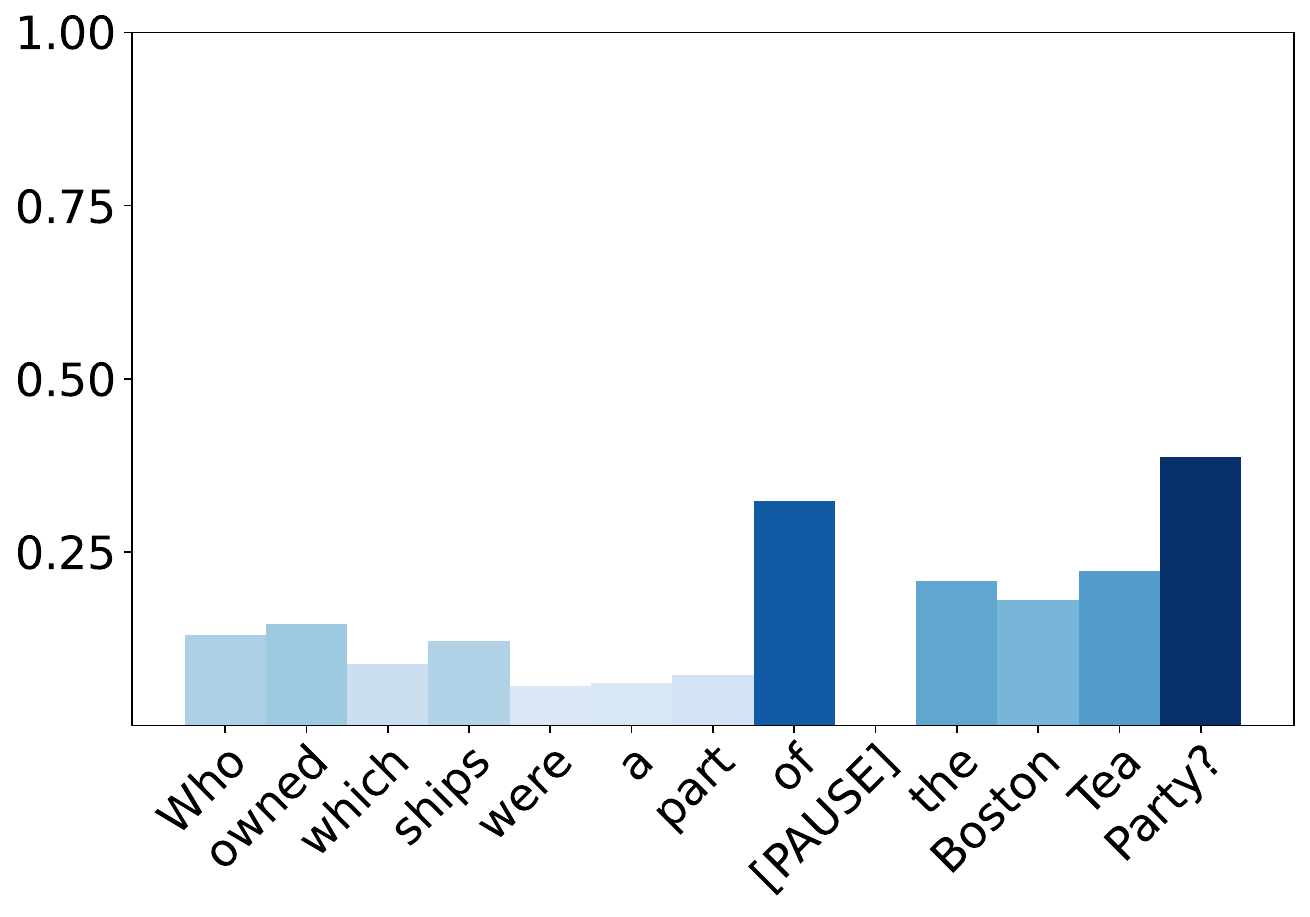}
            \caption[]%
            {{\small After adding \tframed[line width=0.5bp,fill=vred]{\textcolor{white}{\texttt{\textbf{[PAUSE]}}}} tokens} to paraphrase 4.}    
            \label{fig:mean and std of net44}
        \end{subfigure}
        \hfill
        \vskip\baselineskip
        \begin{subfigure}[b]{0.45\textwidth}   
            \centering 
            \includegraphics[width=\textwidth,height=3cm]{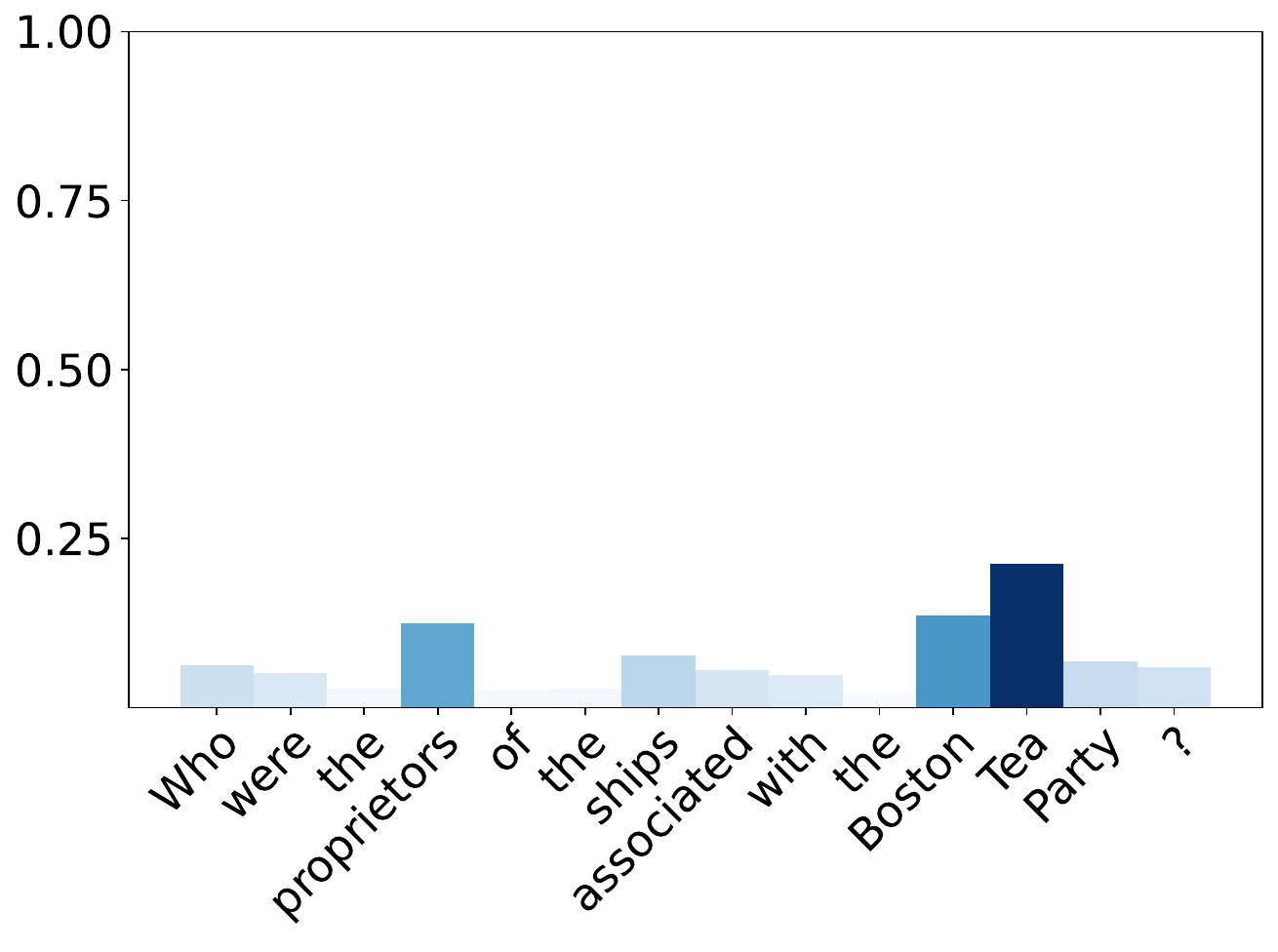}
            \caption[]%
            {{\small Before adding \tframed[line width=0.5bp,fill=vred]{\textcolor{white}{\texttt{\textbf{[PAUSE]}}}} tokens} to paraphrase 5.}    
            \label{fig:mean and std of net44}
        \end{subfigure}
        \hfill
        \begin{subfigure}[b]{0.45\textwidth}   
            \centering 
            \includegraphics[width=\textwidth,height=3cm]{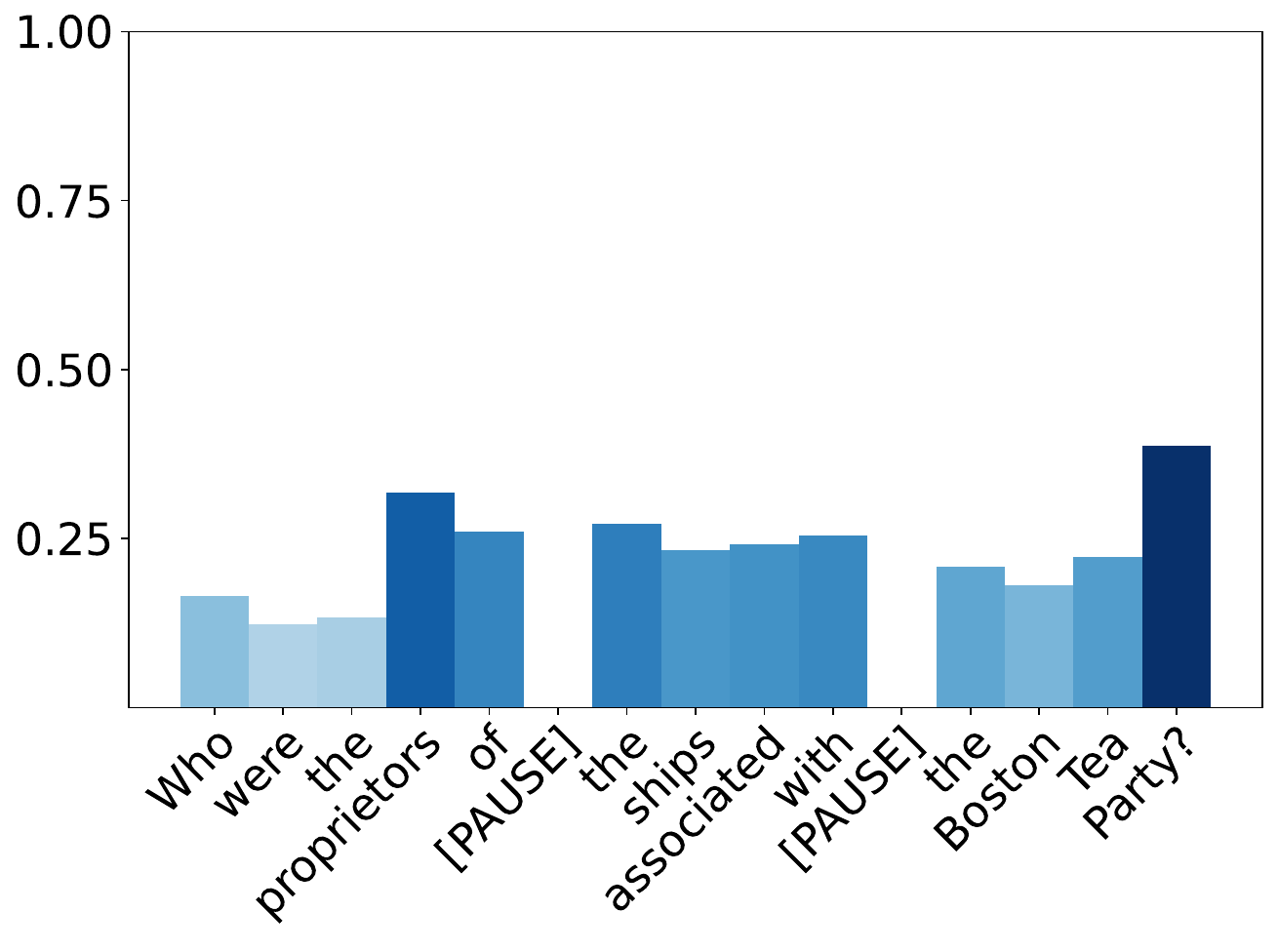}
            \caption[]%
            {{\small After adding \tframed[line width=0.5bp,fill=vred]{\textcolor{white}{\texttt{\textbf{[PAUSE]}}}} tokens} to paraphrase 5.}    
            \label{fig:mean and std of net44}
        \end{subfigure}
        \caption[]%
            {{\small The phrase \textbf{Boston Tea} gets more importance score after adding \tframed[line width=0.5bp,fill=vred]{\textcolor{white}{\texttt{\textbf{[PAUSE]}}}} token for bloomz.}}   
        \label{fig:bloomz}
\end{figure*}

\begin{figure*}[!ht]
        \centering
        \begin{subfigure}[b]{0.45\textwidth}
            \centering
            \includegraphics[width=\textwidth,height=3cm]{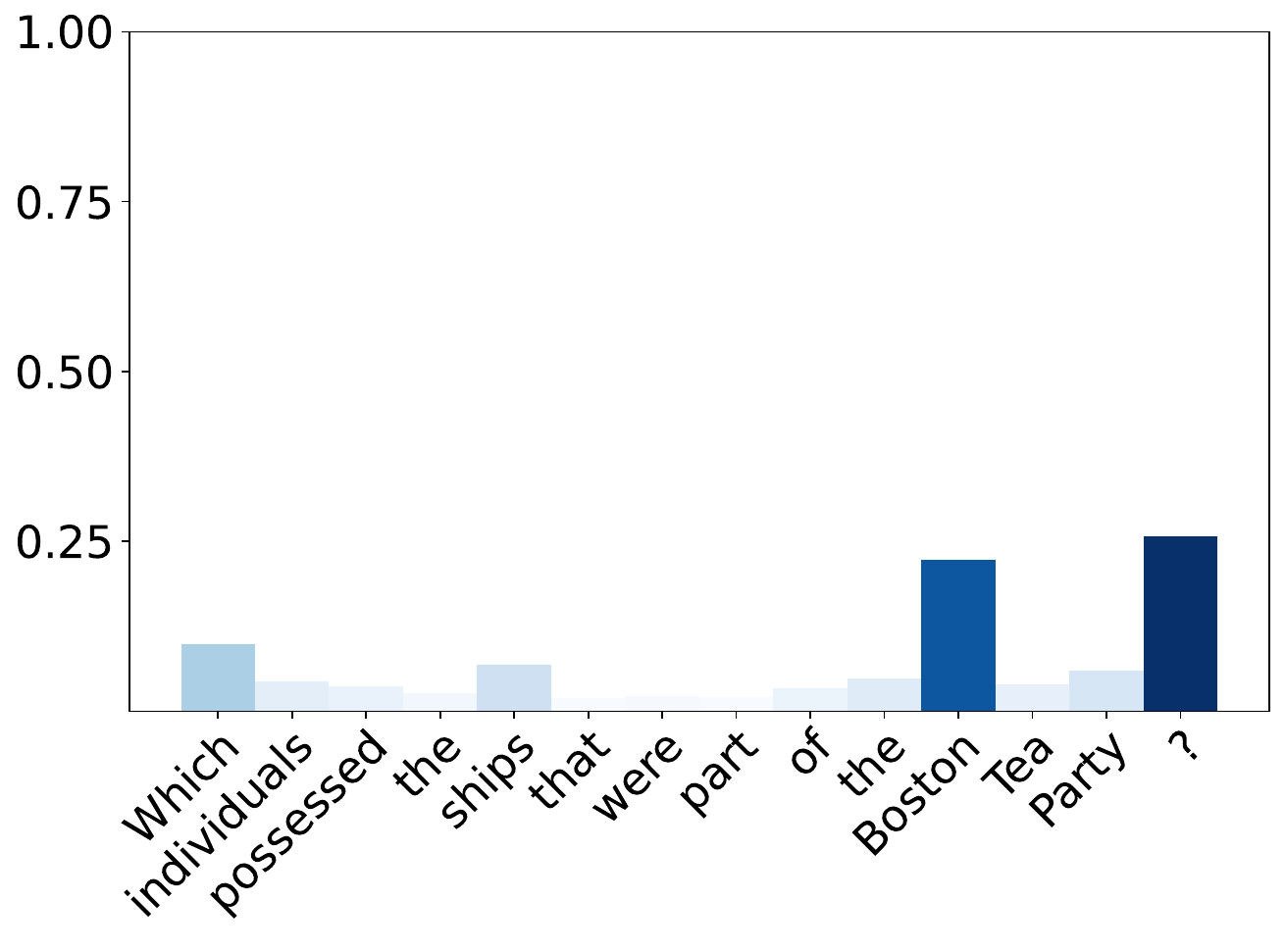}
            \caption[]%
            {{\small Before adding \tframed[line width=0.5bp,fill=vred]{\textcolor{white}{\texttt{\textbf{[PAUSE]}}}} tokens} to original prompt.}
            \label{fig:mean and std of net14}
        \end{subfigure}
        \hfill
        \begin{subfigure}[b]{0.45\textwidth}  
            \centering 
            \includegraphics[width=\textwidth,height=3cm]{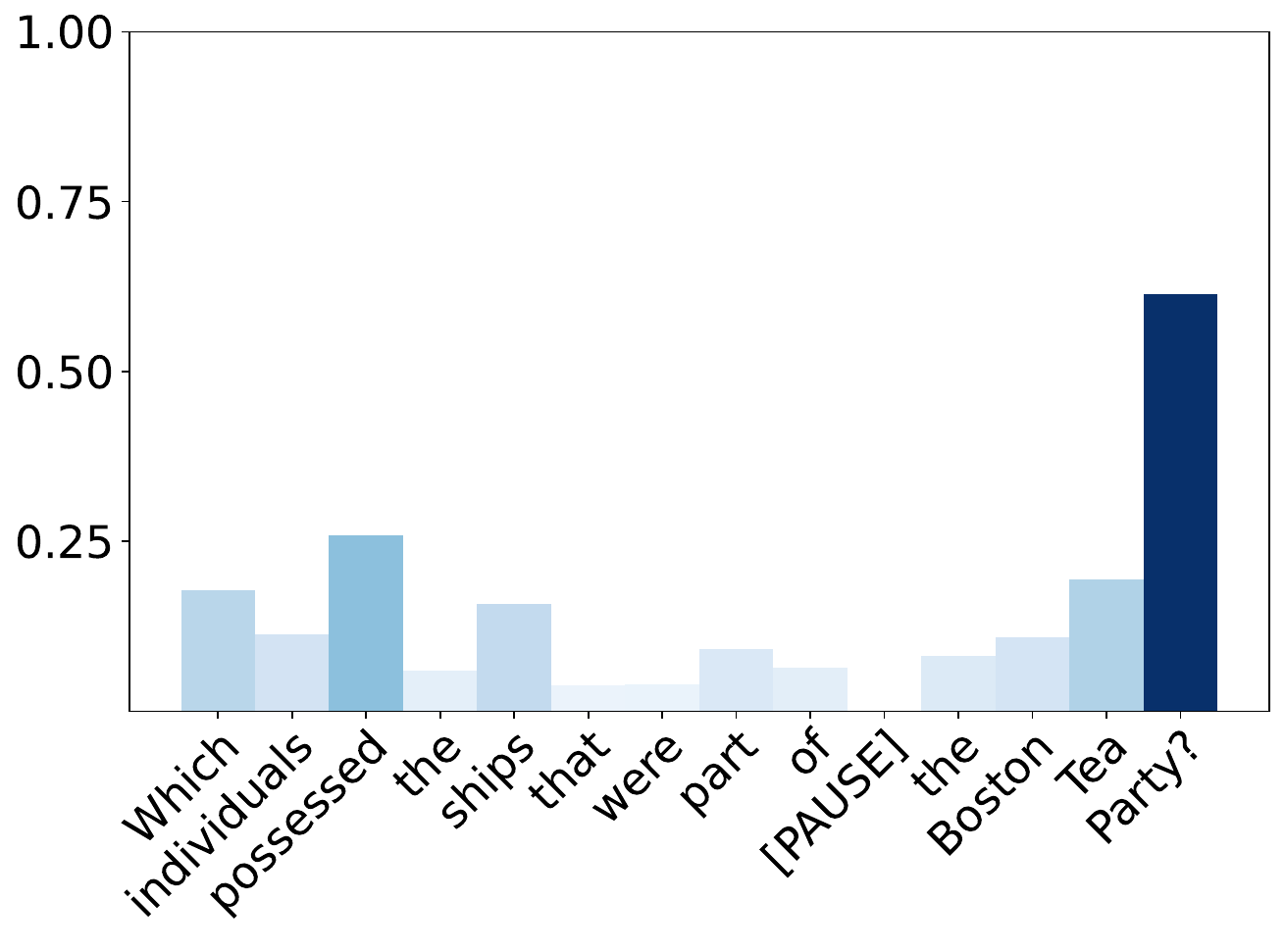}
            \caption[]%
            {{\small After adding \tframed[line width=0.5bp,fill=vred]{\textcolor{white}{\texttt{\textbf{[PAUSE]}}}} tokens} to original prompt.}    
            \label{fig:mean and std of net24}
        \end{subfigure}
        \hfill
        \vskip\baselineskip
        \begin{subfigure}[b]{0.45\textwidth}   
            \centering 
            \includegraphics[width=\textwidth,height=3cm]{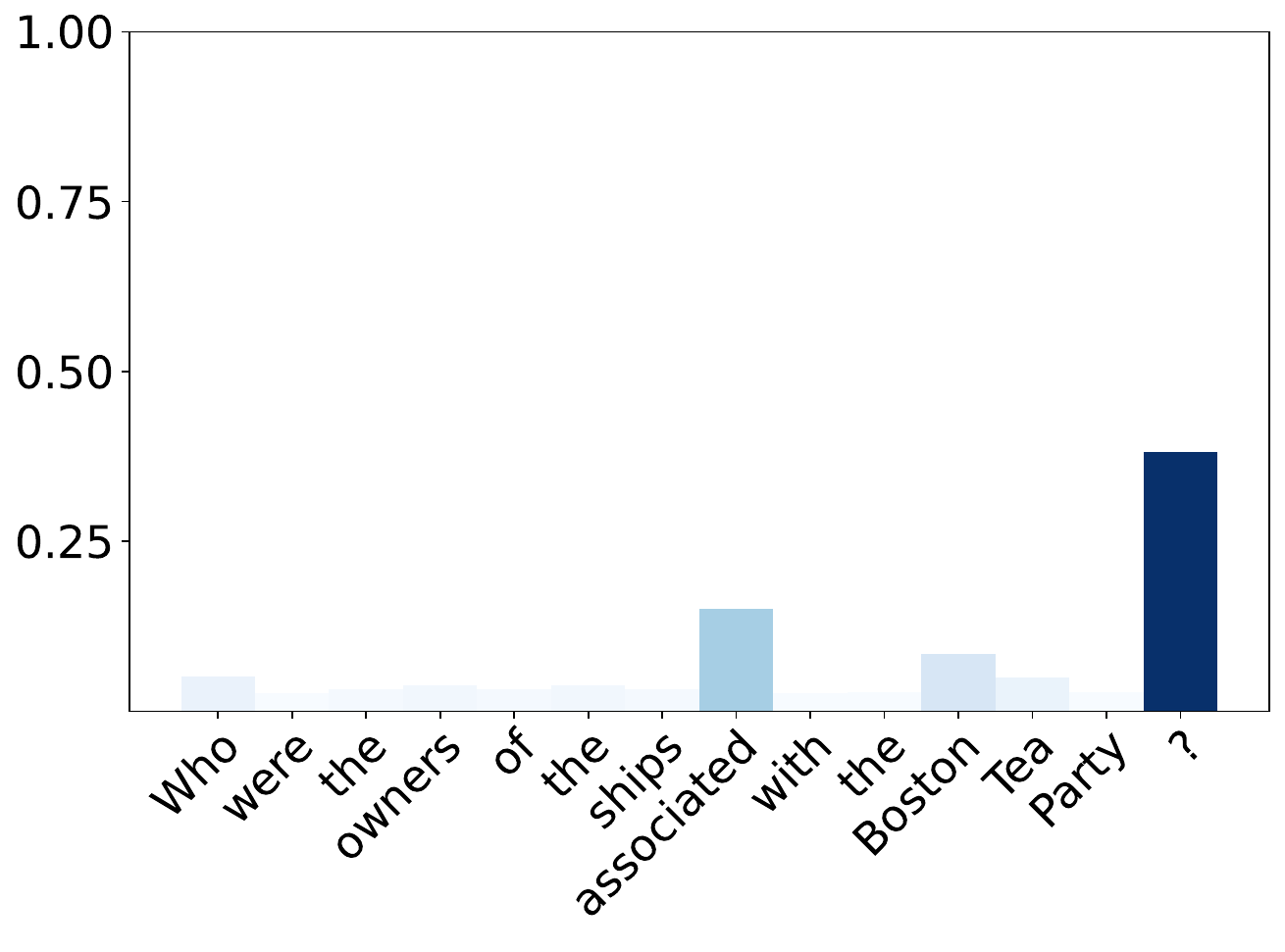}
            \caption[]%
            {{\small Before adding \tframed[line width=0.5bp,fill=vred]{\textcolor{white}{\texttt{\textbf{[PAUSE]}}}} tokens} to paraphrase 1.}    
            \label{fig:mean and std of net34}
        \end{subfigure}
        \hfill
        \begin{subfigure}[b]{0.45\textwidth}   
            \centering 
            \includegraphics[width=\textwidth,height=3cm]{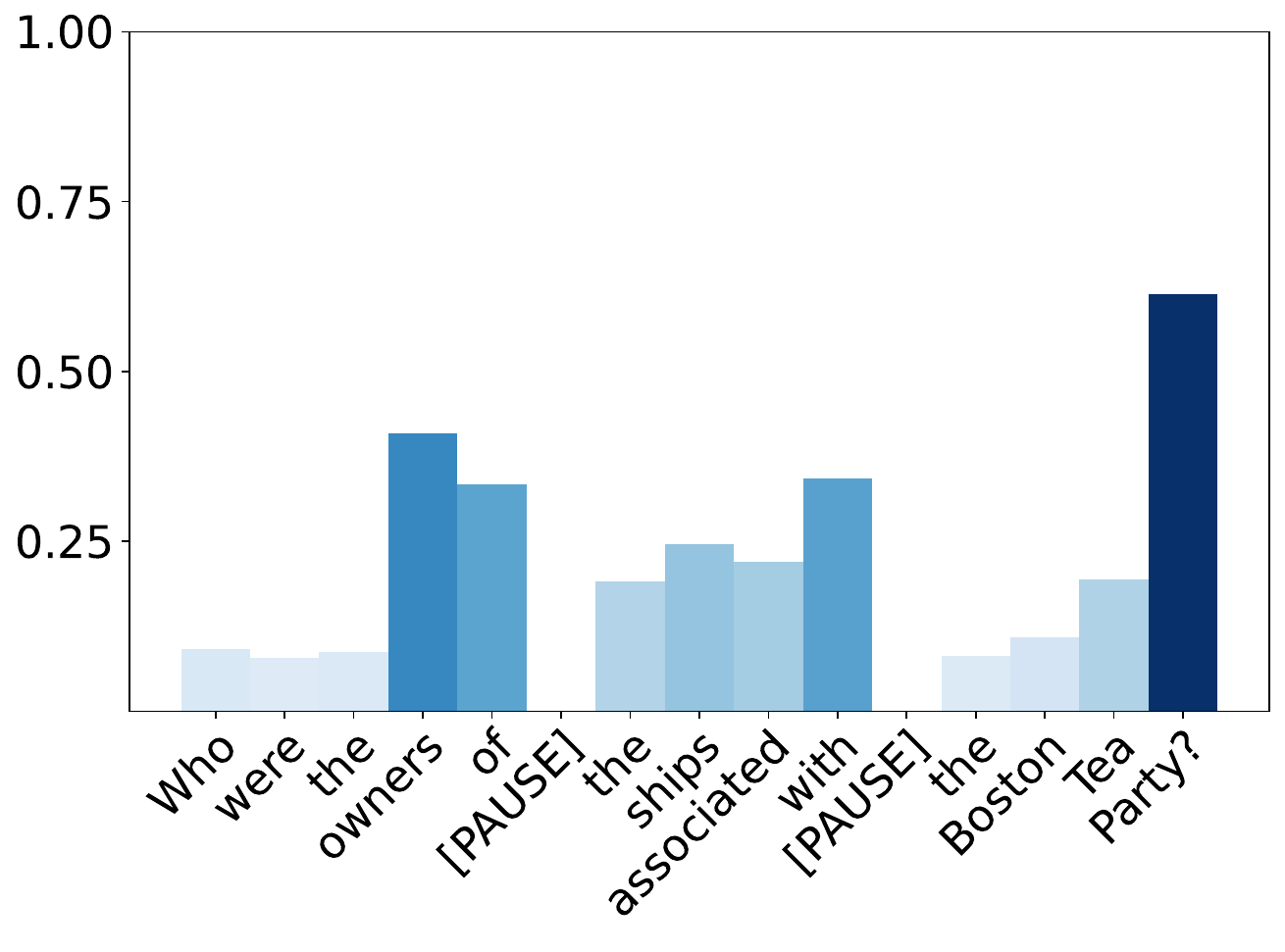}
            \caption[]%
            {{\small After adding \tframed[line width=0.5bp,fill=vred]{\textcolor{white}{\texttt{\textbf{[PAUSE]}}}} tokens} to paraphrase 1.}    
            \label{fig:mean and std of net44}
        \end{subfigure}
        \hfill
        \vskip\baselineskip
        \begin{subfigure}[b]{0.45\textwidth}   
            \centering 
            \includegraphics[width=\textwidth,height=3cm]{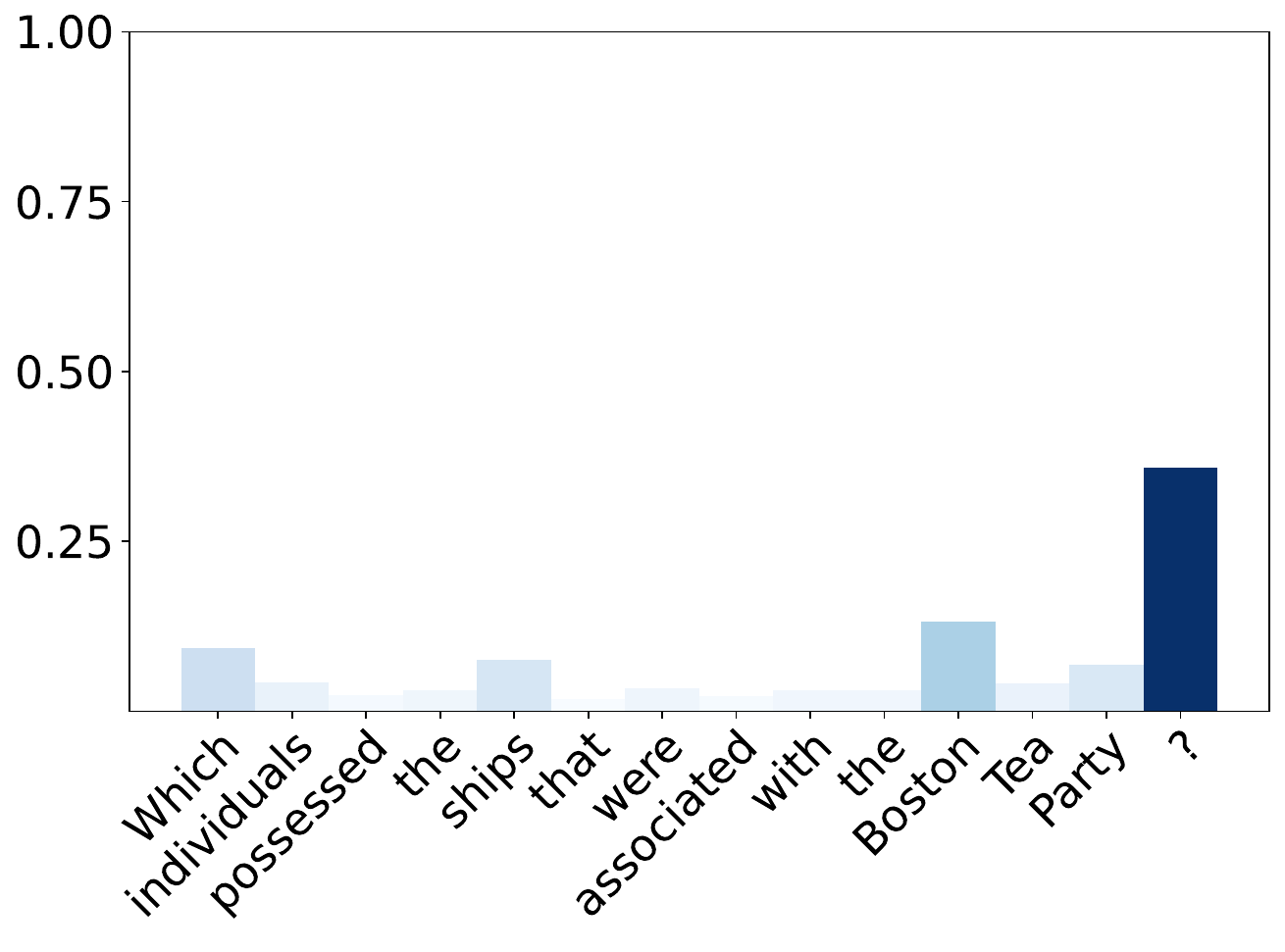}
            \caption[]%
            {{\small Before adding \tframed[line width=0.5bp,fill=vred]{\textcolor{white}{\texttt{\textbf{[PAUSE]}}}} tokens} to paraphrase 2.}
            \label{fig:mean and std of net34}
        \end{subfigure}
        \hfill
        \begin{subfigure}[b]{0.45\textwidth}   
            \centering 
            \includegraphics[width=\textwidth,height=3cm]{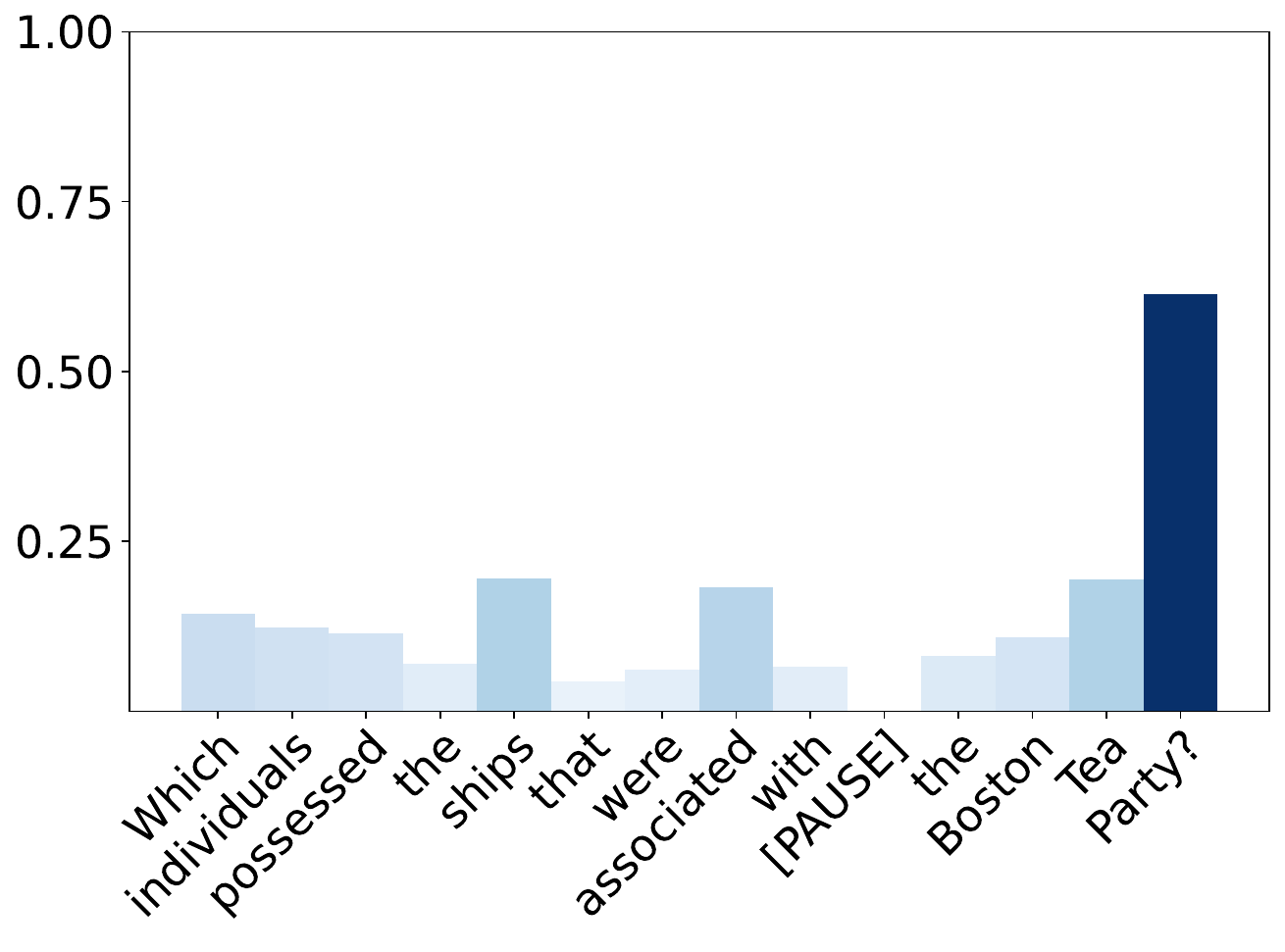}
            \caption[]%
            {{\small After adding \tframed[line width=0.5bp,fill=vred]{\textcolor{white}{\texttt{\textbf{[PAUSE]}}}} tokens} to paraphrase 2.} 
            \label{fig:mean and std of net44}
        \end{subfigure}
        \hfill
        \vskip\baselineskip
        \begin{subfigure}[b]{0.45\textwidth}   
            \centering 
            \includegraphics[width=\textwidth,height=3cm]{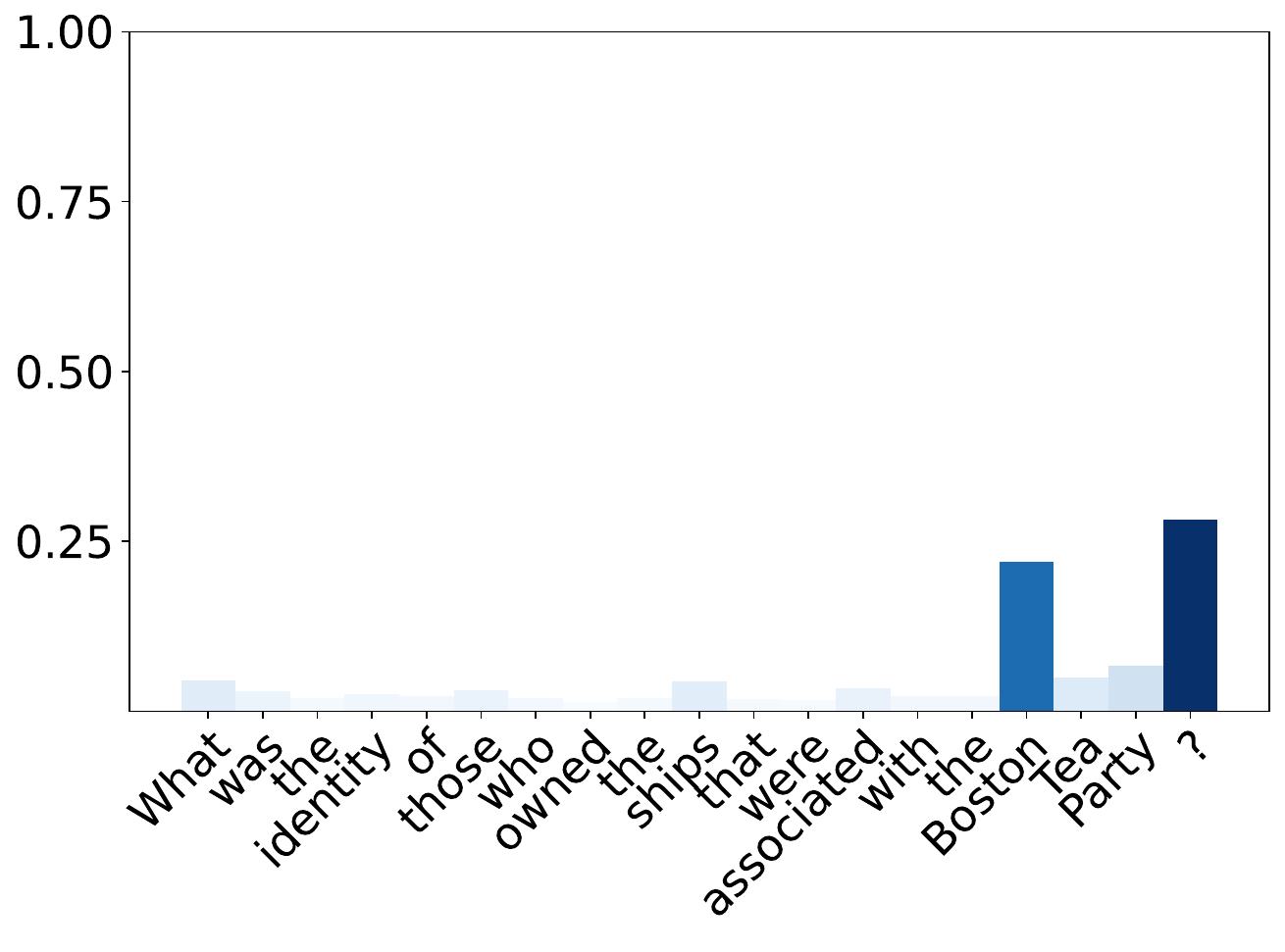}
            \caption[]%
            {{\small Before adding \tframed[line width=0.5bp,fill=vred]{\textcolor{white}{\texttt{\textbf{[PAUSE]}}}} tokens} to paraphrase 3.}
            \label{fig:mean and std of net44}
        \end{subfigure}
        \hfill
        \begin{subfigure}[b]{0.45\textwidth}   
            \centering 
            \includegraphics[width=\textwidth,height=3cm]{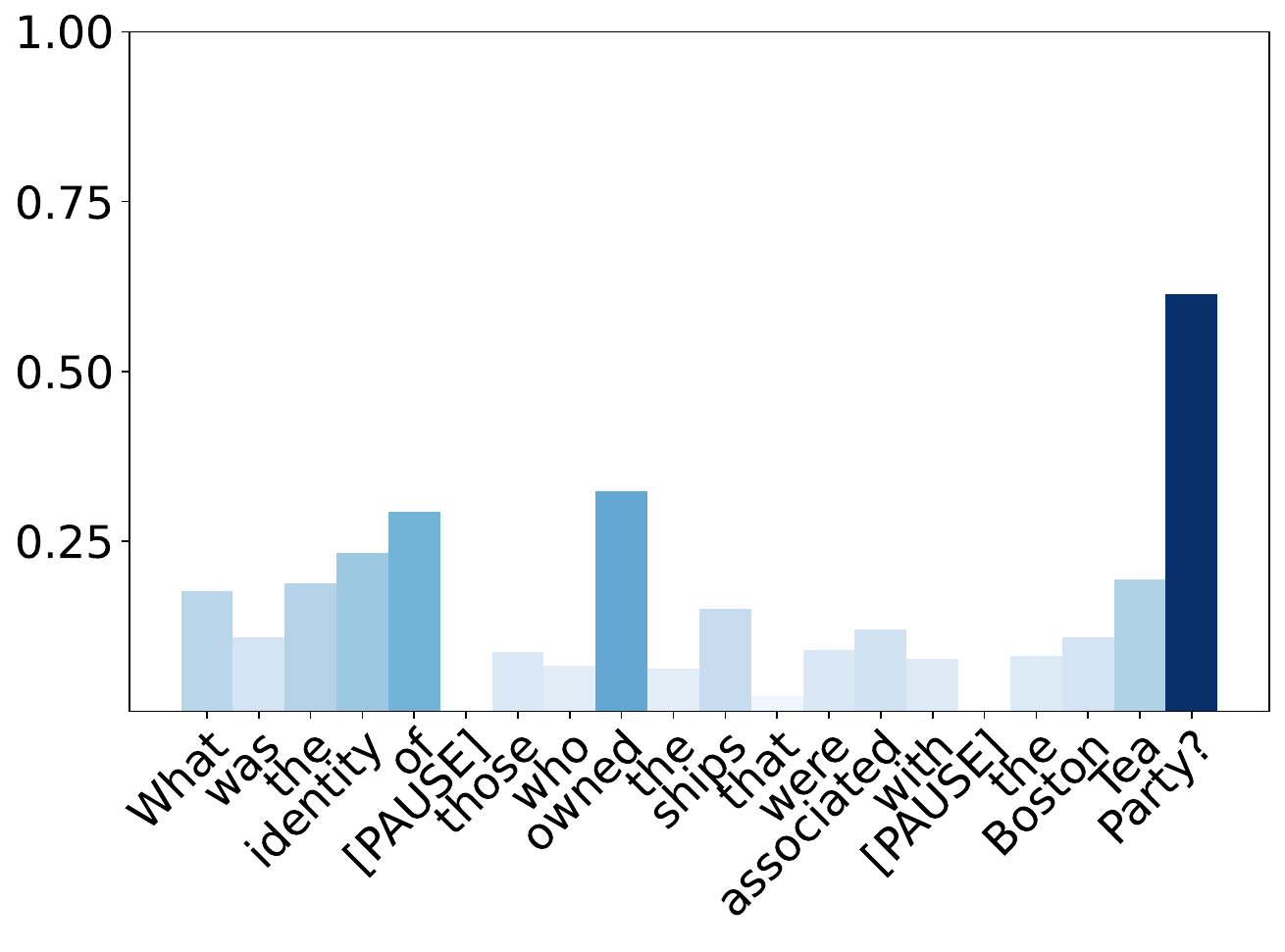}
            \caption[]%
            {{\small After adding \tframed[line width=0.5bp,fill=vred]{\textcolor{white}{\texttt{\textbf{[PAUSE]}}}} tokens} to paraphrase 3.}    
            \label{fig:mean and std of net44}
        \end{subfigure}
        \hfill
        \vskip\baselineskip
        \begin{subfigure}[b]{0.45\textwidth}   
            \centering 
            \includegraphics[width=\textwidth,height=3cm]{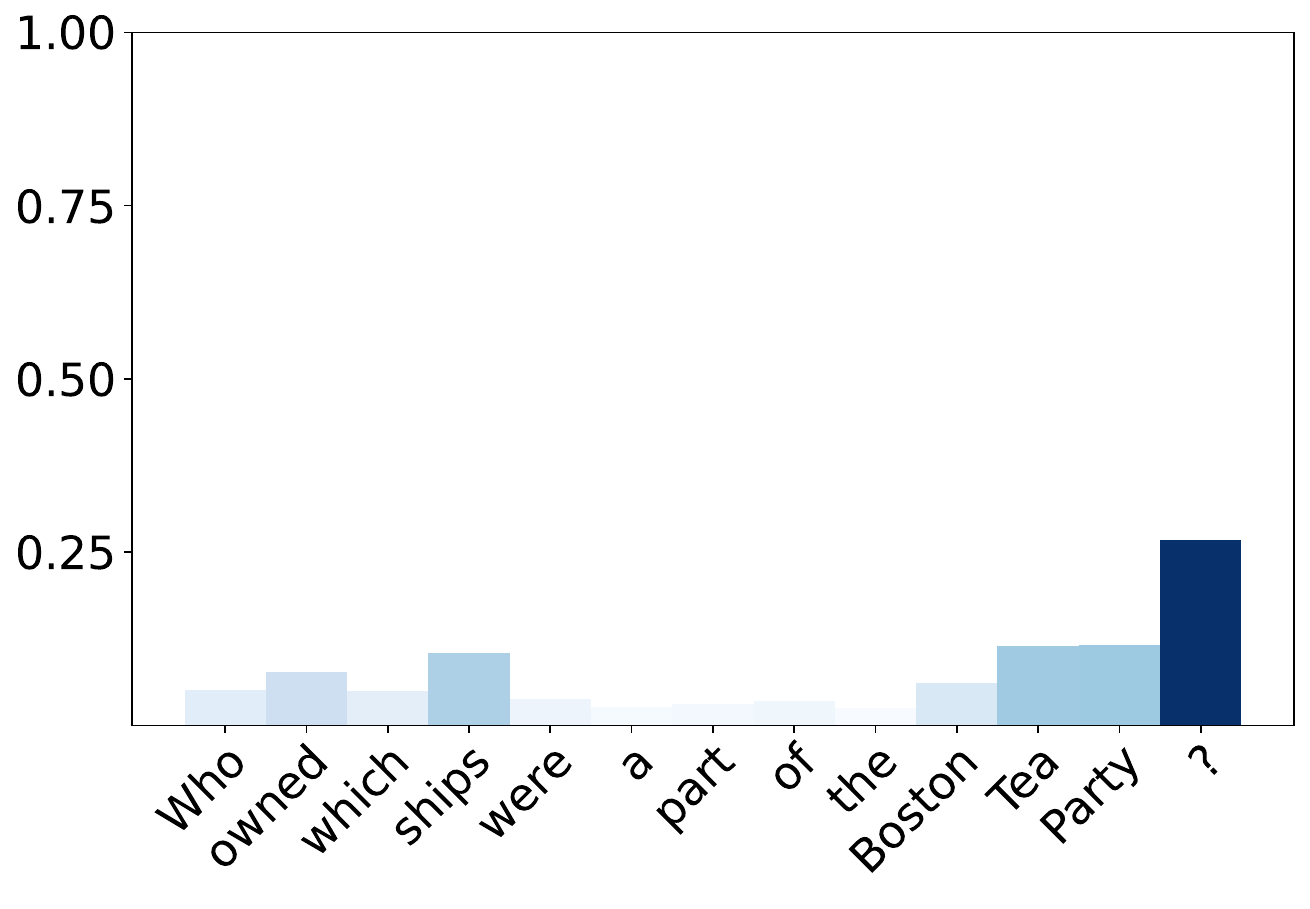}
            \caption[]%
            {{\small Before adding \tframed[line width=0.5bp,fill=vred]{\textcolor{white}{\texttt{\textbf{[PAUSE]}}}} tokens} to paraphrase 4.}    
            \label{fig:mean and std of net44}
        \end{subfigure}
        \hfill
        \begin{subfigure}[b]{0.45\textwidth}   
            \centering 
            \includegraphics[width=\textwidth,height=3cm]{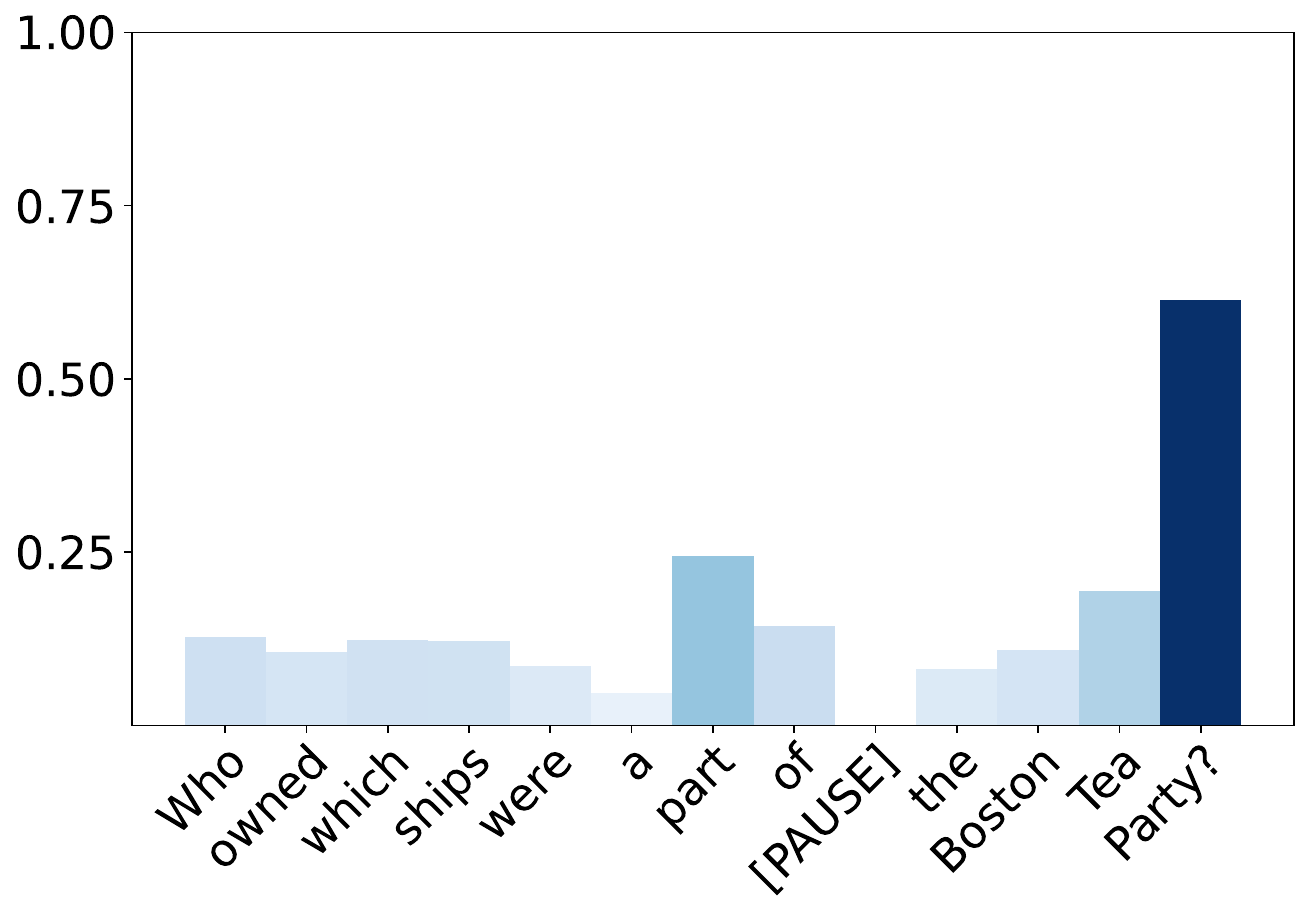}
            \caption[]%
            {{\small After adding \tframed[line width=0.5bp,fill=vred]{\textcolor{white}{\texttt{\textbf{[PAUSE]}}}} tokens} to paraphrase 4.}    
            \label{fig:mean and std of net44}
        \end{subfigure}
        \hfill
        \vskip\baselineskip
        \begin{subfigure}[b]{0.45\textwidth}   
            \centering 
            \includegraphics[width=\textwidth,height=3cm]{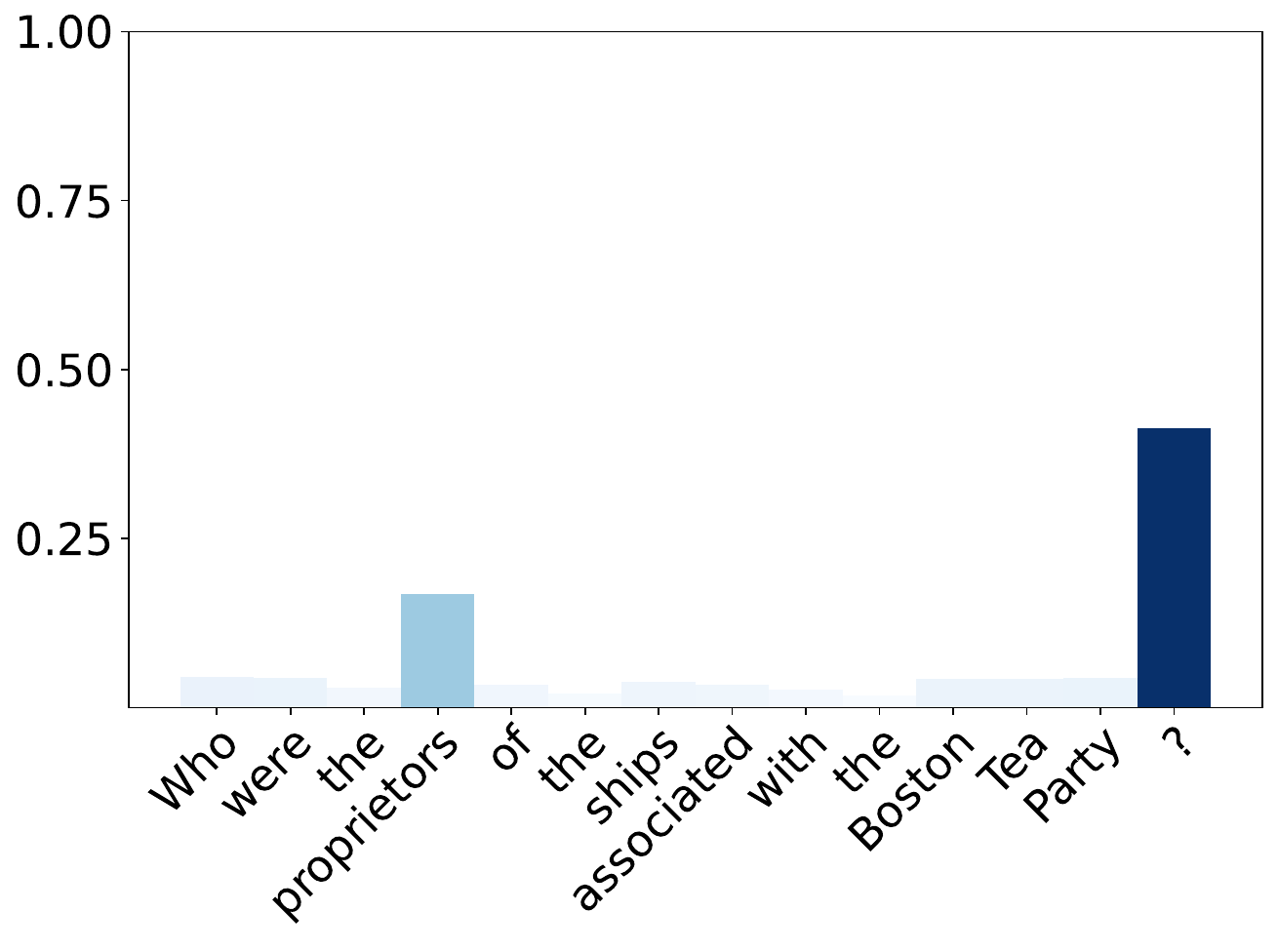}
            \caption[]%
            {{\small Before adding \tframed[line width=0.5bp,fill=vred]{\textcolor{white}{\texttt{\textbf{[PAUSE]}}}} tokens} to paraphrase 5.}    
            \label{fig:mean and std of net44}
        \end{subfigure}
        \hfill
        \begin{subfigure}[b]{0.45\textwidth}   
            \centering 
            \includegraphics[width=\textwidth,height=3cm]{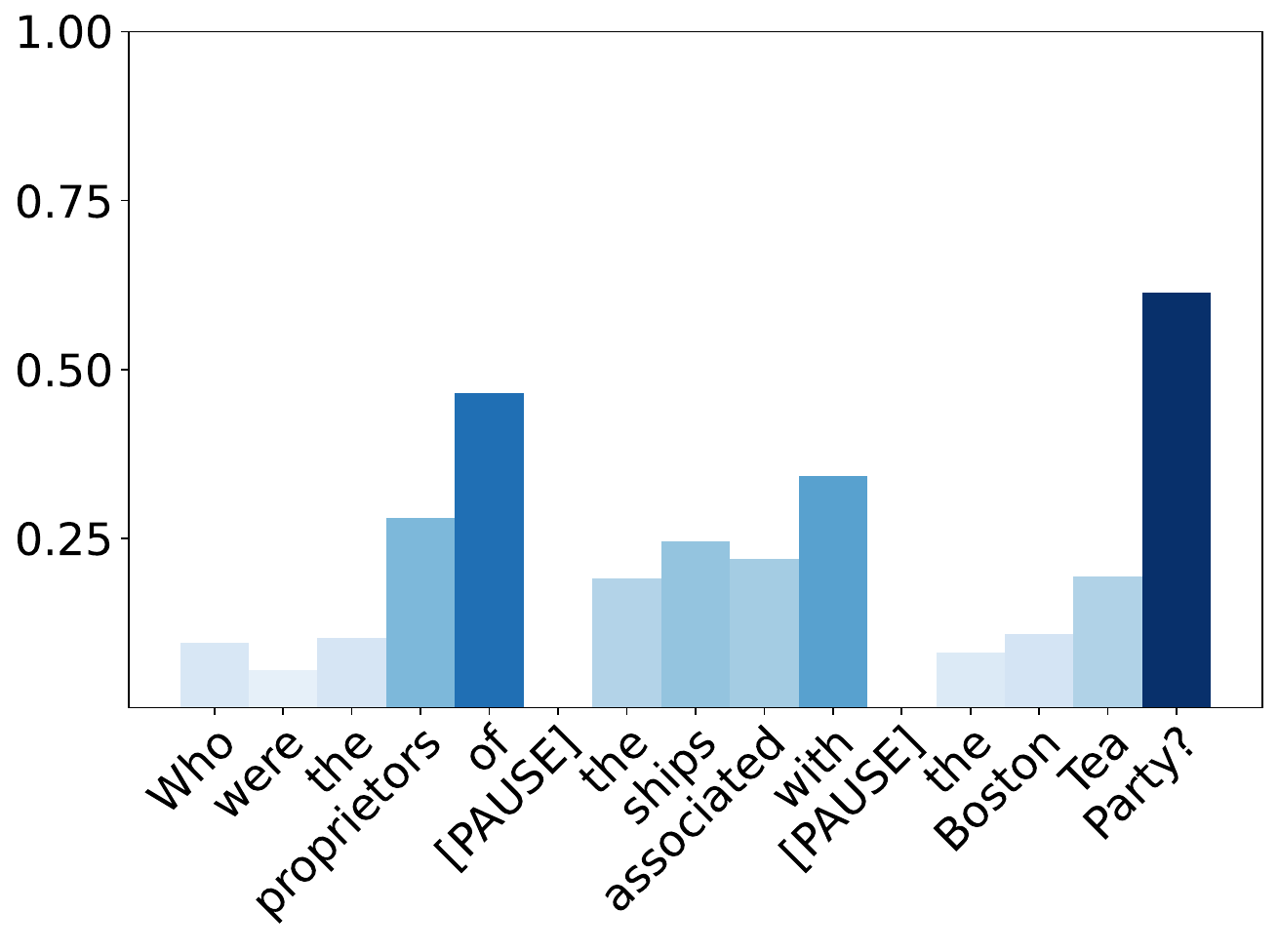}
            \caption[]%
            {{\small After adding \tframed[line width=0.5bp,fill=vred]{\textcolor{white}{\texttt{\textbf{[PAUSE]}}}} tokens} to paraphrase 5.}    
            \label{fig:mean and std of net44}
        \end{subfigure}
        \caption[]%
            {{\small The phrase \textbf{Boston Tea} gets more importance score after adding \tframed[line width=0.5bp,fill=vred]{\textcolor{white}{\texttt{\textbf{[PAUSE]}}}} token for dolly.}}   
        \label{fig:dolly}
\end{figure*}

\begin{figure*}[!ht]
        \centering
        \begin{subfigure}[b]{0.45\textwidth}
            \centering
            \includegraphics[width=\textwidth,height=3cm]{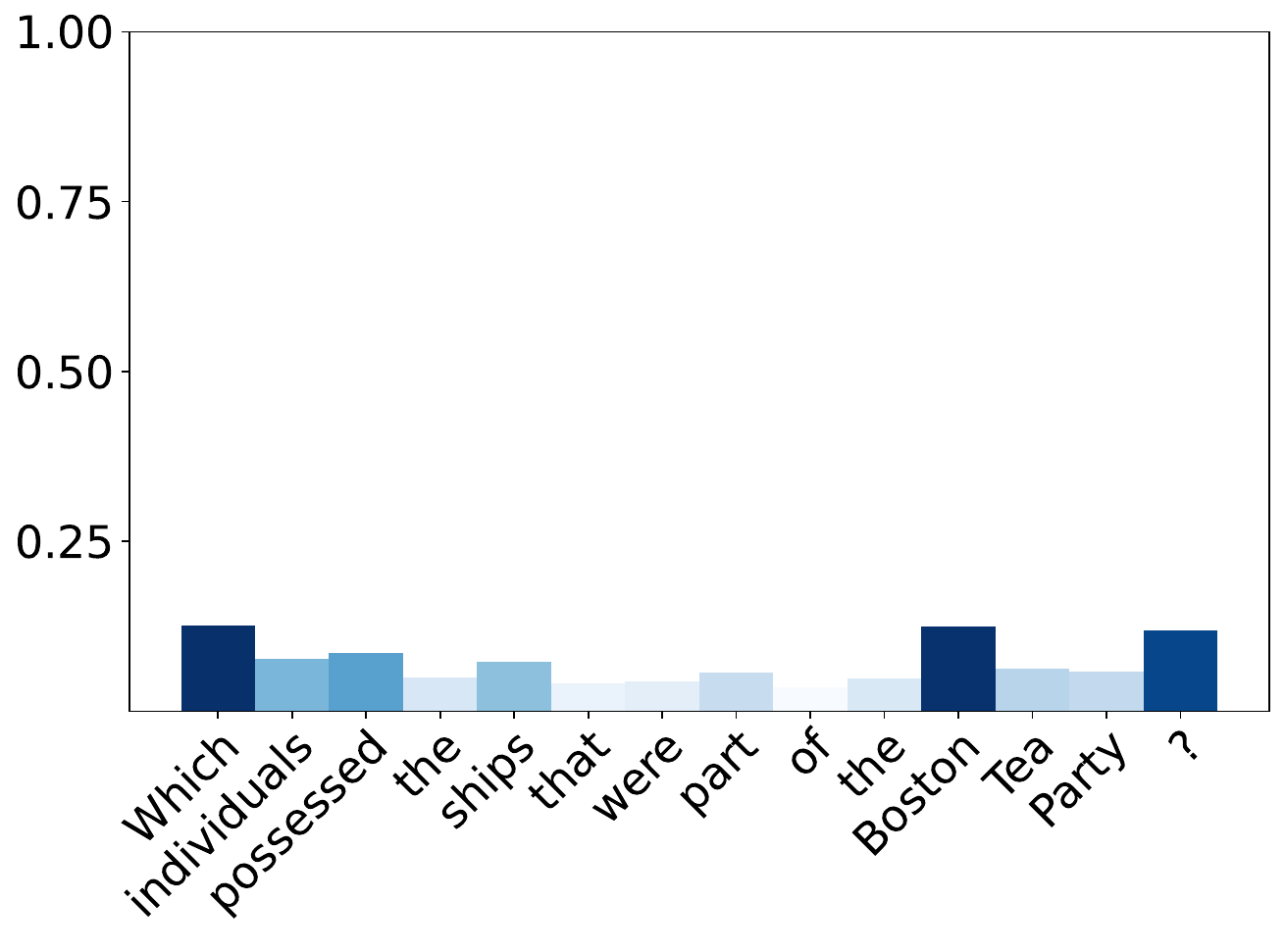}
            \caption[]%
            {{\small Before adding \tframed[line width=0.5bp,fill=vred]{\textcolor{white}{\texttt{\textbf{[PAUSE]}}}} tokens} to original prompt.}
            \label{fig:mean and std of net14}
        \end{subfigure}
        \hfill
        \begin{subfigure}[b]{0.45\textwidth}  
            \centering 
            \includegraphics[width=\textwidth,height=3cm]{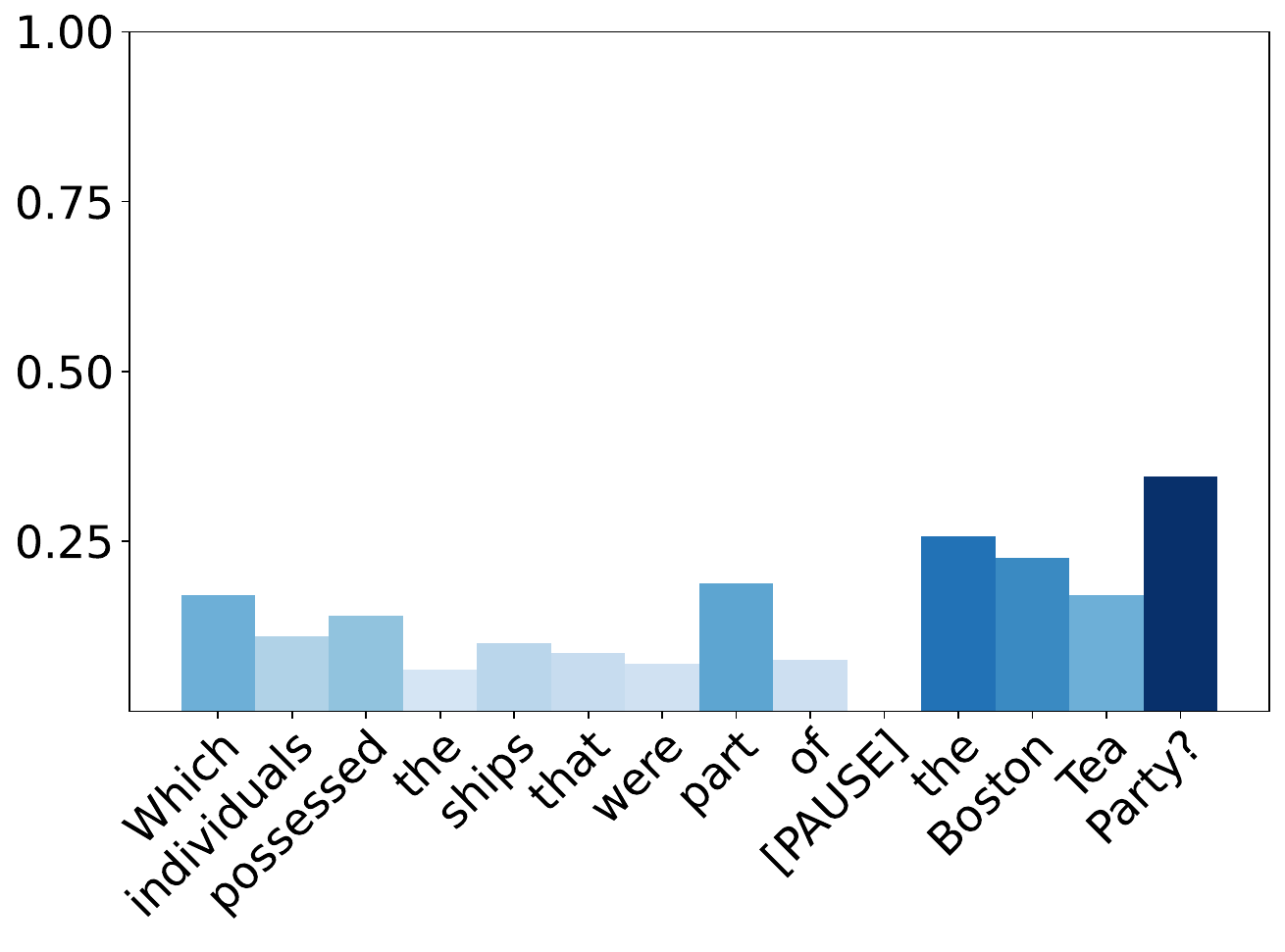}
            \caption[]%
            {{\small After adding \tframed[line width=0.5bp,fill=vred]{\textcolor{white}{\texttt{\textbf{[PAUSE]}}}} tokens} to original prompt.}    
            \label{fig:mean and std of net24}
        \end{subfigure}
        \hfill
        \vskip\baselineskip
        \begin{subfigure}[b]{0.45\textwidth}   
            \centering 
            \includegraphics[width=\textwidth,height=3cm]{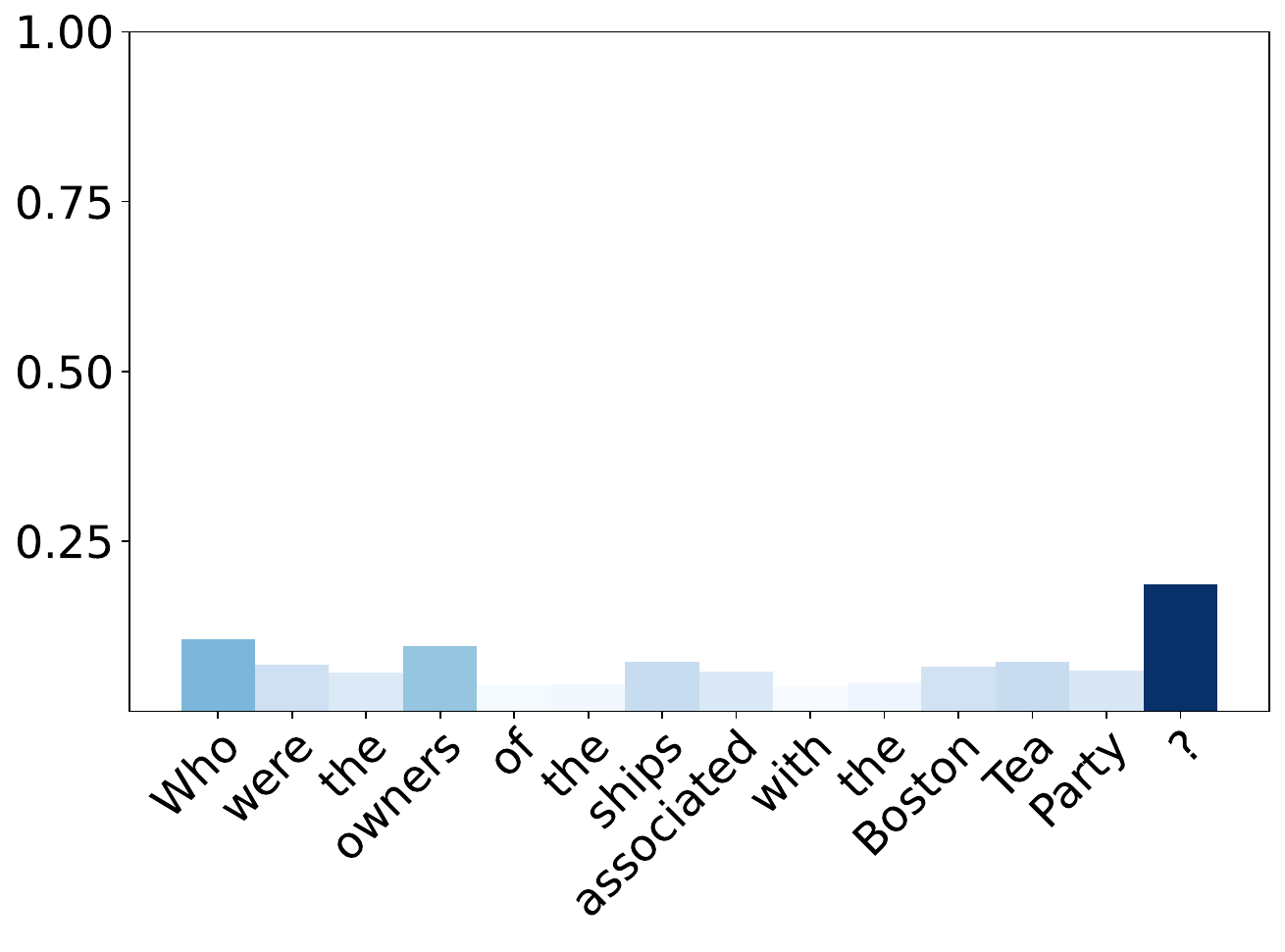}
            \caption[]%
            {{\small Before adding \tframed[line width=0.5bp,fill=vred]{\textcolor{white}{\texttt{\textbf{[PAUSE]}}}} tokens} to paraphrase 1.}    
            \label{fig:mean and std of net34}
        \end{subfigure}
        \hfill
        \begin{subfigure}[b]{0.45\textwidth}   
            \centering 
            \includegraphics[width=\textwidth,height=3cm]{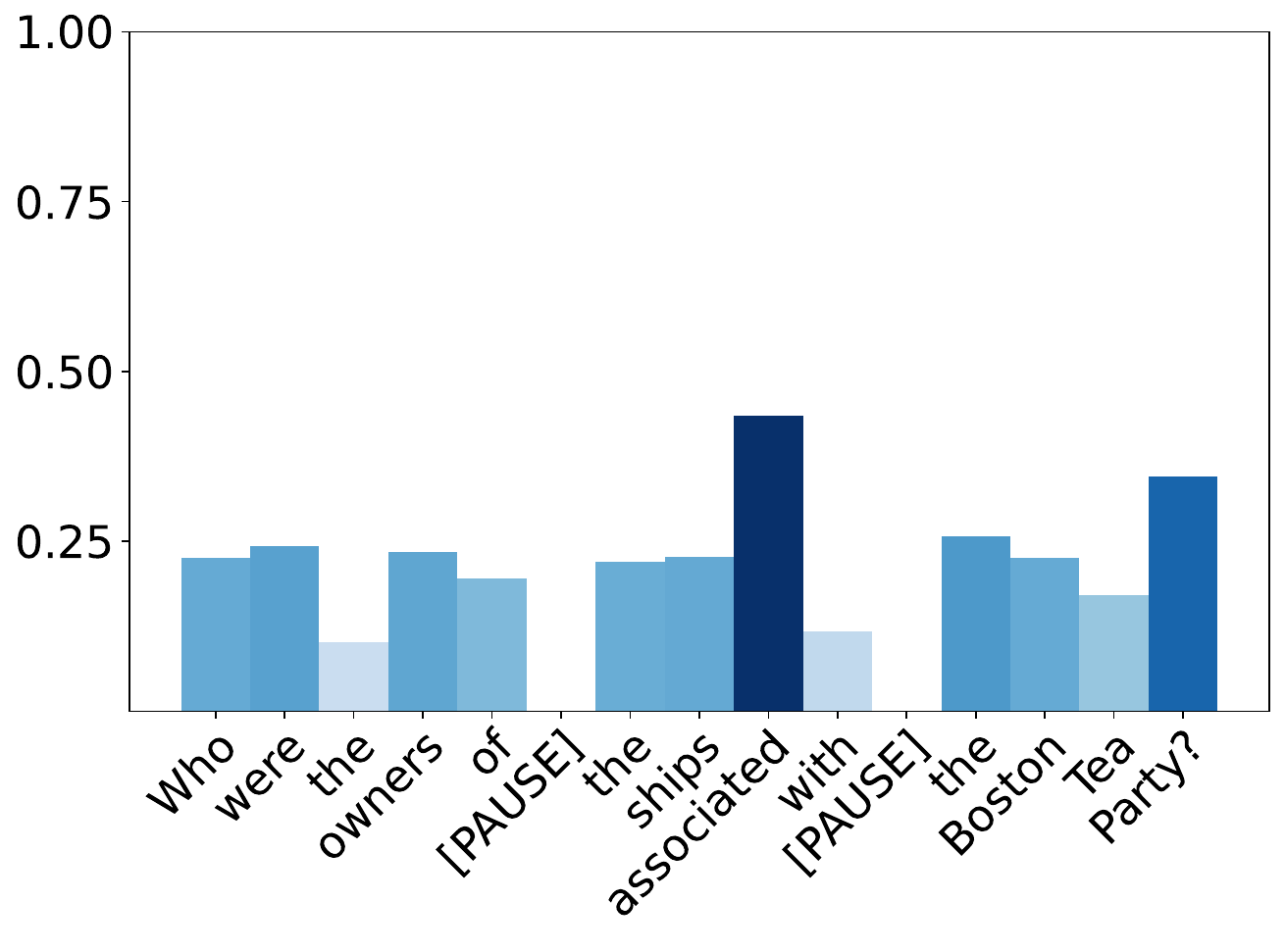}
            \caption[]%
            {{\small After adding \tframed[line width=0.5bp,fill=vred]{\textcolor{white}{\texttt{\textbf{[PAUSE]}}}} tokens} to paraphrase 1.}    
            \label{fig:mean and std of net44}
        \end{subfigure}
        \hfill
        \vskip\baselineskip
        \begin{subfigure}[b]{0.45\textwidth}   
            \centering 
            \includegraphics[width=\textwidth,height=3cm]{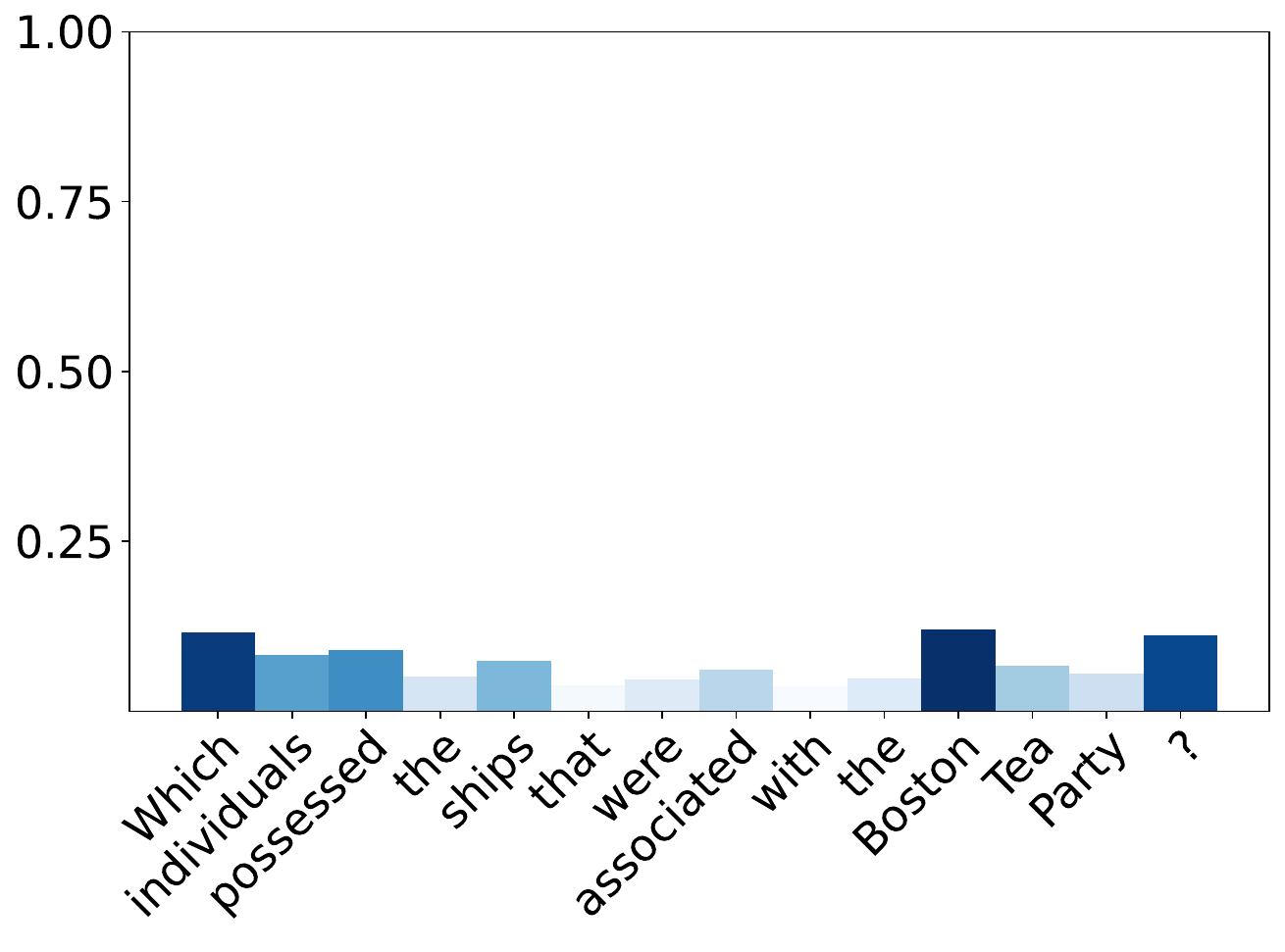}
            \caption[]%
            {{\small Before adding \tframed[line width=0.5bp,fill=vred]{\textcolor{white}{\texttt{\textbf{[PAUSE]}}}} tokens} to paraphrase 2.}
            \label{fig:mean and std of net34}
        \end{subfigure}
        \hfill
        \begin{subfigure}[b]{0.45\textwidth}   
            \centering 
            \includegraphics[width=\textwidth,height=3cm]{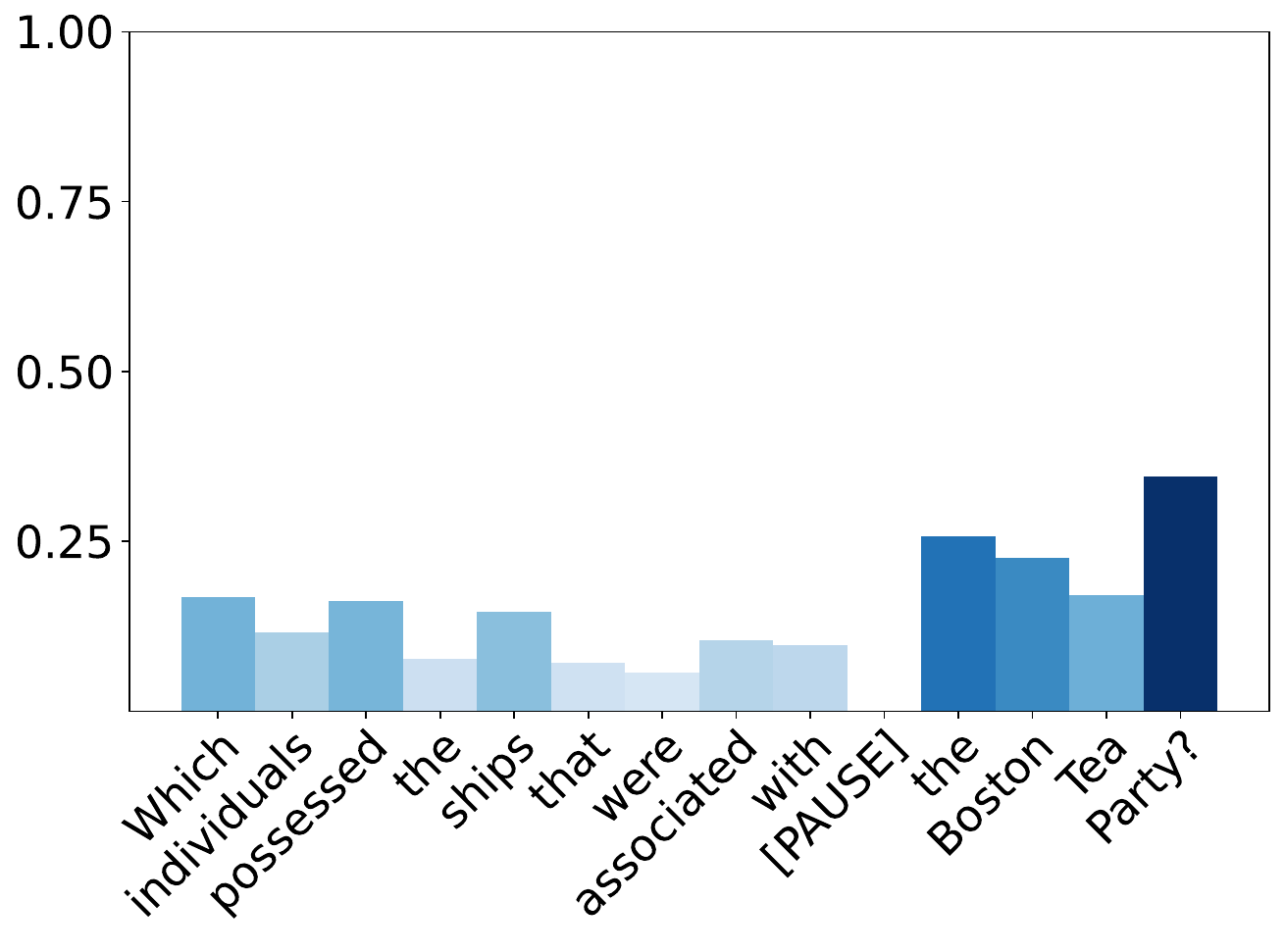}
            \caption[]%
            {{\small After adding \tframed[line width=0.5bp,fill=vred]{\textcolor{white}{\texttt{\textbf{[PAUSE]}}}} tokens} to paraphrase 2.} 
            \label{fig:mean and std of net44}
        \end{subfigure}
        \hfill
        \vskip\baselineskip
        \begin{subfigure}[b]{0.45\textwidth}   
            \centering 
            \includegraphics[width=\textwidth,height=3cm]{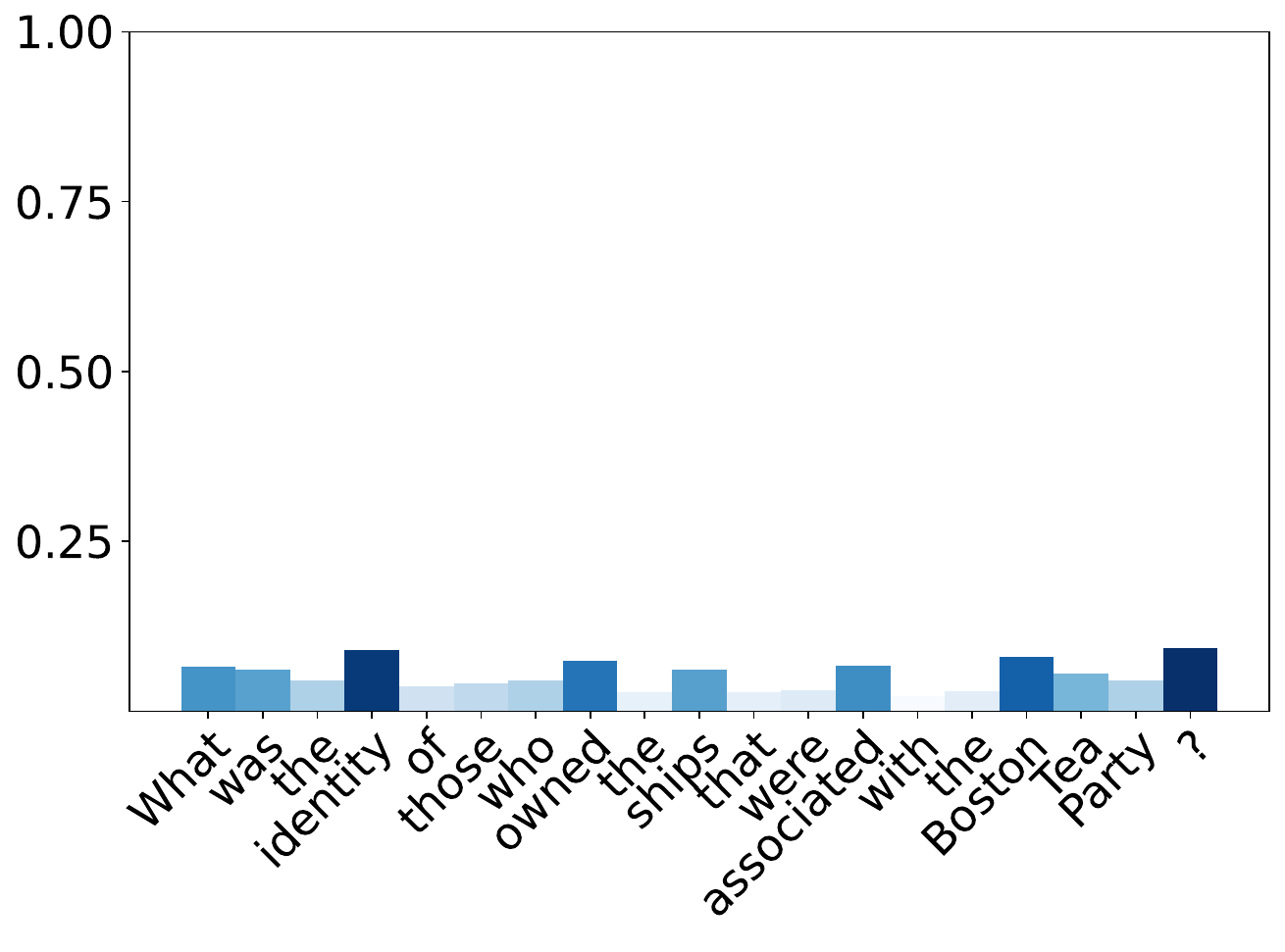}
            \caption[]%
            {{\small Before adding \tframed[line width=0.5bp,fill=vred]{\textcolor{white}{\texttt{\textbf{[PAUSE]}}}} tokens} to paraphrase 3.}
            \label{fig:mean and std of net44}
        \end{subfigure}
        \hfill
        \begin{subfigure}[b]{0.45\textwidth}   
            \centering 
            \includegraphics[width=\textwidth,height=3cm]{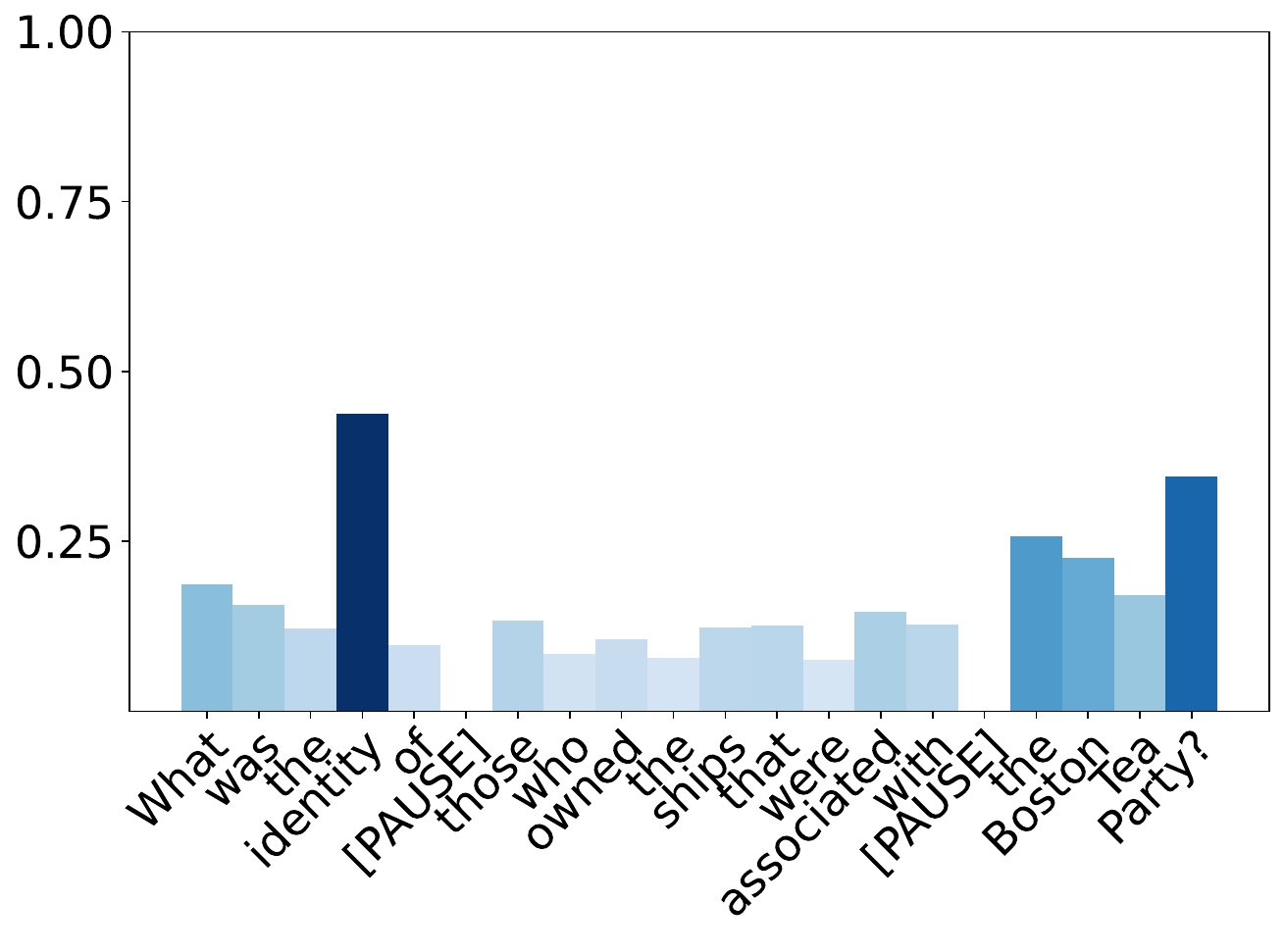}
            \caption[]%
            {{\small After adding \tframed[line width=0.5bp,fill=vred]{\textcolor{white}{\texttt{\textbf{[PAUSE]}}}} tokens} to paraphrase 3.}    
            \label{fig:mean and std of net44}
        \end{subfigure}
        \hfill
        \vskip\baselineskip
        \begin{subfigure}[b]{0.45\textwidth}   
            \centering 
            \includegraphics[width=\textwidth,height=3cm]{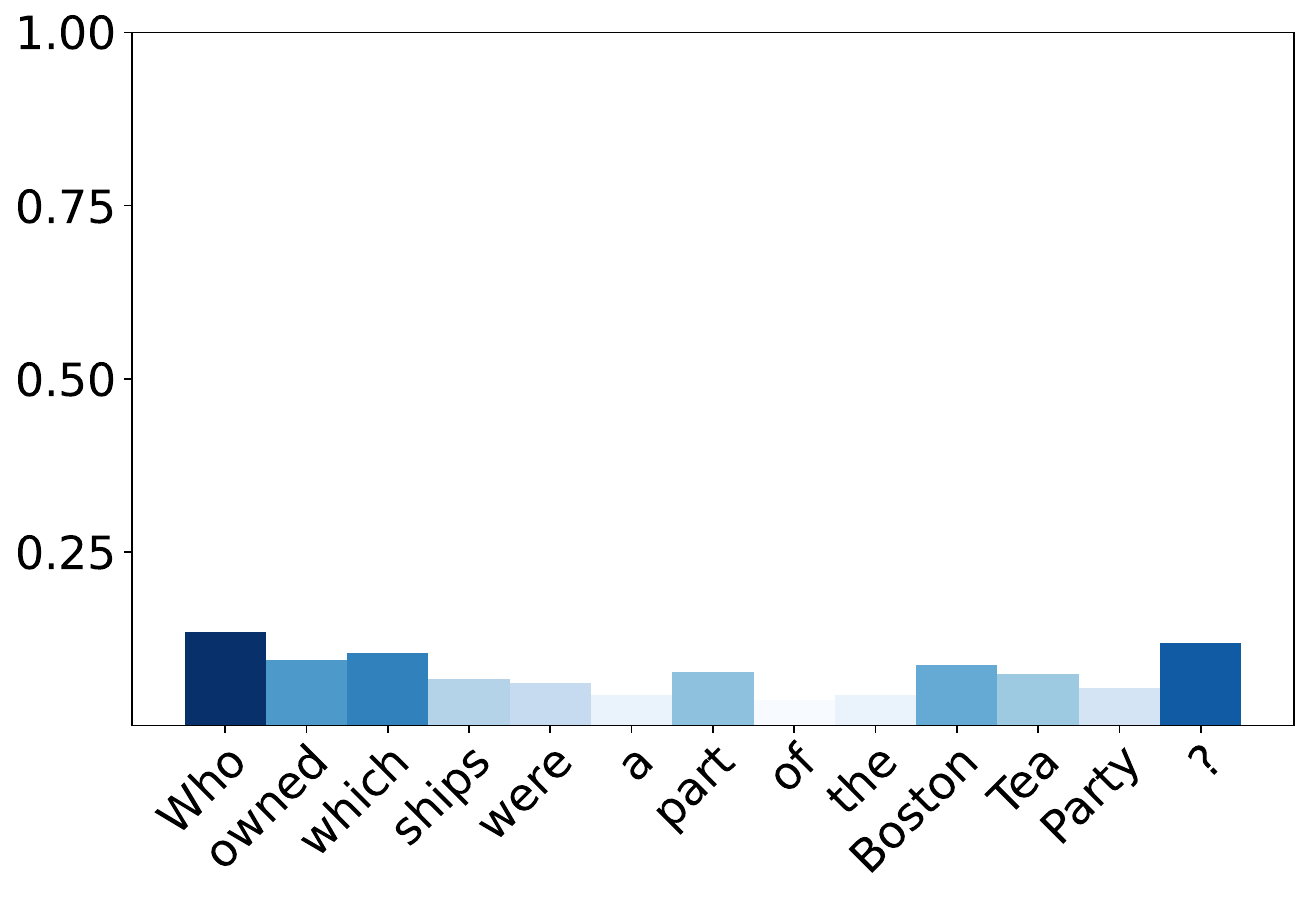}
            \caption[]%
            {{\small Before adding \tframed[line width=0.5bp,fill=vred]{\textcolor{white}{\texttt{\textbf{[PAUSE]}}}} tokens} to paraphrase 4.}    
            \label{fig:mean and std of net44}
        \end{subfigure}
        \hfill
        \begin{subfigure}[b]{0.45\textwidth}   
            \centering 
            \includegraphics[width=\textwidth,height=3cm]{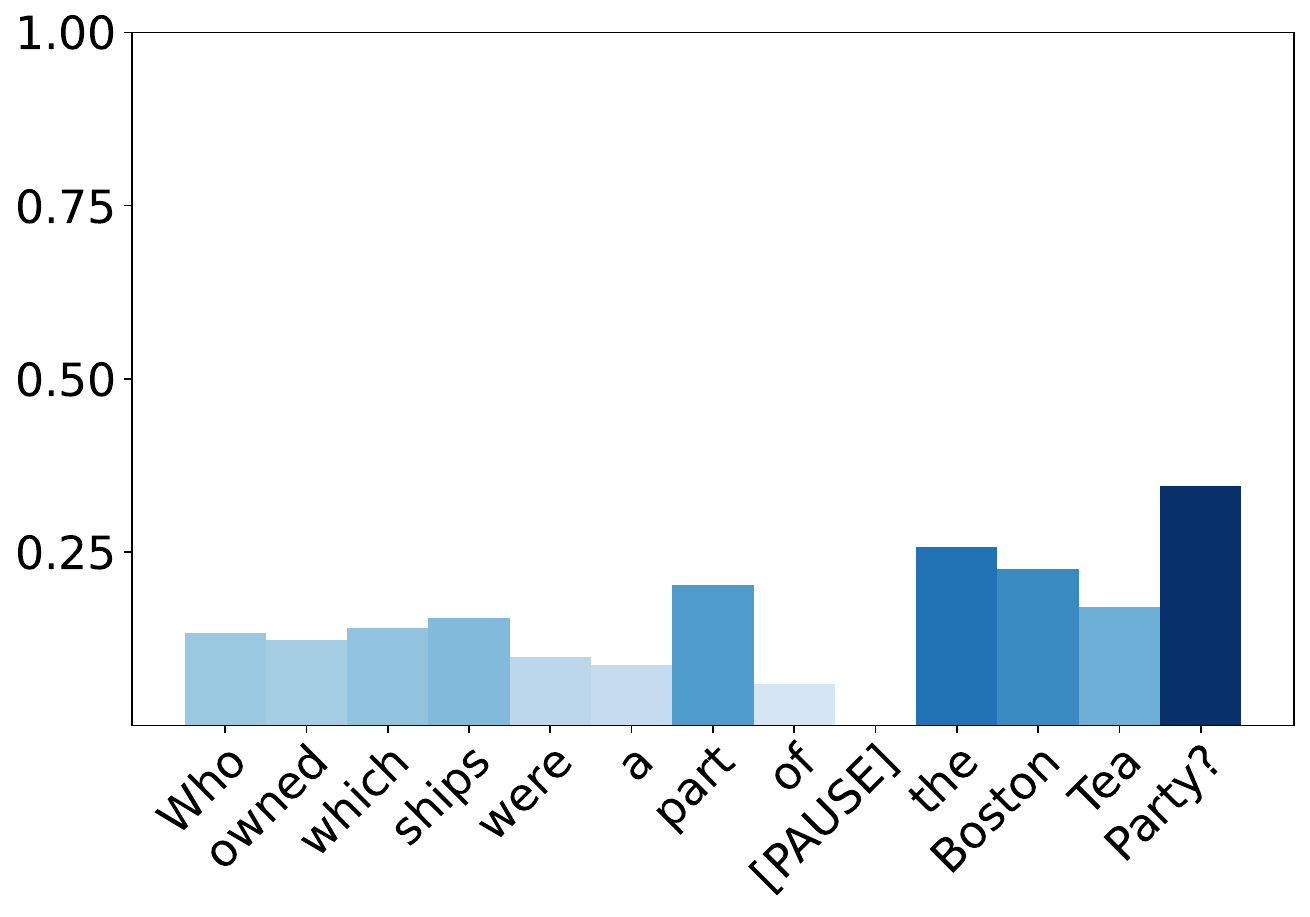}
            \caption[]%
            {{\small After adding \tframed[line width=0.5bp,fill=vred]{\textcolor{white}{\texttt{\textbf{[PAUSE]}}}} tokens} to paraphrase 4.}    
            \label{fig:mean and std of net44}
        \end{subfigure}
        \hfill
        \vskip\baselineskip
        \begin{subfigure}[b]{0.45\textwidth}   
            \centering 
            \includegraphics[width=\textwidth,height=3cm]{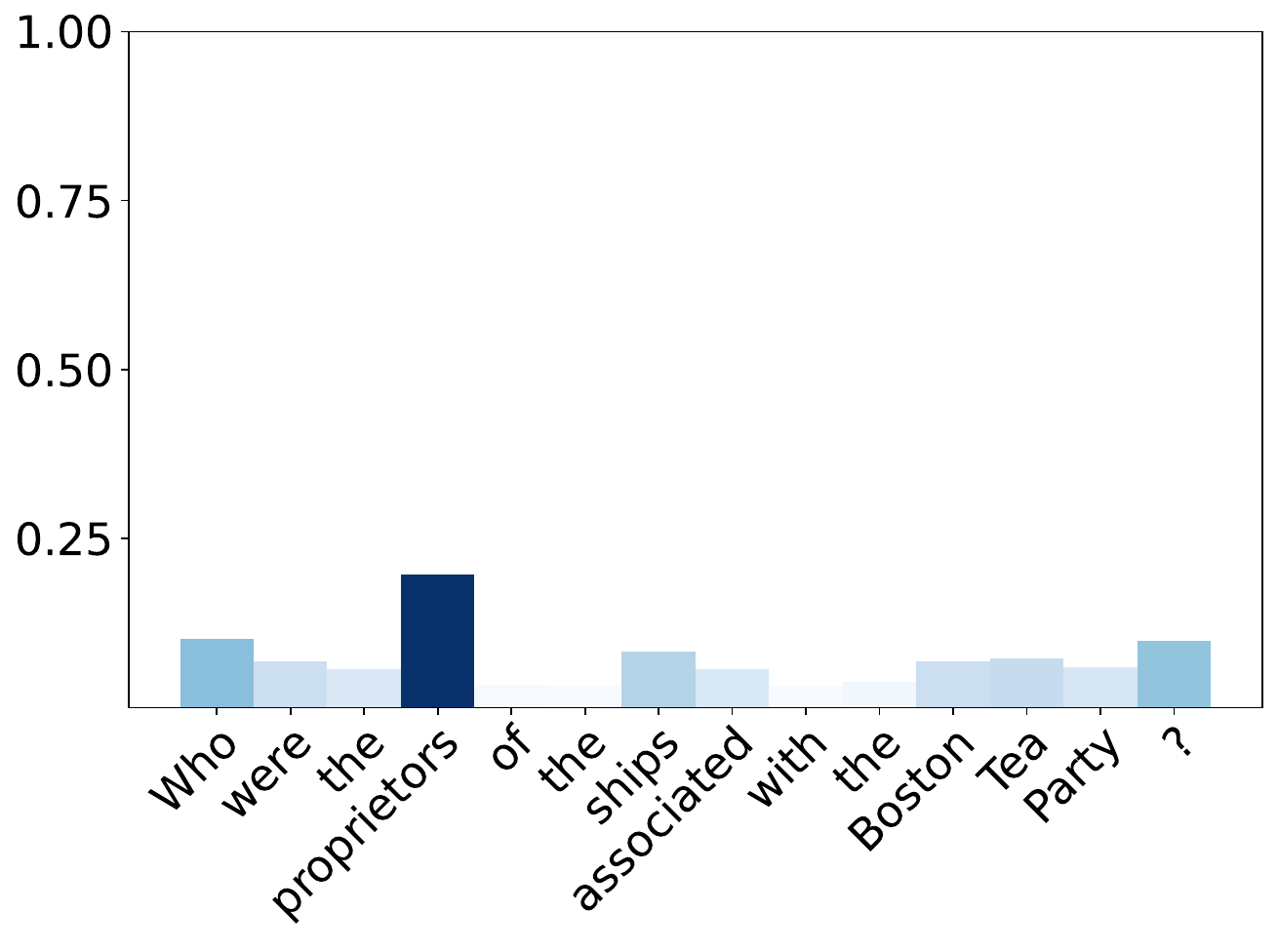}
            \caption[]%
            {{\small Before adding \tframed[line width=0.5bp,fill=vred]{\textcolor{white}{\texttt{\textbf{[PAUSE]}}}} tokens} to paraphrase 5.}    
            \label{fig:mean and std of net44}
        \end{subfigure}
        \hfill
        \begin{subfigure}[b]{0.45\textwidth}   
            \centering 
            \includegraphics[width=\textwidth,height=3cm]{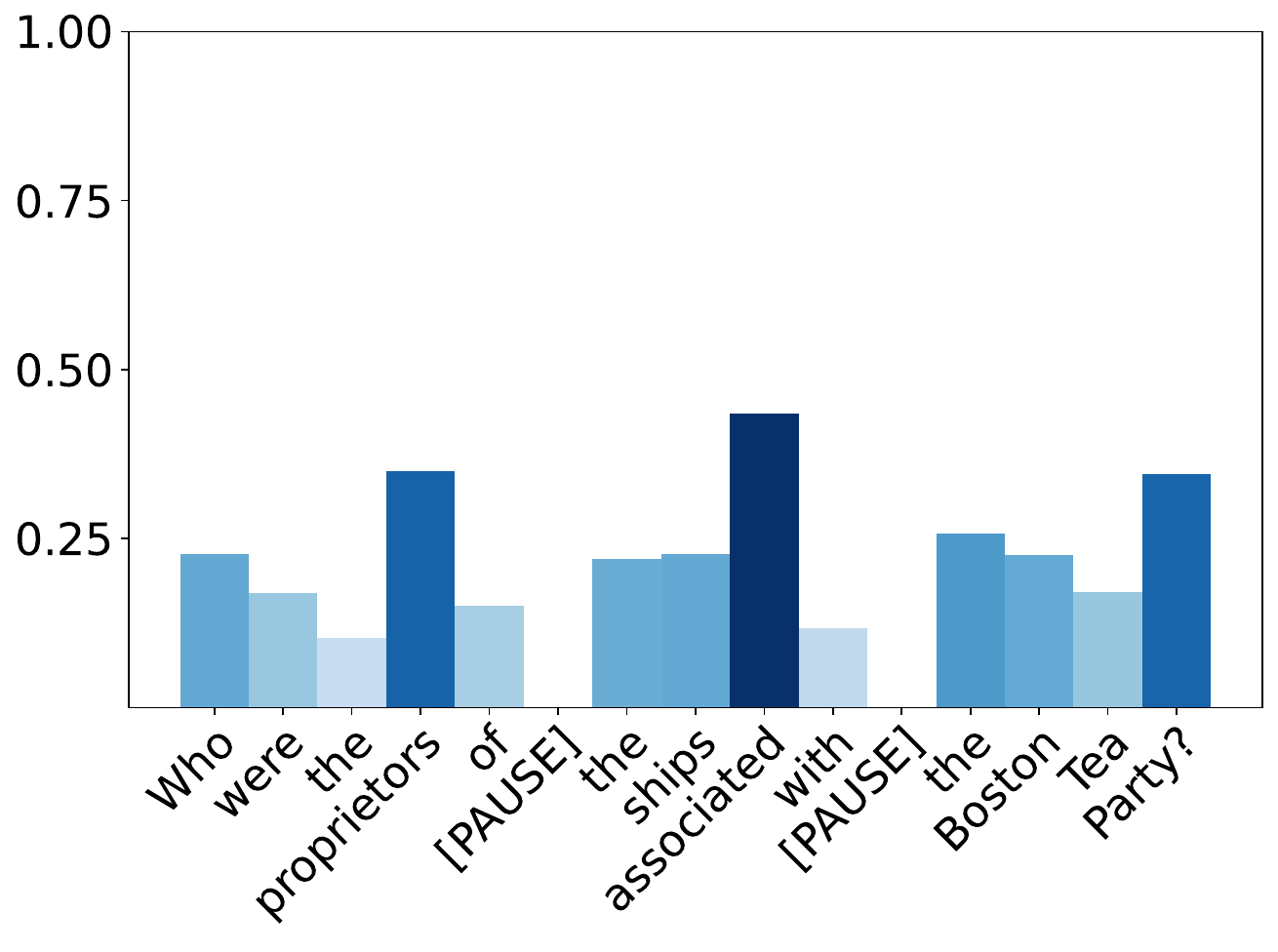}
            \caption[]%
            {{\small After adding \tframed[line width=0.5bp,fill=vred]{\textcolor{white}{\texttt{\textbf{[PAUSE]}}}} tokens} to paraphrase 5.}    
            \label{fig:mean and std of net44}
        \end{subfigure}
        \caption[]%
            {{\small The phrase \textbf{Boston Tea} gets more importance score after adding \tframed[line width=0.5bp,fill=vred]{\textcolor{white}{\texttt{\textbf{[PAUSE]}}}} token for Falcon.}}   
        \label{fig:Falcon}
\end{figure*}

\begin{figure*}[!ht]
        \centering
        \begin{subfigure}[b]{0.45\textwidth}
            \centering
            \includegraphics[width=\textwidth,height=3cm]{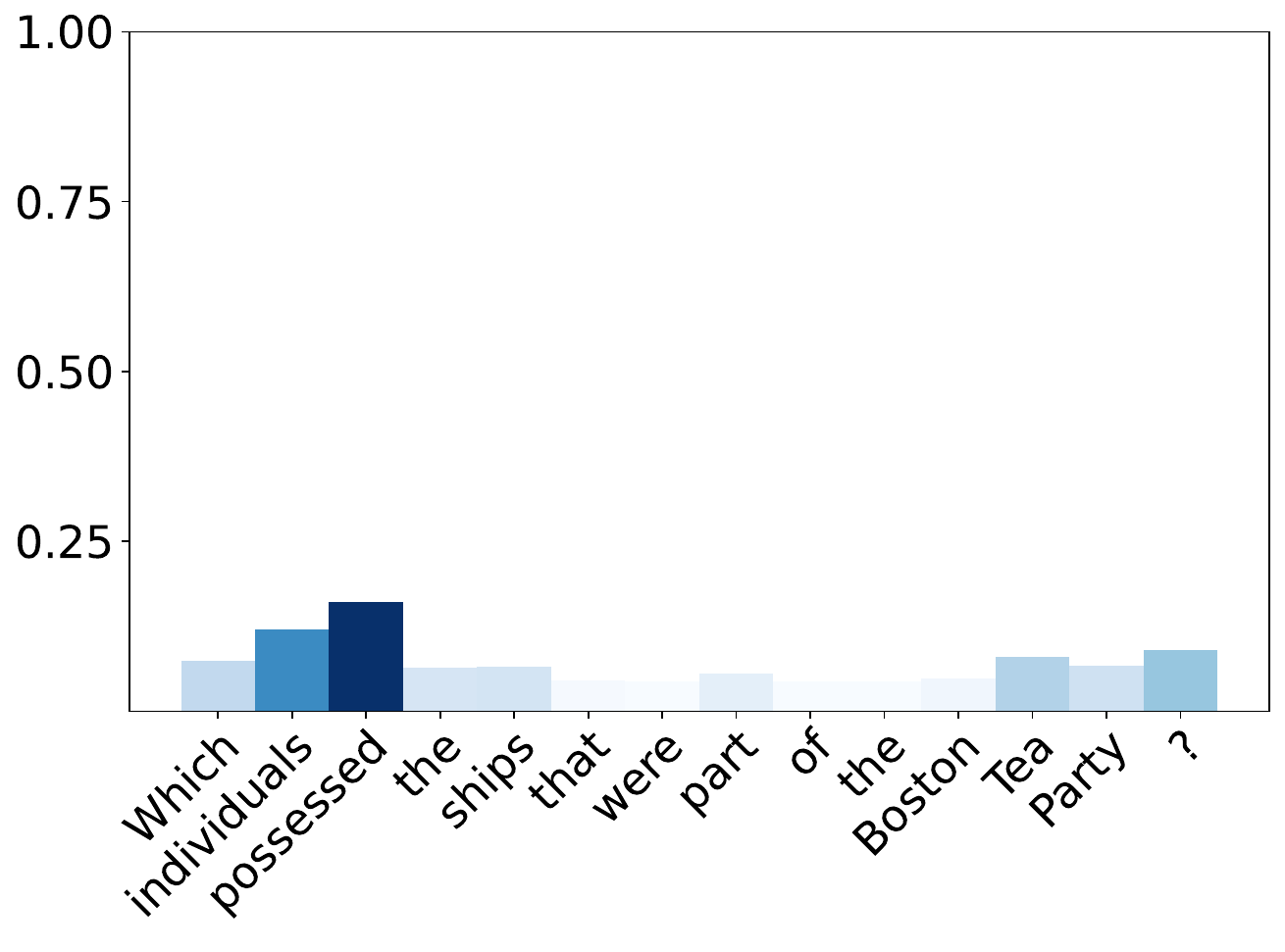}
            \caption[]%
            {{\small Before adding \tframed[line width=0.5bp,fill=vred]{\textcolor{white}{\texttt{\textbf{[PAUSE]}}}} tokens} to original prompt.}
            \label{fig:mean and std of net14}
        \end{subfigure}
        \hfill
        \begin{subfigure}[b]{0.45\textwidth}  
            \centering 
            \includegraphics[width=\textwidth,height=3cm]{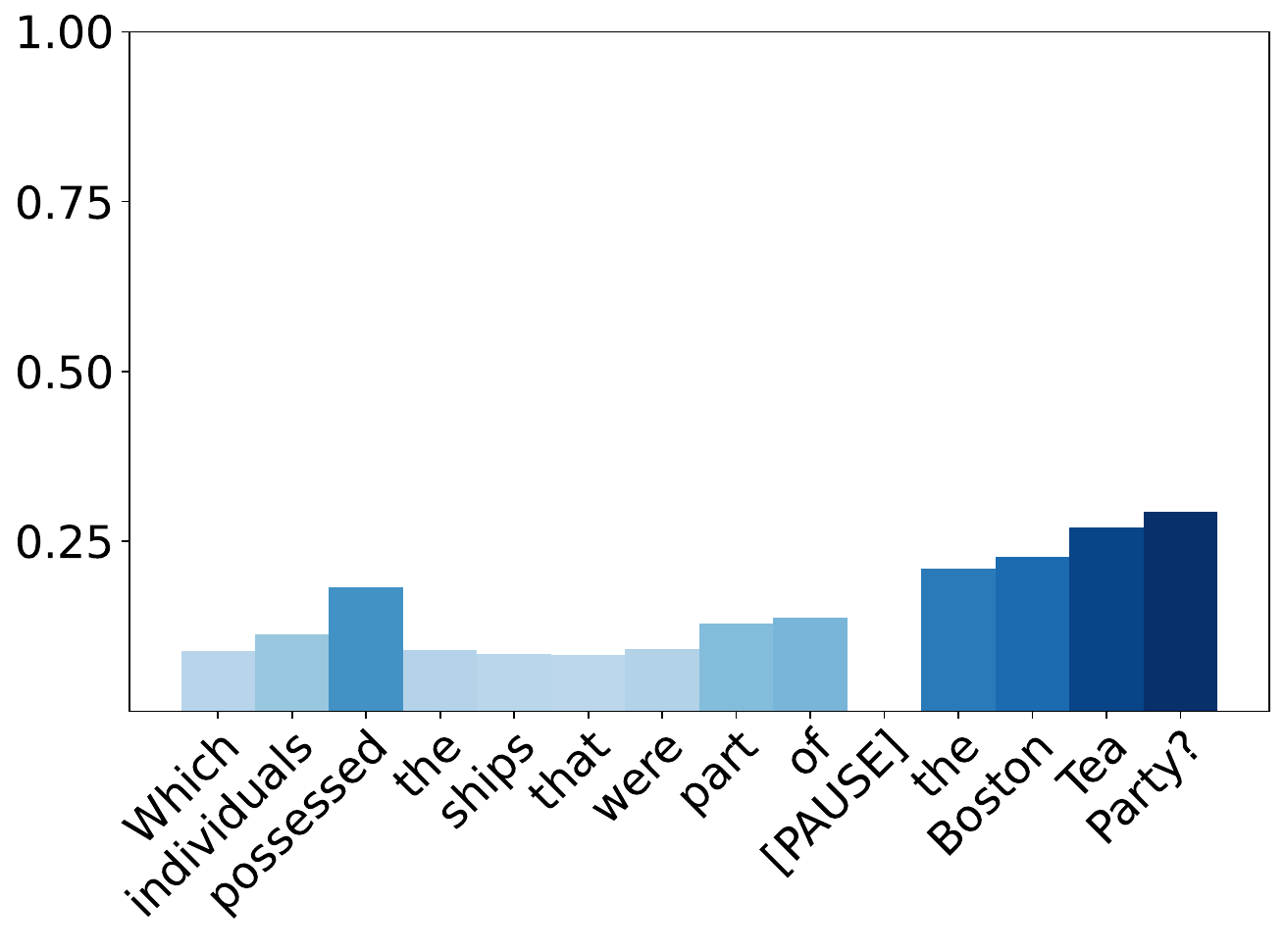}
            \caption[]%
            {{\small After adding \tframed[line width=0.5bp,fill=vred]{\textcolor{white}{\texttt{\textbf{[PAUSE]}}}} tokens} to original prompt.}    
            \label{fig:mean and std of net24}
        \end{subfigure}
        \hfill
        \vskip\baselineskip
        \begin{subfigure}[b]{0.45\textwidth}   
            \centering 
            \includegraphics[width=\textwidth,height=3cm]{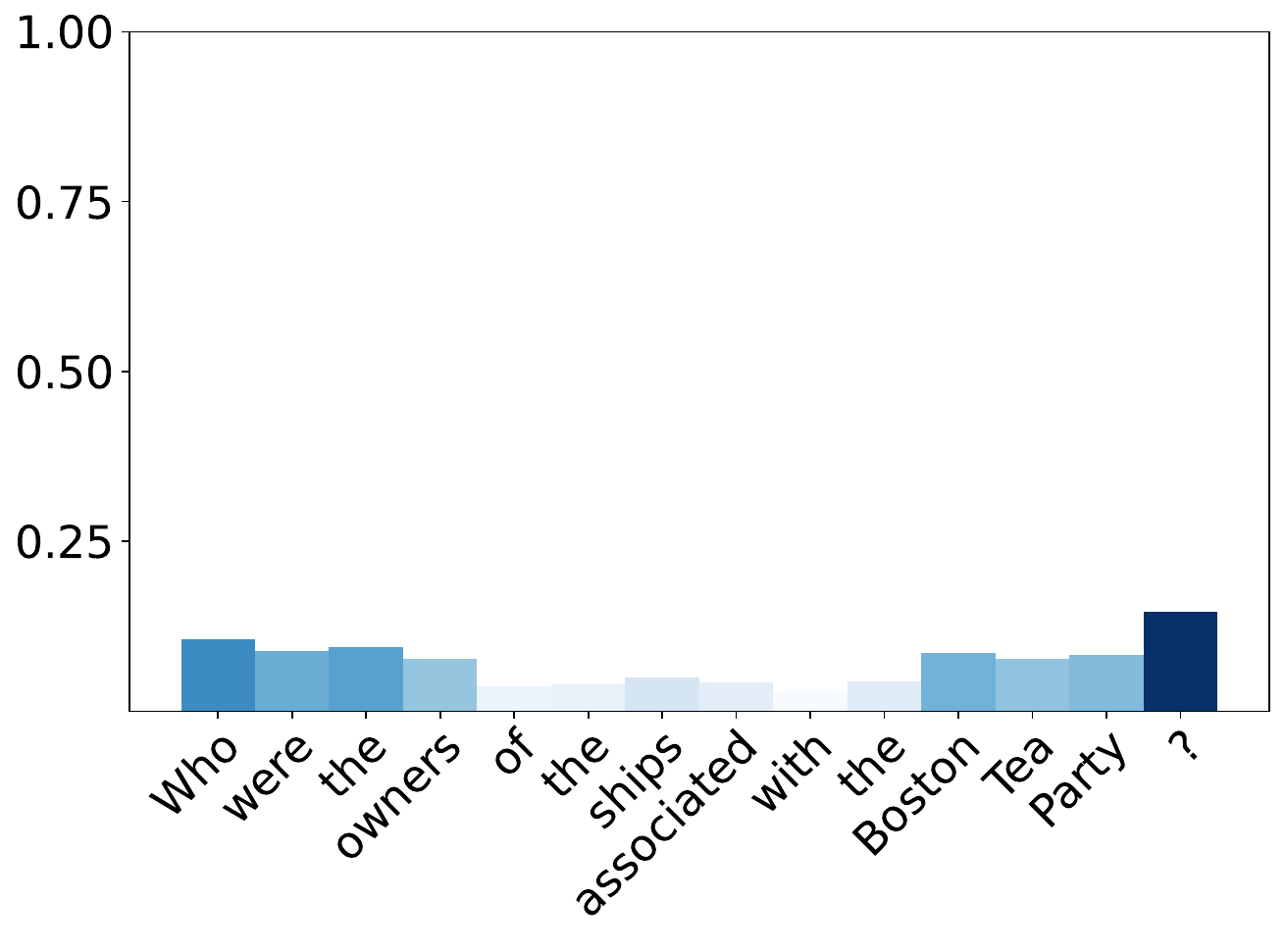}
            \caption[]%
            {{\small Before adding \tframed[line width=0.5bp,fill=vred]{\textcolor{white}{\texttt{\textbf{[PAUSE]}}}} tokens} to paraphrase 1.}    
            \label{fig:mean and std of net34}
        \end{subfigure}
        \hfill
        \begin{subfigure}[b]{0.45\textwidth}   
            \centering 
            \includegraphics[width=\textwidth,height=3cm]{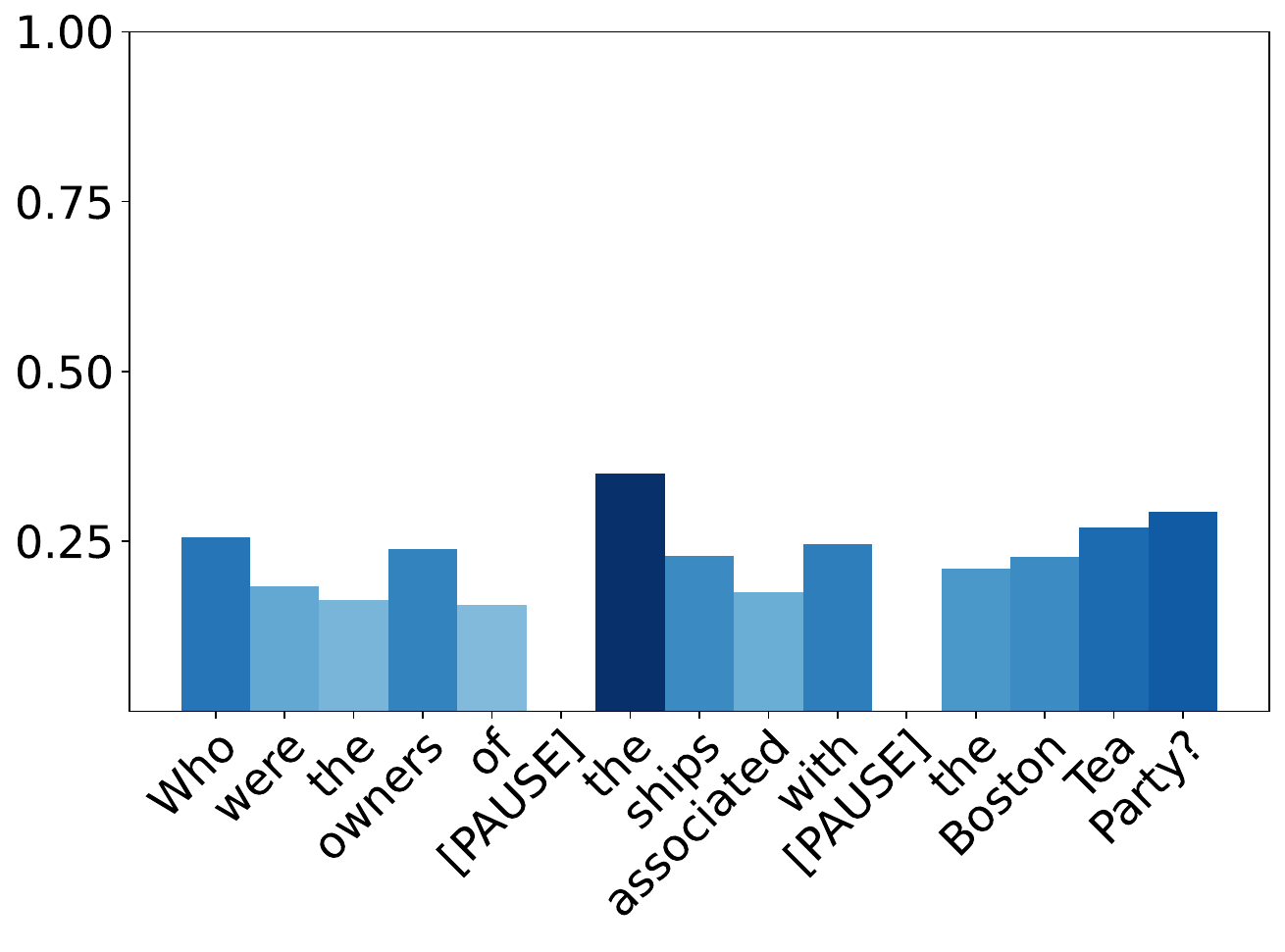}
            \caption[]%
            {{\small After adding \tframed[line width=0.5bp,fill=vred]{\textcolor{white}{\texttt{\textbf{[PAUSE]}}}} tokens} to paraphrase 1.}    
            \label{fig:mean and std of net44}
        \end{subfigure}
        \hfill
        \vskip\baselineskip
        \begin{subfigure}[b]{0.45\textwidth}   
            \centering 
            \includegraphics[width=\textwidth,height=3cm]{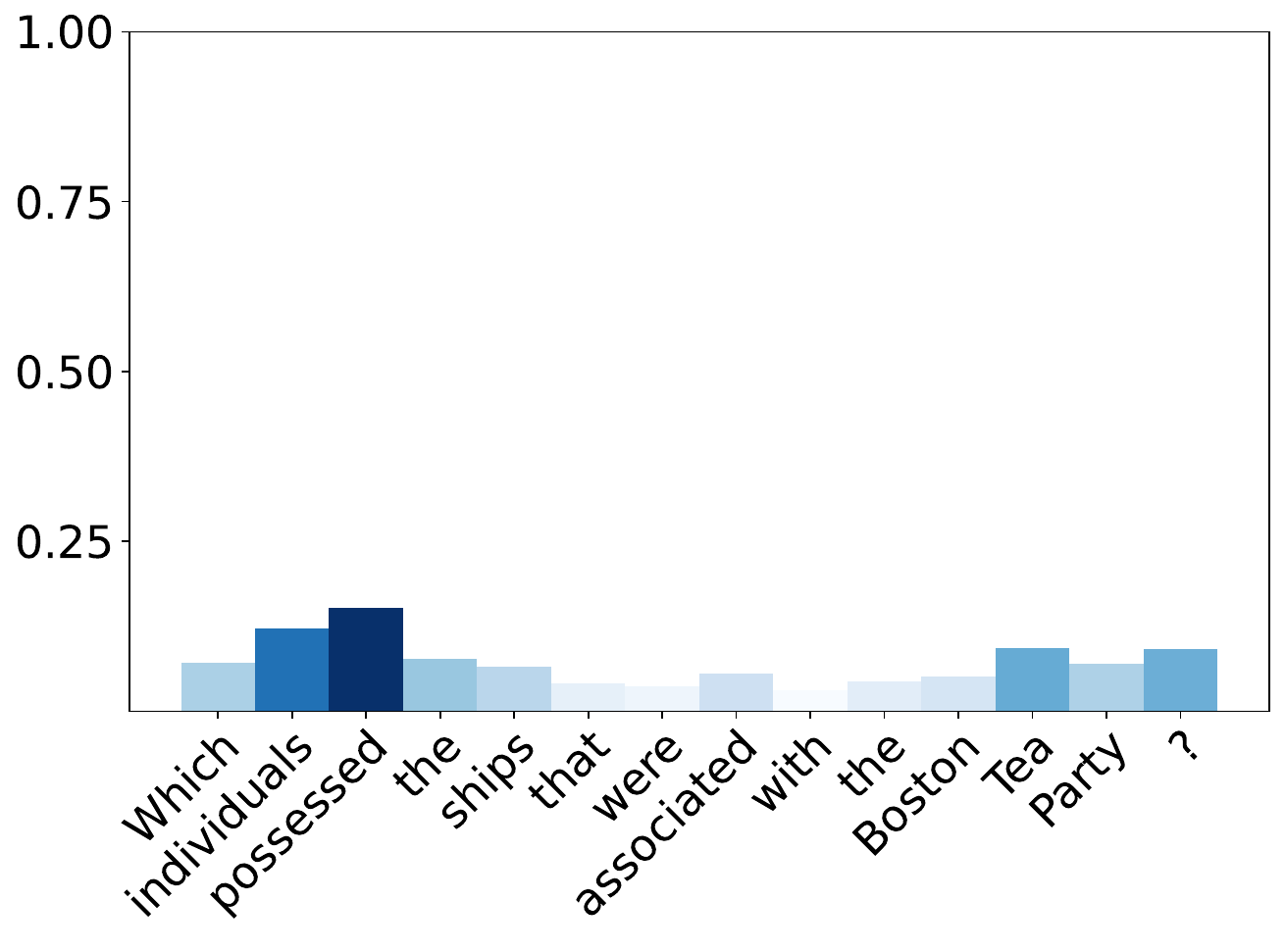}
            \caption[]%
            {{\small Before adding \tframed[line width=0.5bp,fill=vred]{\textcolor{white}{\texttt{\textbf{[PAUSE]}}}} tokens} to paraphrase 2.}
            \label{fig:mean and std of net34}
        \end{subfigure}
        \hfill
        \begin{subfigure}[b]{0.45\textwidth}   
            \centering 
            \includegraphics[width=\textwidth,height=3cm]{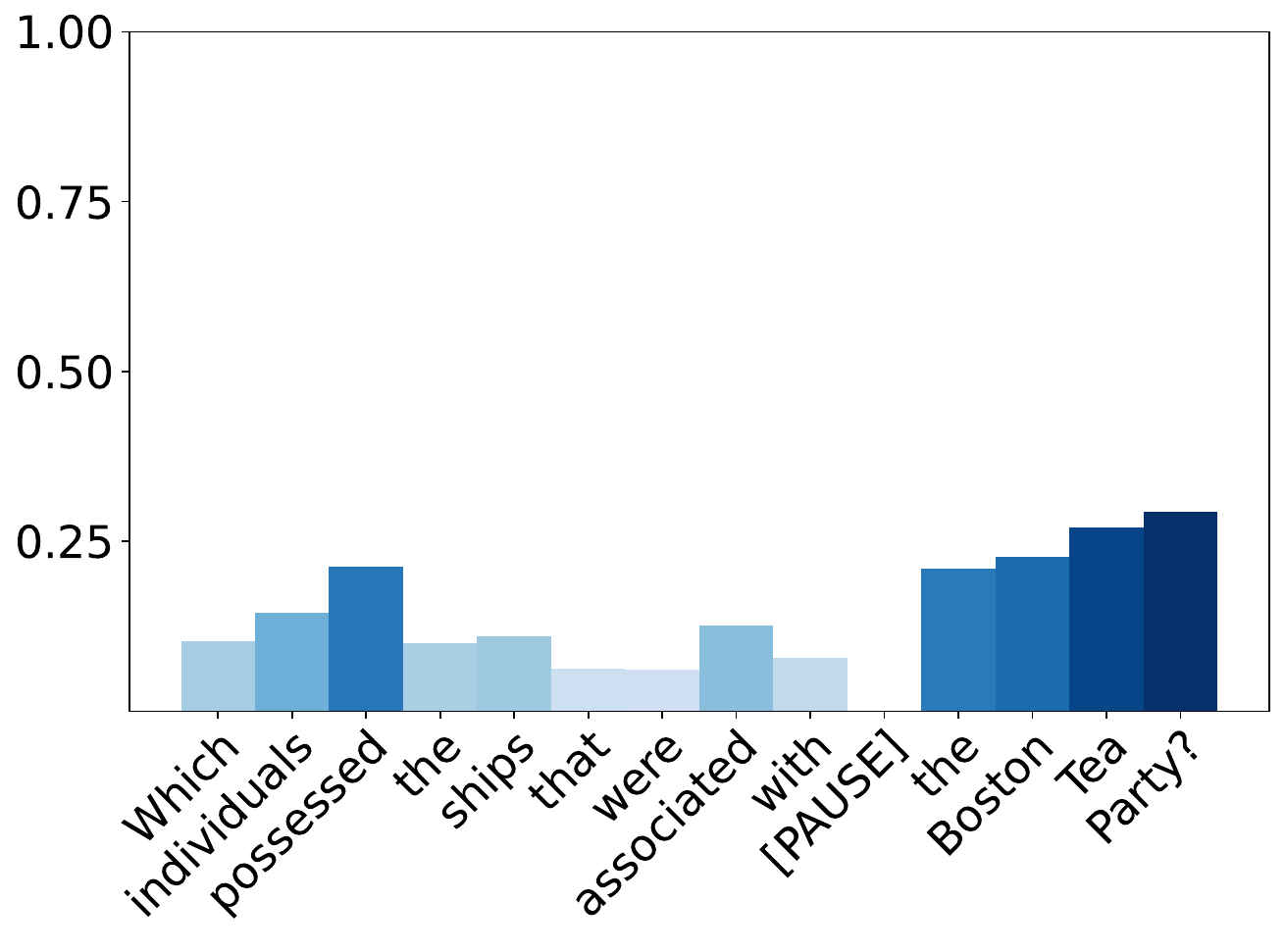}
            \caption[]%
            {{\small After adding \tframed[line width=0.5bp,fill=vred]{\textcolor{white}{\texttt{\textbf{[PAUSE]}}}} tokens} to paraphrase 2.} 
            \label{fig:mean and std of net44}
        \end{subfigure}
        \hfill
        \vskip\baselineskip
        \begin{subfigure}[b]{0.45\textwidth}   
            \centering 
            \includegraphics[width=\textwidth,height=3cm]{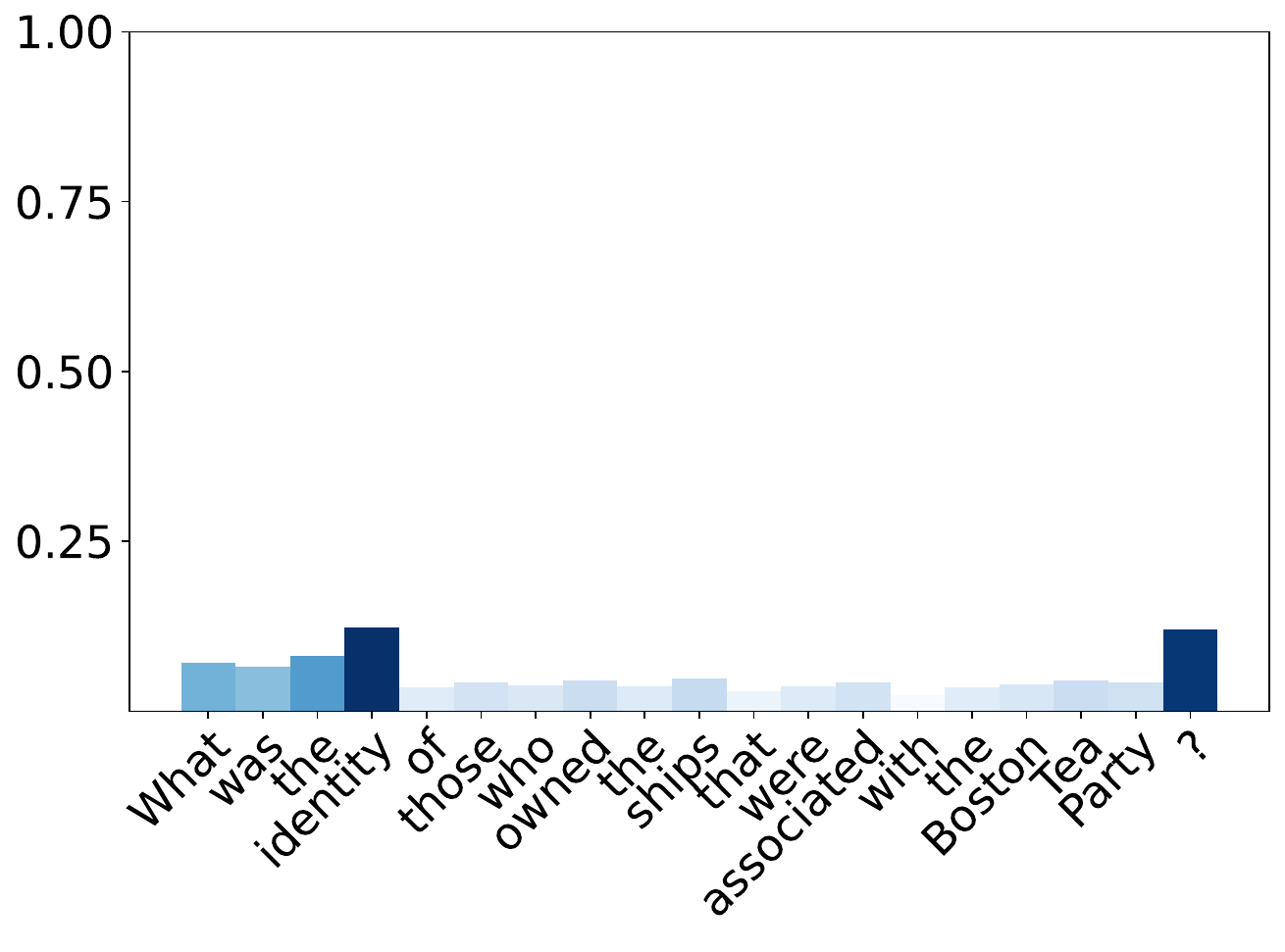}
            \caption[]%
            {{\small Before adding \tframed[line width=0.5bp,fill=vred]{\textcolor{white}{\texttt{\textbf{[PAUSE]}}}} tokens} to paraphrase 3.}
            \label{fig:mean and std of net44}
        \end{subfigure}
        \hfill
        \begin{subfigure}[b]{0.45\textwidth}   
            \centering 
            \includegraphics[width=\textwidth,height=3cm]{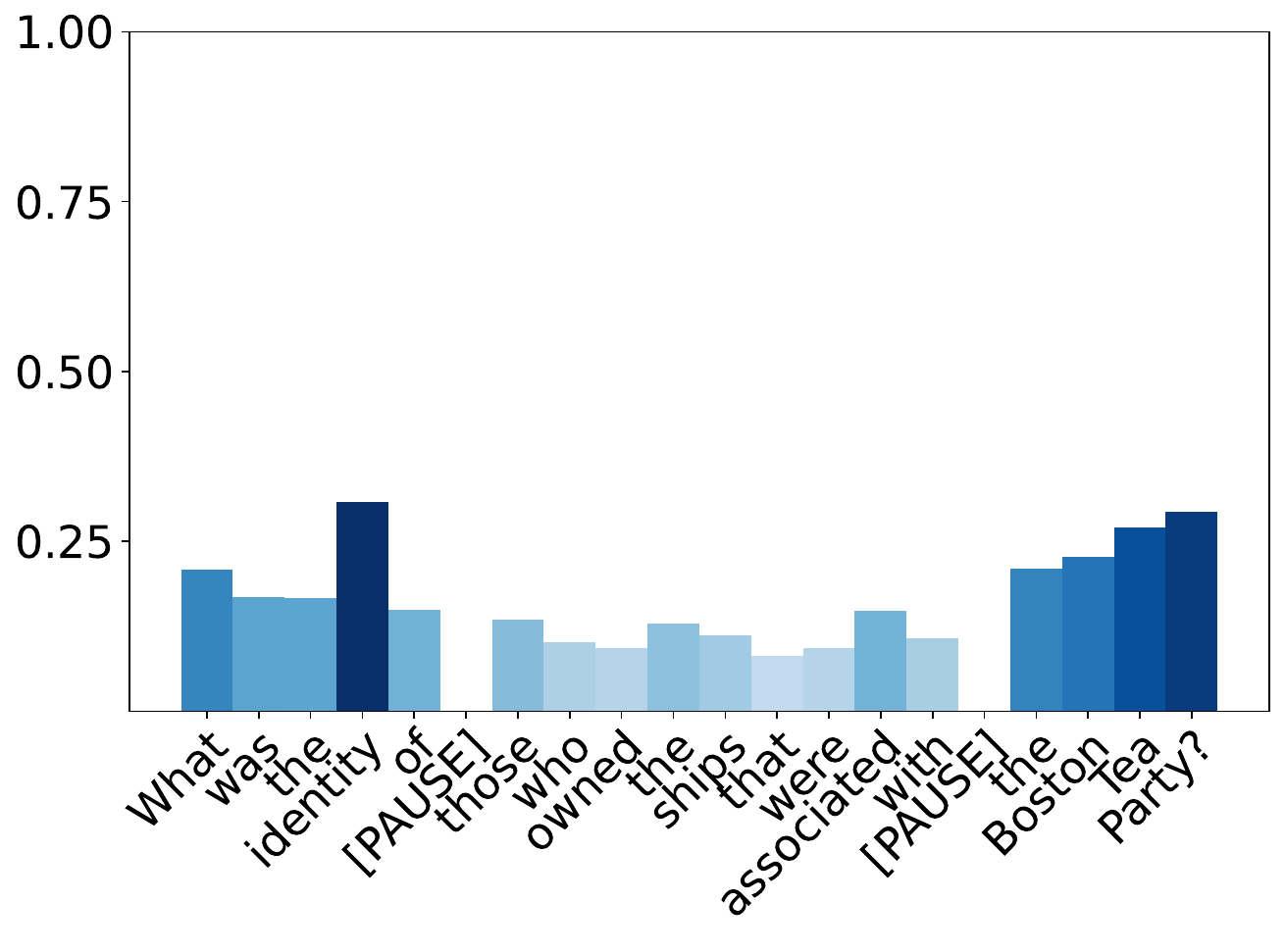}
            \caption[]%
            {{\small After adding \tframed[line width=0.5bp,fill=vred]{\textcolor{white}{\texttt{\textbf{[PAUSE]}}}} tokens} to paraphrase 3.}    
            \label{fig:mean and std of net44}
        \end{subfigure}
        \hfill
        \vskip\baselineskip
        \begin{subfigure}[b]{0.45\textwidth}   
            \centering 
            \includegraphics[width=\textwidth,height=3cm]{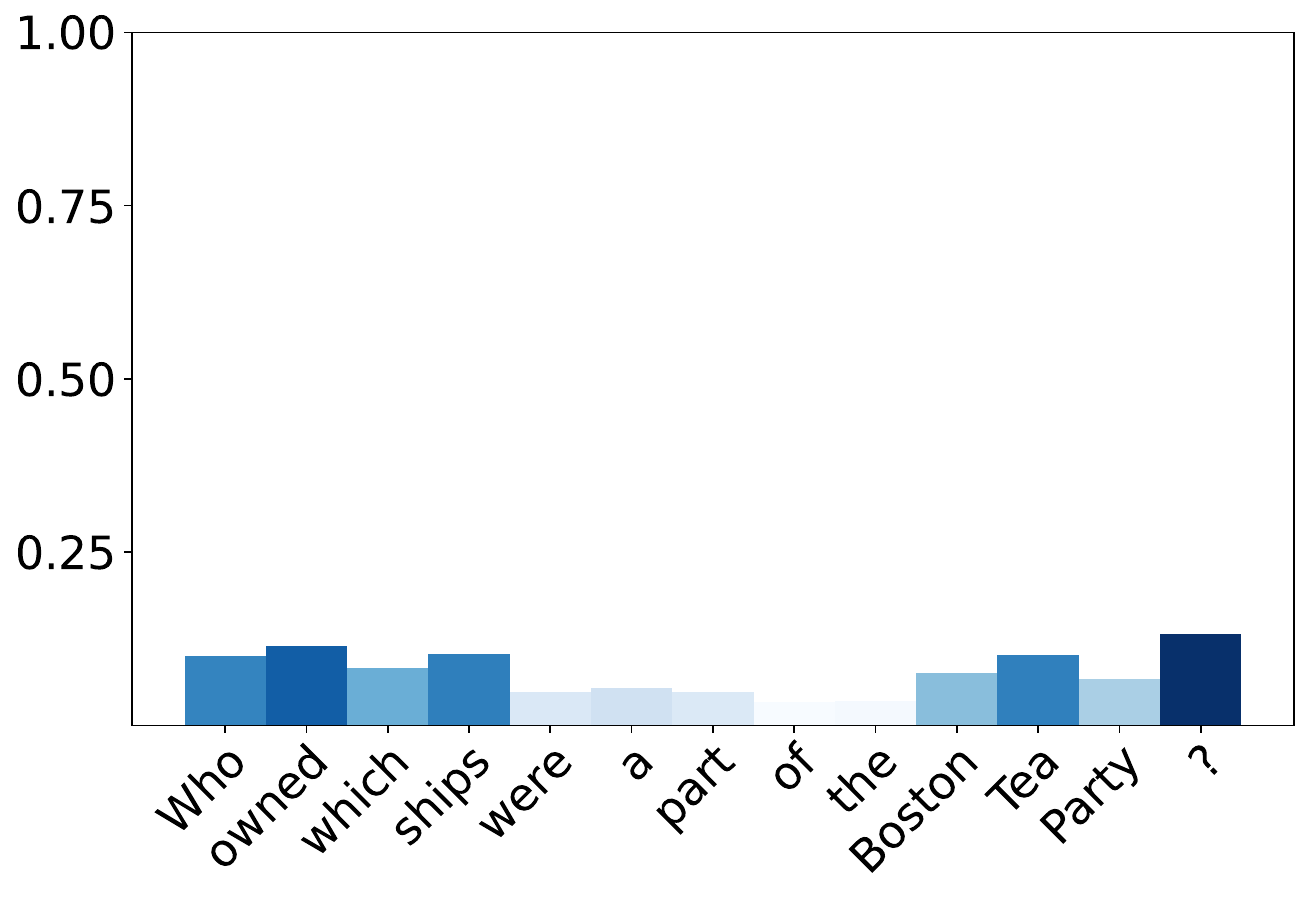}
            \caption[]%
            {{\small Before adding \tframed[line width=0.5bp,fill=vred]{\textcolor{white}{\texttt{\textbf{[PAUSE]}}}} tokens} to paraphrase 4.}    
            \label{fig:mean and std of net44}
        \end{subfigure}
        \hfill
        \begin{subfigure}[b]{0.45\textwidth}   
            \centering 
            \includegraphics[width=\textwidth,height=3cm]{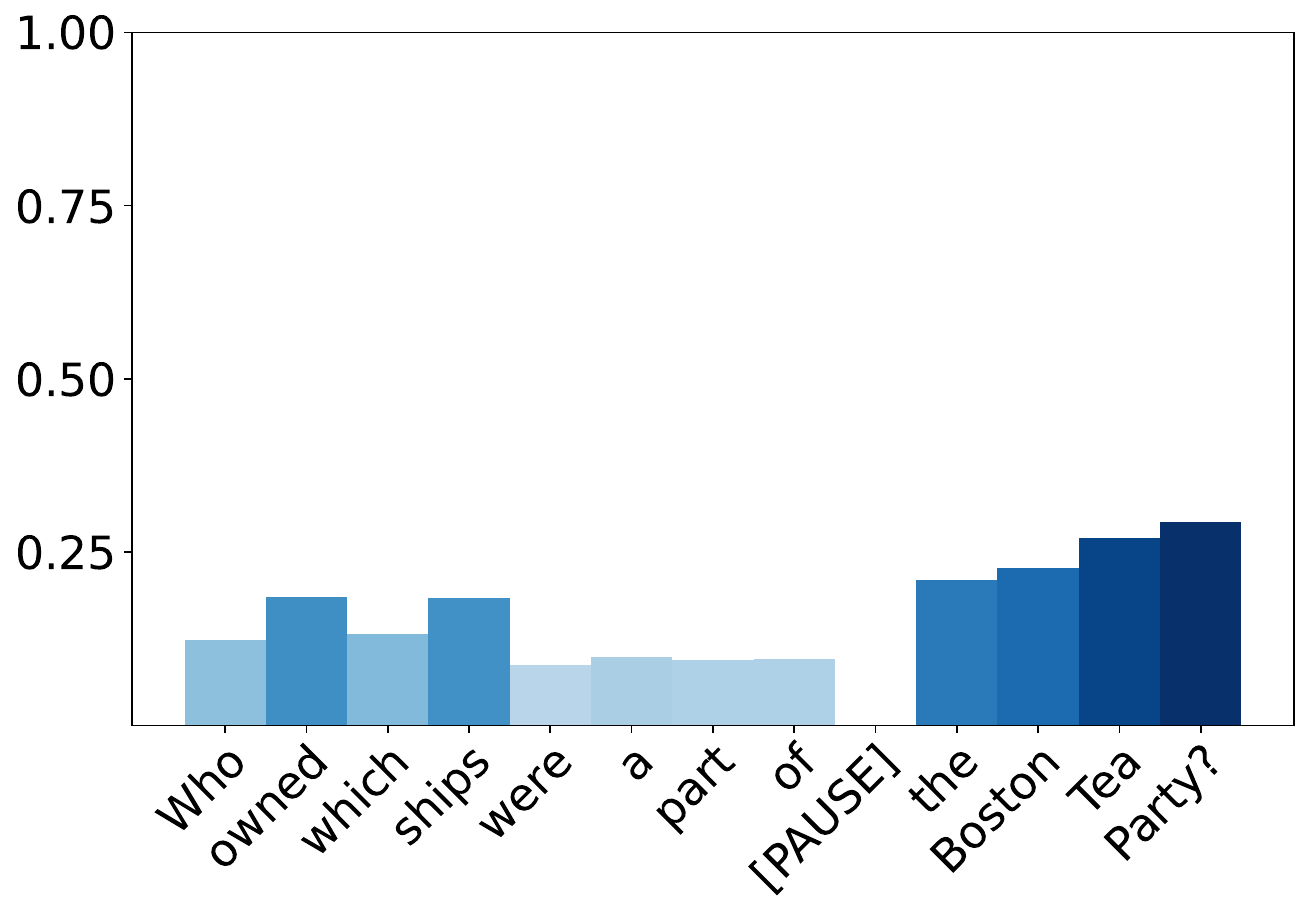}
            \caption[]%
            {{\small After adding \tframed[line width=0.5bp,fill=vred]{\textcolor{white}{\texttt{\textbf{[PAUSE]}}}} tokens} to paraphrase 4.}    
            \label{fig:mean and std of net44}
        \end{subfigure}
        \hfill
        \vskip\baselineskip
        \begin{subfigure}[b]{0.45\textwidth}   
            \centering 
            \includegraphics[width=\textwidth,height=3cm]{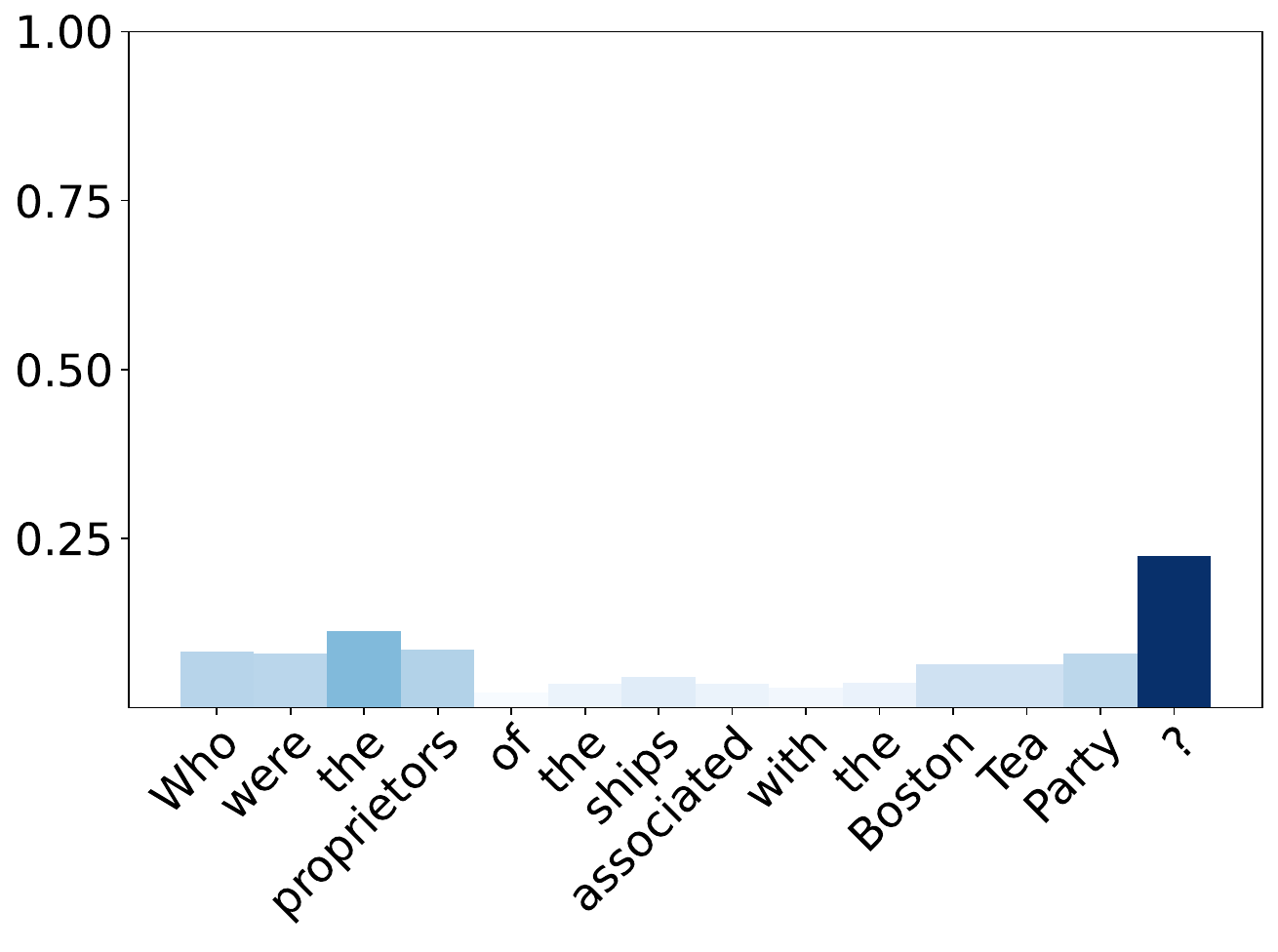}
            \caption[]%
            {{\small Before adding \tframed[line width=0.5bp,fill=vred]{\textcolor{white}{\texttt{\textbf{[PAUSE]}}}} tokens} to paraphrase 5.}    
            \label{fig:mean and std of net44}
        \end{subfigure}
        \hfill
        \begin{subfigure}[b]{0.45\textwidth}   
            \centering 
            \includegraphics[width=\textwidth,height=3cm]{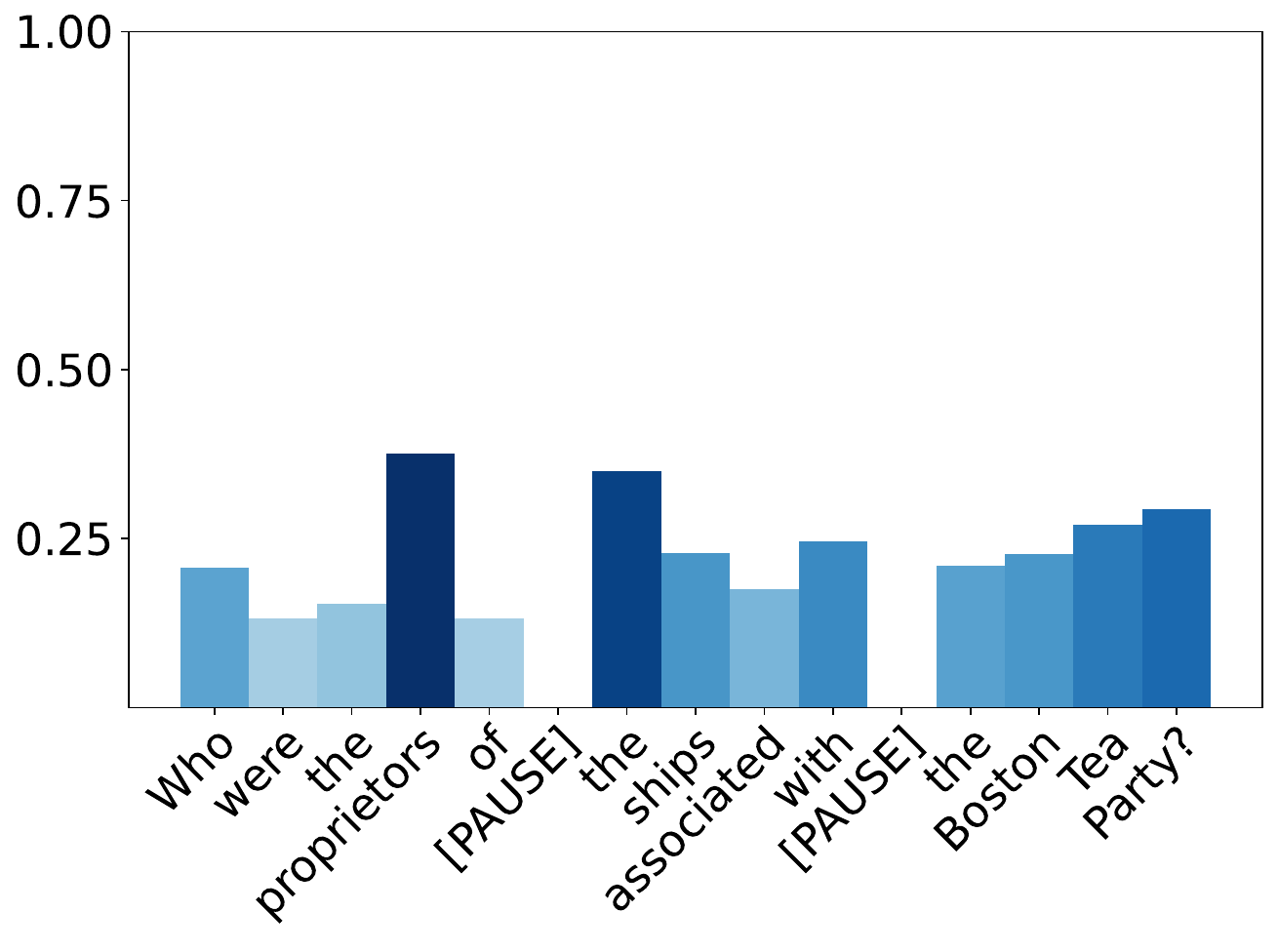}
            \caption[]%
            {{\small After adding \tframed[line width=0.5bp,fill=vred]{\textcolor{white}{\texttt{\textbf{[PAUSE]}}}} tokens} to paraphrase 5.}    
            \label{fig:mean and std of net44}
        \end{subfigure}
        \caption[]%
            {{\small The phrase \textbf{Boston Tea} gets more importance score after adding \tframed[line width=0.5bp,fill=vred]{\textcolor{white}{\texttt{\textbf{[PAUSE]}}}} token for FLAN-T5.}}   
        \label{fig:FLAN}
\end{figure*}

\begin{figure*}[!ht]
        \centering
        \begin{subfigure}[b]{0.45\textwidth}
            \centering
            \includegraphics[width=\textwidth,height=3cm]{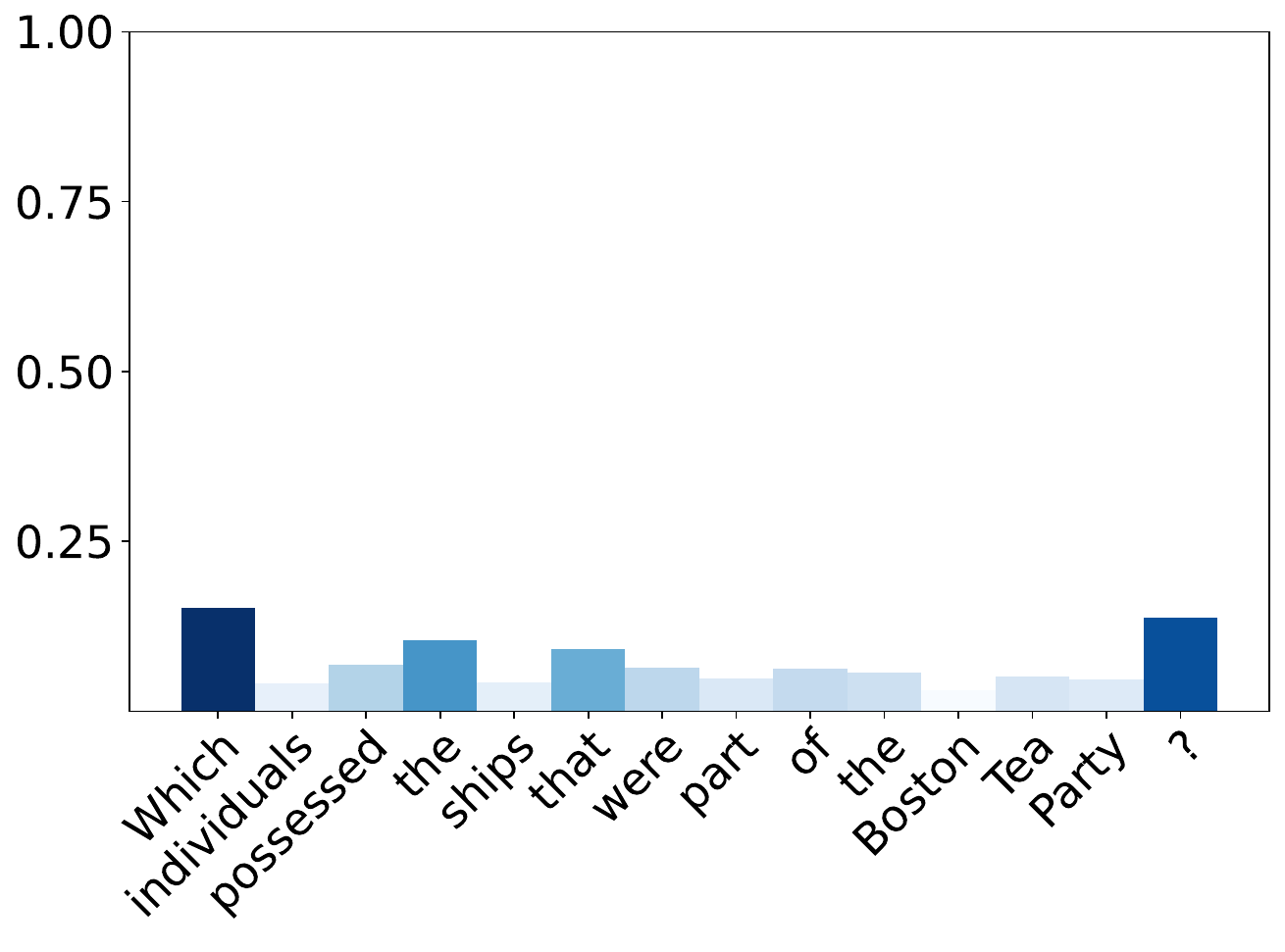}
            \caption[]%
            {{\small Before adding \tframed[line width=0.5bp,fill=vred]{\textcolor{white}{\texttt{\textbf{[PAUSE]}}}} tokens} to original prompt.}
            \label{fig:mean and std of net14}
        \end{subfigure}
        \hfill
        \begin{subfigure}[b]{0.45\textwidth}  
            \centering 
            \includegraphics[width=\textwidth,height=3cm]{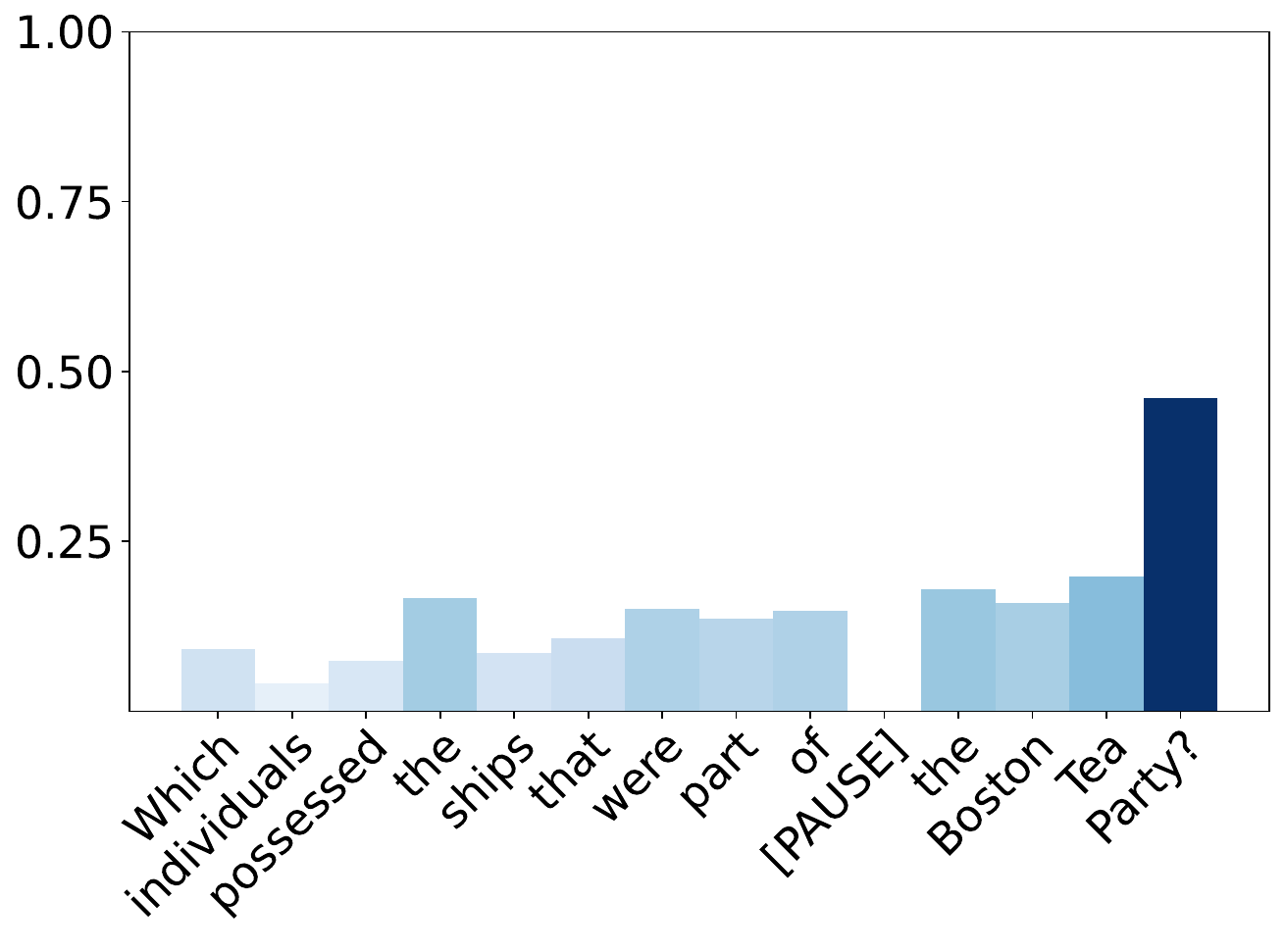}
            \caption[]%
            {{\small After adding \tframed[line width=0.5bp,fill=vred]{\textcolor{white}{\texttt{\textbf{[PAUSE]}}}} tokens} to original prompt.}    
            \label{fig:mean and std of net24}
        \end{subfigure}
        \hfill
        \vskip\baselineskip
        \begin{subfigure}[b]{0.45\textwidth}   
            \centering 
            \includegraphics[width=\textwidth,height=3cm]{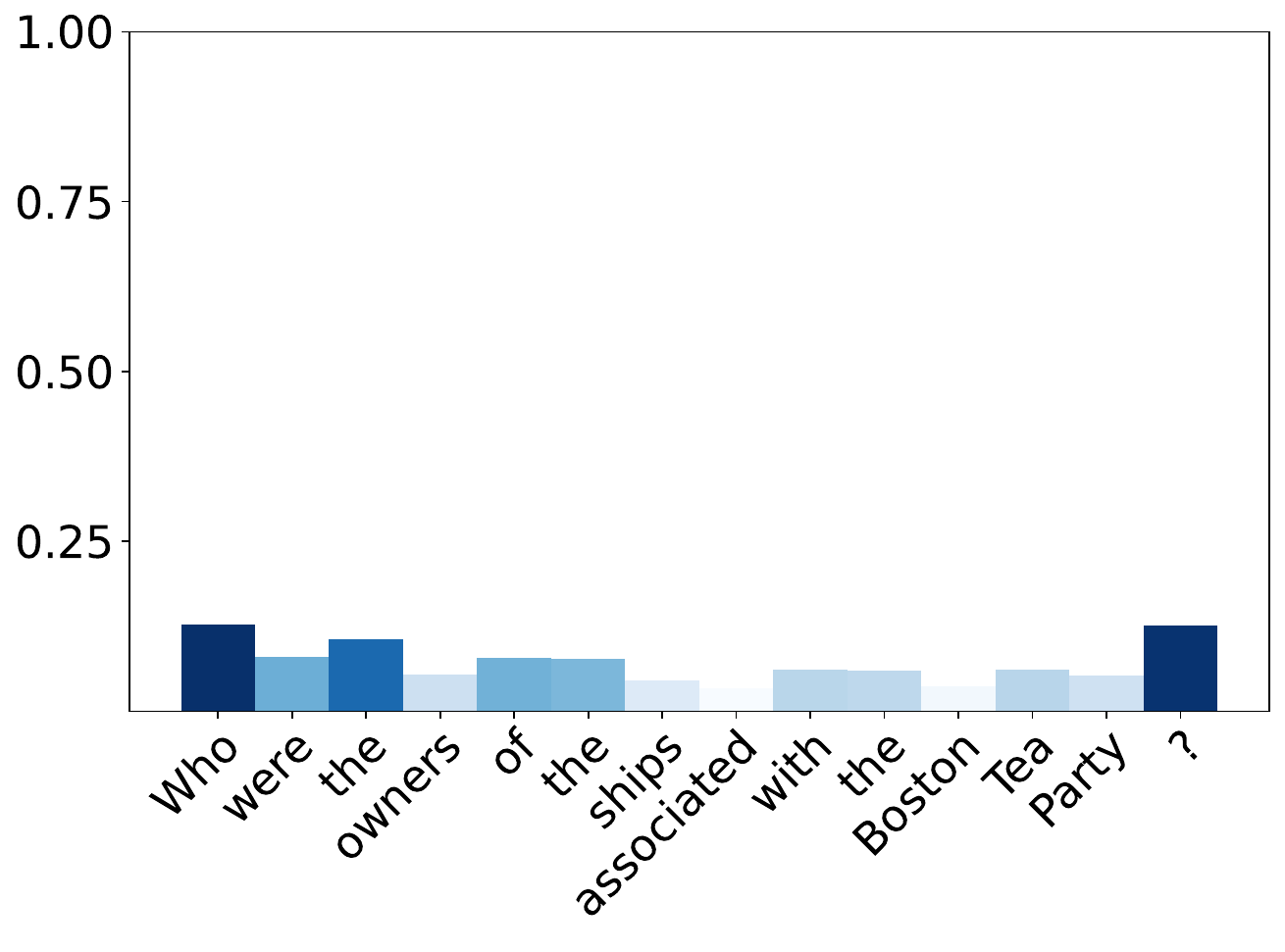}
            \caption[]%
            {{\small Before adding \tframed[line width=0.5bp,fill=vred]{\textcolor{white}{\texttt{\textbf{[PAUSE]}}}} tokens} to paraphrase 1.}    
            \label{fig:mean and std of net34}
        \end{subfigure}
        \hfill
        \begin{subfigure}[b]{0.45\textwidth}   
            \centering 
            \includegraphics[width=\textwidth,height=3cm]{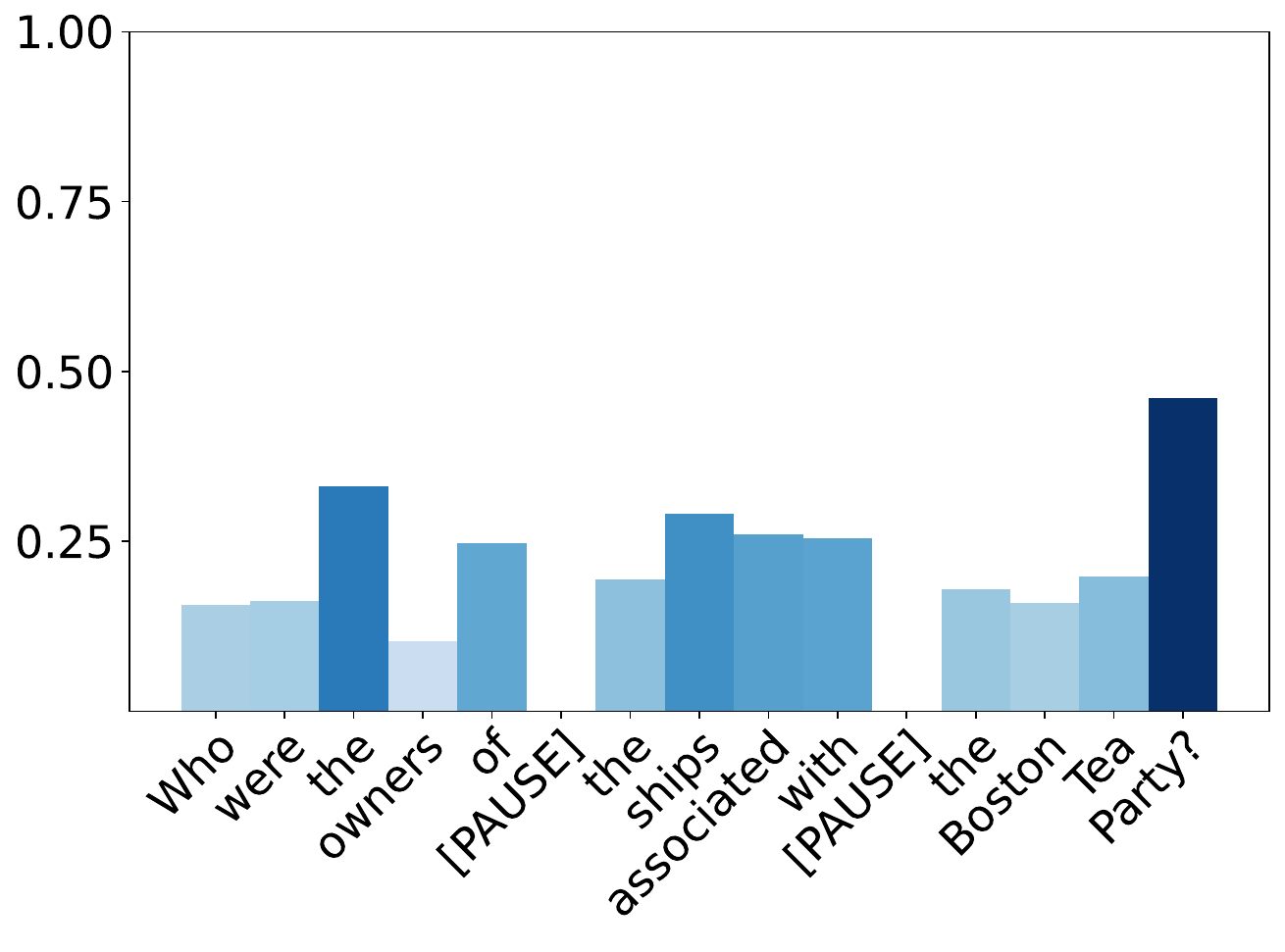}
            \caption[]%
            {{\small After adding \tframed[line width=0.5bp,fill=vred]{\textcolor{white}{\texttt{\textbf{[PAUSE]}}}} tokens} to paraphrase 1.}    
            \label{fig:mean and std of net44}
        \end{subfigure}
        \hfill
        \vskip\baselineskip
        \begin{subfigure}[b]{0.45\textwidth}   
            \centering 
            \includegraphics[width=\textwidth,height=3cm]{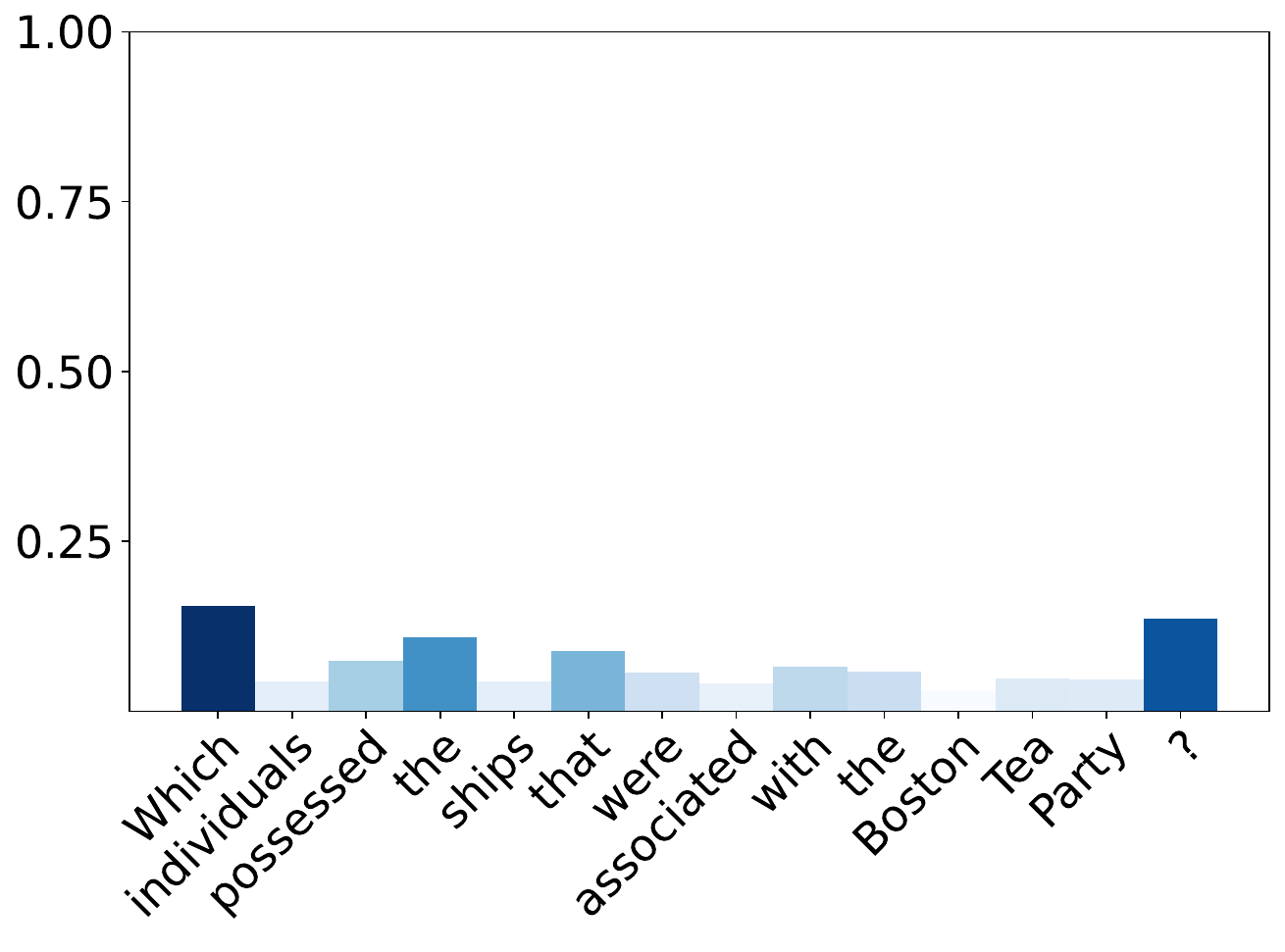}
            \caption[]%
            {{\small Before adding \tframed[line width=0.5bp,fill=vred]{\textcolor{white}{\texttt{\textbf{[PAUSE]}}}} tokens} to paraphrase 2.}
            \label{fig:mean and std of net34}
        \end{subfigure}
        \hfill
        \begin{subfigure}[b]{0.45\textwidth}   
            \centering 
            \includegraphics[width=\textwidth,height=3cm]{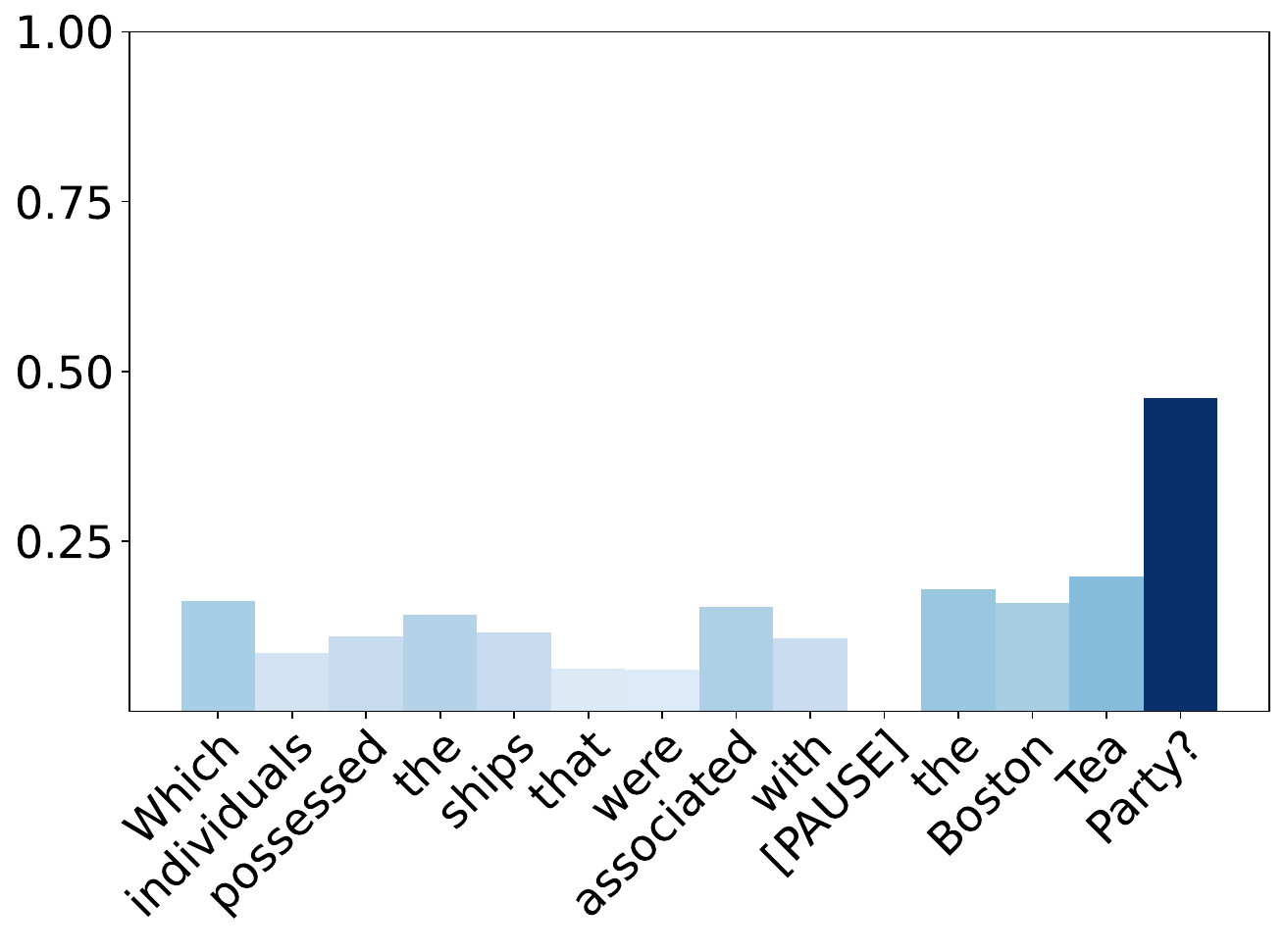}
            \caption[]%
            {{\small After adding \tframed[line width=0.5bp,fill=vred]{\textcolor{white}{\texttt{\textbf{[PAUSE]}}}} tokens} to paraphrase 2.} 
            \label{fig:mean and std of net44}
        \end{subfigure}
        \hfill
        \vskip\baselineskip
        \begin{subfigure}[b]{0.45\textwidth}   
            \centering 
            \includegraphics[width=\textwidth,height=3cm]{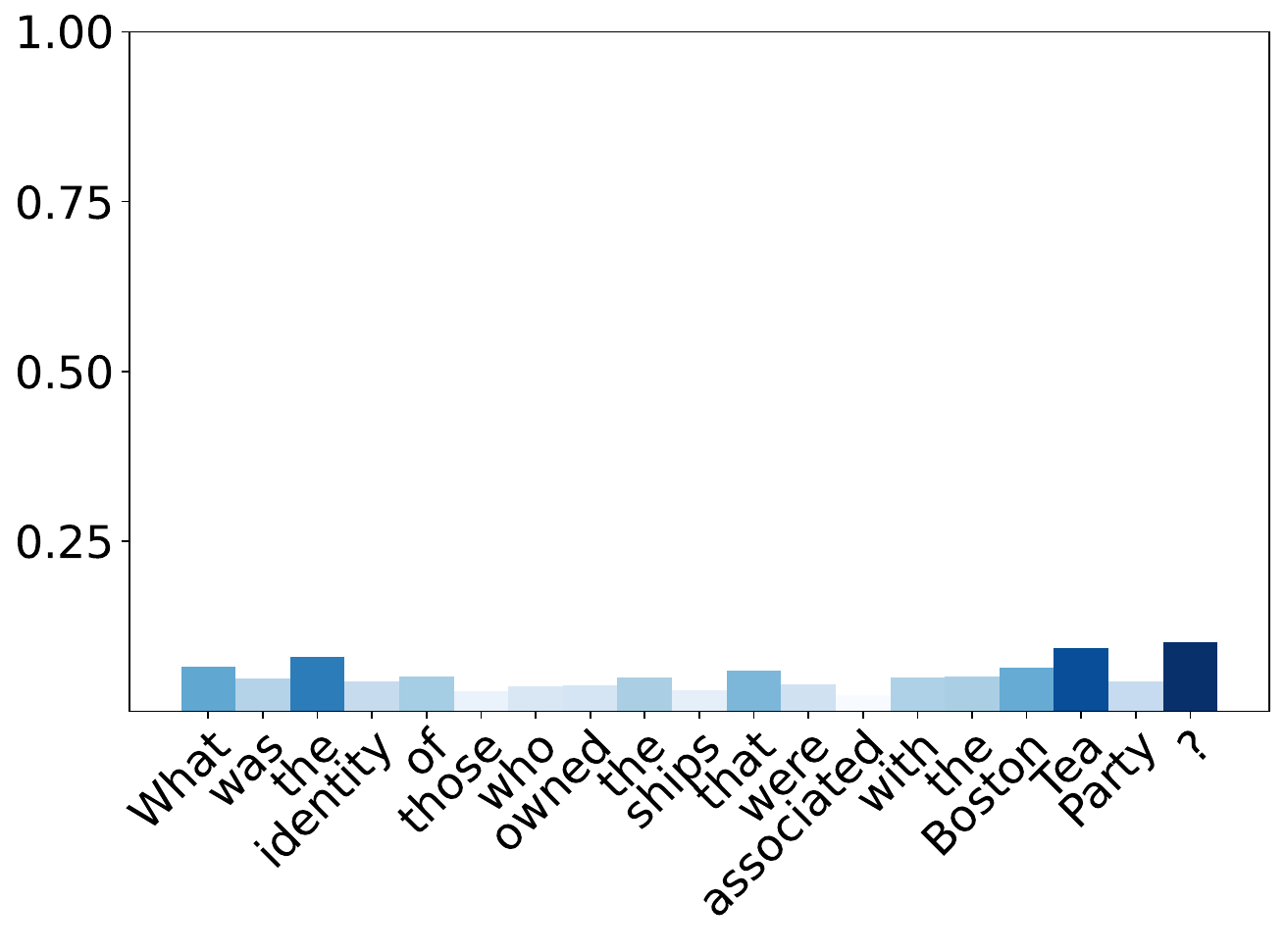}
            \caption[]%
            {{\small Before adding \tframed[line width=0.5bp,fill=vred]{\textcolor{white}{\texttt{\textbf{[PAUSE]}}}} tokens} to paraphrase 3.}
            \label{fig:mean and std of net44}
        \end{subfigure}
        \hfill
        \begin{subfigure}[b]{0.45\textwidth}   
            \centering 
            \includegraphics[width=\textwidth,height=3cm]{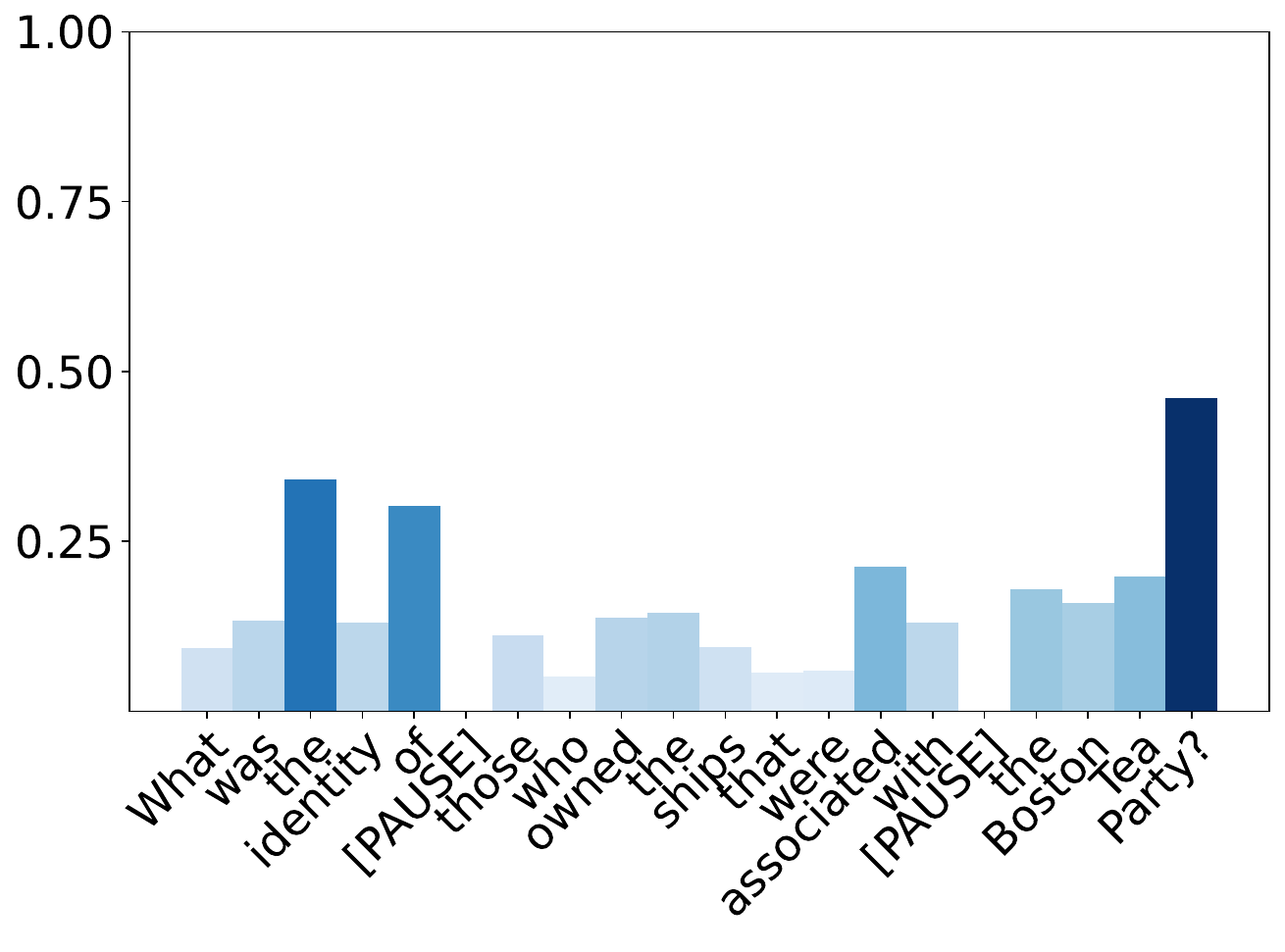}
            \caption[]%
            {{\small After adding \tframed[line width=0.5bp,fill=vred]{\textcolor{white}{\texttt{\textbf{[PAUSE]}}}} tokens} to paraphrase 3.}    
            \label{fig:mean and std of net44}
        \end{subfigure}
        \hfill
        \vskip\baselineskip
        \begin{subfigure}[b]{0.45\textwidth}   
            \centering 
            \includegraphics[width=\textwidth,height=3cm]{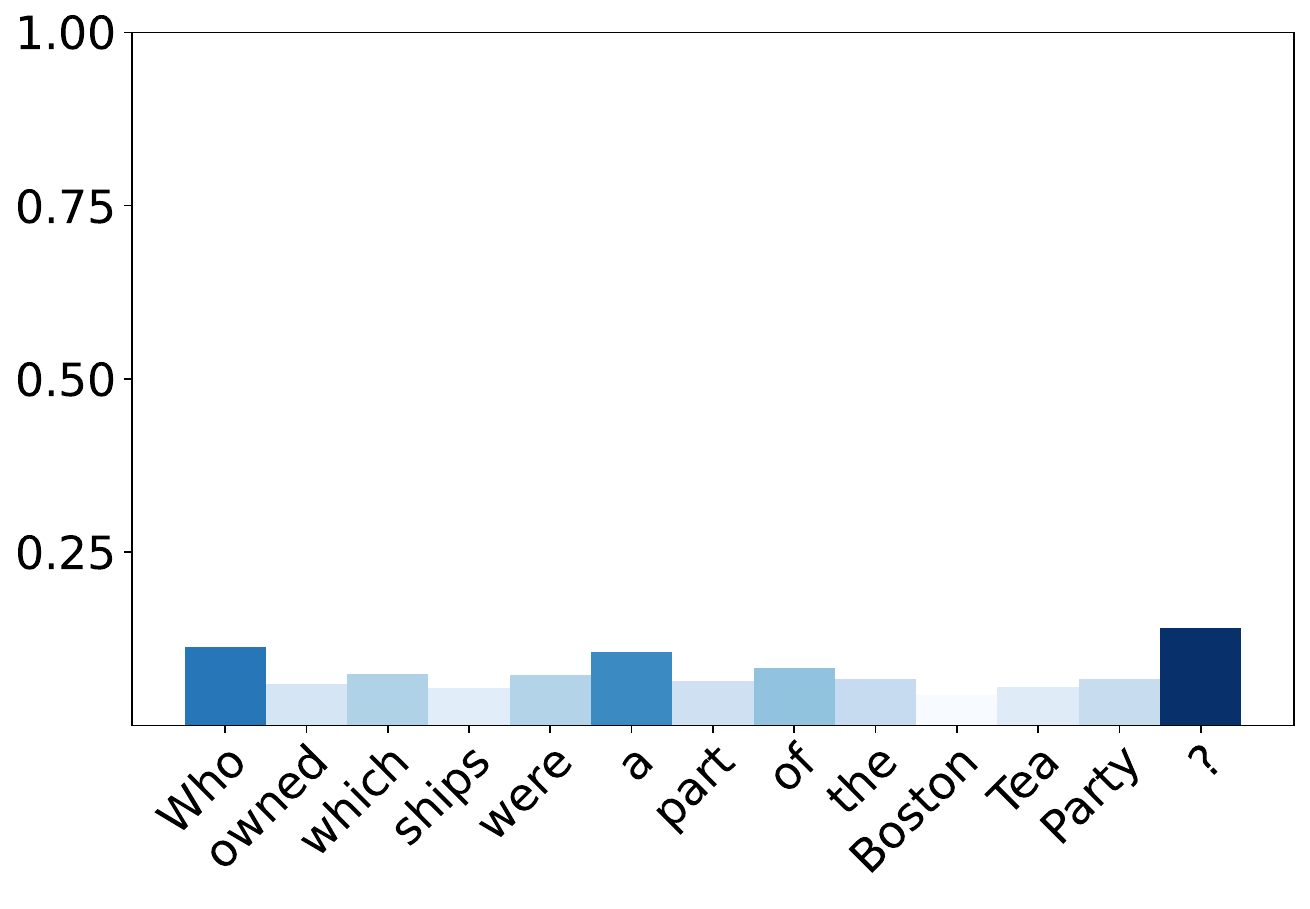}
            \caption[]%
            {{\small Before adding \tframed[line width=0.5bp,fill=vred]{\textcolor{white}{\texttt{\textbf{[PAUSE]}}}} tokens} to paraphrase 4.}    
            \label{fig:mean and std of net44}
        \end{subfigure}
        \hfill
        \begin{subfigure}[b]{0.45\textwidth}   
            \centering 
            \includegraphics[width=\textwidth,height=3cm]{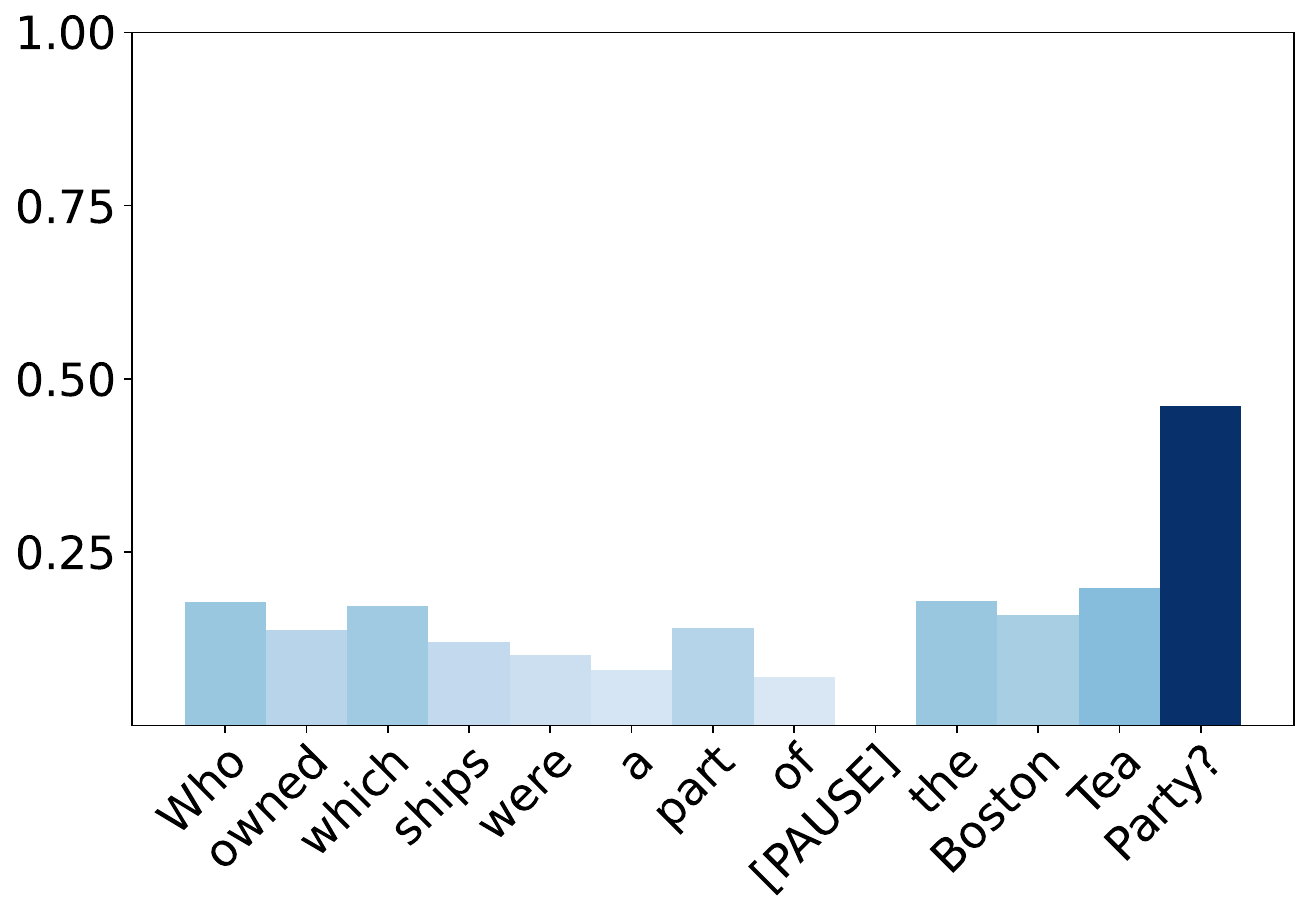}
            \caption[]%
            {{\small After adding \tframed[line width=0.5bp,fill=vred]{\textcolor{white}{\texttt{\textbf{[PAUSE]}}}} tokens} to paraphrase 4.}    
            \label{fig:mean and std of net44}
        \end{subfigure}
        \hfill
        \vskip\baselineskip
        \begin{subfigure}[b]{0.45\textwidth}   
            \centering 
            \includegraphics[width=\textwidth,height=3cm]{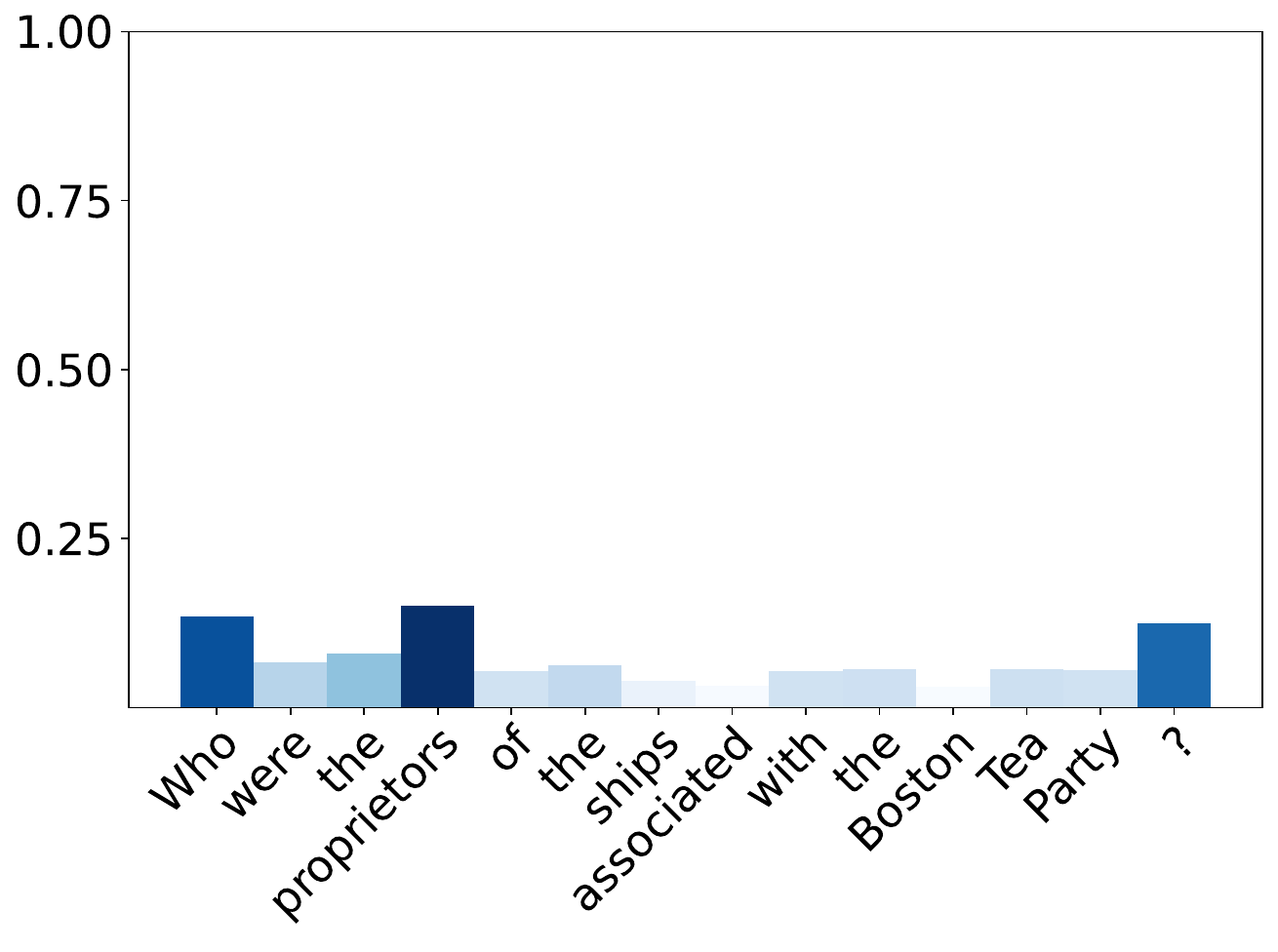}
            \caption[]%
            {{\small Before adding \tframed[line width=0.5bp,fill=vred]{\textcolor{white}{\texttt{\textbf{[PAUSE]}}}} tokens} to paraphrase 5.}    
            \label{fig:mean and std of net44}
        \end{subfigure}
        \hfill
        \begin{subfigure}[b]{0.45\textwidth}   
            \centering 
            \includegraphics[width=\textwidth,height=3cm]{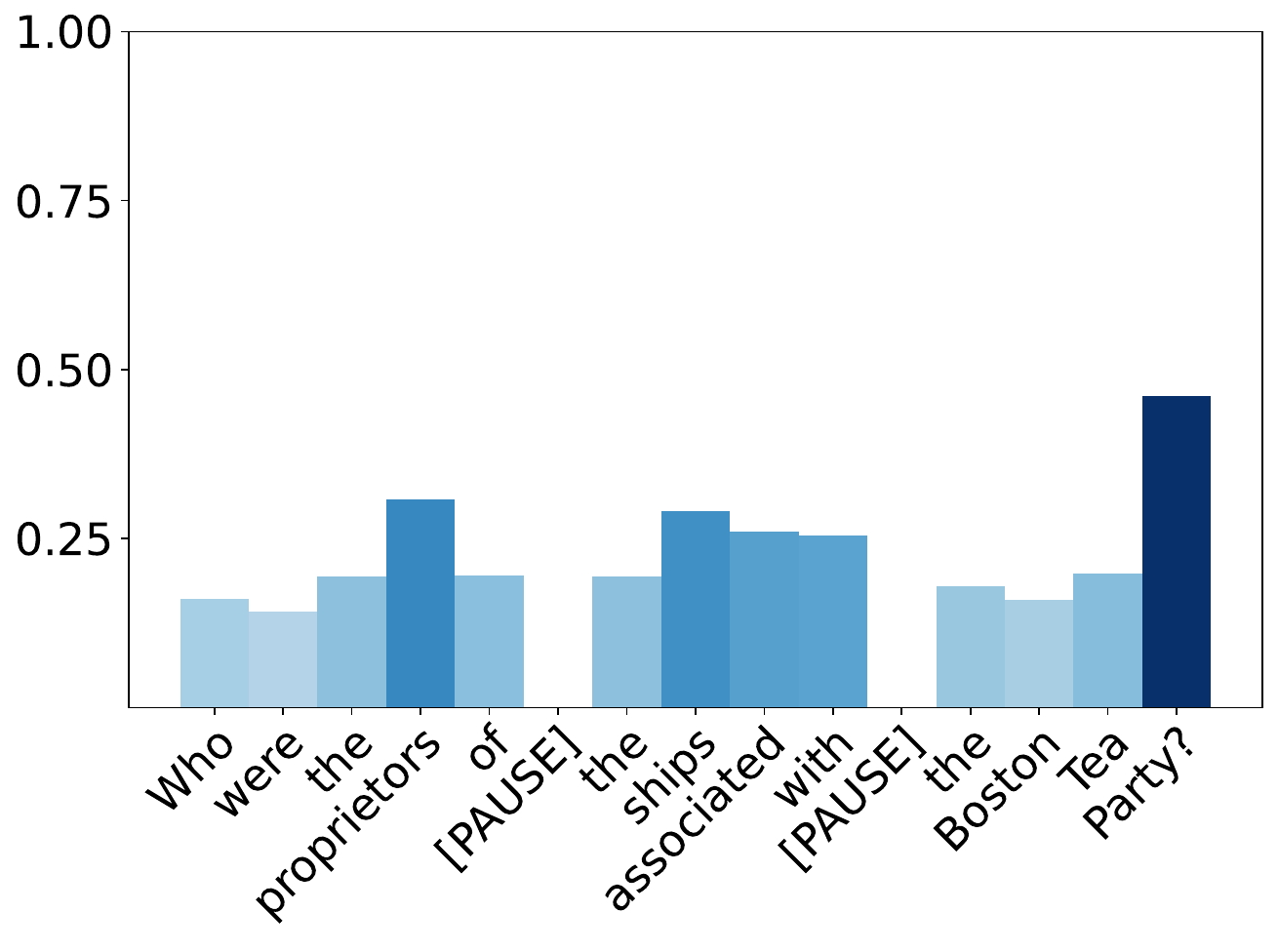}
            \caption[]%
            {{\small After adding \tframed[line width=0.5bp,fill=vred]{\textcolor{white}{\texttt{\textbf{[PAUSE]}}}} tokens} to paraphrase 5.}    
            \label{fig:mean and std of net44}
        \end{subfigure}
        \caption[]%
            {{\small The phrase \textbf{Boston Tea} gets more importance score after adding \tframed[line width=0.5bp,fill=vred]{\textcolor{white}{\texttt{\textbf{[PAUSE]}}}} token for GPT Neo.}}   
        \label{fig:GPT Neo}
\end{figure*}

\begin{figure*}[!ht]
        \centering
        \begin{subfigure}[b]{0.45\textwidth}
            \centering
            \includegraphics[width=\textwidth,height=3cm]{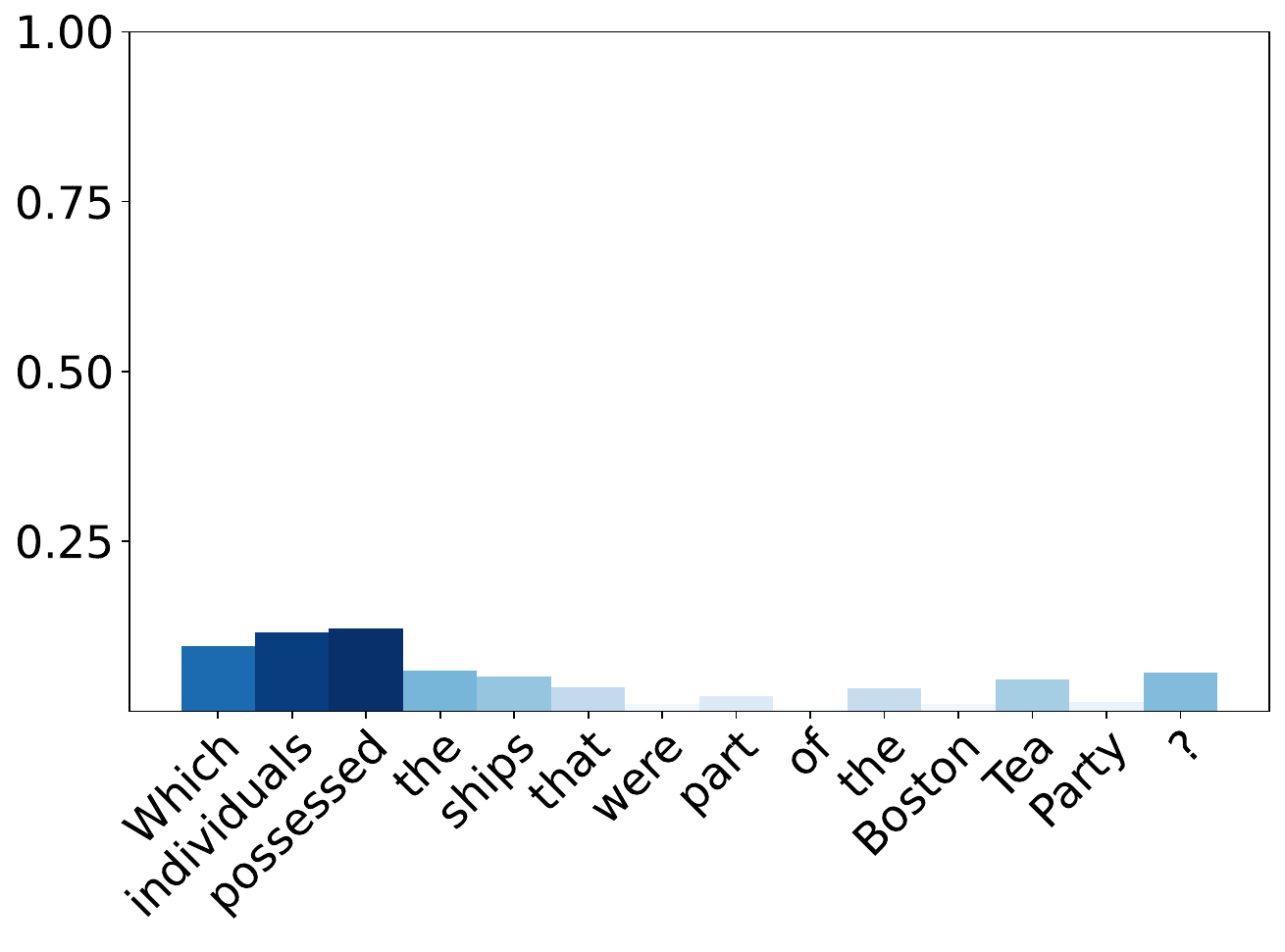}
            \caption[]%
            {{\small Before adding \tframed[line width=0.5bp,fill=vred]{\textcolor{white}{\texttt{\textbf{[PAUSE]}}}} tokens} to original prompt.}
            \label{fig:mean and std of net14}
        \end{subfigure}
        \hfill
        \begin{subfigure}[b]{0.45\textwidth}  
            \centering 
            \includegraphics[width=\textwidth,height=3cm]{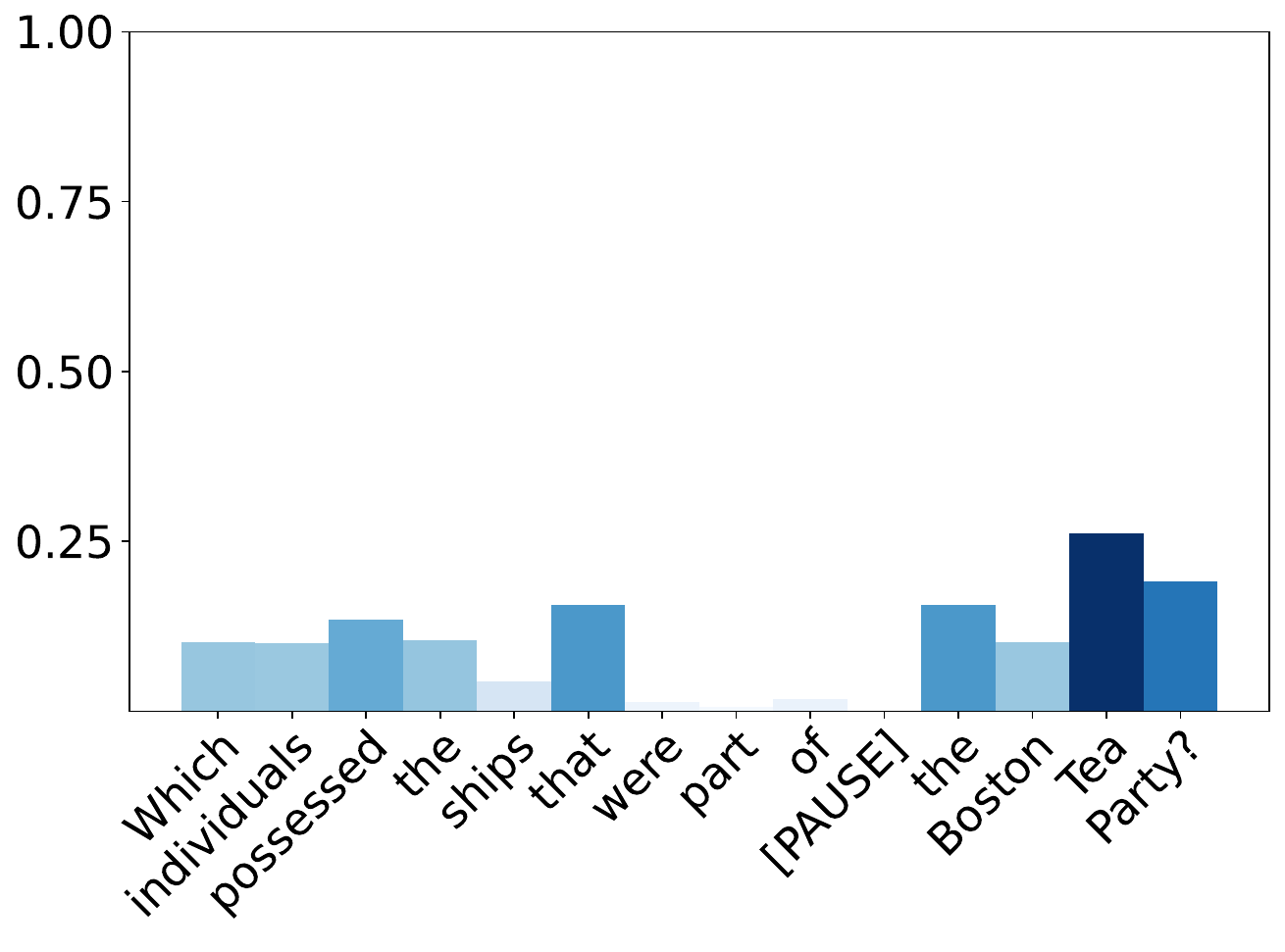}
            \caption[]%
            {{\small After adding \tframed[line width=0.5bp,fill=vred]{\textcolor{white}{\texttt{\textbf{[PAUSE]}}}} tokens} to original prompt.}    
            \label{fig:mean and std of net24}
        \end{subfigure}
        \hfill
        \vskip\baselineskip
        \begin{subfigure}[b]{0.45\textwidth}   
            \centering 
            \includegraphics[width=\textwidth,height=3cm]{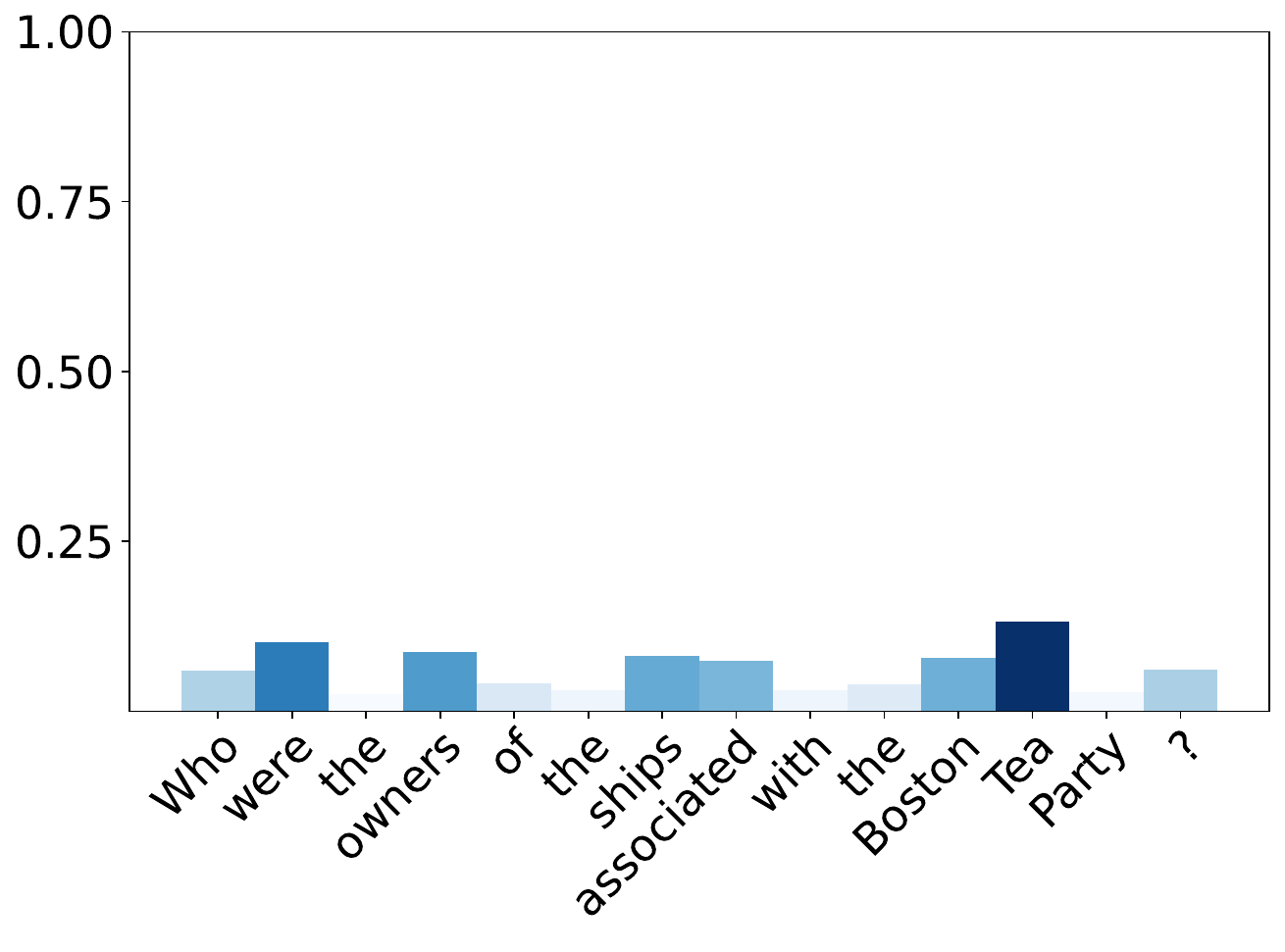}
            \caption[]%
            {{\small Before adding \tframed[line width=0.5bp,fill=vred]{\textcolor{white}{\texttt{\textbf{[PAUSE]}}}} tokens} to paraphrase 1.}    
            \label{fig:mean and std of net34}
        \end{subfigure}
        \hfill
        \begin{subfigure}[b]{0.45\textwidth}   
            \centering 
            \includegraphics[width=\textwidth,height=3cm]{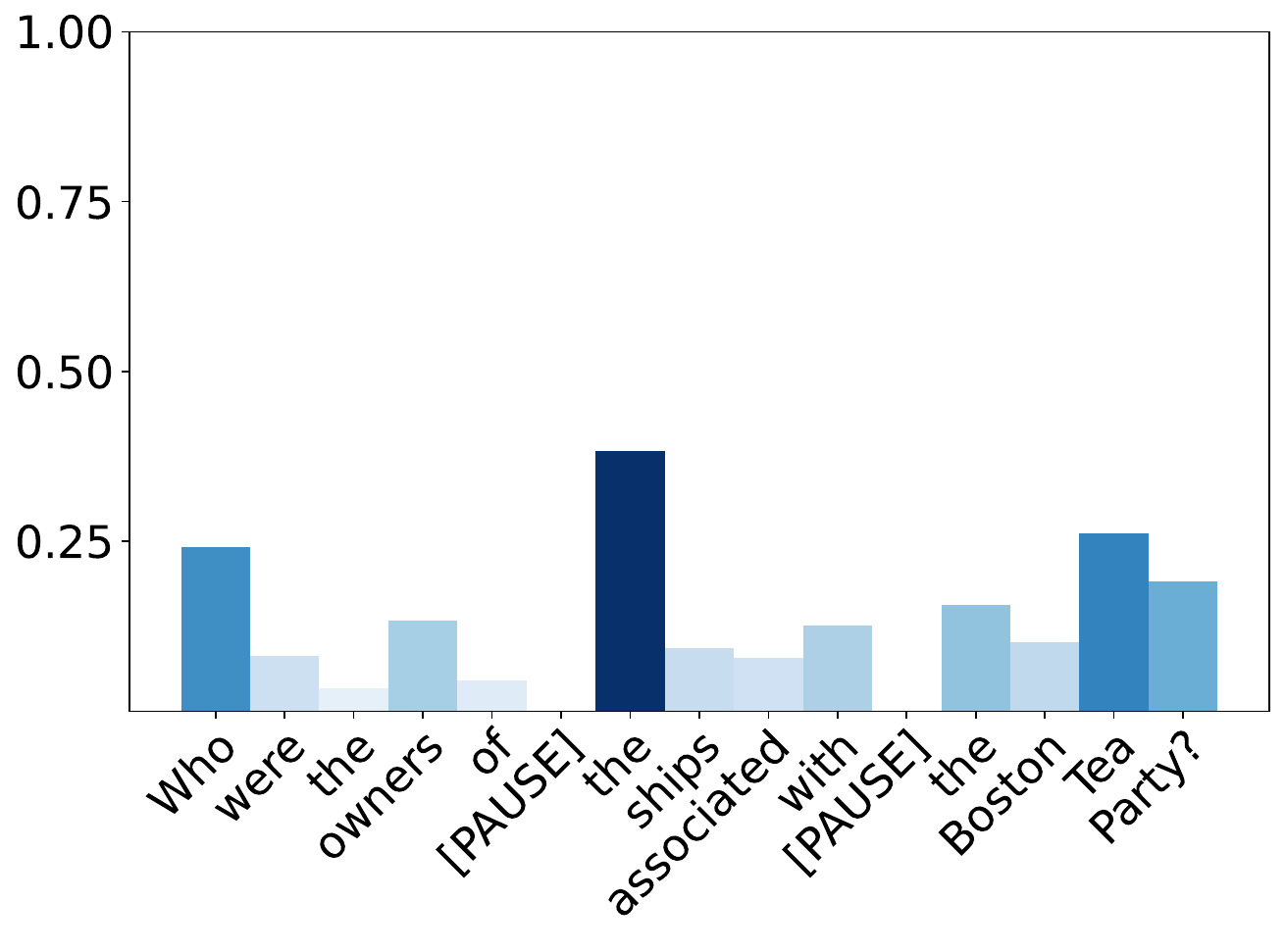}
            \caption[]%
            {{\small After adding \tframed[line width=0.5bp,fill=vred]{\textcolor{white}{\texttt{\textbf{[PAUSE]}}}} tokens} to paraphrase 1.}    
            \label{fig:mean and std of net44}
        \end{subfigure}
        \hfill
        \vskip\baselineskip
        \begin{subfigure}[b]{0.45\textwidth}   
            \centering 
            \includegraphics[width=\textwidth,height=3cm]{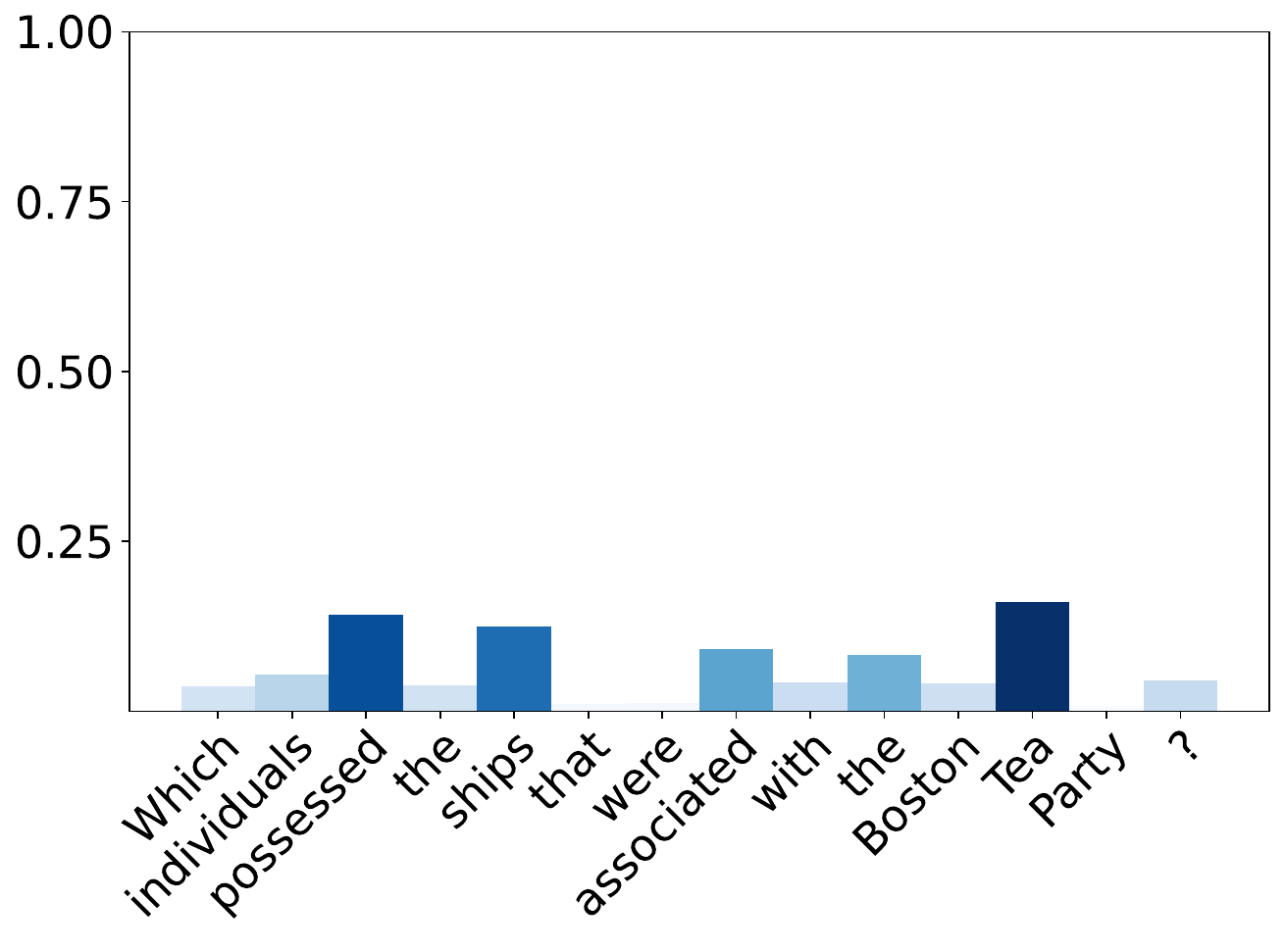}
            \caption[]%
            {{\small Before adding \tframed[line width=0.5bp,fill=vred]{\textcolor{white}{\texttt{\textbf{[PAUSE]}}}} tokens} to paraphrase 2.}
            \label{fig:mean and std of net34}
        \end{subfigure}
        \hfill
        \begin{subfigure}[b]{0.45\textwidth}   
            \centering 
            \includegraphics[width=\textwidth,height=3cm]{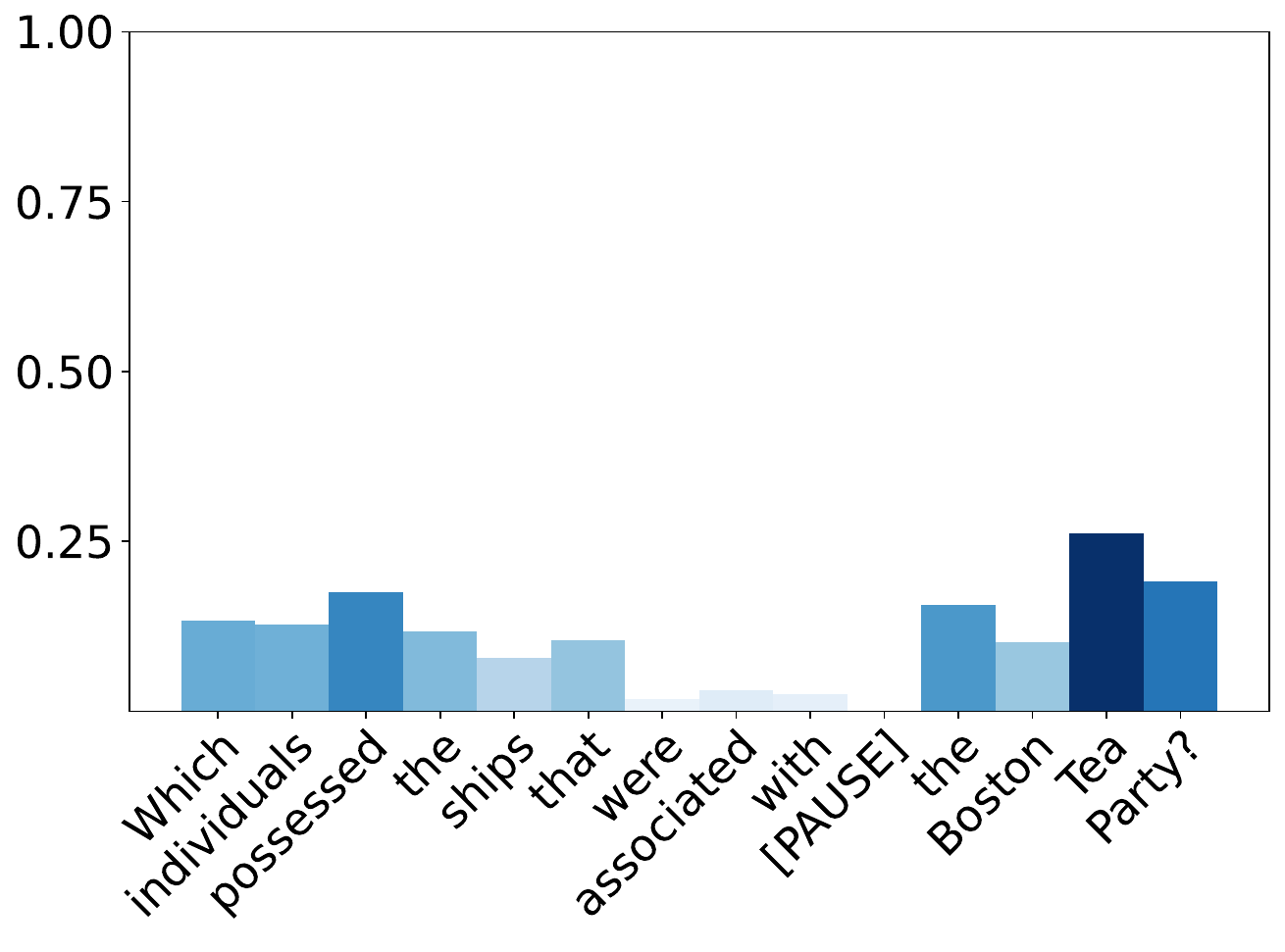}
            \caption[]%
            {{\small After adding \tframed[line width=0.5bp,fill=vred]{\textcolor{white}{\texttt{\textbf{[PAUSE]}}}} tokens} to paraphrase 2.} 
            \label{fig:mean and std of net44}
        \end{subfigure}
        \hfill
        \vskip\baselineskip
        \begin{subfigure}[b]{0.45\textwidth}   
            \centering 
            \includegraphics[width=\textwidth,height=3cm]{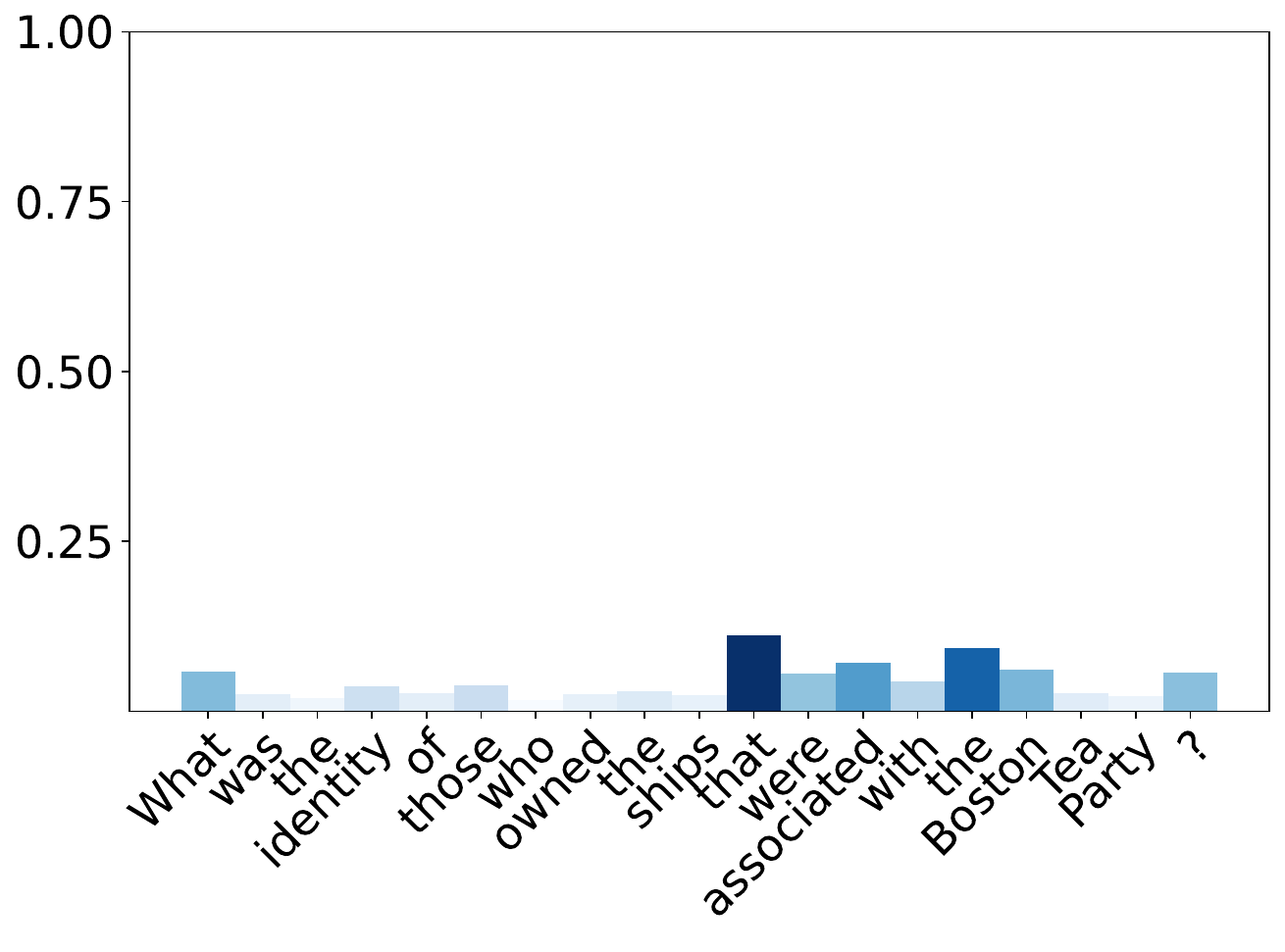}
            \caption[]%
            {{\small Before adding \tframed[line width=0.5bp,fill=vred]{\textcolor{white}{\texttt{\textbf{[PAUSE]}}}} tokens} to paraphrase 3.}
            \label{fig:mean and std of net44}
        \end{subfigure}
        \hfill
        \begin{subfigure}[b]{0.45\textwidth}   
            \centering 
            \includegraphics[width=\textwidth,height=3cm]{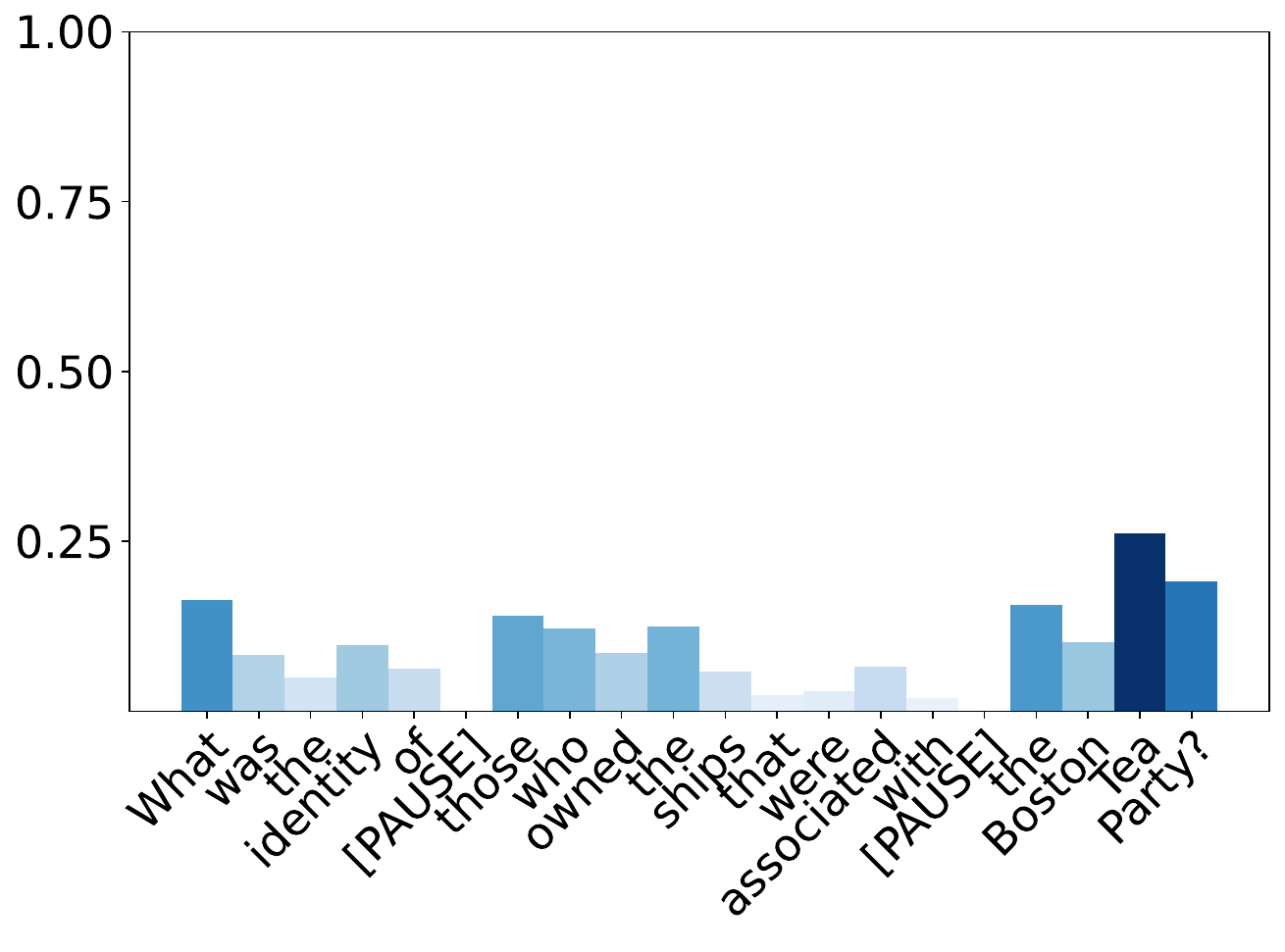}
            \caption[]%
            {{\small After adding \tframed[line width=0.5bp,fill=vred]{\textcolor{white}{\texttt{\textbf{[PAUSE]}}}} tokens} to paraphrase 3.}    
            \label{fig:mean and std of net44}
        \end{subfigure}
        \hfill
        \vskip\baselineskip
        \begin{subfigure}[b]{0.45\textwidth}   
            \centering 
            \includegraphics[width=\textwidth,height=3cm]{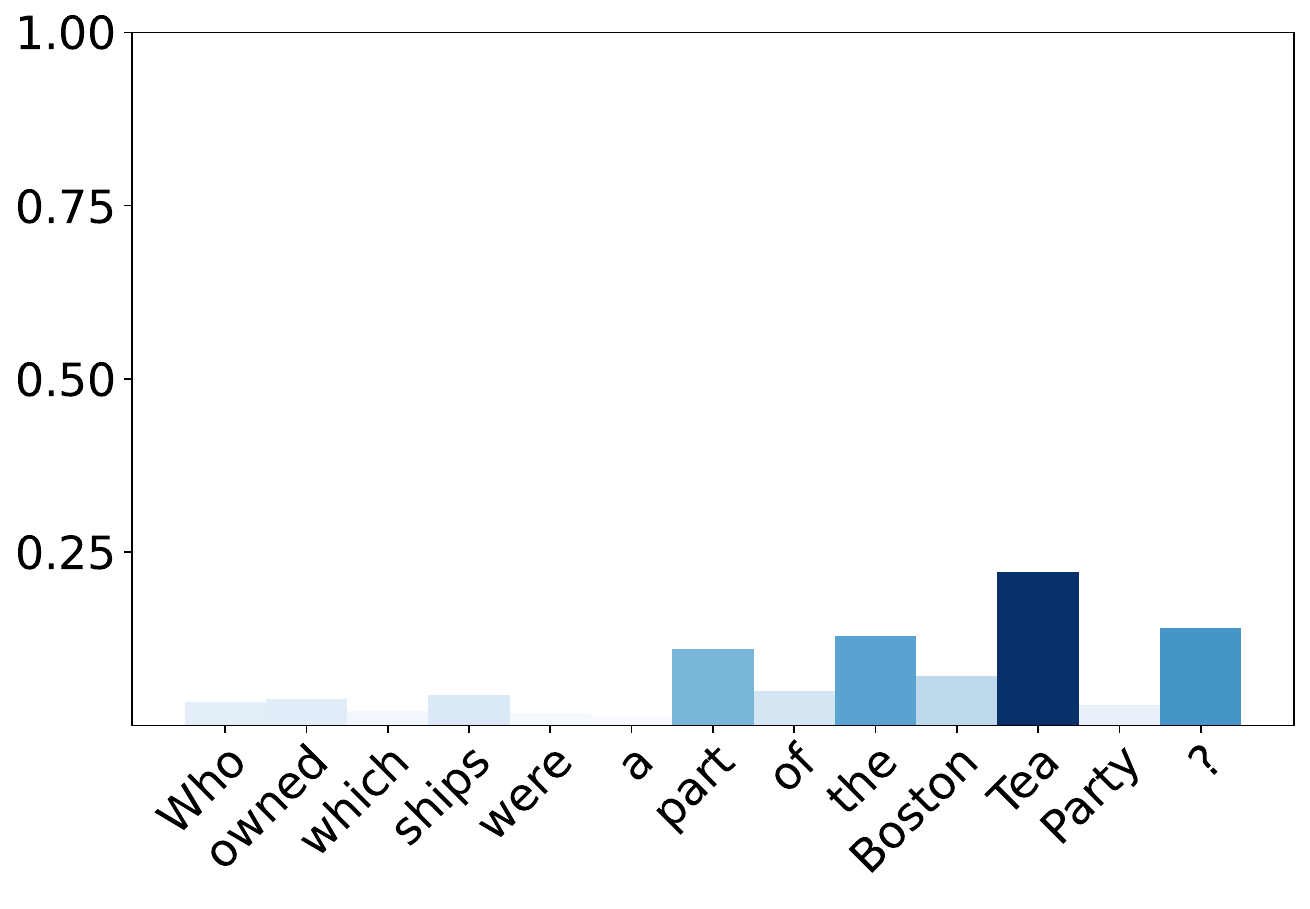}
            \caption[]%
            {{\small Before adding \tframed[line width=0.5bp,fill=vred]{\textcolor{white}{\texttt{\textbf{[PAUSE]}}}} tokens} to paraphrase 4.}    
            \label{fig:mean and std of net44}
        \end{subfigure}
        \hfill
        \begin{subfigure}[b]{0.45\textwidth}   
            \centering 
            \includegraphics[width=\textwidth,height=3cm]{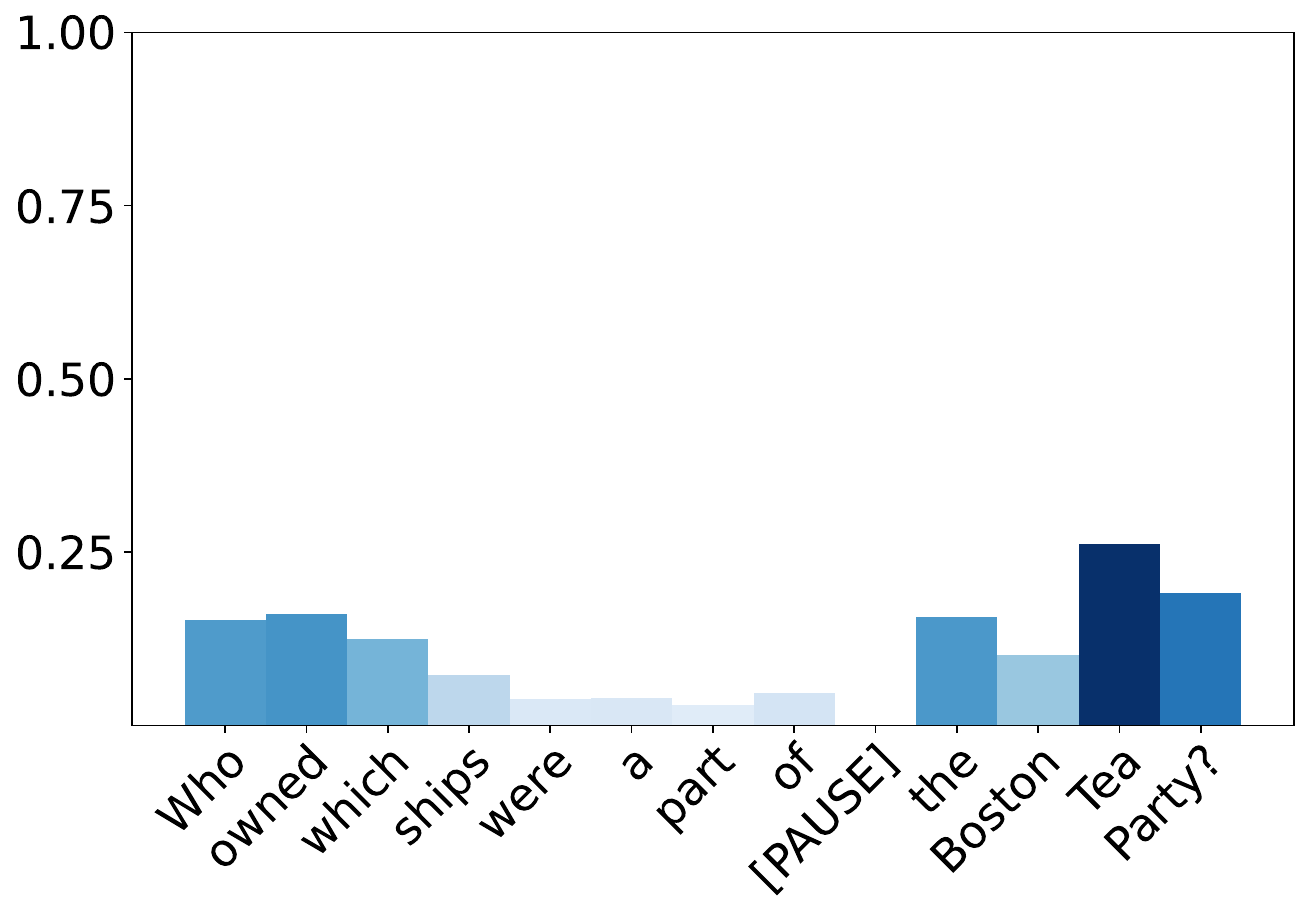}
            \caption[]%
            {{\small After adding \tframed[line width=0.5bp,fill=vred]{\textcolor{white}{\texttt{\textbf{[PAUSE]}}}} tokens} to paraphrase 4.}    
            \label{fig:mean and std of net44}
        \end{subfigure}
        \hfill
        \vskip\baselineskip
        \begin{subfigure}[b]{0.45\textwidth}   
            \centering 
            \includegraphics[width=\textwidth,height=3cm]{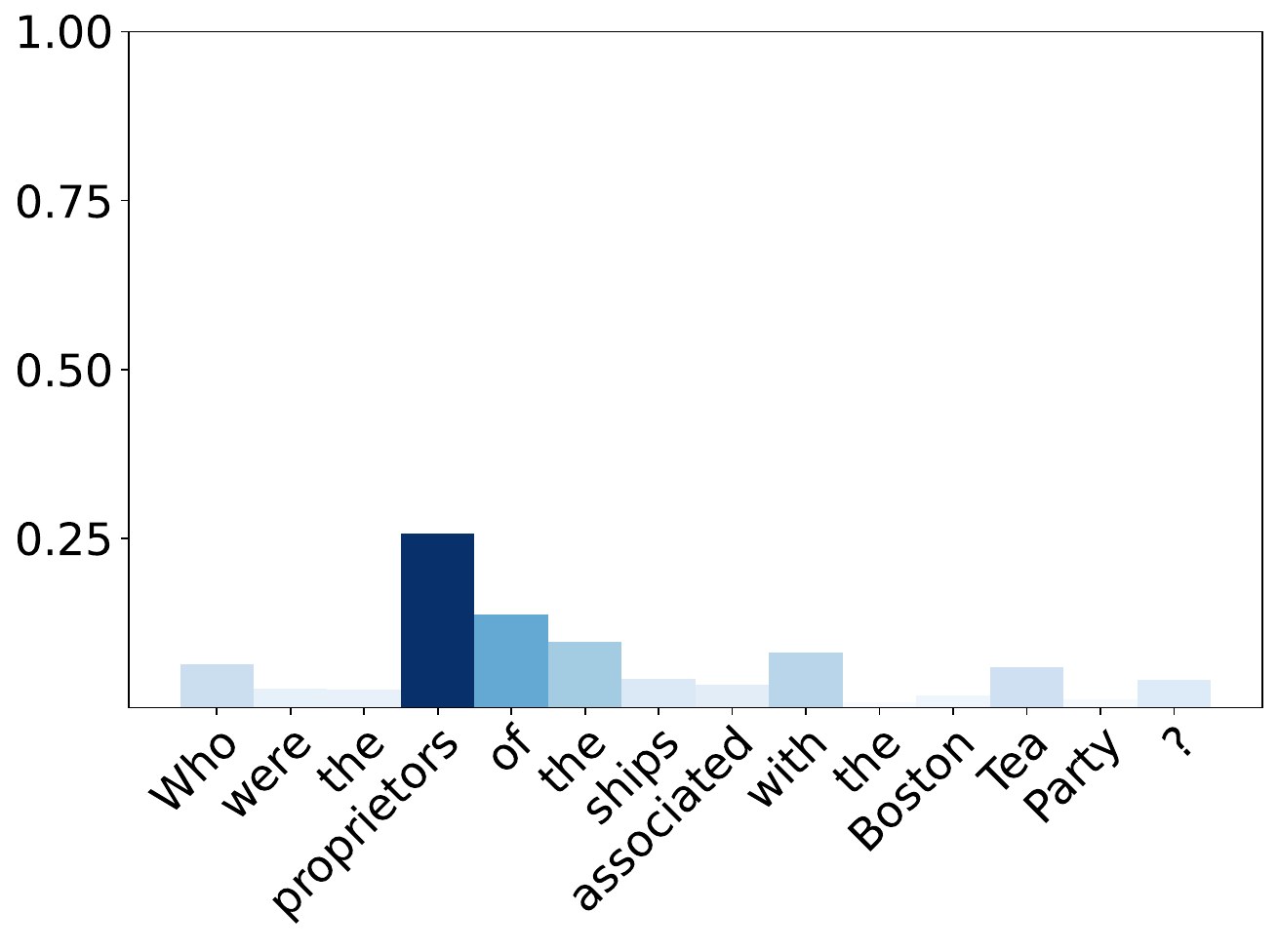}
            \caption[]%
            {{\small Before adding \tframed[line width=0.5bp,fill=vred]{\textcolor{white}{\texttt{\textbf{[PAUSE]}}}} tokens} to paraphrase 5.}    
            \label{fig:mean and std of net44}
        \end{subfigure}
        \hfill
        \begin{subfigure}[b]{0.45\textwidth}   
            \centering 
            \includegraphics[width=\textwidth,height=3cm]{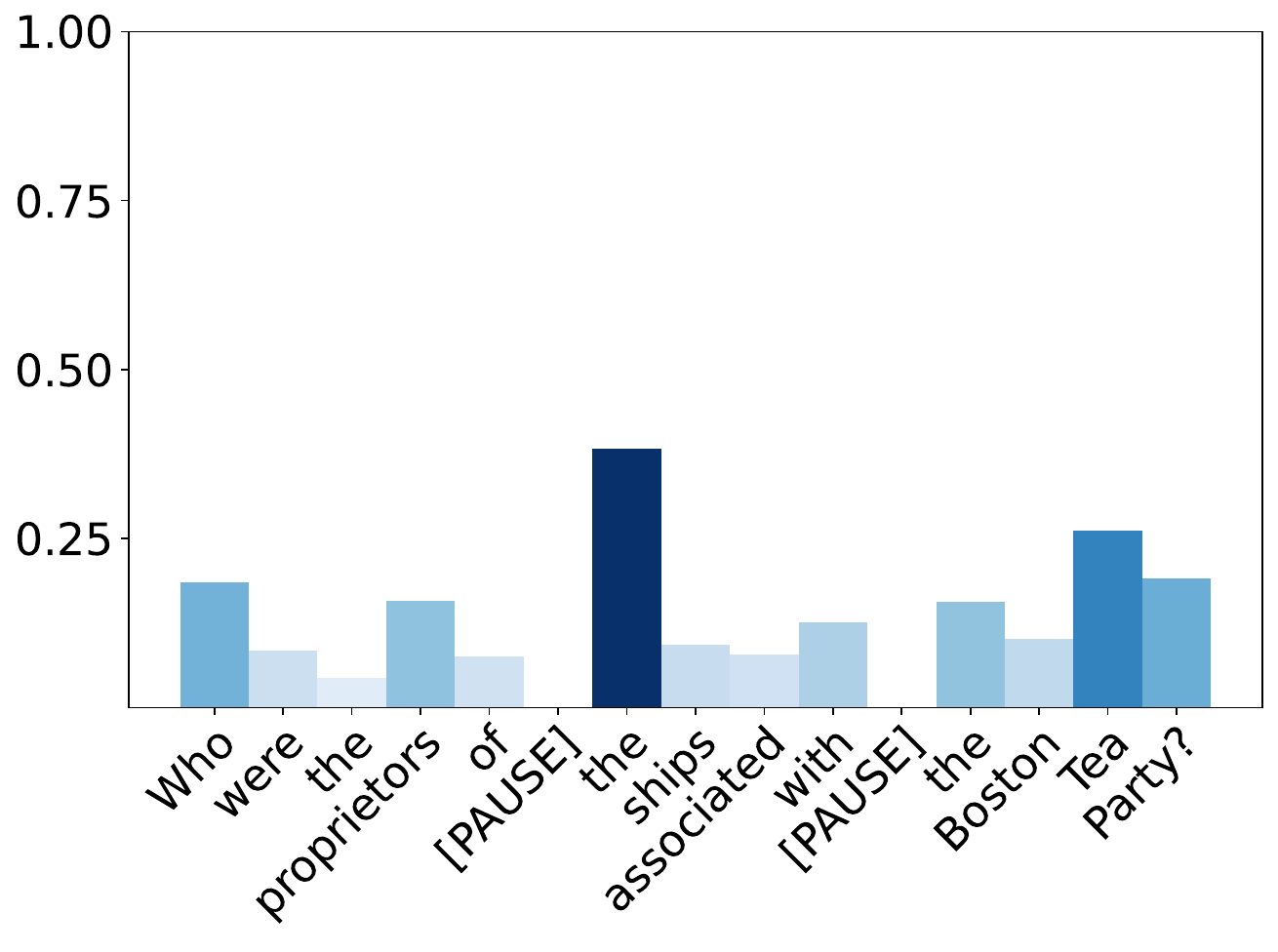}
            \caption[]%
            {{\small After adding \tframed[line width=0.5bp,fill=vred]{\textcolor{white}{\texttt{\textbf{[PAUSE]}}}} tokens} to paraphrase 5.}    
            \label{fig:mean and std of net44}
        \end{subfigure}
        \caption[]%
            {{\small The phrase \textbf{Boston Tea} gets more importance score after adding \tframed[line width=0.5bp,fill=vred]{\textcolor{white}{\texttt{\textbf{[PAUSE]}}}} token for Llama2.}}   
        \label{fig:Llama2}
\end{figure*}

\begin{figure*}[!ht]
        \centering
        \begin{subfigure}[b]{0.45\textwidth}
            \centering
            \includegraphics[width=\textwidth,height=3cm]{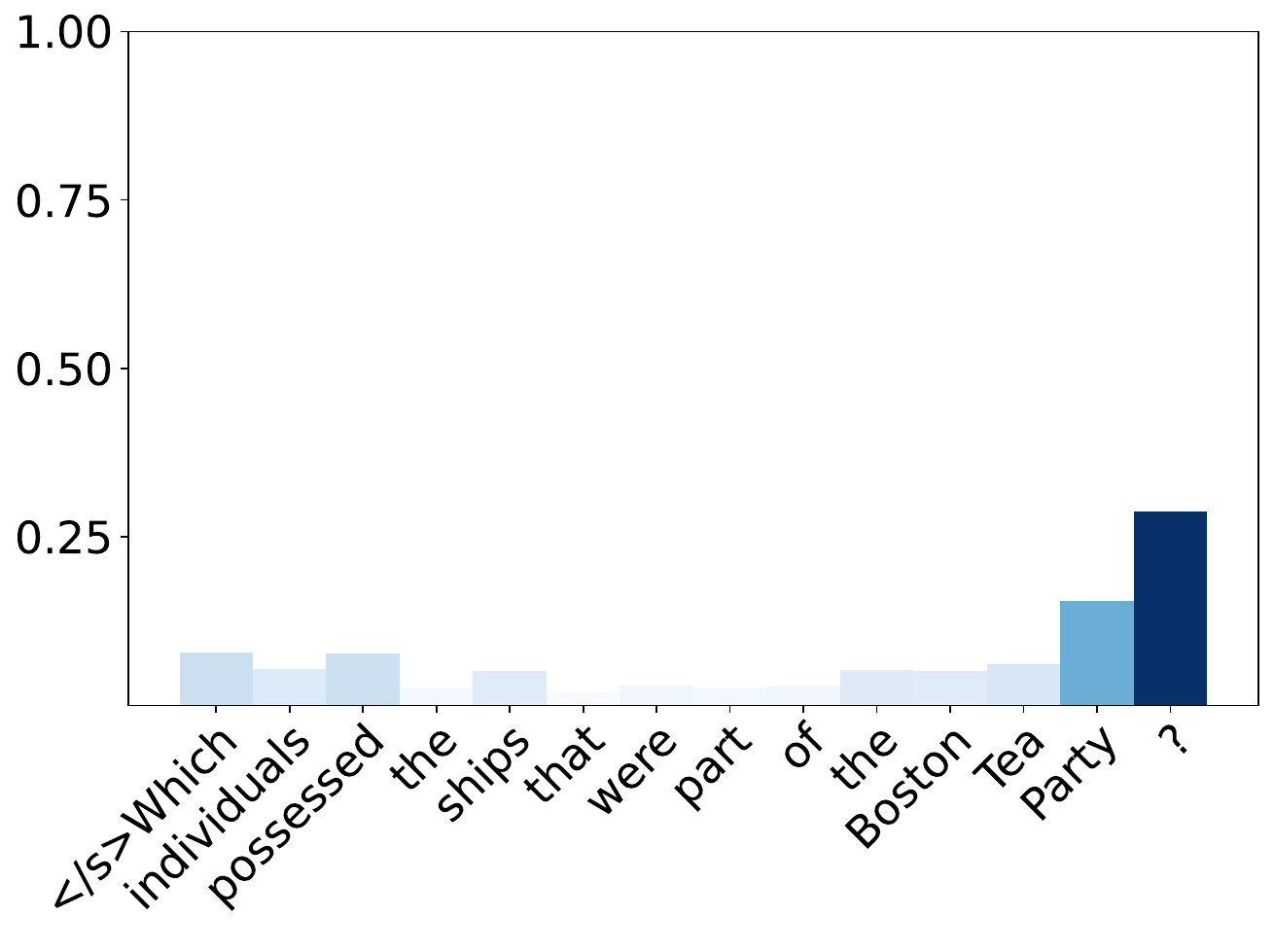}
            \caption[]%
            {{\small Before adding \tframed[line width=0.5bp,fill=vred]{\textcolor{white}{\texttt{\textbf{[PAUSE]}}}} tokens} to original prompt.}
            \label{fig:mean and std of net14}
        \end{subfigure}
        \hfill
        \begin{subfigure}[b]{0.45\textwidth}  
            \centering 
            \includegraphics[width=\textwidth,height=3cm]{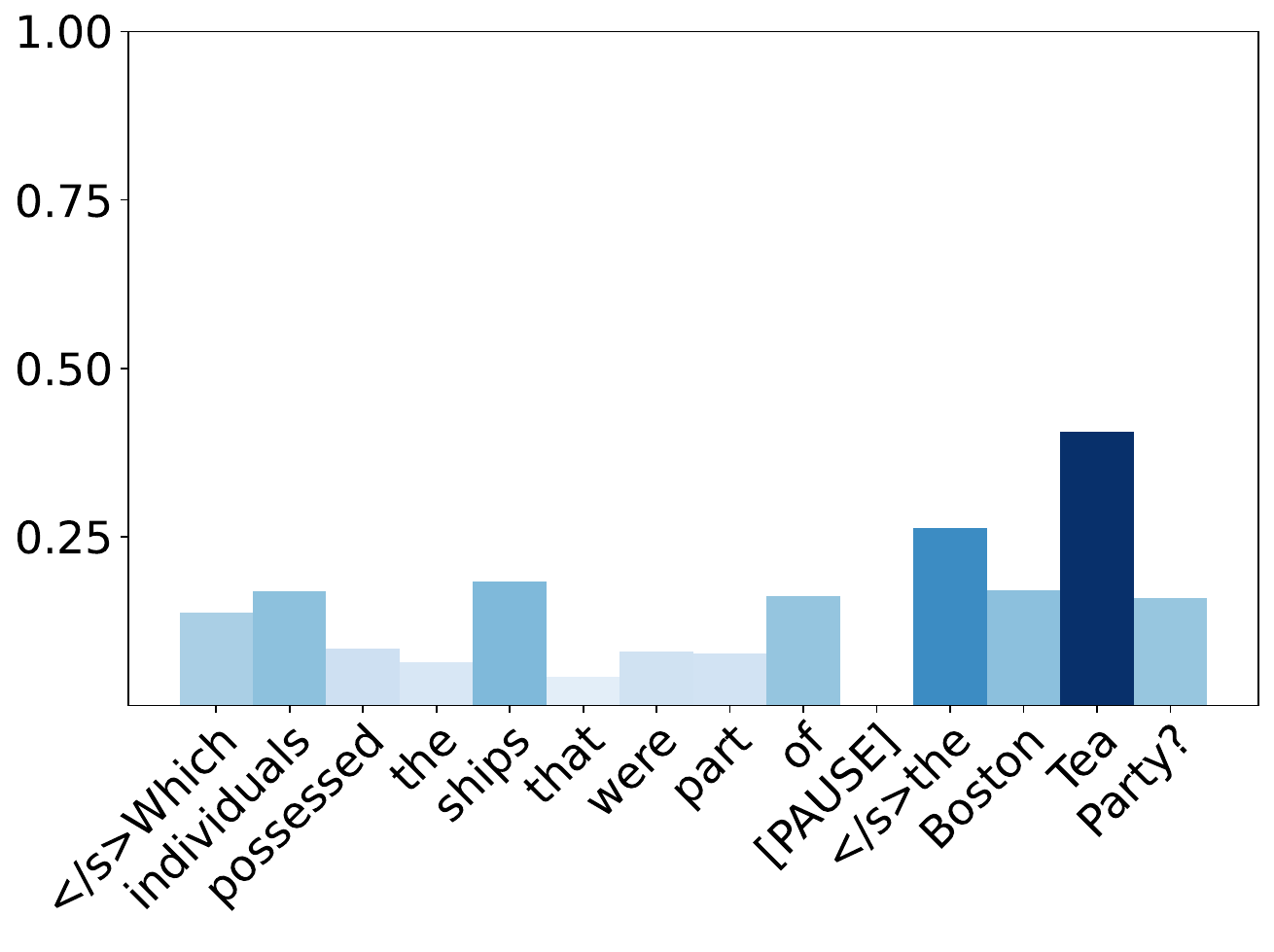}
            \caption[]%
            {{\small After adding \tframed[line width=0.5bp,fill=vred]{\textcolor{white}{\texttt{\textbf{[PAUSE]}}}} tokens} to original prompt.}    
            \label{fig:mean and std of net24}
        \end{subfigure}
        \hfill
        \vskip\baselineskip
        \begin{subfigure}[b]{0.45\textwidth}   
            \centering 
            \includegraphics[width=\textwidth,height=3cm]{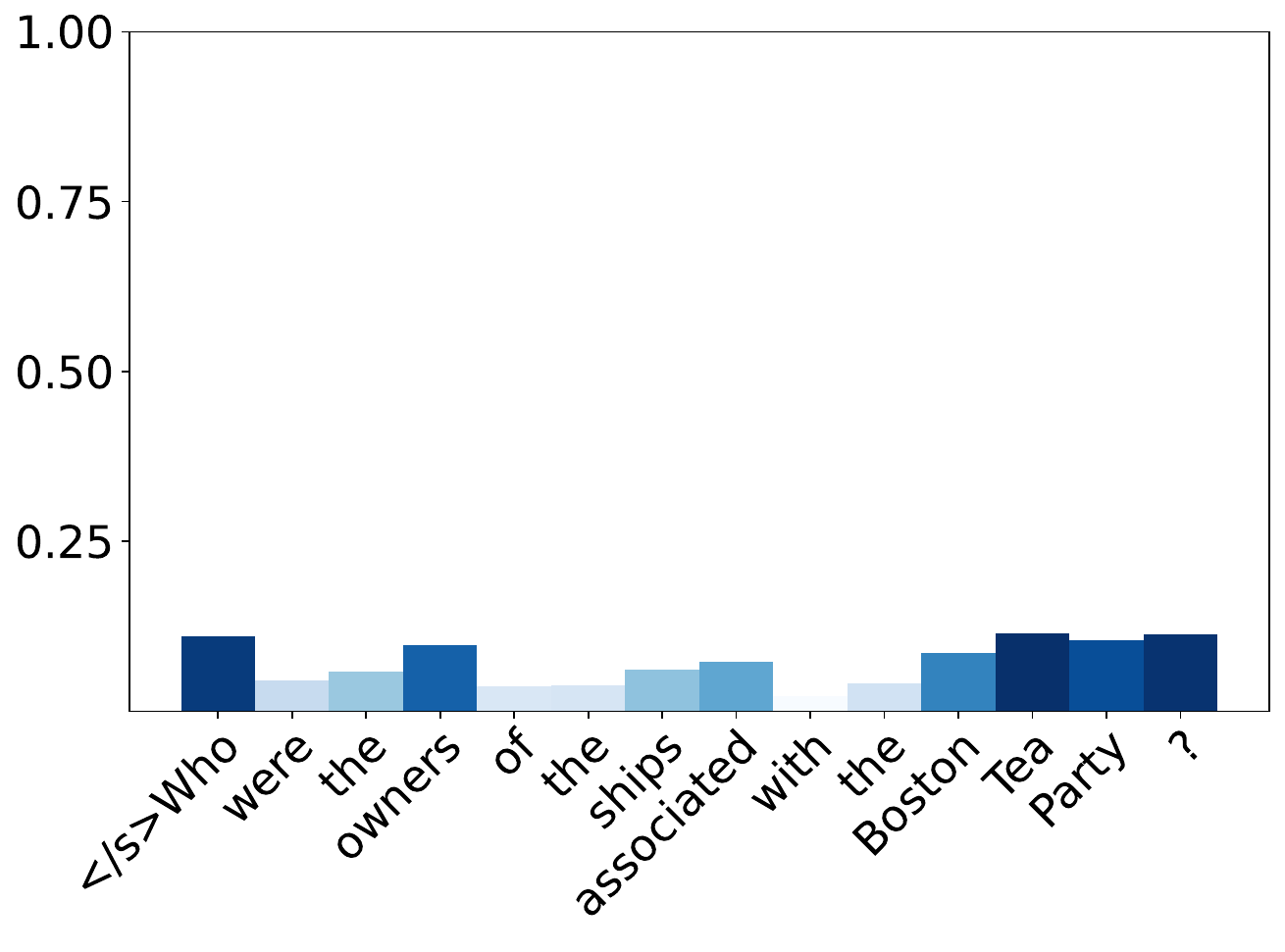}
            \caption[]%
            {{\small Before adding \tframed[line width=0.5bp,fill=vred]{\textcolor{white}{\texttt{\textbf{[PAUSE]}}}} tokens} to paraphrase 1.}    
            \label{fig:mean and std of net34}
        \end{subfigure}
        \hfill
        \begin{subfigure}[b]{0.45\textwidth}   
            \centering 
            \includegraphics[width=\textwidth,height=3cm]{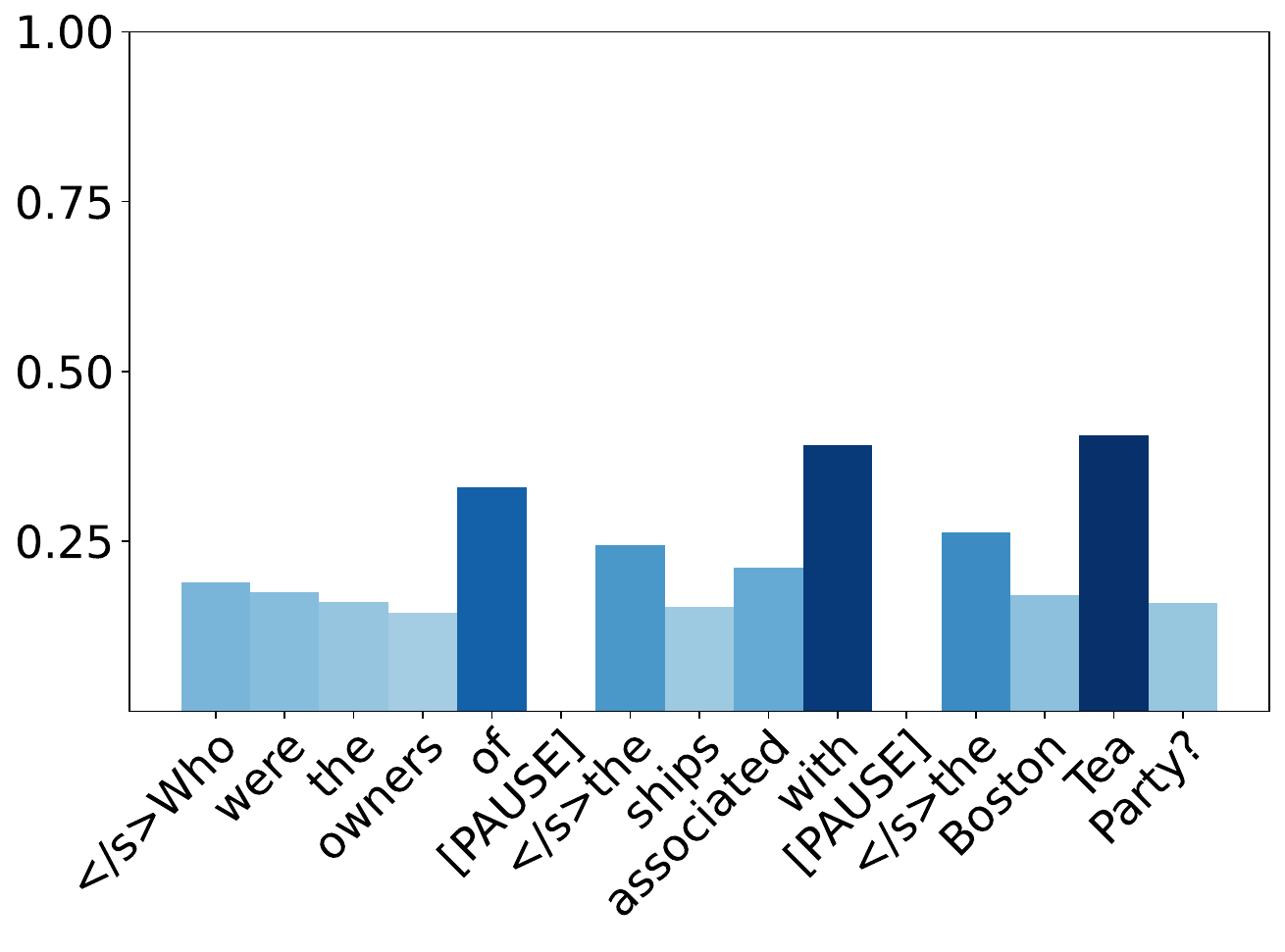}
            \caption[]%
            {{\small After adding \tframed[line width=0.5bp,fill=vred]{\textcolor{white}{\texttt{\textbf{[PAUSE]}}}} tokens} to paraphrase 1.}    
            \label{fig:mean and std of net44}
        \end{subfigure}
        \hfill
        \vskip\baselineskip
        \begin{subfigure}[b]{0.45\textwidth}   
            \centering 
            \includegraphics[width=\textwidth,height=3cm]{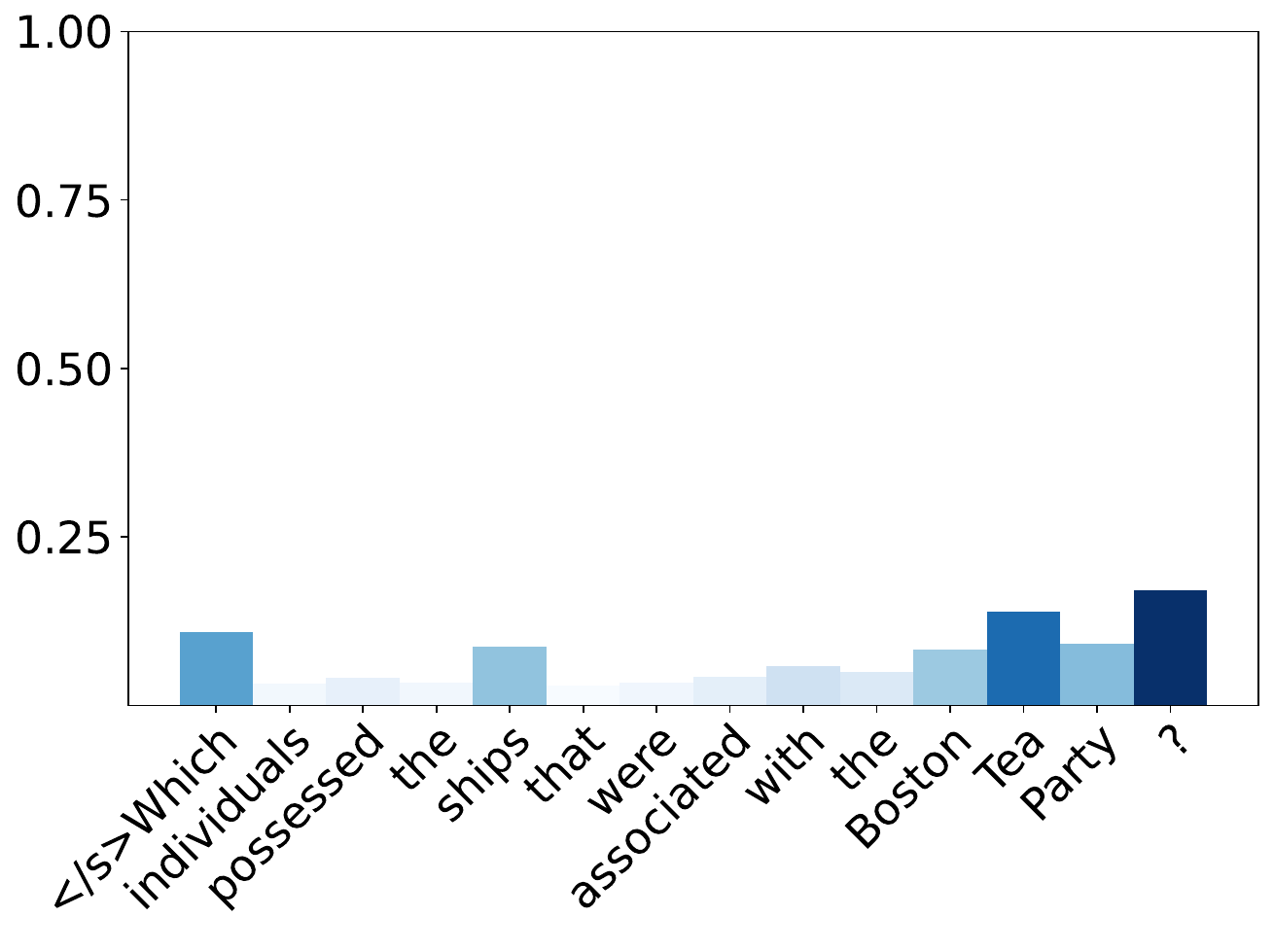}
            \caption[]%
            {{\small Before adding \tframed[line width=0.5bp,fill=vred]{\textcolor{white}{\texttt{\textbf{[PAUSE]}}}} tokens} to paraphrase 2.}
            \label{fig:mean and std of net34}
        \end{subfigure}
        \hfill
        \begin{subfigure}[b]{0.45\textwidth}   
            \centering 
            \includegraphics[width=\textwidth,height=3cm]{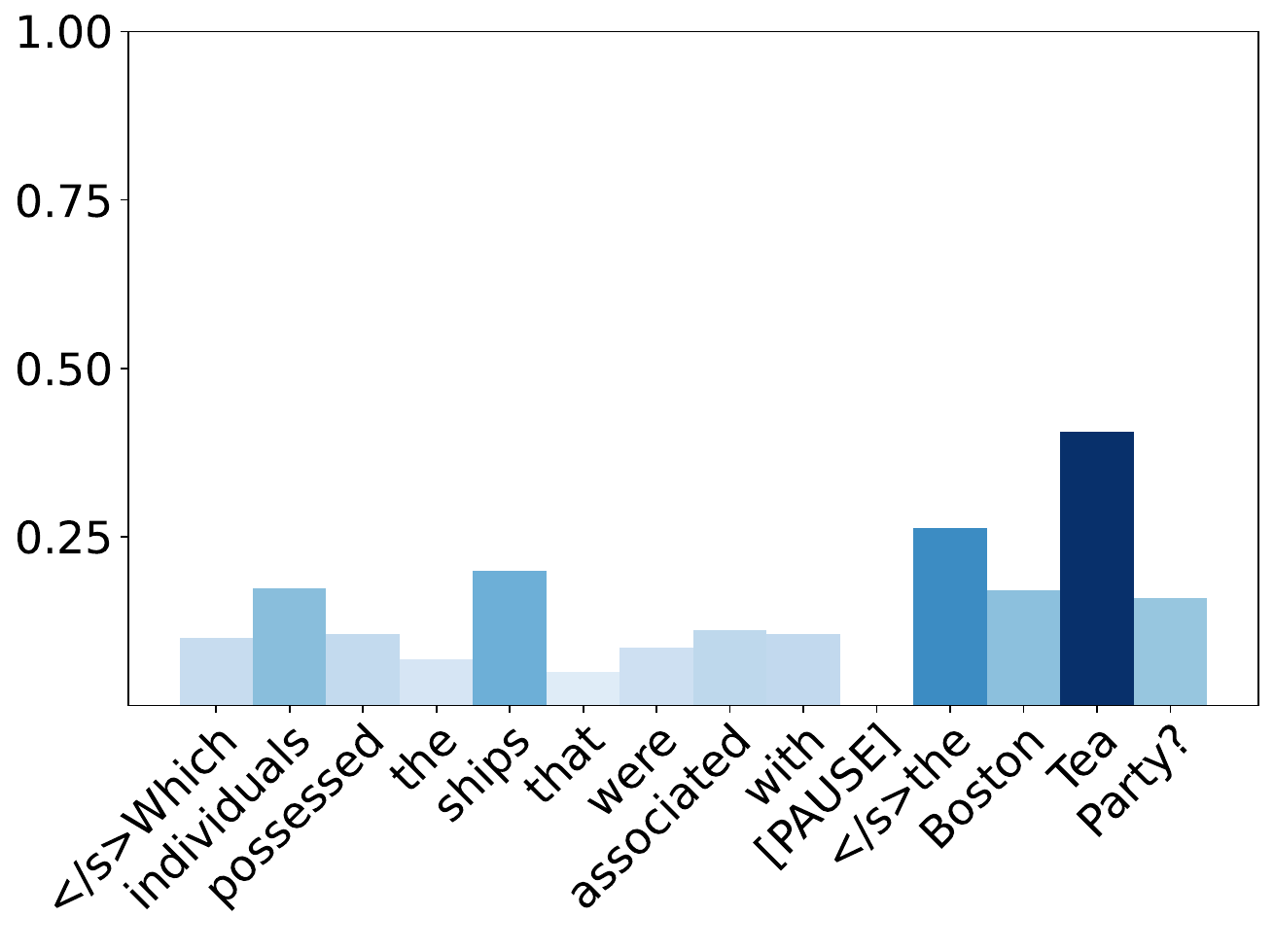}
            \caption[]%
            {{\small After adding \tframed[line width=0.5bp,fill=vred]{\textcolor{white}{\texttt{\textbf{[PAUSE]}}}} tokens} to paraphrase 2.} 
            \label{fig:mean and std of net44}
        \end{subfigure}
        \hfill
        \vskip\baselineskip
        \begin{subfigure}[b]{0.45\textwidth}   
            \centering 
            \includegraphics[width=\textwidth,height=3cm]{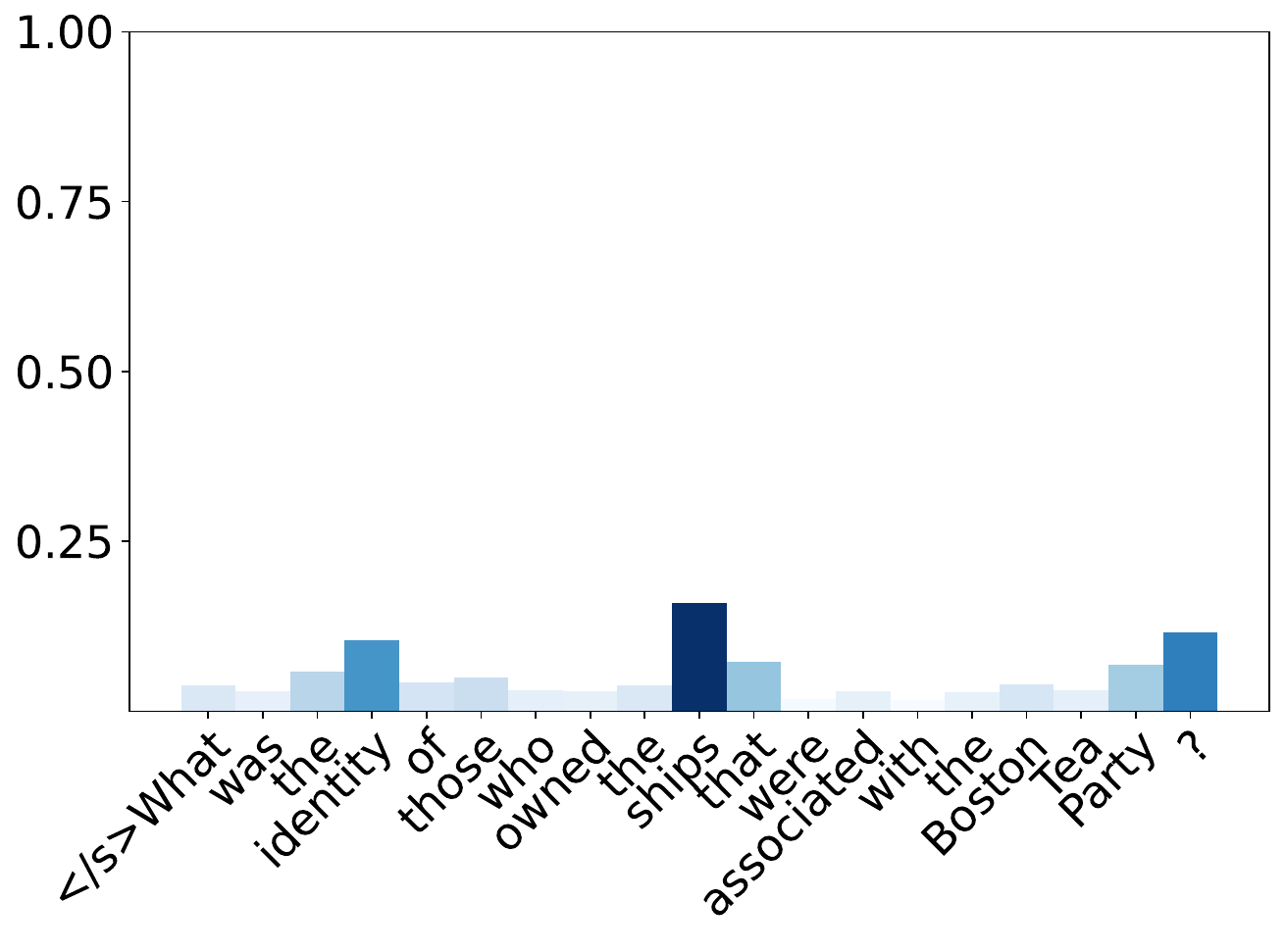}
            \caption[]%
            {{\small Before adding \tframed[line width=0.5bp,fill=vred]{\textcolor{white}{\texttt{\textbf{[PAUSE]}}}} tokens} to paraphrase 3.}
            \label{fig:mean and std of net44}
        \end{subfigure}
        \hfill
        \begin{subfigure}[b]{0.45\textwidth}   
            \centering 
            \includegraphics[width=\textwidth,height=3cm]{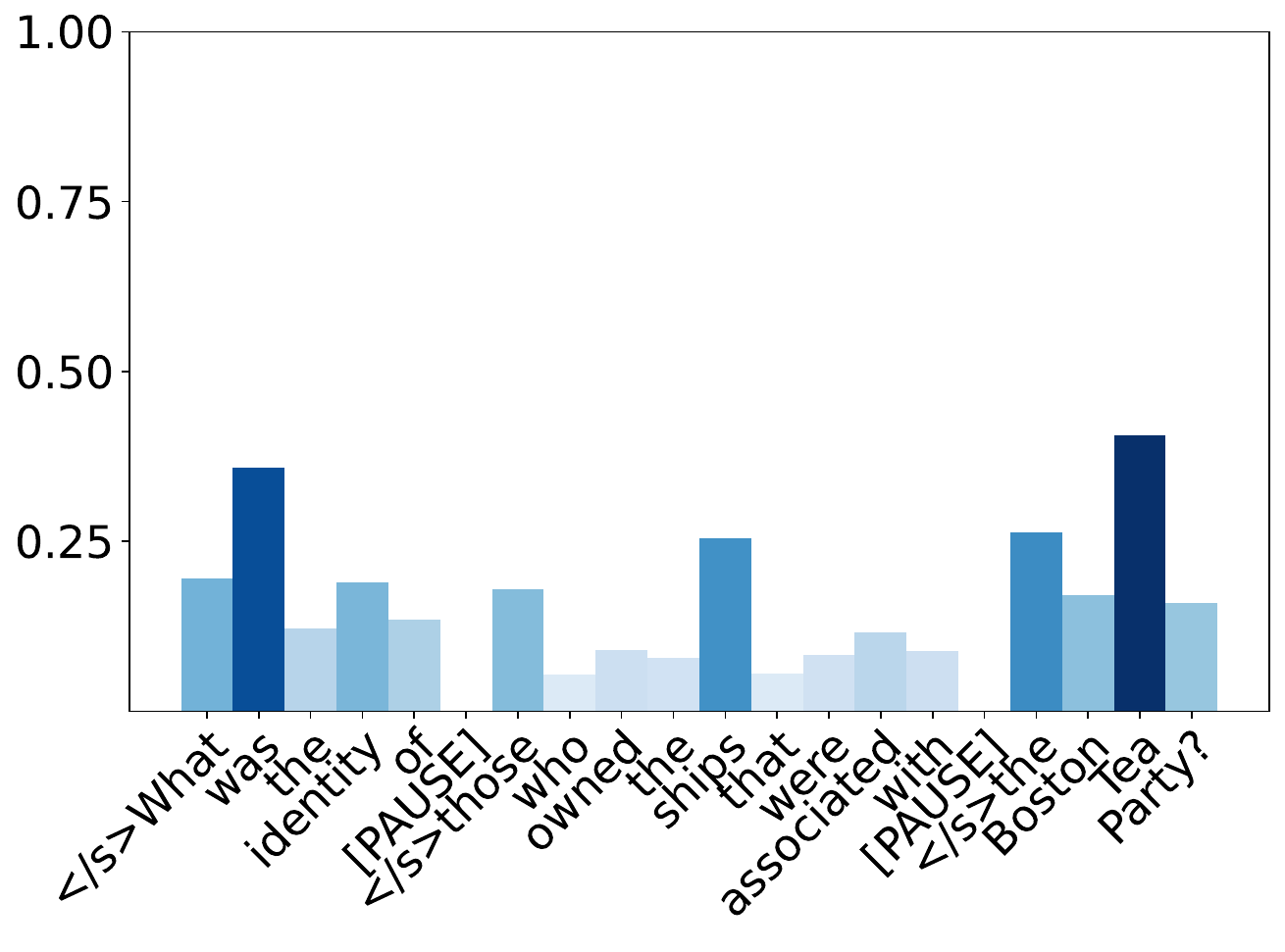}
            \caption[]%
            {{\small After adding \tframed[line width=0.5bp,fill=vred]{\textcolor{white}{\texttt{\textbf{[PAUSE]}}}} tokens} to paraphrase 3.}    
            \label{fig:mean and std of net44}
        \end{subfigure}
        \hfill
        \vskip\baselineskip
        \begin{subfigure}[b]{0.45\textwidth}   
            \centering 
            \includegraphics[width=\textwidth,height=3cm]{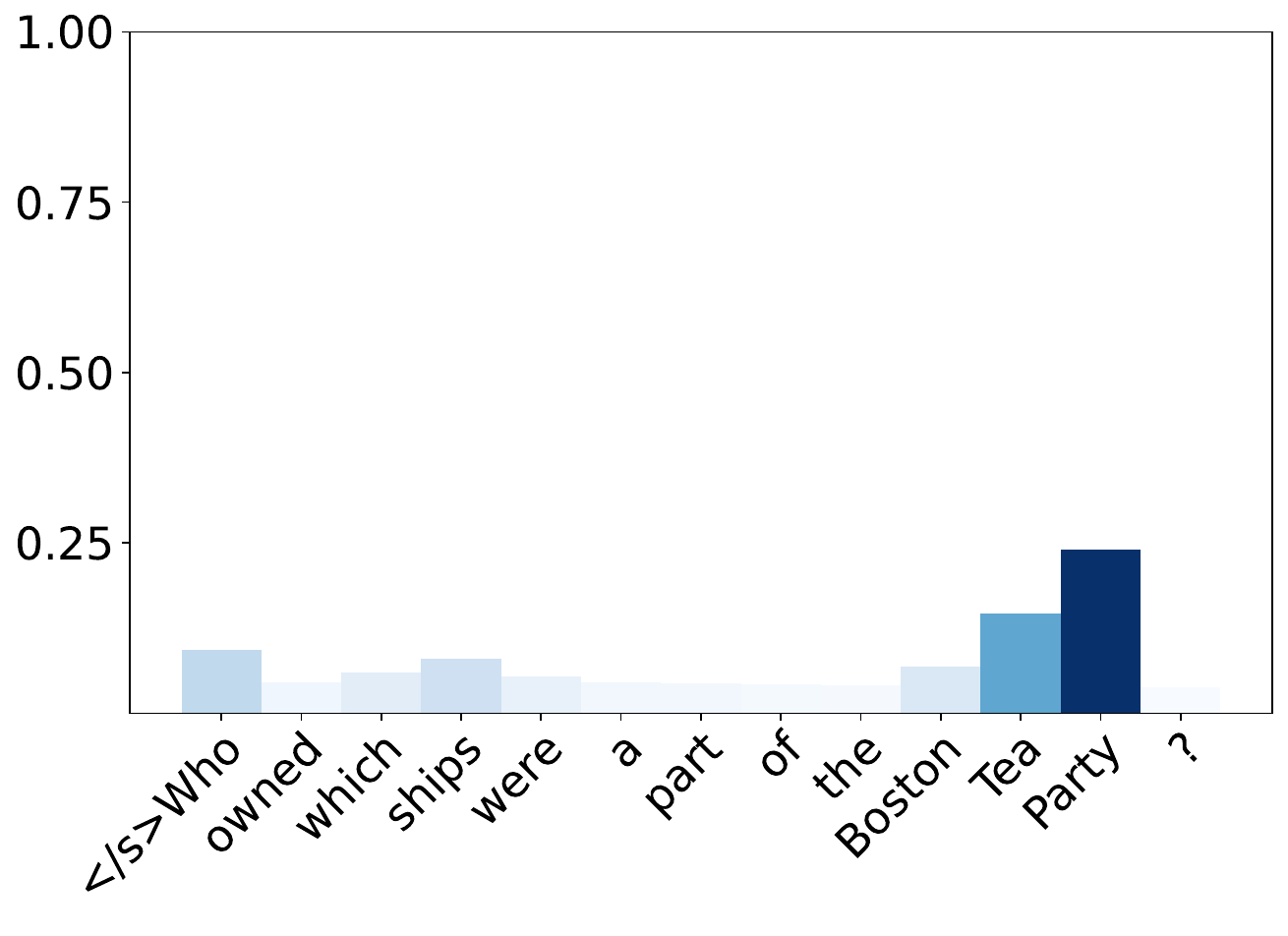}
            \caption[]%
            {{\small Before adding \tframed[line width=0.5bp,fill=vred]{\textcolor{white}{\texttt{\textbf{[PAUSE]}}}} tokens} to paraphrase 4.}    
            \label{fig:mean and std of net44}
        \end{subfigure}
        \hfill
        \begin{subfigure}[b]{0.45\textwidth}   
            \centering 
            \includegraphics[width=\textwidth,height=3cm]{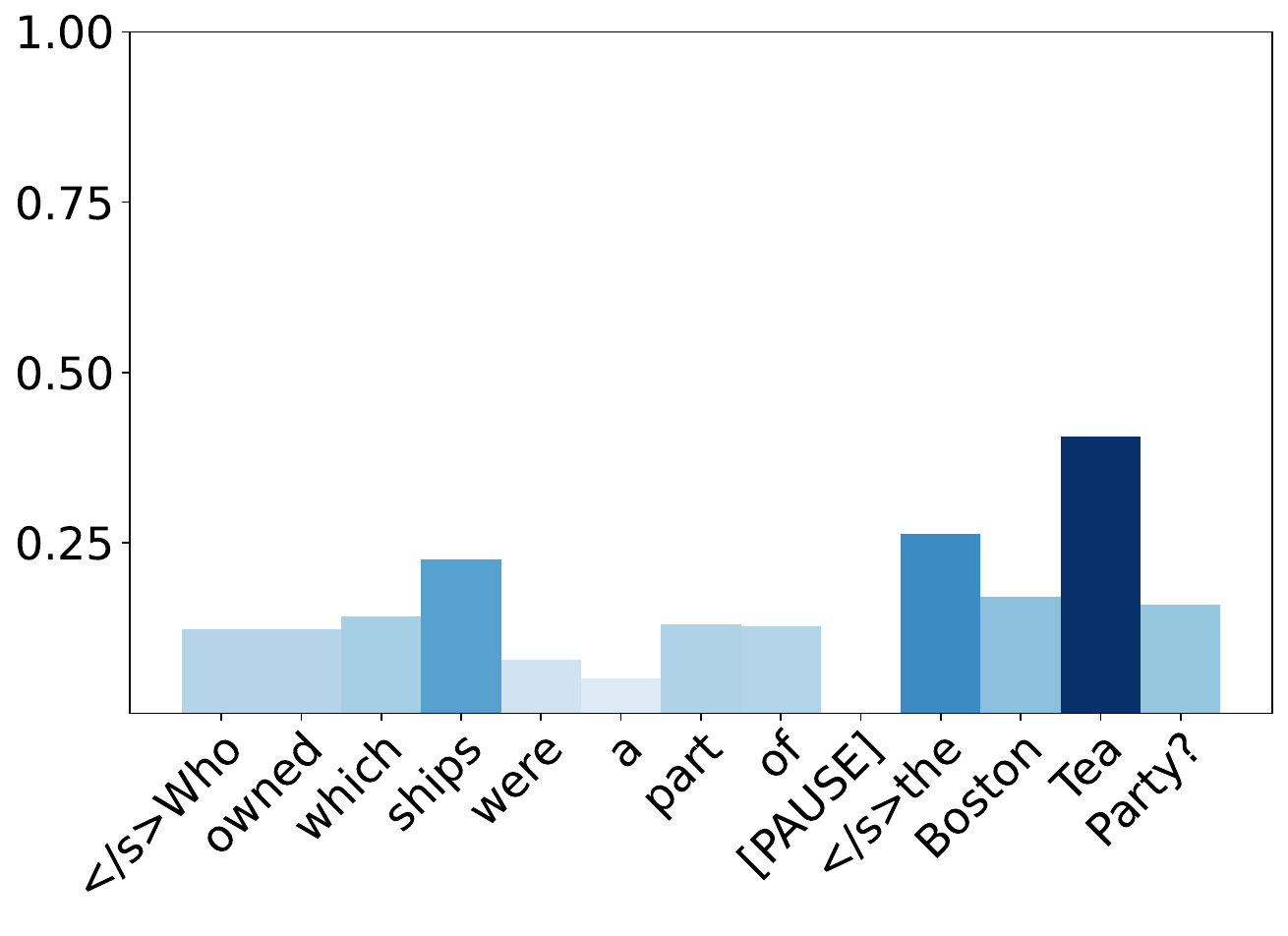}
            \caption[]%
            {{\small After adding \tframed[line width=0.5bp,fill=vred]{\textcolor{white}{\texttt{\textbf{[PAUSE]}}}} tokens} to paraphrase 4.}    
            \label{fig:mean and std of net44}
        \end{subfigure}
        \hfill
        \vskip\baselineskip
        \begin{subfigure}[b]{0.45\textwidth}   
            \centering 
            \includegraphics[width=\textwidth,height=3cm]{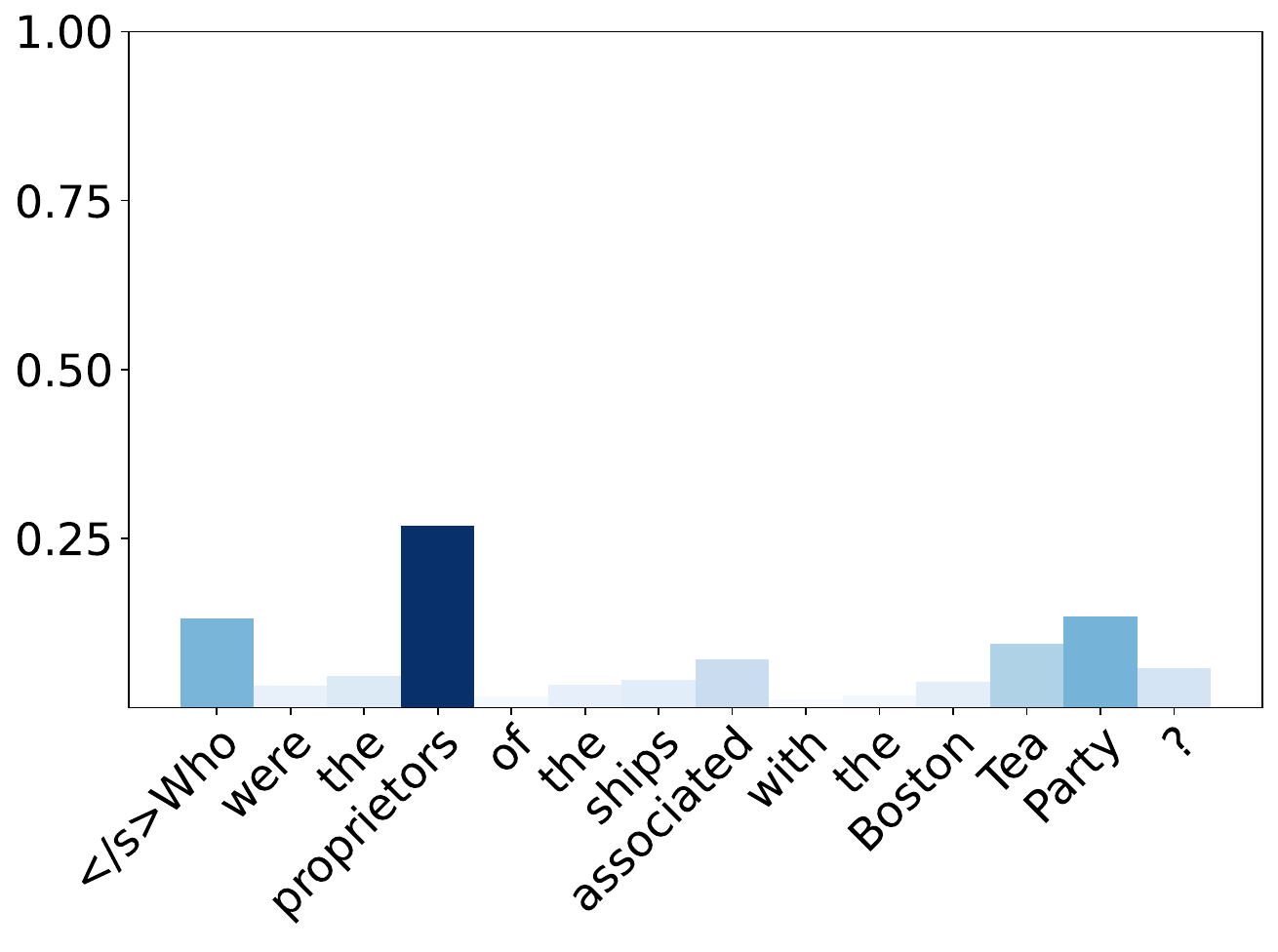}
            \caption[]%
            {{\small Before adding \tframed[line width=0.5bp,fill=vred]{\textcolor{white}{\texttt{\textbf{[PAUSE]}}}} tokens} to paraphrase 5.}    
            \label{fig:mean and std of net44}
        \end{subfigure}
        \hfill
        \begin{subfigure}[b]{0.45\textwidth}   
            \centering 
            \includegraphics[width=\textwidth,height=3cm]{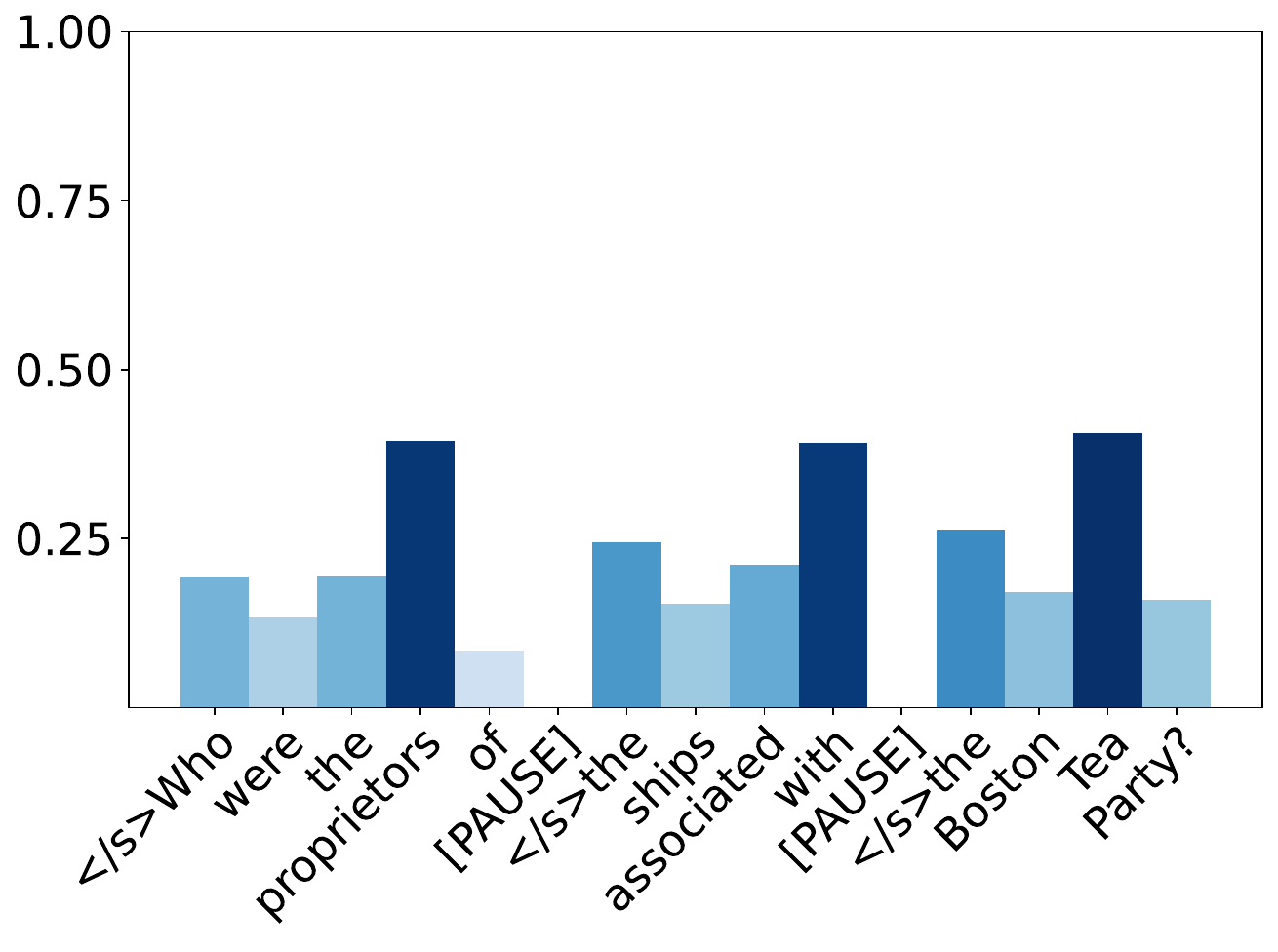}
            \caption[]%
            {{\small After adding \tframed[line width=0.5bp,fill=vred]{\textcolor{white}{\texttt{\textbf{[PAUSE]}}}} tokens} to paraphrase 5.}    
            \label{fig:mean and std of net44}
        \end{subfigure}
        \caption[]%
            {{\small The phrase \textbf{Boston Tea} gets more importance score after adding \tframed[line width=0.5bp,fill=vred]{\textcolor{white}{\texttt{\textbf{[PAUSE]}}}} token for OPT.}}   
        \label{fig:OPT}
\end{figure*}

\begin{figure*}[!ht]
        \centering
        \begin{subfigure}[b]{0.45\textwidth}
            \centering
            \includegraphics[width=\textwidth,height=3cm]{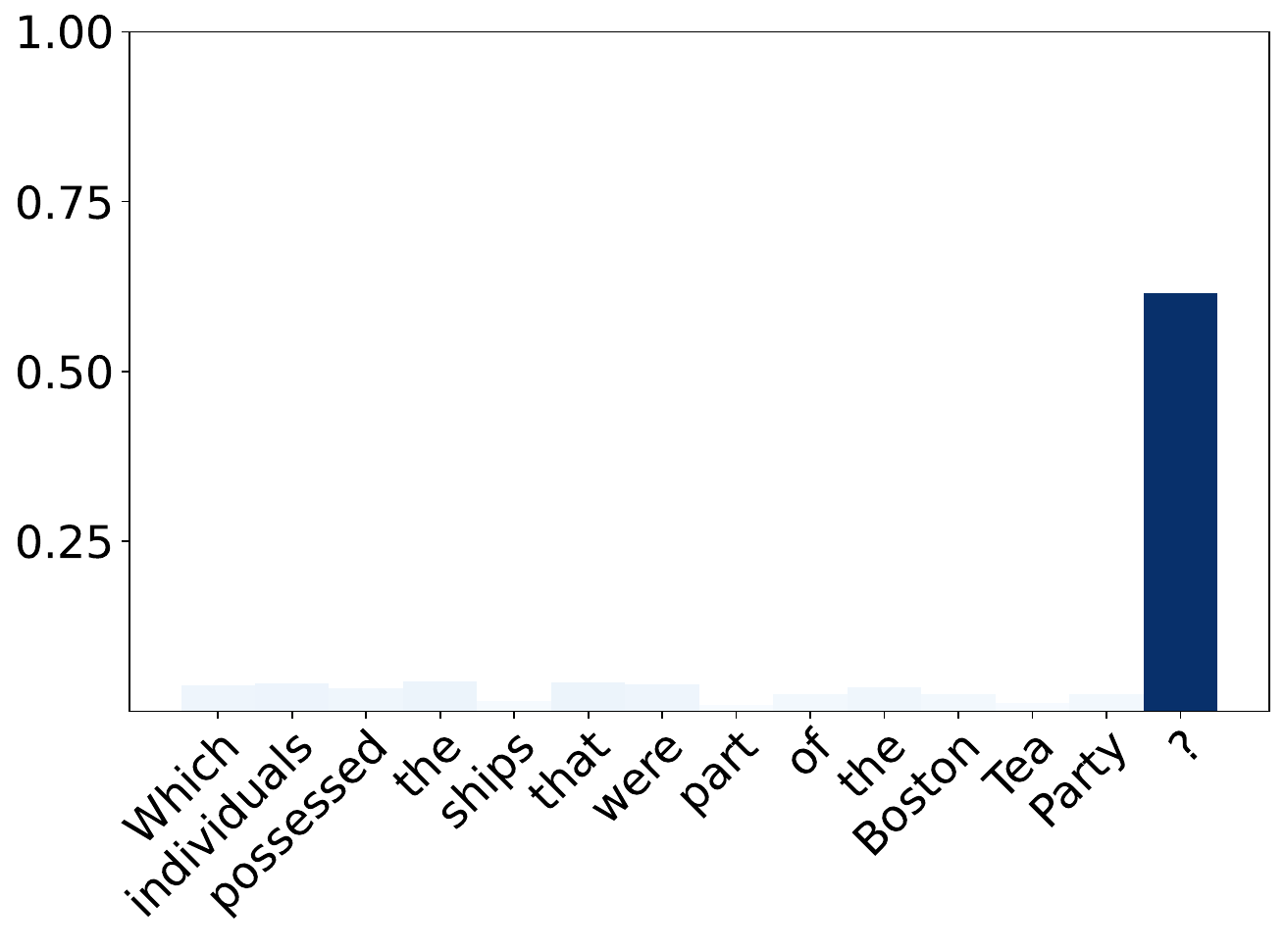}
            \caption[]%
            {{\small Before adding \tframed[line width=0.5bp,fill=vred]{\textcolor{white}{\texttt{\textbf{[PAUSE]}}}} tokens} to original prompt.}
            \label{fig:mean and std of net14}
        \end{subfigure}
        \hfill
        \begin{subfigure}[b]{0.45\textwidth}  
            \centering 
            \includegraphics[width=\textwidth,height=3cm]{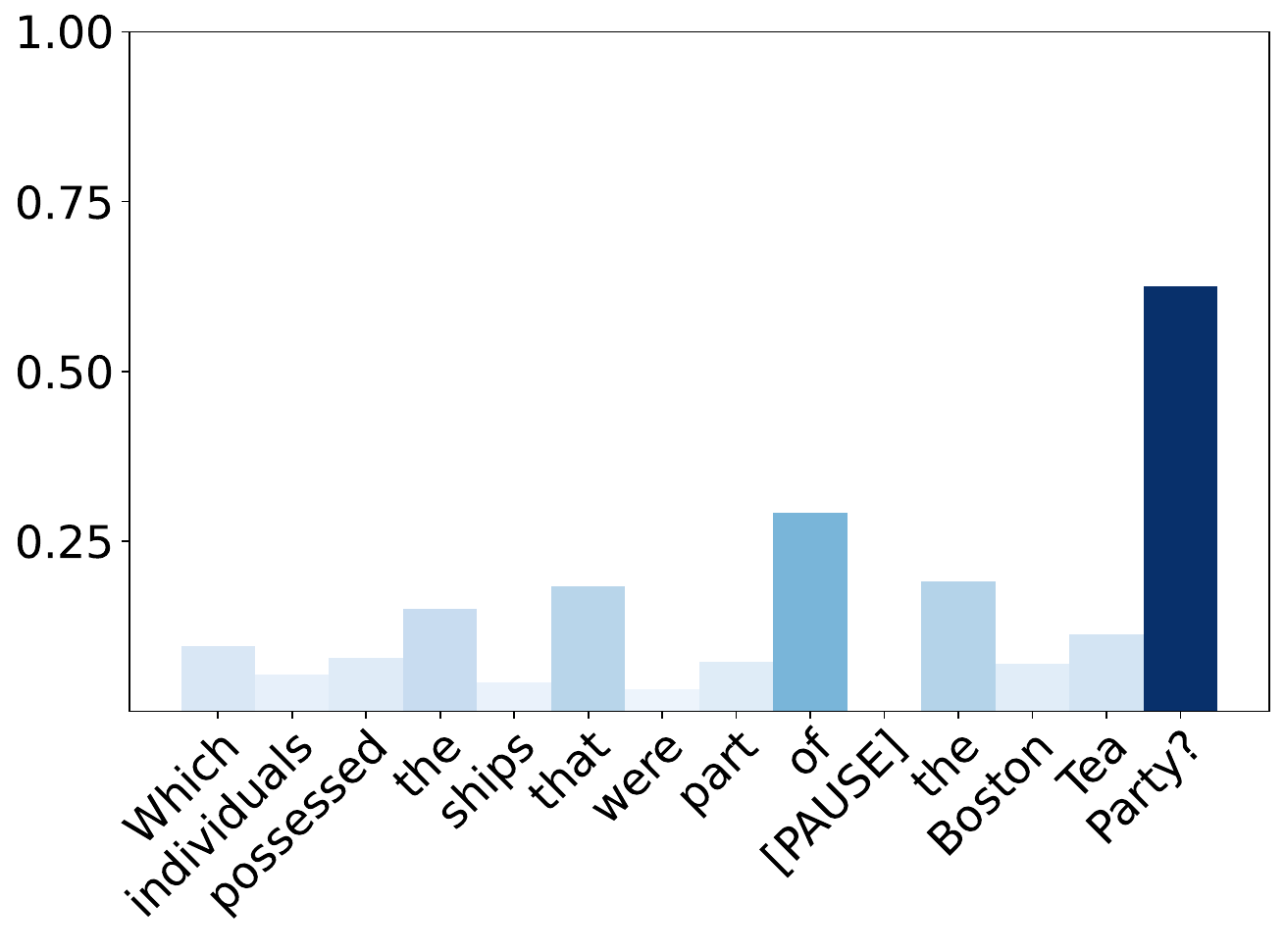}
            \caption[]%
            {{\small After adding \tframed[line width=0.5bp,fill=vred]{\textcolor{white}{\texttt{\textbf{[PAUSE]}}}} tokens} to original prompt.}    
            \label{fig:mean and std of net24}
        \end{subfigure}
        \hfill
        \vskip\baselineskip
        \begin{subfigure}[b]{0.45\textwidth}   
            \centering 
            \includegraphics[width=\textwidth,height=3cm]{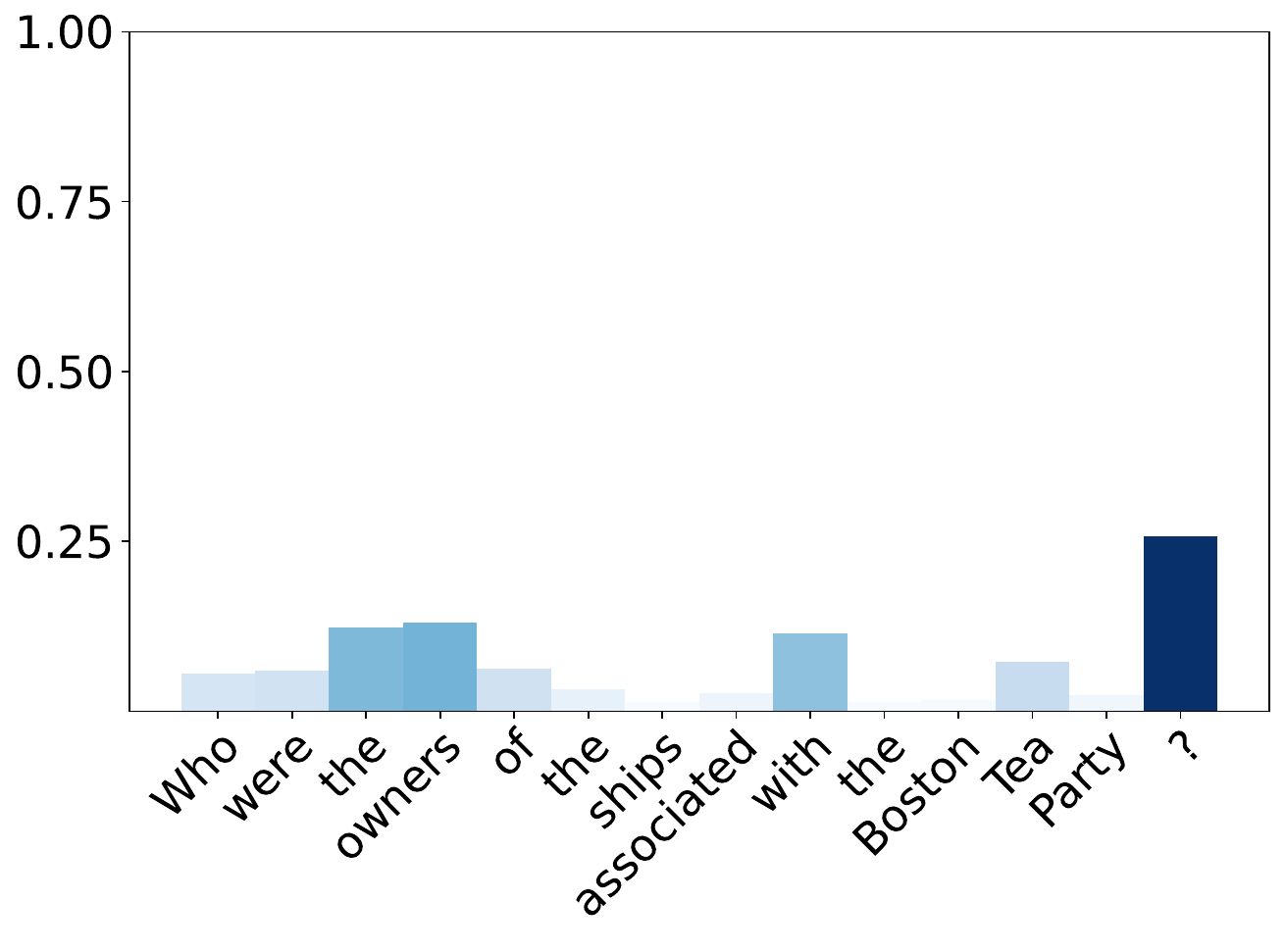}
            \caption[]%
            {{\small Before adding \tframed[line width=0.5bp,fill=vred]{\textcolor{white}{\texttt{\textbf{[PAUSE]}}}} tokens} to paraphrase 1.}    
            \label{fig:mean and std of net34}
        \end{subfigure}
        \hfill
        \begin{subfigure}[b]{0.45\textwidth}   
            \centering 
            \includegraphics[width=\textwidth,height=3cm]{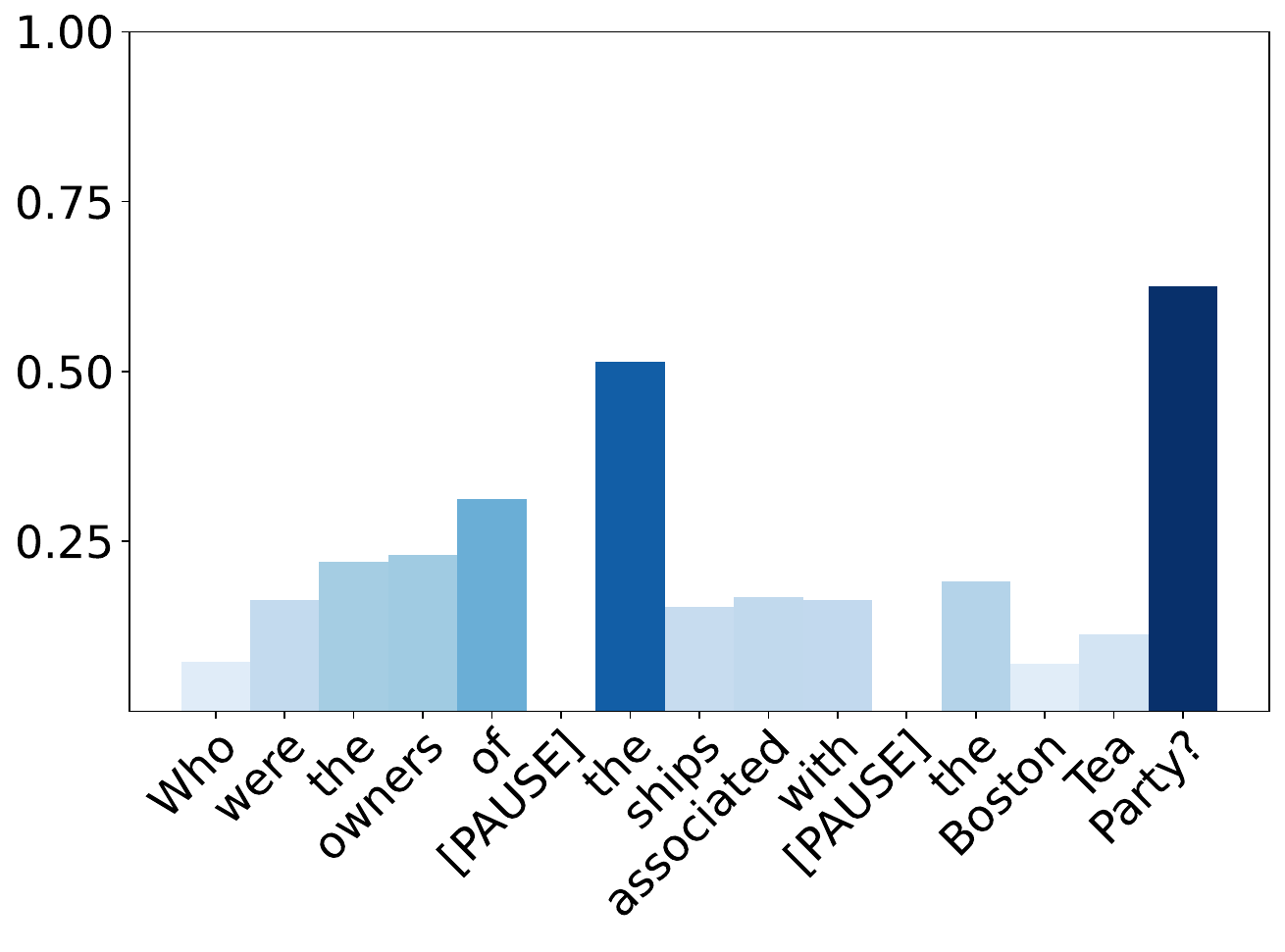}
            \caption[]%
            {{\small After adding \tframed[line width=0.5bp,fill=vred]{\textcolor{white}{\texttt{\textbf{[PAUSE]}}}} tokens} to paraphrase 1.}    
            \label{fig:mean and std of net44}
        \end{subfigure}
        \hfill
        \vskip\baselineskip
        \begin{subfigure}[b]{0.45\textwidth}   
            \centering 
            \includegraphics[width=\textwidth,height=3cm]{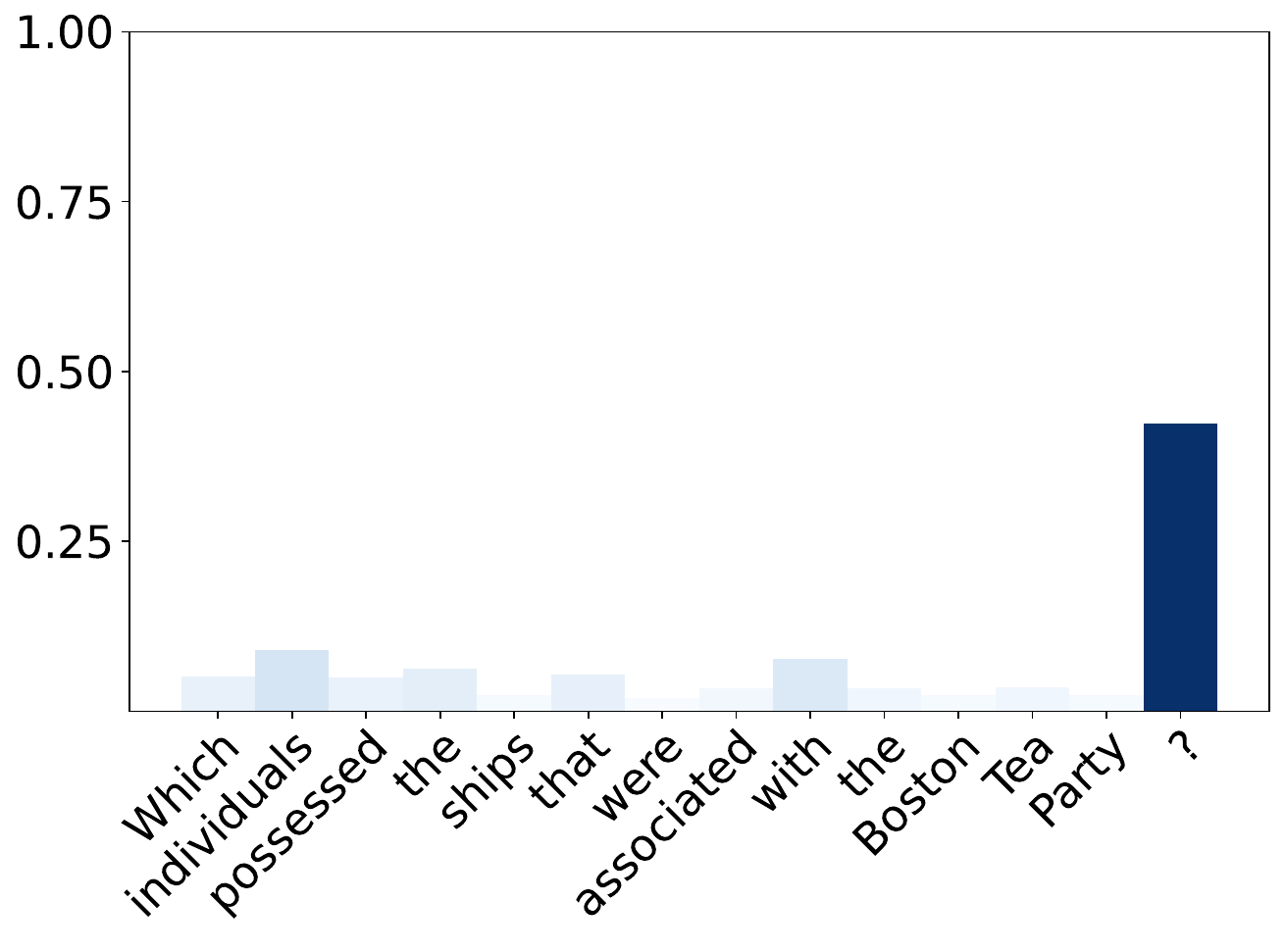}
            \caption[]%
            {{\small Before adding \tframed[line width=0.5bp,fill=vred]{\textcolor{white}{\texttt{\textbf{[PAUSE]}}}} tokens} to paraphrase 2.}
            \label{fig:mean and std of net34}
        \end{subfigure}
        \hfill
        \begin{subfigure}[b]{0.45\textwidth}   
            \centering 
            \includegraphics[width=\textwidth,height=3cm]{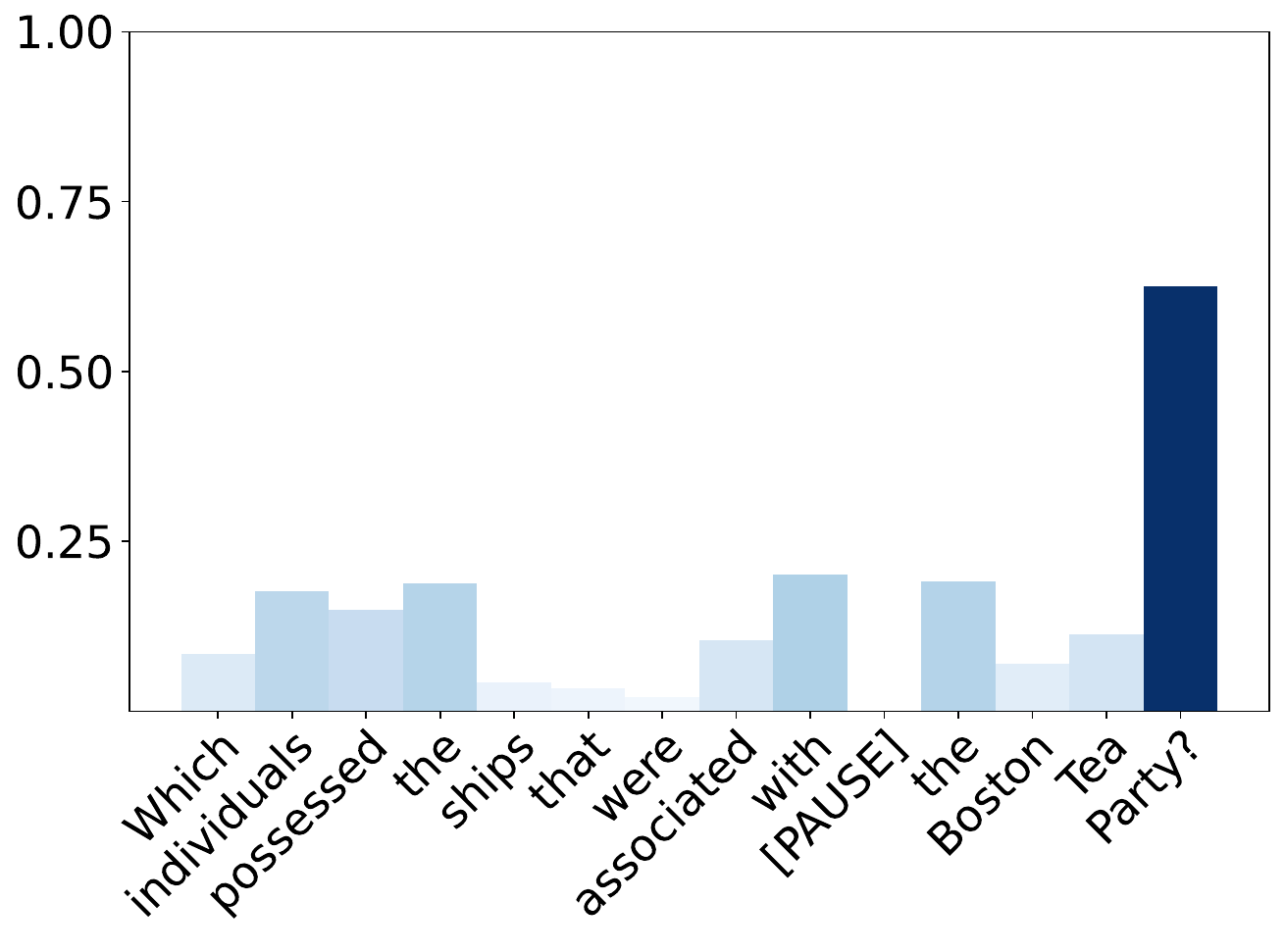}
            \caption[]%
            {{\small After adding \tframed[line width=0.5bp,fill=vred]{\textcolor{white}{\texttt{\textbf{[PAUSE]}}}} tokens} to paraphrase 2.} 
            \label{fig:mean and std of net44}
        \end{subfigure}
        \hfill
        \vskip\baselineskip
        \begin{subfigure}[b]{0.45\textwidth}   
            \centering 
            \includegraphics[width=\textwidth,height=3cm]{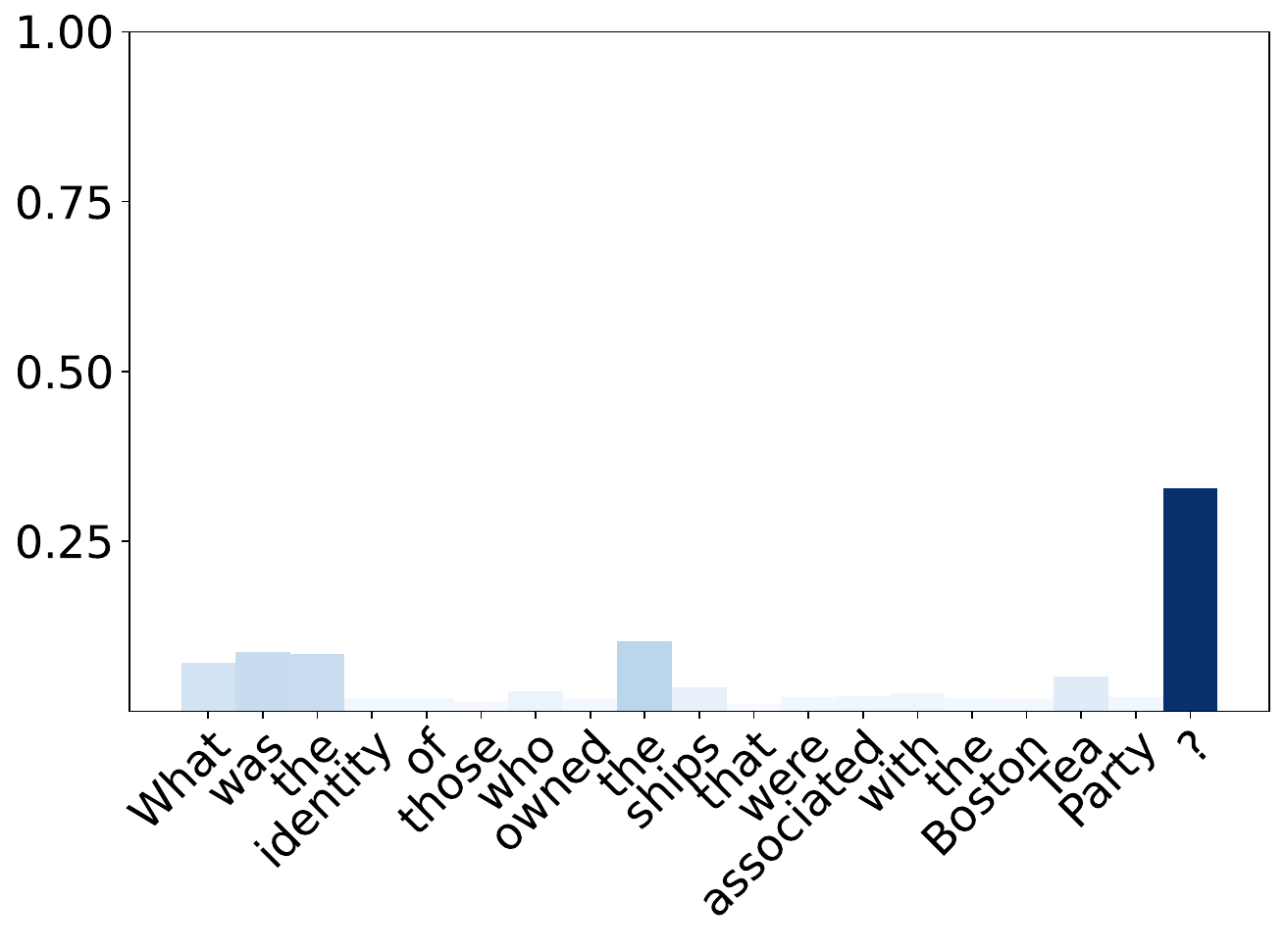}
            \caption[]%
            {{\small Before adding \tframed[line width=0.5bp,fill=vred]{\textcolor{white}{\texttt{\textbf{[PAUSE]}}}} tokens} to paraphrase 3.}
            \label{fig:mean and std of net44}
        \end{subfigure}
        \hfill
        \begin{subfigure}[b]{0.45\textwidth}   
            \centering 
            \includegraphics[width=\textwidth,height=3cm]{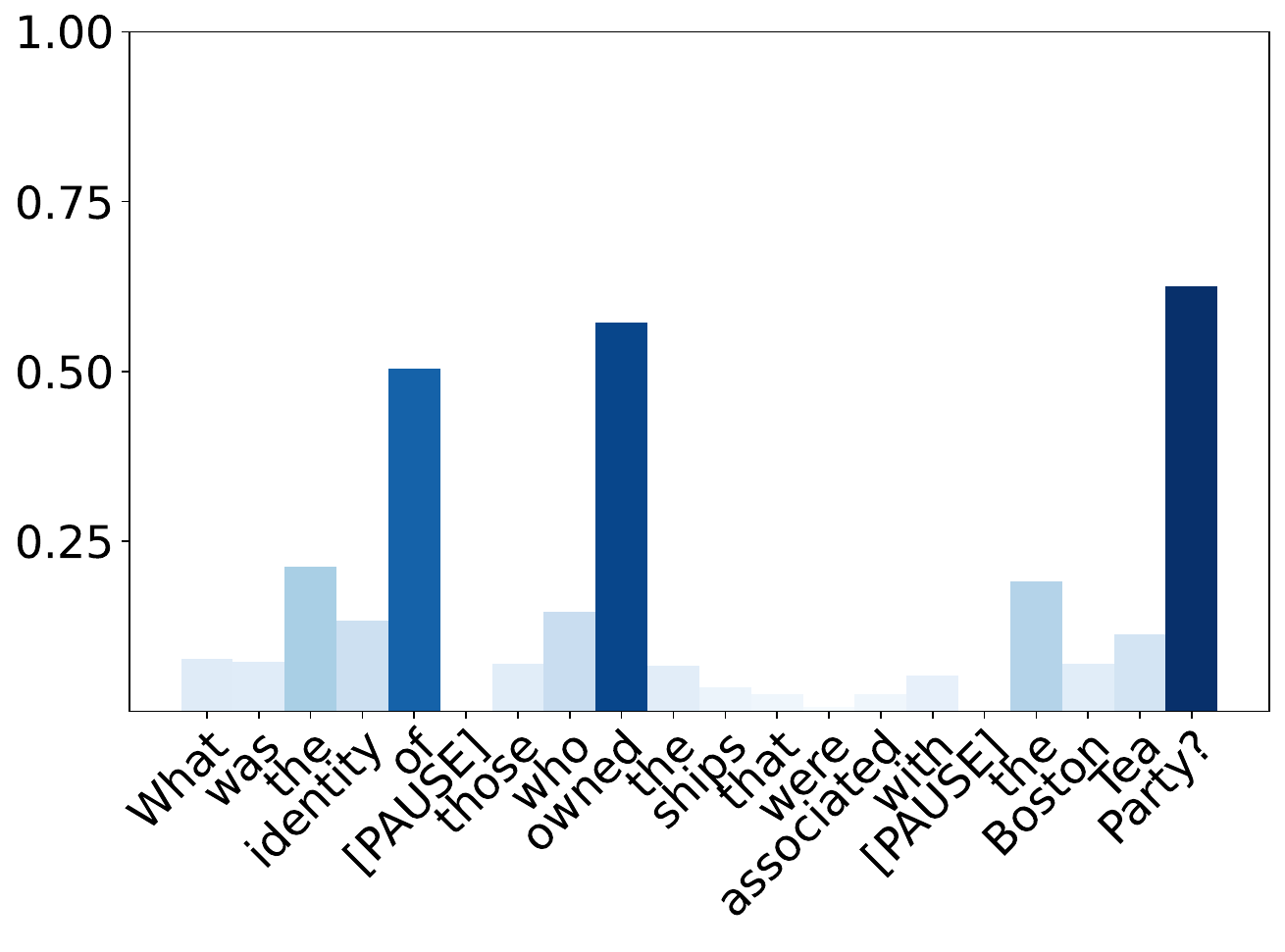}
            \caption[]%
            {{\small After adding \tframed[line width=0.5bp,fill=vred]{\textcolor{white}{\texttt{\textbf{[PAUSE]}}}} tokens} to paraphrase 3.}    
            \label{fig:mean and std of net44}
        \end{subfigure}
        \hfill
        \vskip\baselineskip
        \begin{subfigure}[b]{0.45\textwidth}   
            \centering 
            \includegraphics[width=\textwidth,height=3cm]{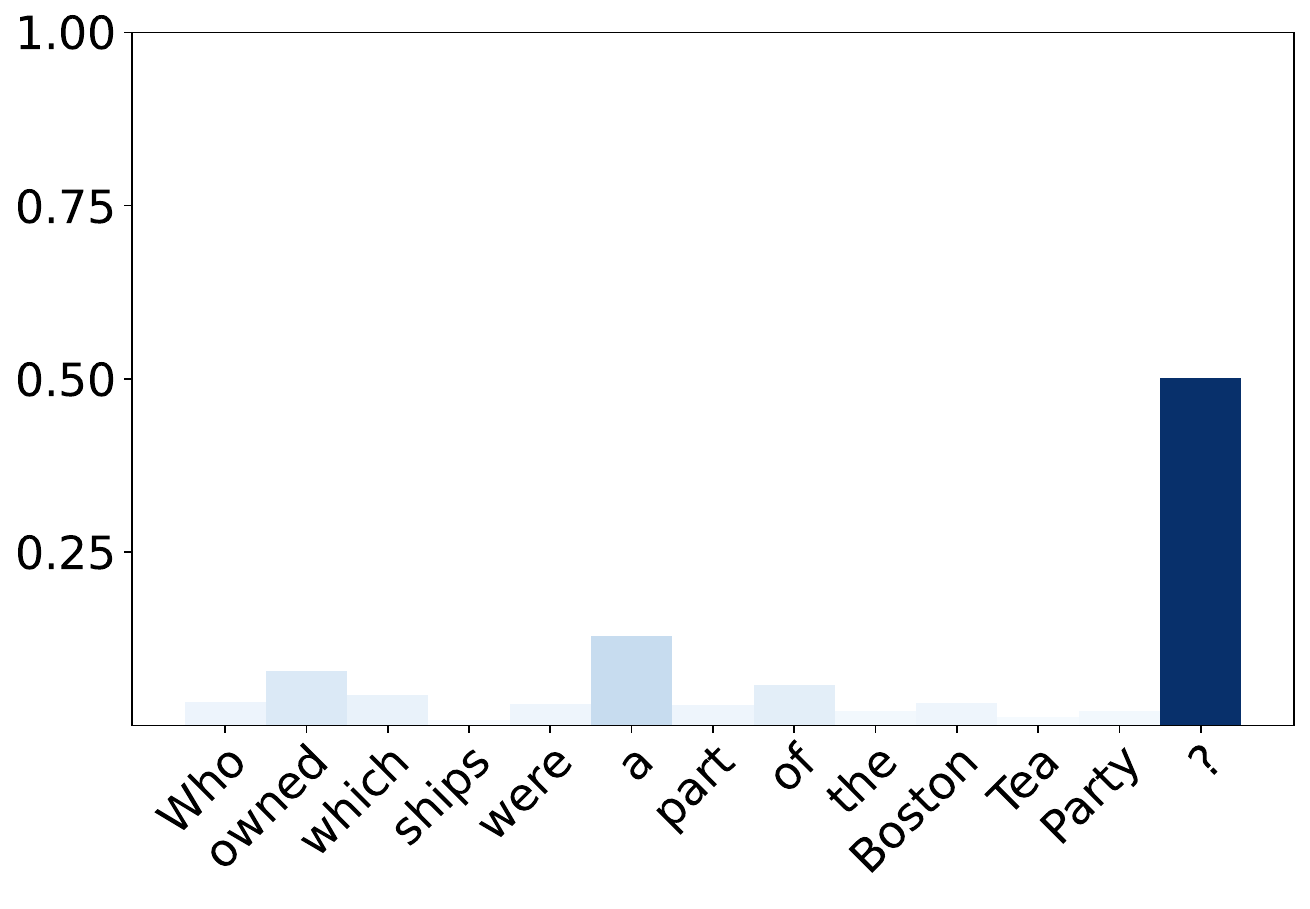}
            \caption[]%
            {{\small Before adding \tframed[line width=0.5bp,fill=vred]{\textcolor{white}{\texttt{\textbf{[PAUSE]}}}} tokens} to paraphrase 4.}    
            \label{fig:mean and std of net44}
        \end{subfigure}
        \hfill
        \begin{subfigure}[b]{0.45\textwidth}   
            \centering 
            \includegraphics[width=\textwidth,height=3cm]{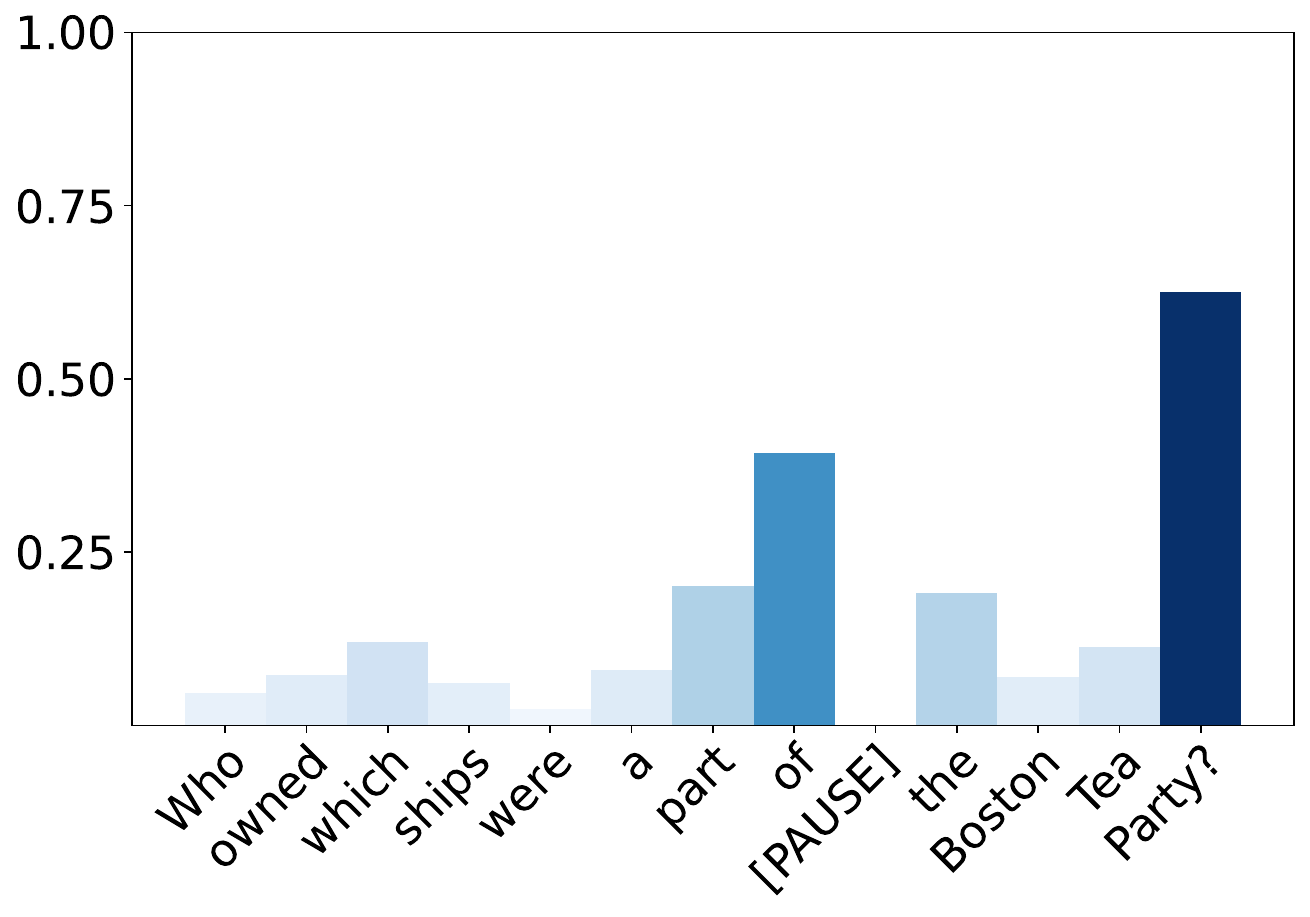}
            \caption[]%
            {{\small After adding \tframed[line width=0.5bp,fill=vred]{\textcolor{white}{\texttt{\textbf{[PAUSE]}}}} tokens} to paraphrase 4.}    
            \label{fig:mean and std of net44}
        \end{subfigure}
        \hfill
        \vskip\baselineskip
        \begin{subfigure}[b]{0.45\textwidth}   
            \centering 
            \includegraphics[width=\textwidth,height=3cm]{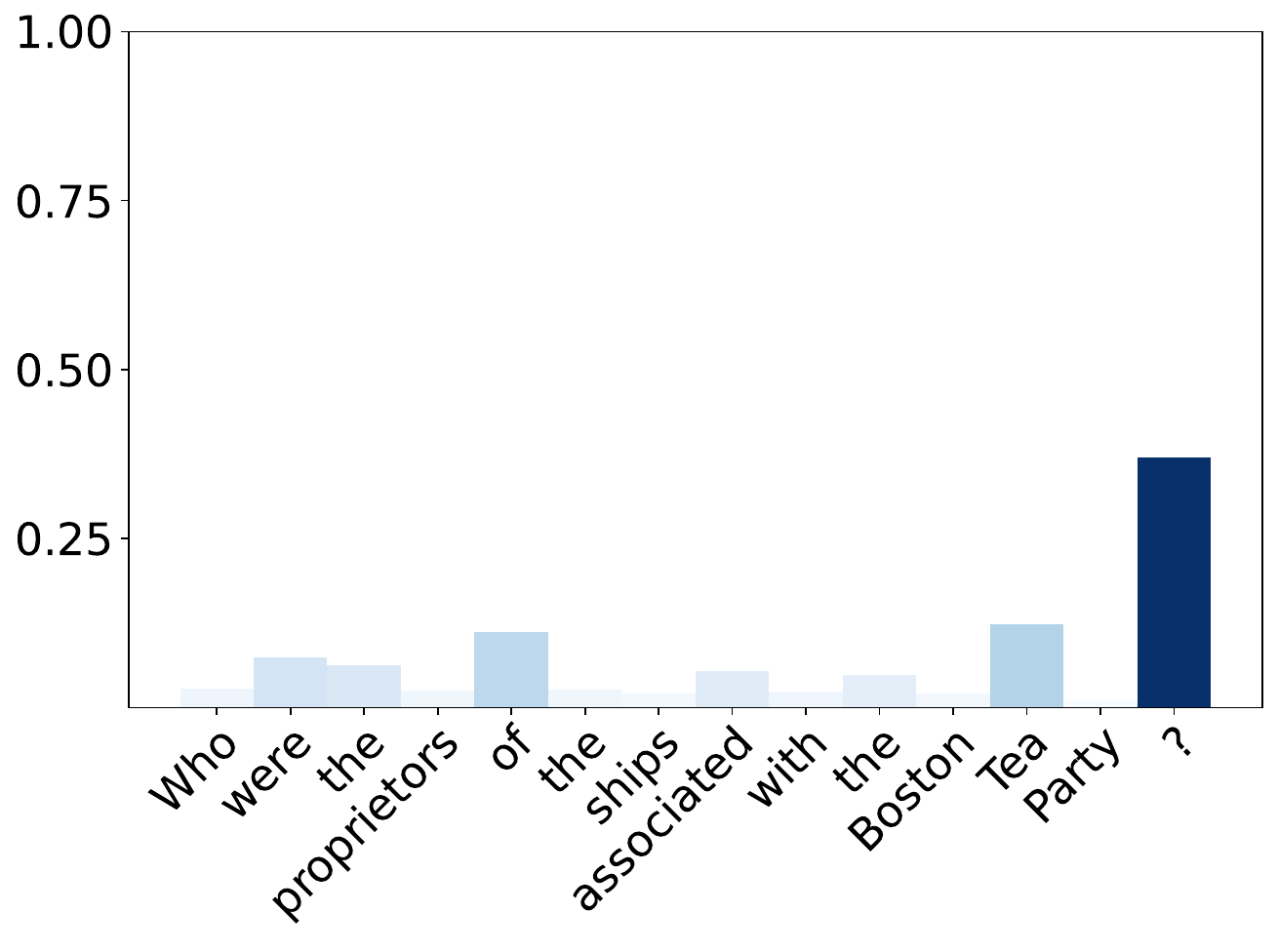}
            \caption[]%
            {{\small Before adding \tframed[line width=0.5bp,fill=vred]{\textcolor{white}{\texttt{\textbf{[PAUSE]}}}} tokens} to paraphrase 5.}    
            \label{fig:mean and std of net44}
        \end{subfigure}
        \hfill
        \begin{subfigure}[b]{0.45\textwidth}   
            \centering 
            \includegraphics[width=\textwidth,height=3cm]{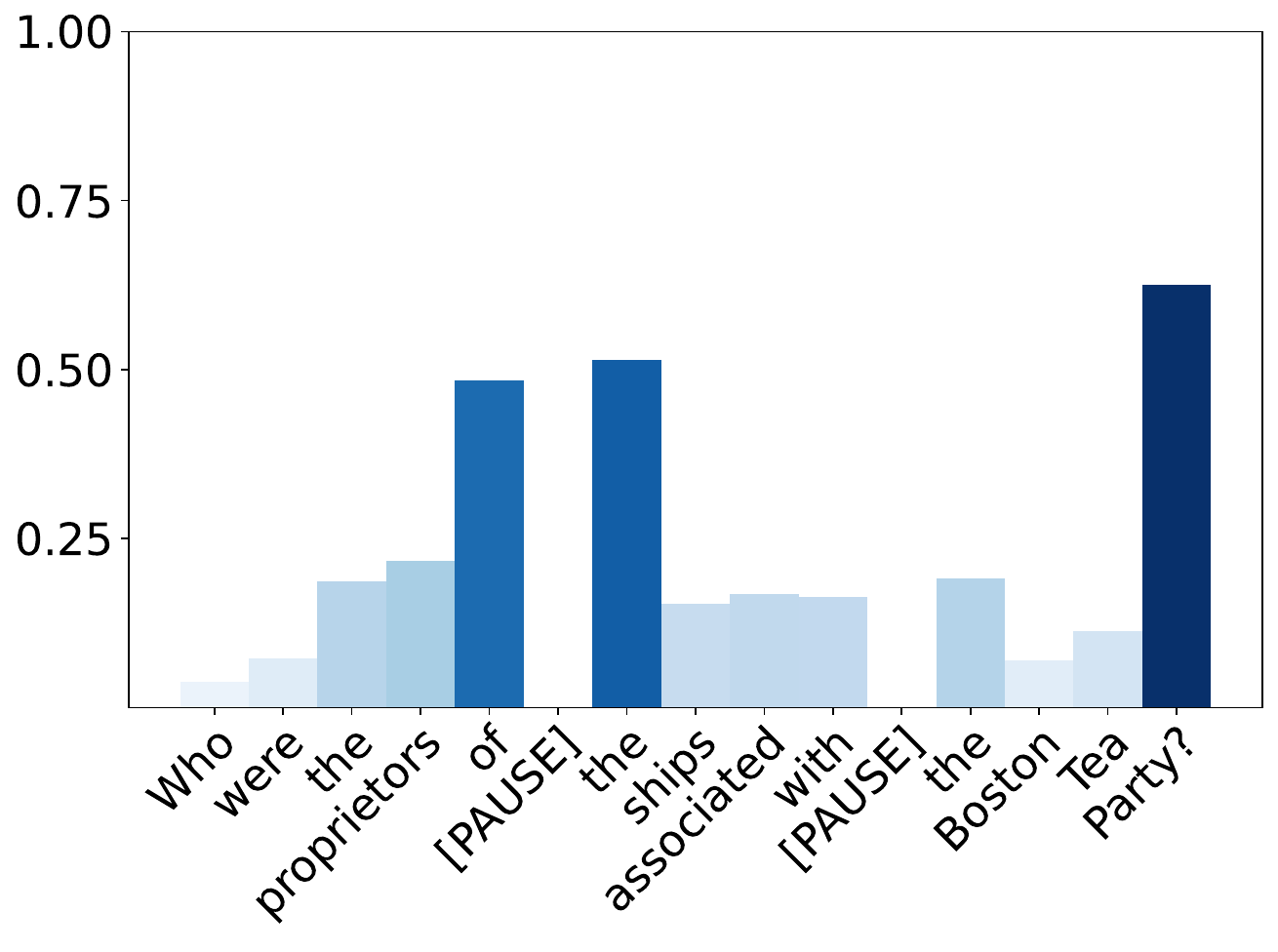}
            \caption[]%
            {{\small After adding \tframed[line width=0.5bp,fill=vred]{\textcolor{white}{\texttt{\textbf{[PAUSE]}}}} tokens} to paraphrase 5.}    
            \label{fig:mean and std of net44}
        \end{subfigure}
        \caption[]%
            {{\small The phrase \textbf{Boston Tea} gets more importance score after adding \tframed[line width=0.5bp,fill=vred]{\textcolor{white}{\texttt{\textbf{[PAUSE]}}}} token for phi-2.}}   
        \label{fig:phi}
\end{figure*}

\begin{figure*}[!ht]
        \centering
        \begin{subfigure}[b]{0.45\textwidth}
            \centering
            \includegraphics[width=\textwidth,height=3cm]{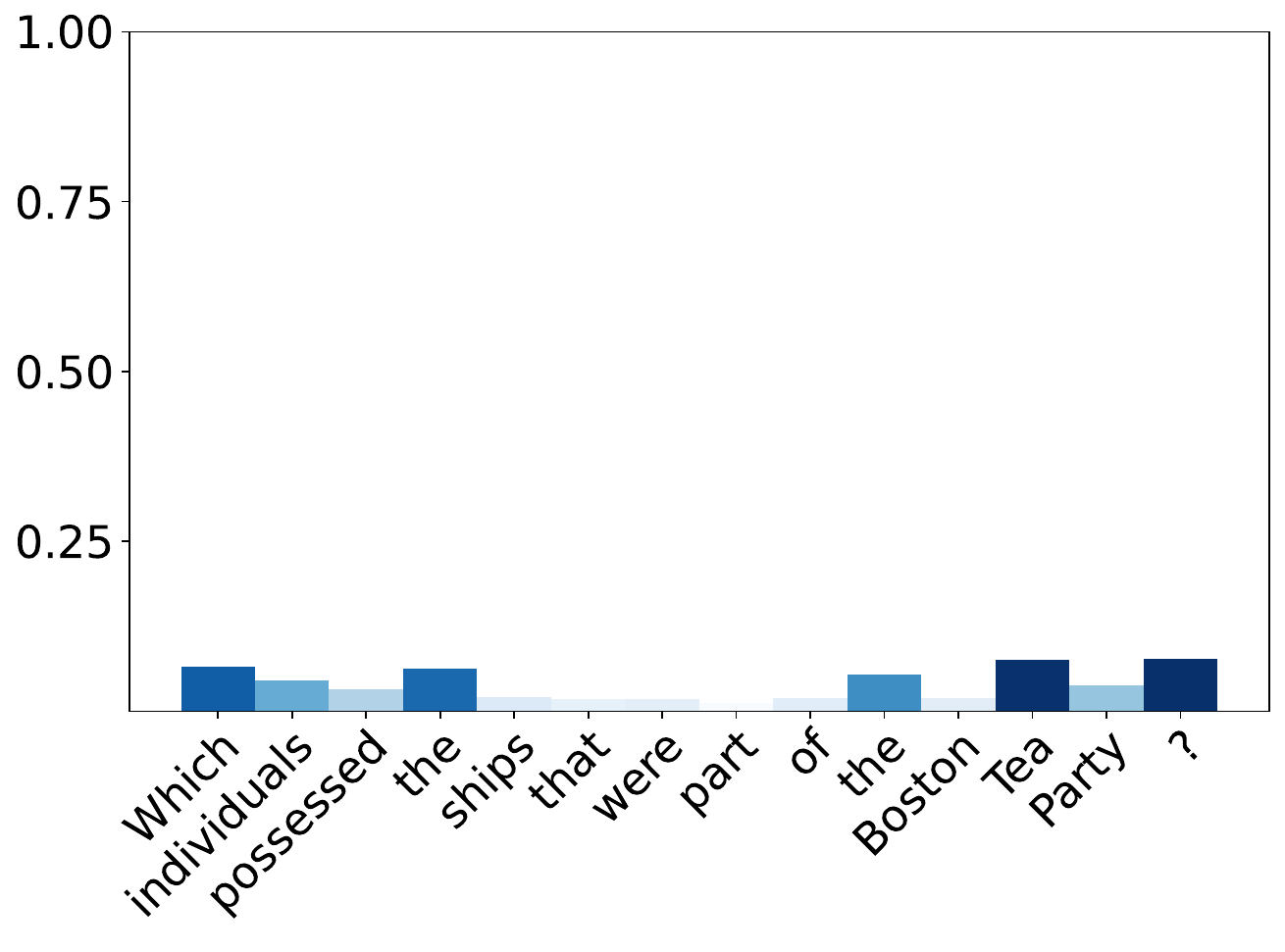}
            \caption[]%
            {{\small Before adding \tframed[line width=0.5bp,fill=vred]{\textcolor{white}{\texttt{\textbf{[PAUSE]}}}} tokens} to original prompt.}
            \label{fig:mean and std of net14}
        \end{subfigure}
        \hfill
        \begin{subfigure}[b]{0.45\textwidth}  
            \centering 
            \includegraphics[width=\textwidth,height=3cm]{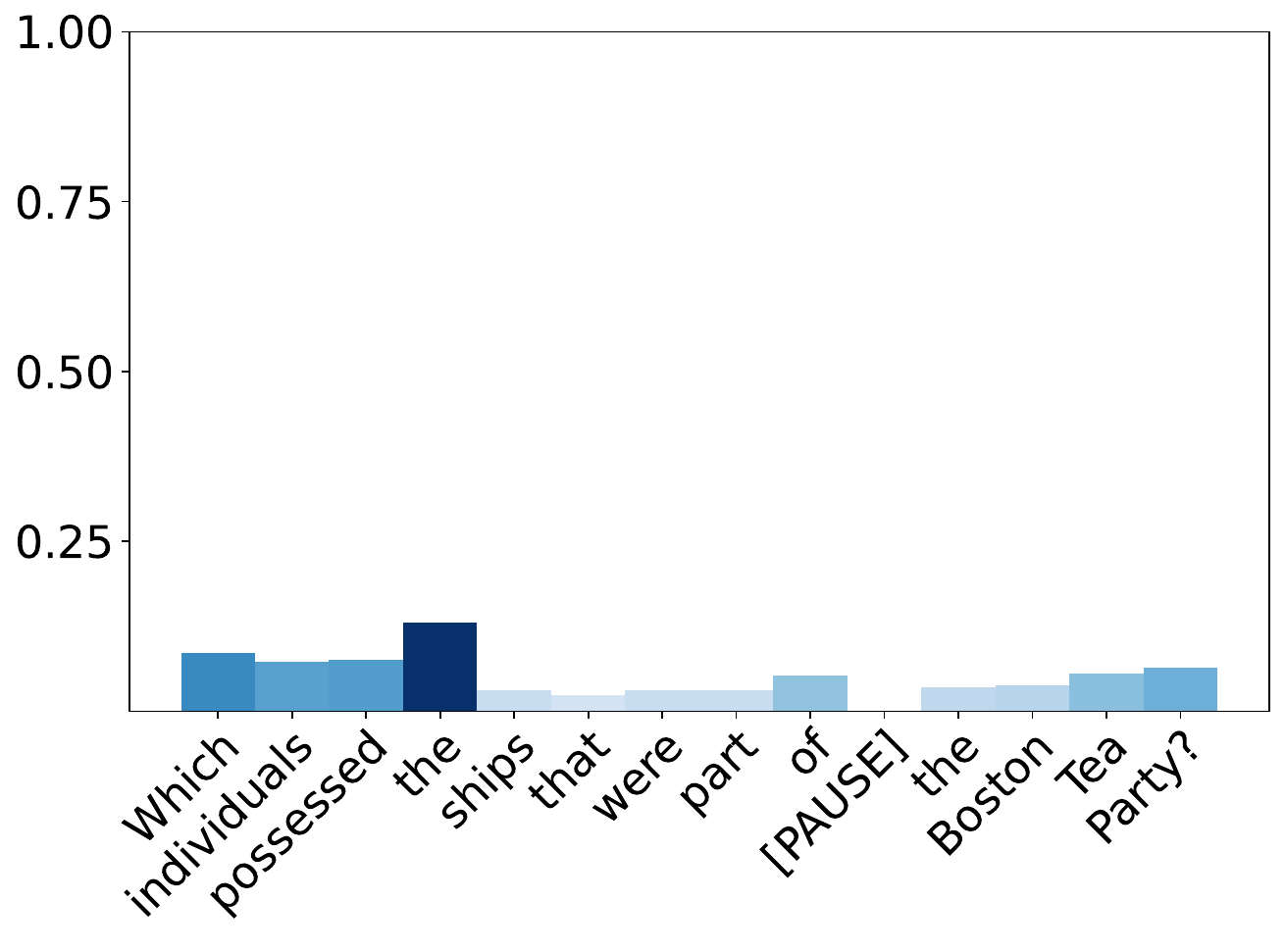}
            \caption[]%
            {{\small After adding \tframed[line width=0.5bp,fill=vred]{\textcolor{white}{\texttt{\textbf{[PAUSE]}}}} tokens} to original prompt.}    
            \label{fig:mean and std of net24}
        \end{subfigure}
        \hfill
        \vskip\baselineskip
        \begin{subfigure}[b]{0.45\textwidth}   
            \centering 
            \includegraphics[width=\textwidth,height=3cm]{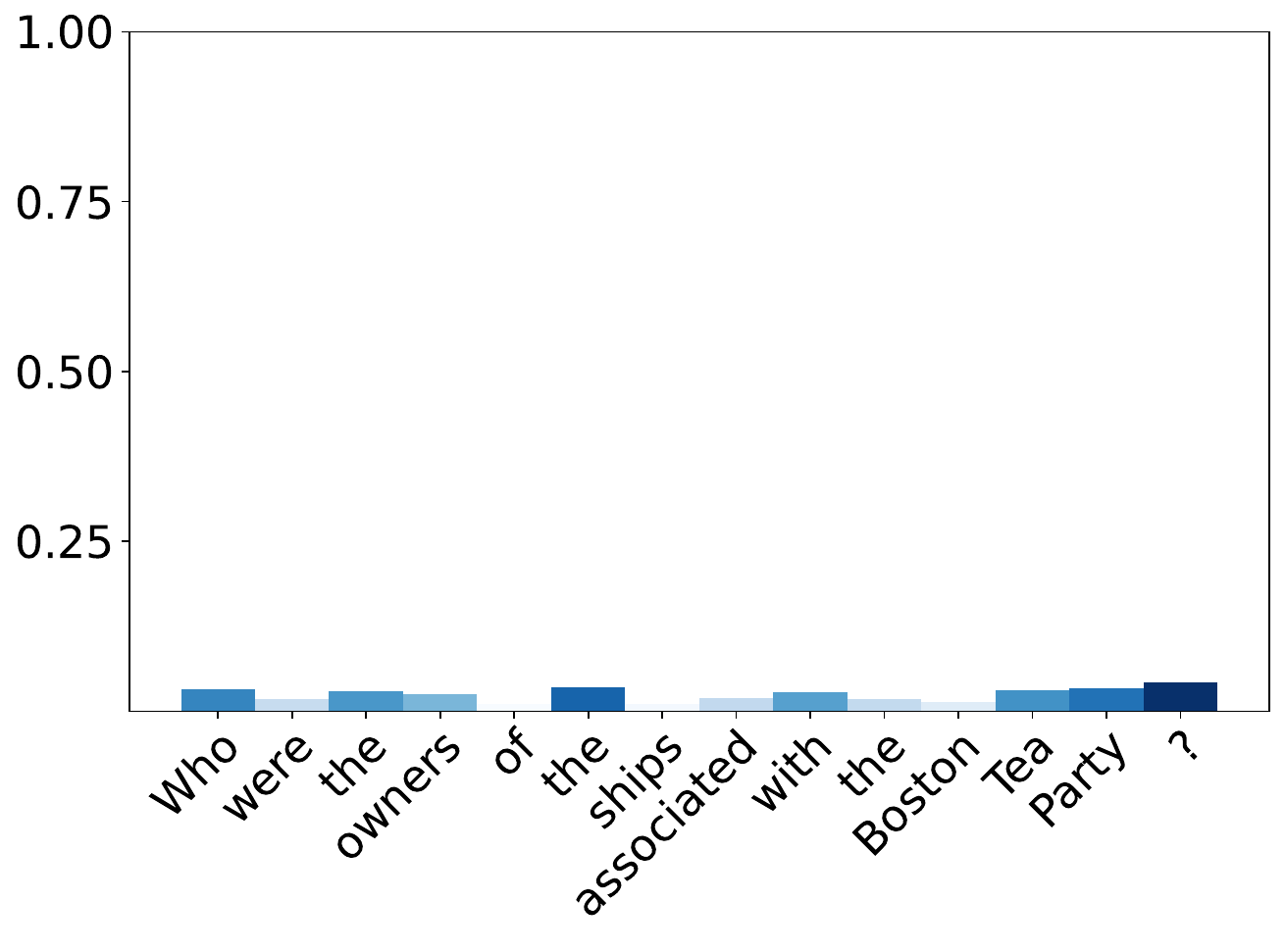}
            \caption[]%
            {{\small Before adding \tframed[line width=0.5bp,fill=vred]{\textcolor{white}{\texttt{\textbf{[PAUSE]}}}} tokens} to paraphrase 1.}    
            \label{fig:mean and std of net34}
        \end{subfigure}
        \hfill
        \begin{subfigure}[b]{0.45\textwidth}   
            \centering 
            \includegraphics[width=\textwidth,height=3cm]{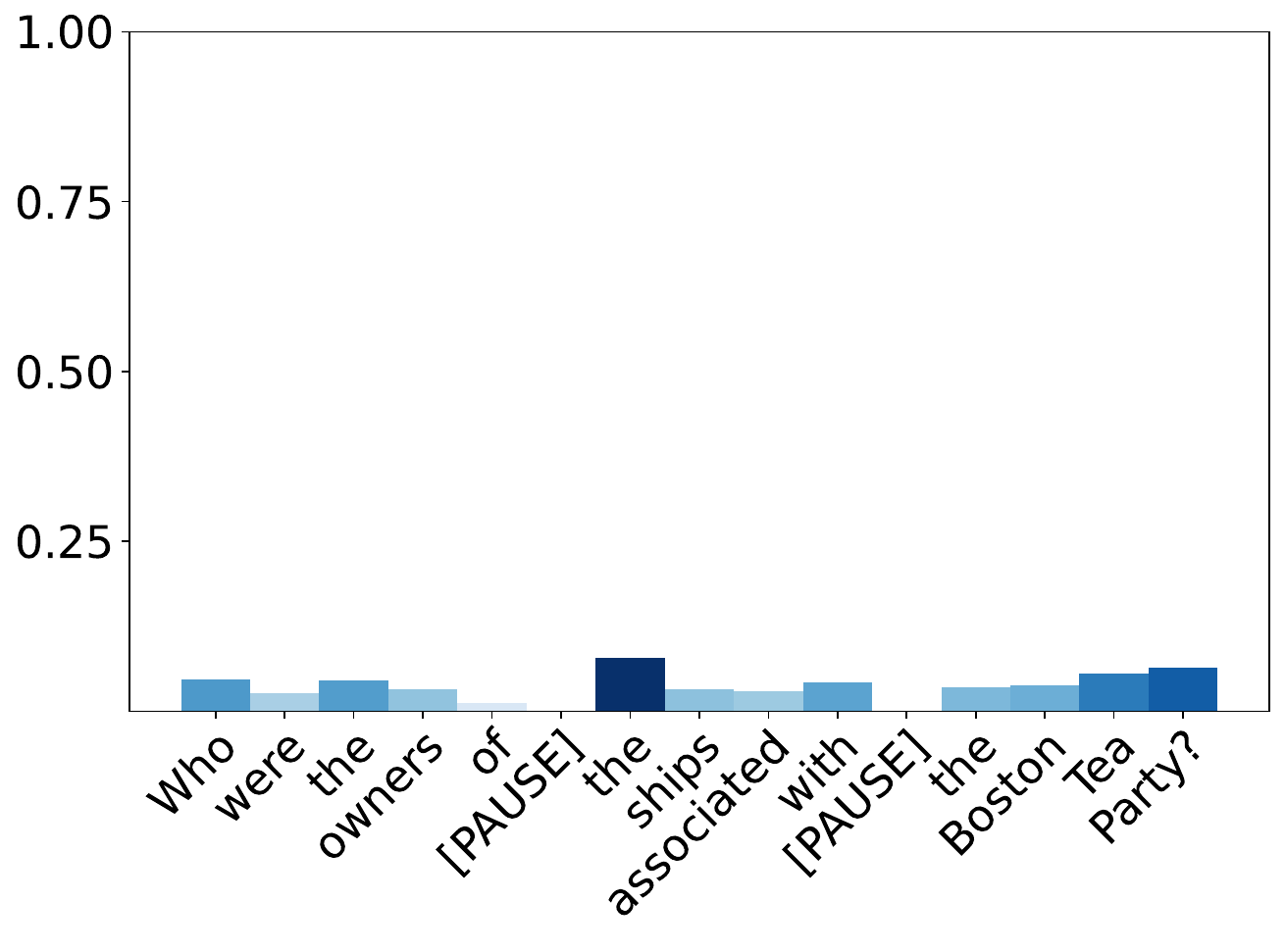}
            \caption[]%
            {{\small After adding \tframed[line width=0.5bp,fill=vred]{\textcolor{white}{\texttt{\textbf{[PAUSE]}}}} tokens} to paraphrase 1.}    
            \label{fig:mean and std of net44}
        \end{subfigure}
        \hfill
        \vskip\baselineskip
        \begin{subfigure}[b]{0.45\textwidth}   
            \centering 
            \includegraphics[width=\textwidth,height=3cm]{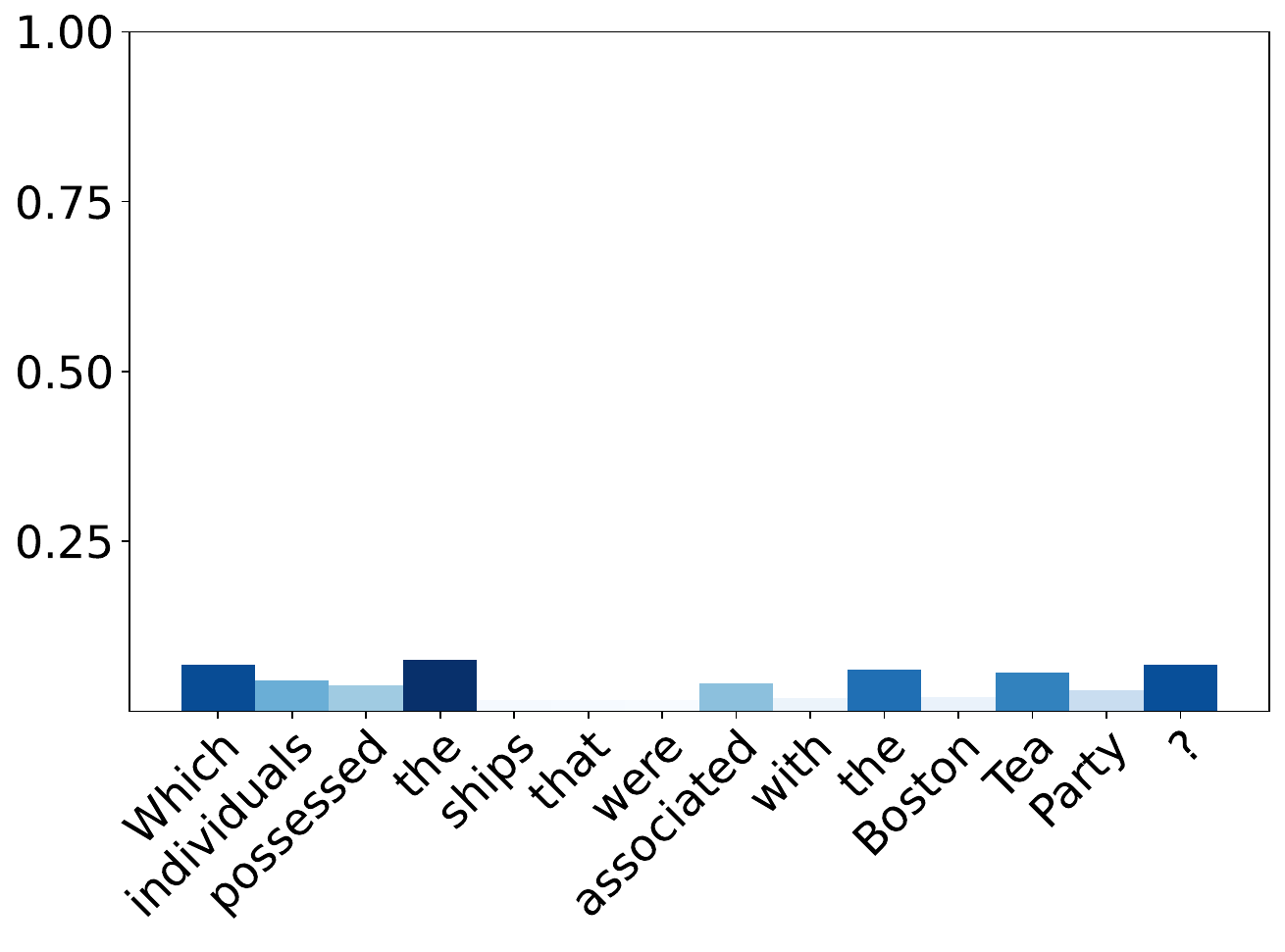}
            \caption[]%
            {{\small Before adding \tframed[line width=0.5bp,fill=vred]{\textcolor{white}{\texttt{\textbf{[PAUSE]}}}} tokens} to paraphrase 2.}
            \label{fig:mean and std of net34}
        \end{subfigure}
        \hfill
        \begin{subfigure}[b]{0.45\textwidth}   
            \centering 
            \includegraphics[width=\textwidth,height=3cm]{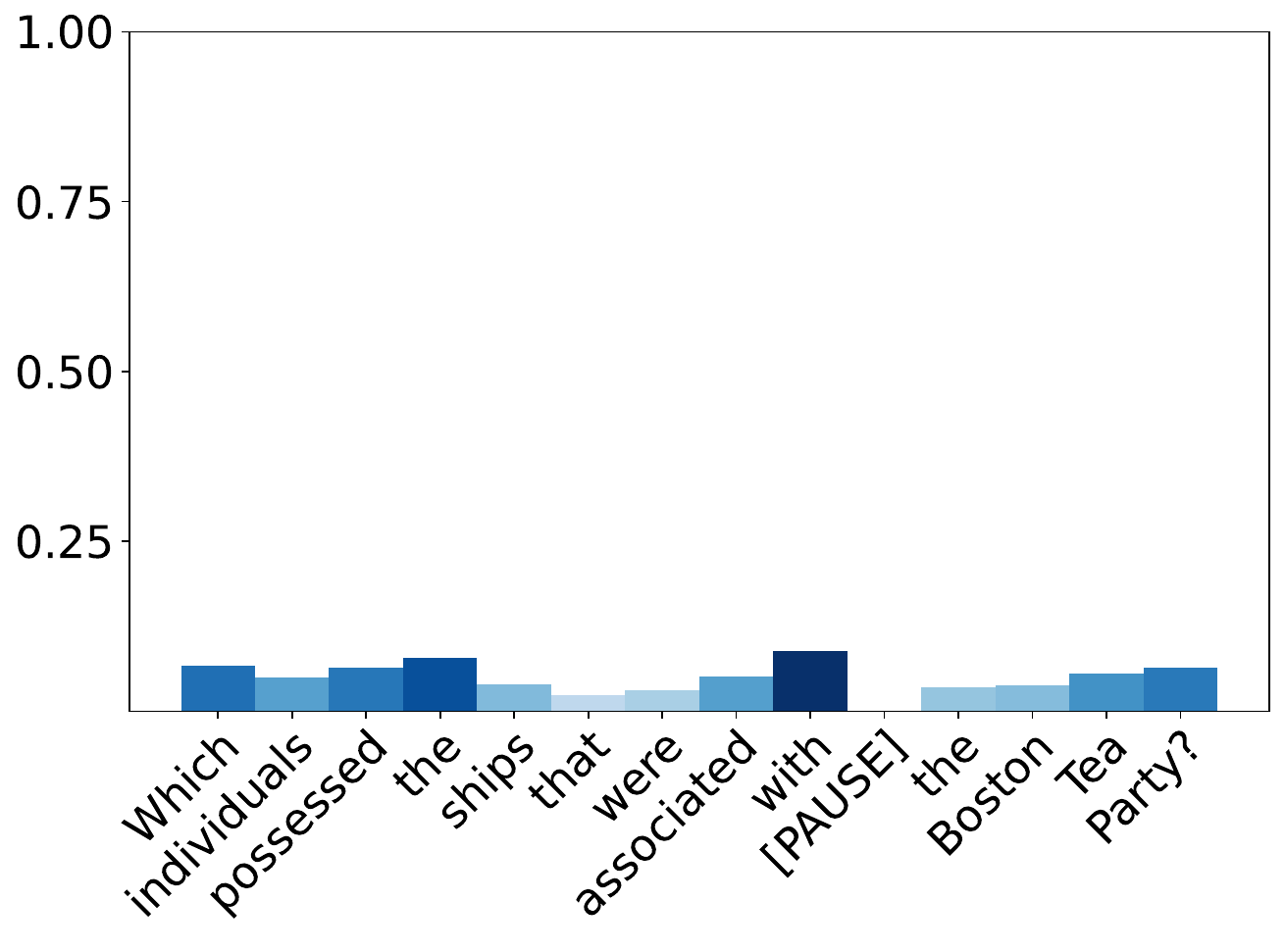}
            \caption[]%
            {{\small After adding \tframed[line width=0.5bp,fill=vred]{\textcolor{white}{\texttt{\textbf{[PAUSE]}}}} tokens} to paraphrase 2.} 
            \label{fig:mean and std of net44}
        \end{subfigure}
        \hfill
        \vskip\baselineskip
        \begin{subfigure}[b]{0.45\textwidth}   
            \centering 
            \includegraphics[width=\textwidth,height=3cm]{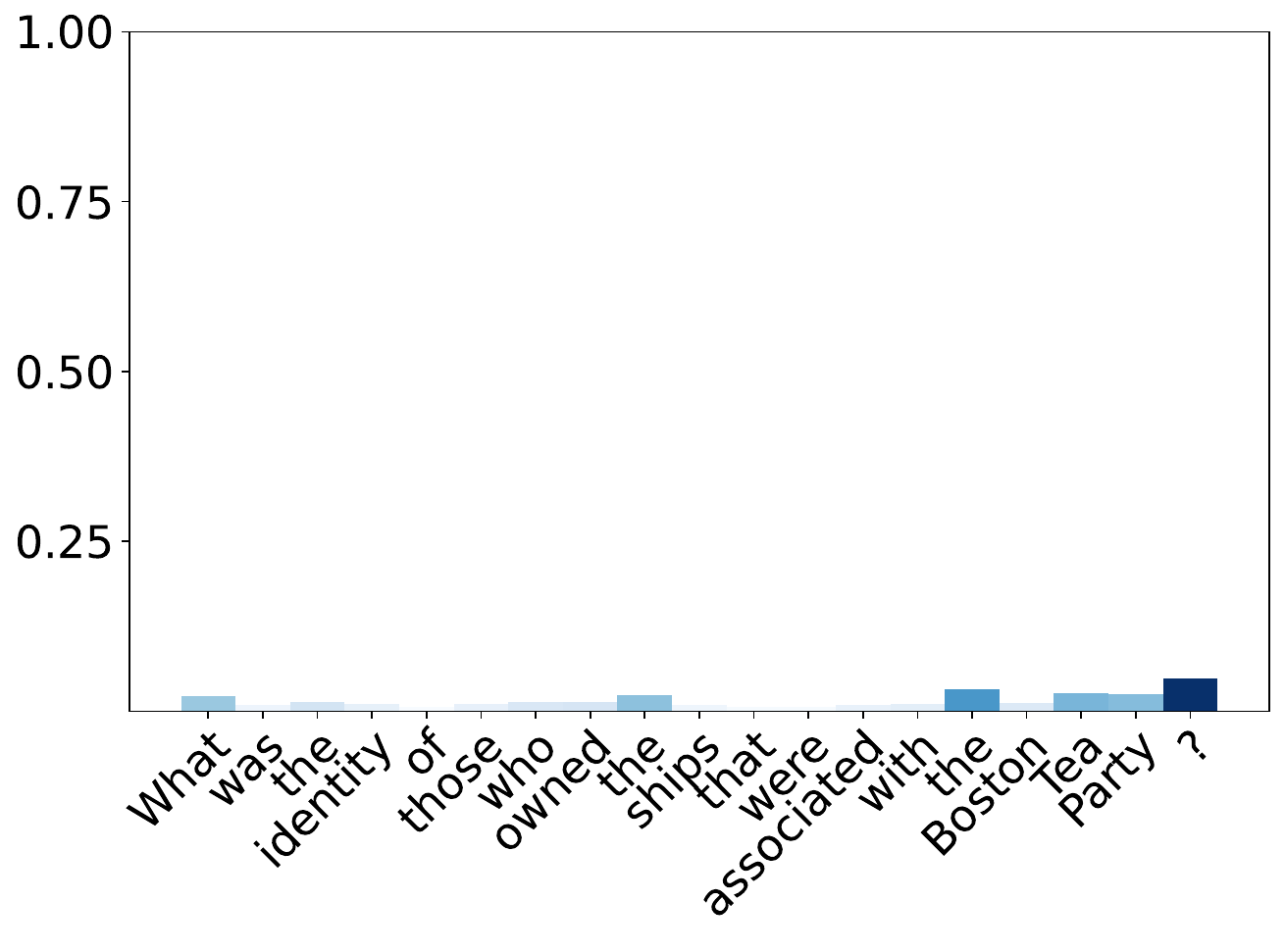}
            \caption[]%
            {{\small Before adding \tframed[line width=0.5bp,fill=vred]{\textcolor{white}{\texttt{\textbf{[PAUSE]}}}} tokens} to paraphrase 3.}
            \label{fig:mean and std of net44}
        \end{subfigure}
        \hfill
        \begin{subfigure}[b]{0.45\textwidth}   
            \centering 
            \includegraphics[width=\textwidth,height=3cm]{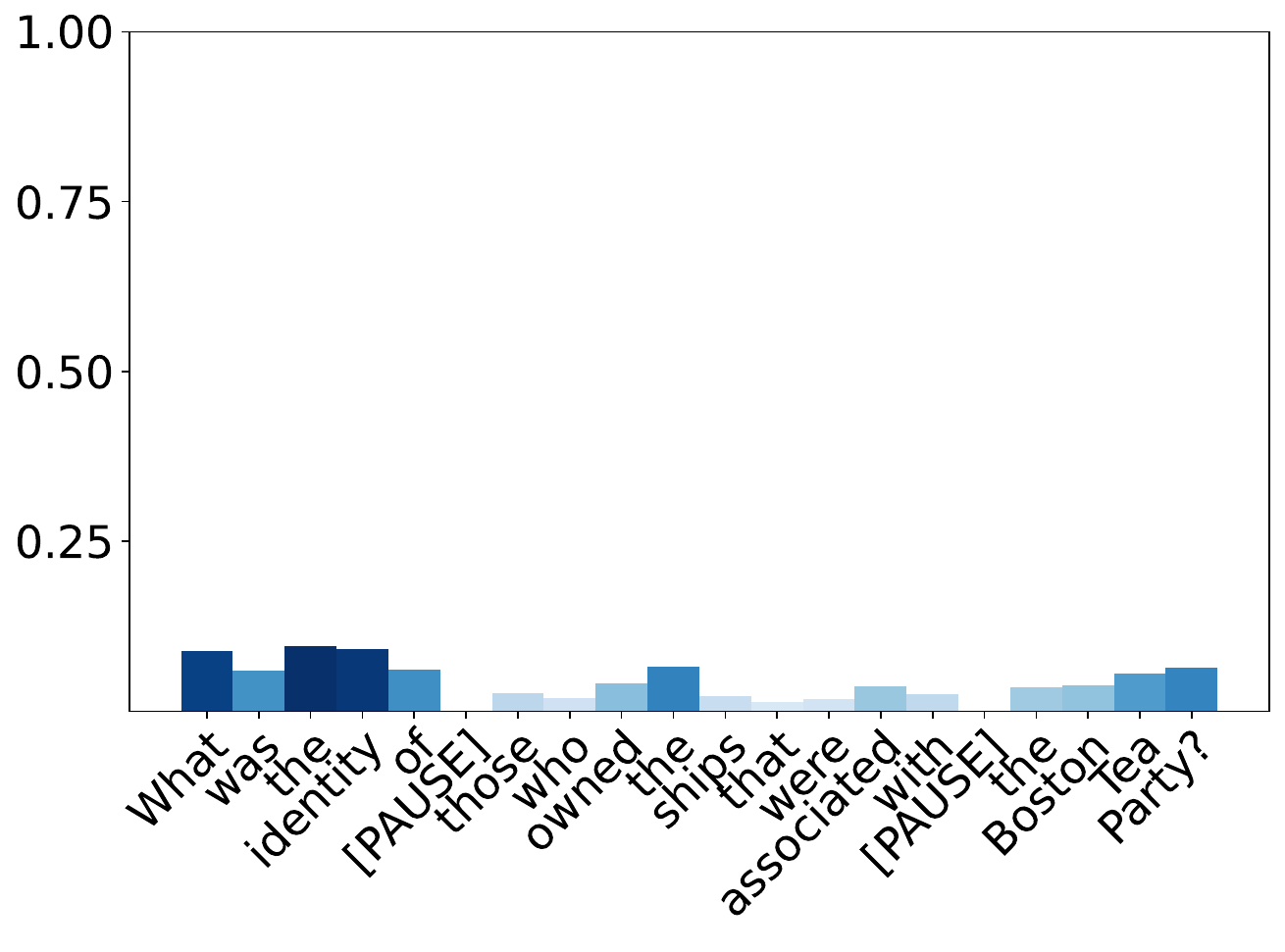}
            \caption[]%
            {{\small After adding \tframed[line width=0.5bp,fill=vred]{\textcolor{white}{\texttt{\textbf{[PAUSE]}}}} tokens} to paraphrase 3.}    
            \label{fig:mean and std of net44}
        \end{subfigure}
        \hfill
        \vskip\baselineskip
        \begin{subfigure}[b]{0.45\textwidth}   
            \centering 
            \includegraphics[width=\textwidth,height=3cm]{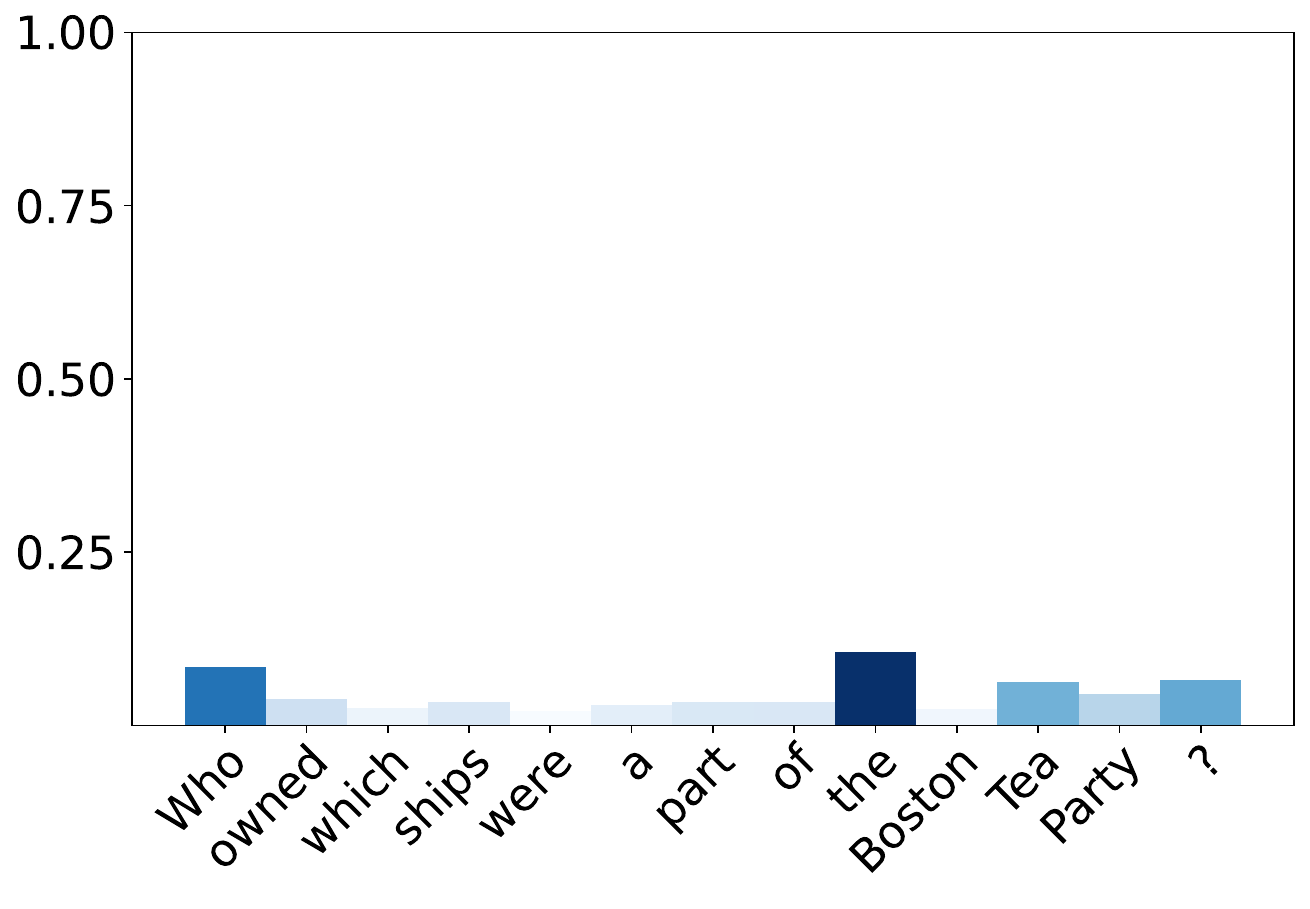}
            \caption[]%
            {{\small Before adding \tframed[line width=0.5bp,fill=vred]{\textcolor{white}{\texttt{\textbf{[PAUSE]}}}} tokens} to paraphrase 4.}    
            \label{fig:mean and std of net44}
        \end{subfigure}
        \hfill
        \begin{subfigure}[b]{0.45\textwidth}   
            \centering 
            \includegraphics[width=\textwidth,height=3cm]{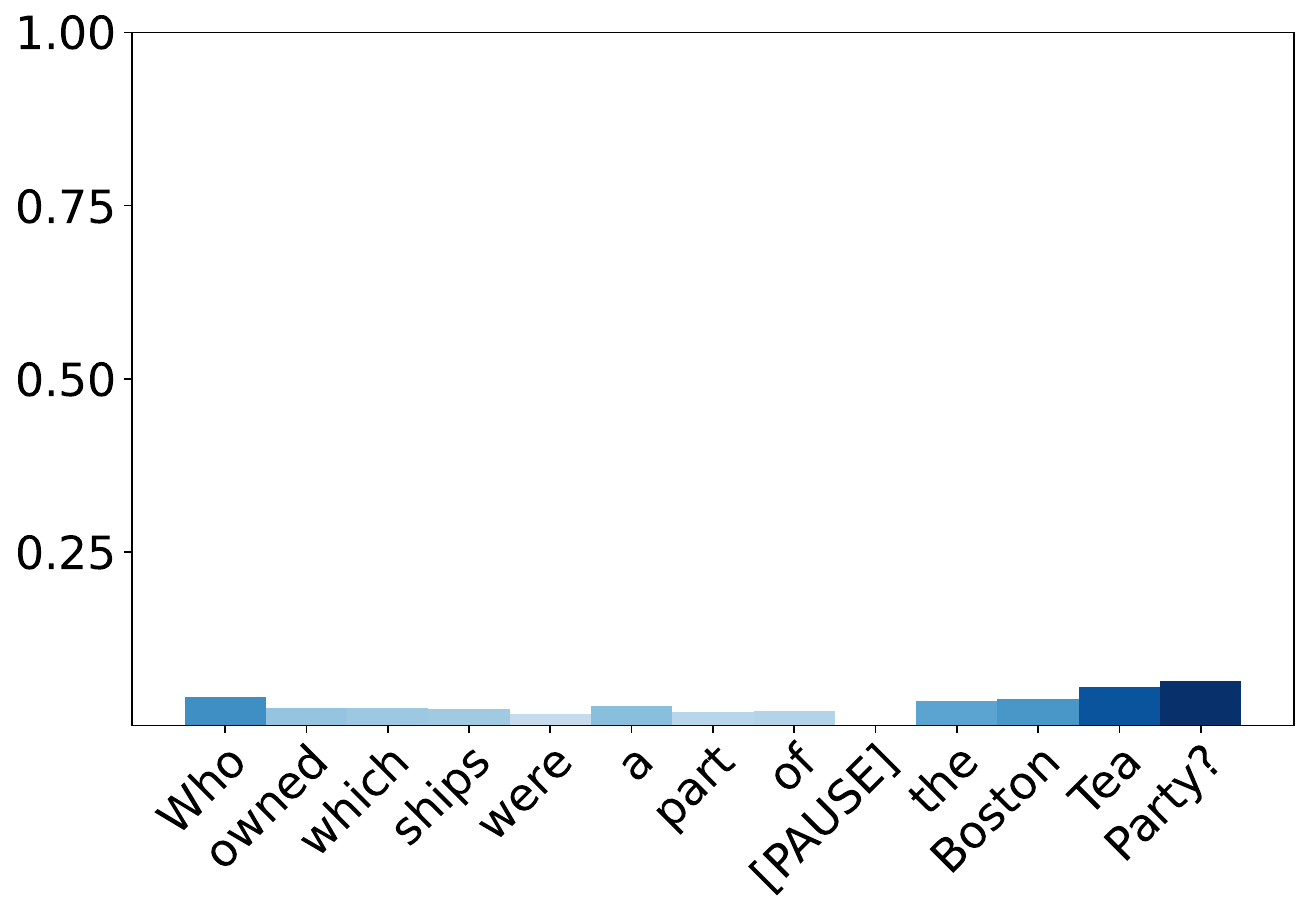}
            \caption[]%
            {{\small After adding \tframed[line width=0.5bp,fill=vred]{\textcolor{white}{\texttt{\textbf{[PAUSE]}}}} tokens} to paraphrase 4.}    
            \label{fig:mean and std of net44}
        \end{subfigure}
        \hfill
        \vskip\baselineskip
        \begin{subfigure}[b]{0.45\textwidth}   
            \centering 
            \includegraphics[width=\textwidth,height=3cm]{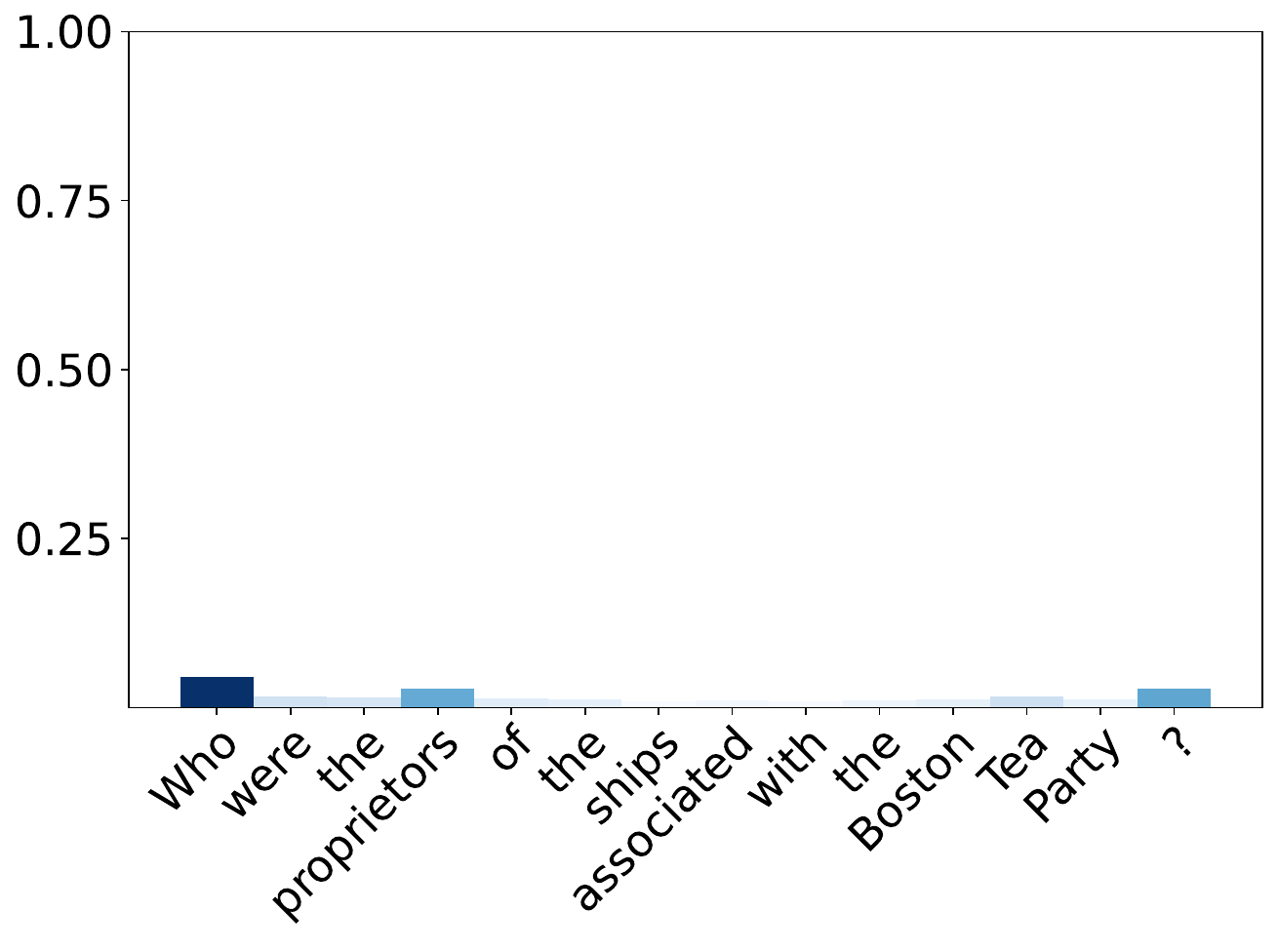}
            \caption[]%
            {{\small Before adding \tframed[line width=0.5bp,fill=vred]{\textcolor{white}{\texttt{\textbf{[PAUSE]}}}} tokens} to paraphrase 5.}    
            \label{fig:mean and std of net44}
        \end{subfigure}
        \hfill
        \begin{subfigure}[b]{0.45\textwidth}   
            \centering 
            \includegraphics[width=\textwidth,height=3cm]{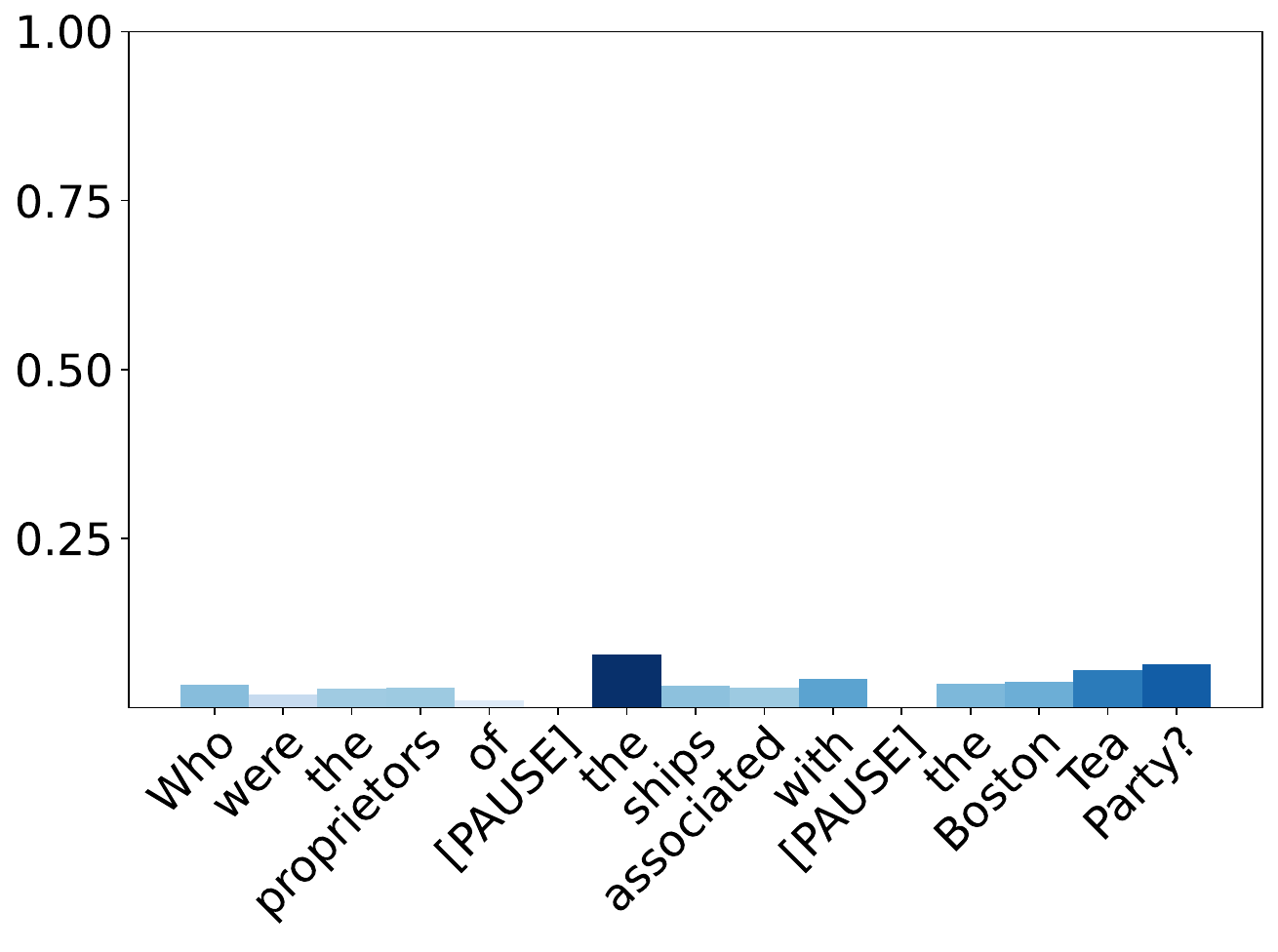}
            \caption[]%
            {{\small After adding \tframed[line width=0.5bp,fill=vred]{\textcolor{white}{\texttt{\textbf{[PAUSE]}}}} tokens} to paraphrase 5.}    
            \label{fig:mean and std of net44}
        \end{subfigure}
        \caption[]%
            {{\small The phrase \textbf{Boston Tea} gets more importance score after adding \tframed[line width=0.5bp,fill=vred]{\textcolor{white}{\texttt{\textbf{[PAUSE]}}}} token for Vicuna.}}   
        \label{fig:Vicuna}
\end{figure*}

\begin{figure*}[!ht]
        \centering
        \begin{subfigure}[b]{0.45\textwidth}
            \centering
            \includegraphics[width=\textwidth,height=3cm]{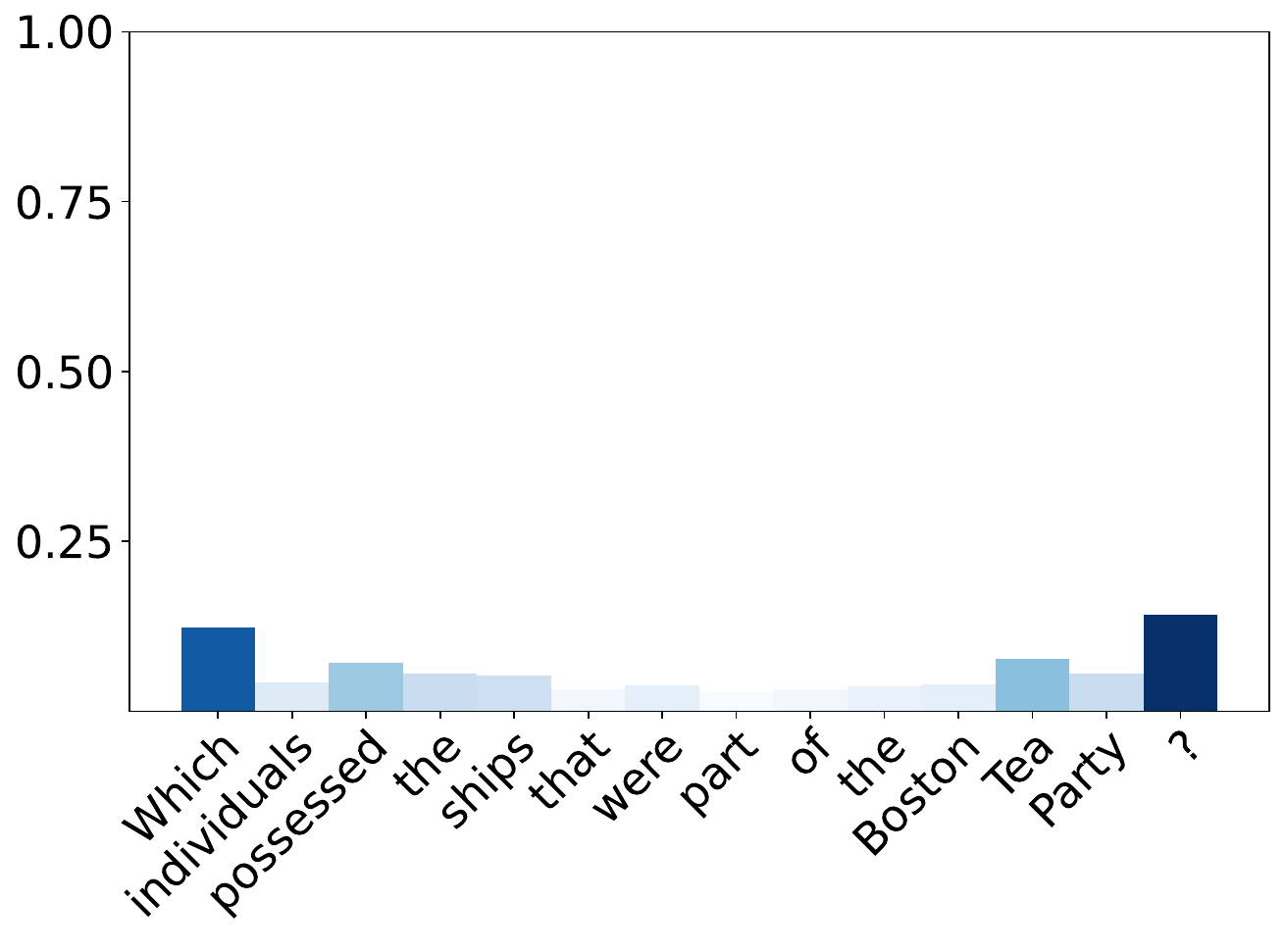}
            \caption[]%
            {{\small Before adding \tframed[line width=0.5bp,fill=vred]{\textcolor{white}{\texttt{\textbf{[PAUSE]}}}} tokens} to original prompt.}
            \label{fig:mean and std of net14}
        \end{subfigure}
        \hfill
        \begin{subfigure}[b]{0.45\textwidth}  
            \centering 
            \includegraphics[width=\textwidth,height=3cm]{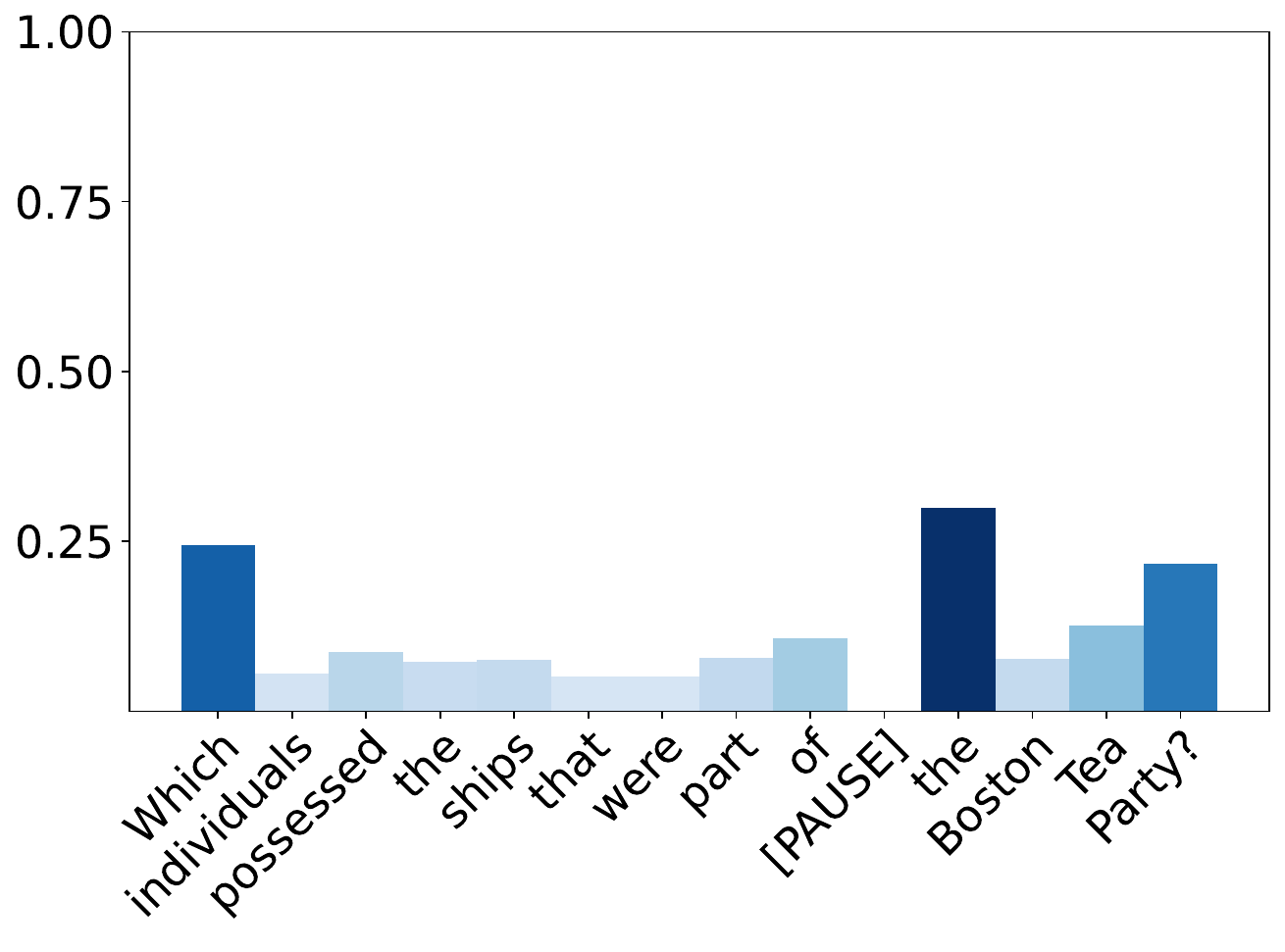}
            \caption[]%
            {{\small After adding \tframed[line width=0.5bp,fill=vred]{\textcolor{white}{\texttt{\textbf{[PAUSE]}}}} tokens} to original prompt.}    
            \label{fig:mean and std of net24}
        \end{subfigure}
        \hfill
        \vskip\baselineskip
        \begin{subfigure}[b]{0.45\textwidth}   
            \centering 
            \includegraphics[width=\textwidth,height=3cm]{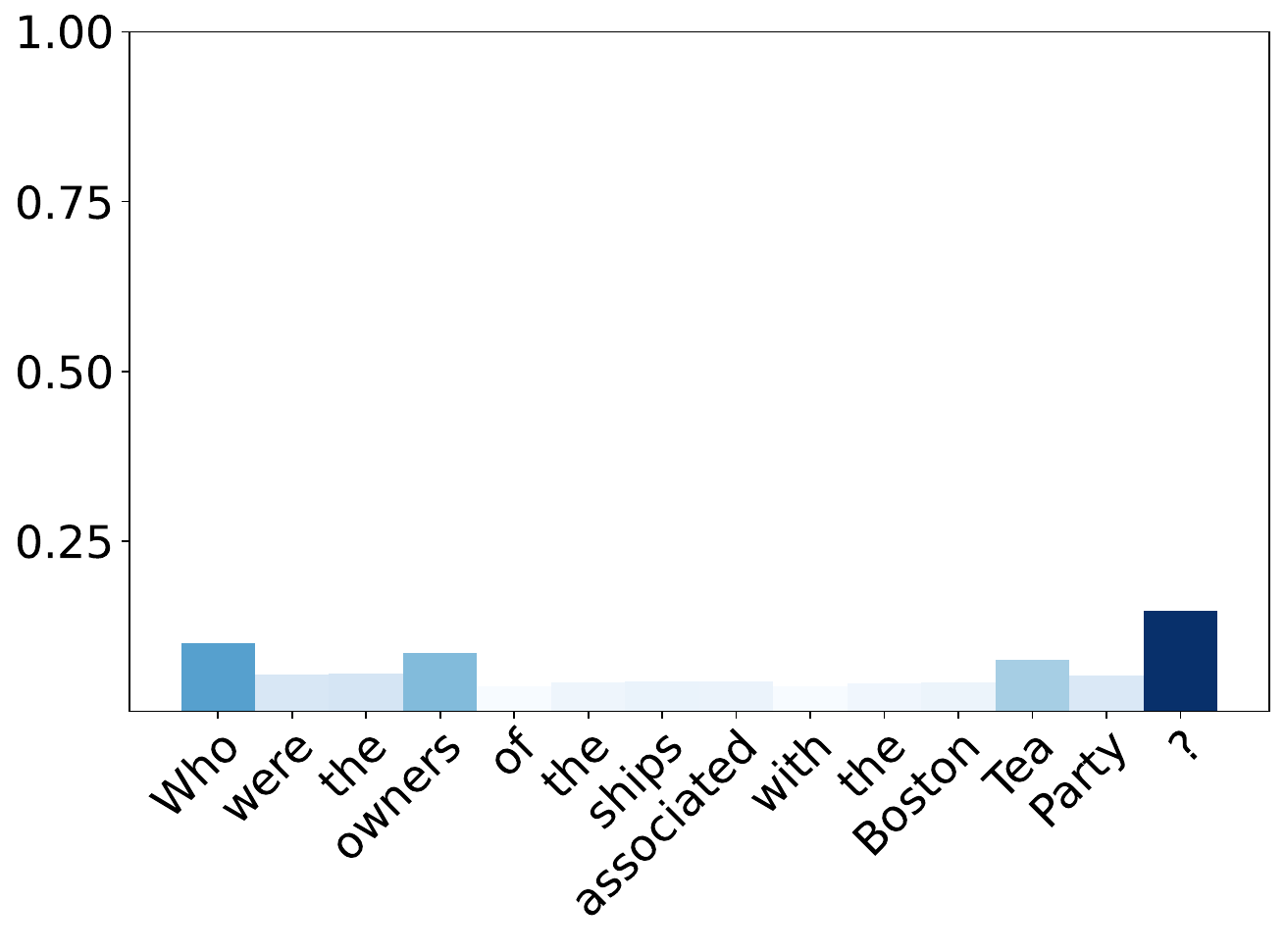}
            \caption[]%
            {{\small Before adding \tframed[line width=0.5bp,fill=vred]{\textcolor{white}{\texttt{\textbf{[PAUSE]}}}} tokens} to paraphrase 1.}    
            \label{fig:mean and std of net34}
        \end{subfigure}
        \hfill
        \begin{subfigure}[b]{0.45\textwidth}   
            \centering 
            \includegraphics[width=\textwidth,height=3cm]{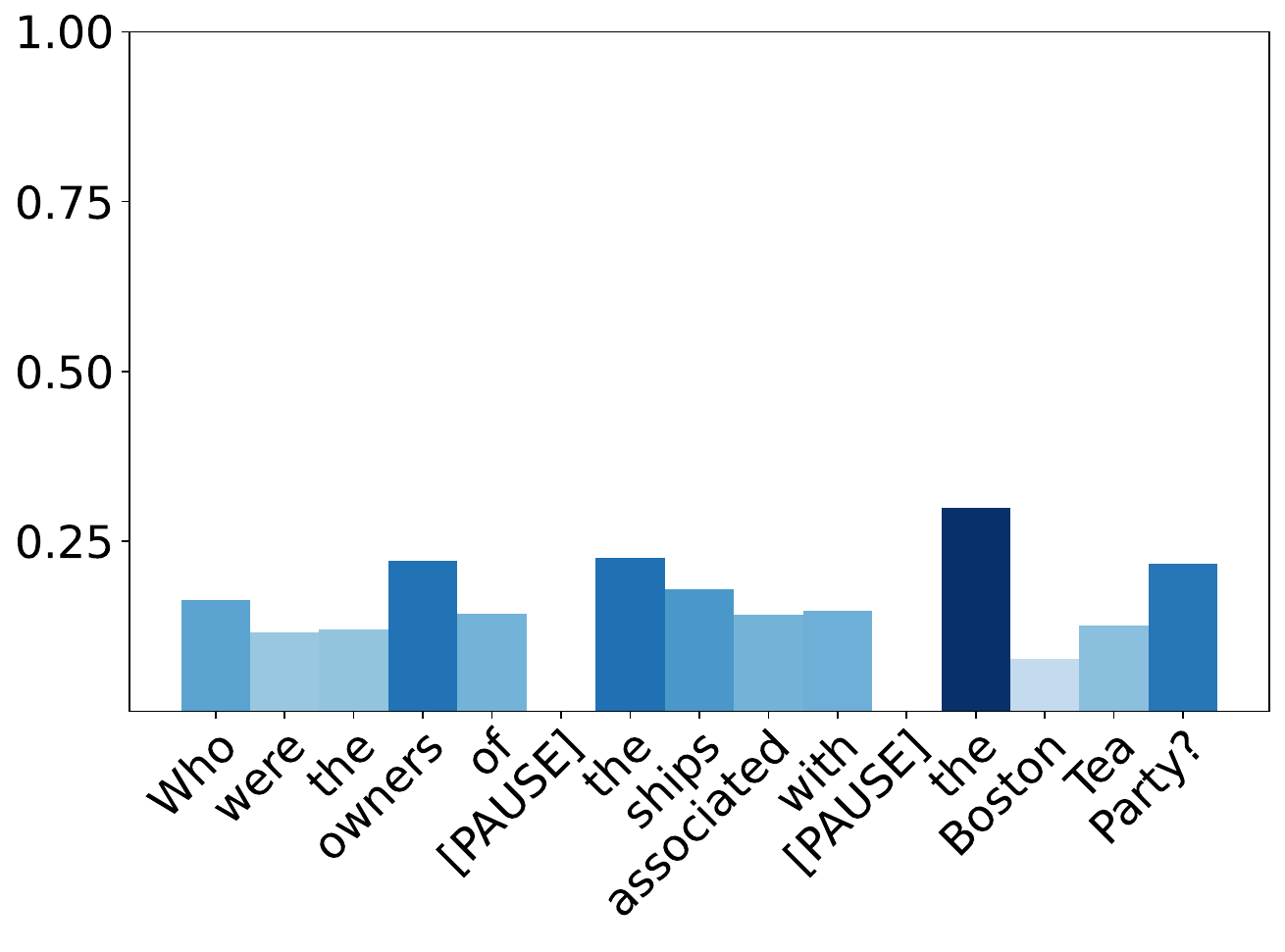}
            \caption[]%
            {{\small After adding \tframed[line width=0.5bp,fill=vred]{\textcolor{white}{\texttt{\textbf{[PAUSE]}}}} tokens} to paraphrase 1.}    
            \label{fig:mean and std of net44}
        \end{subfigure}
        \hfill
        \vskip\baselineskip
        \begin{subfigure}[b]{0.45\textwidth}   
            \centering 
            \includegraphics[width=\textwidth,height=3cm]{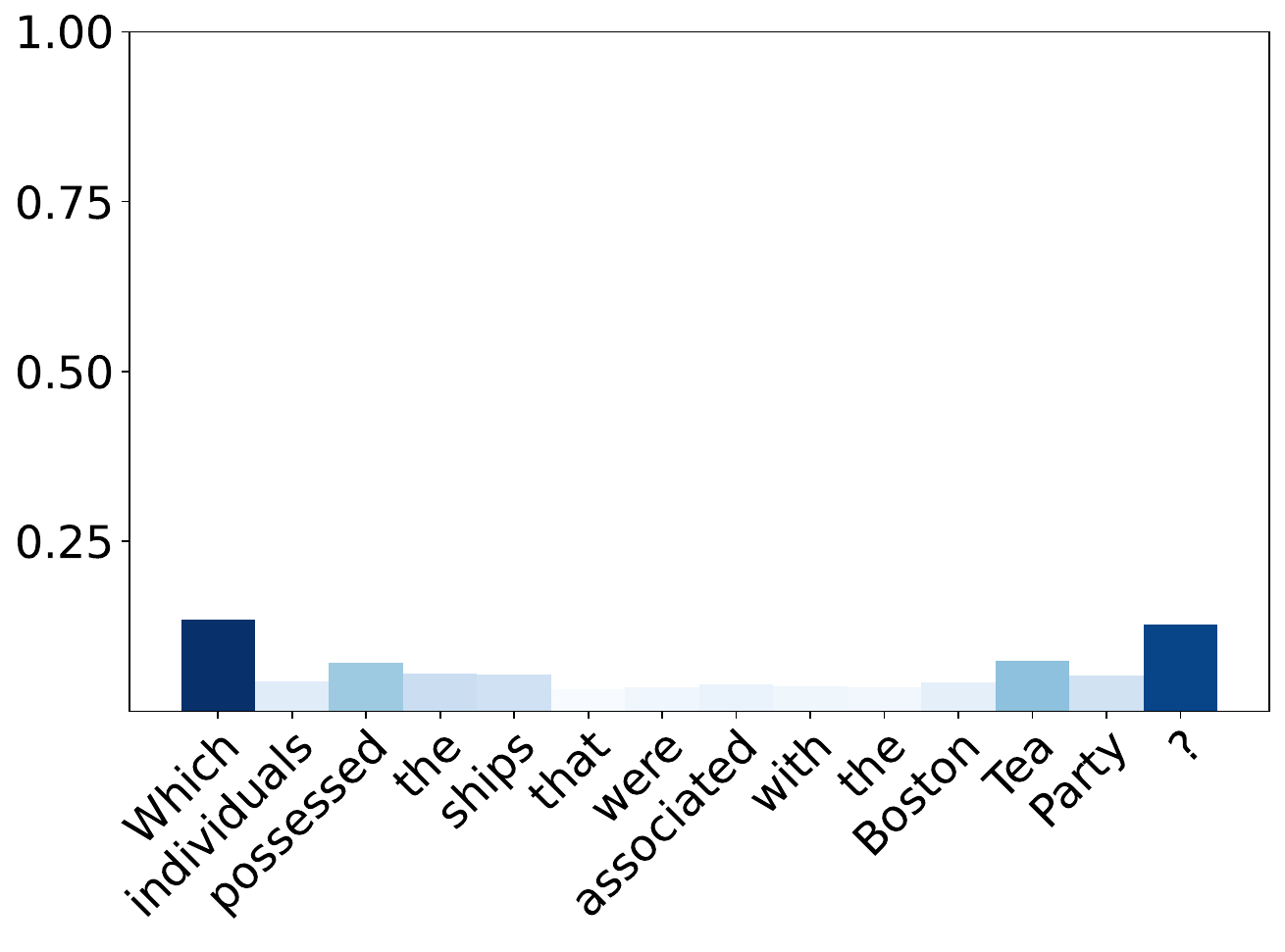}
            \caption[]%
            {{\small Before adding \tframed[line width=0.5bp,fill=vred]{\textcolor{white}{\texttt{\textbf{[PAUSE]}}}} tokens} to paraphrase 2.}
            \label{fig:mean and std of net34}
        \end{subfigure}
        \hfill
        \begin{subfigure}[b]{0.45\textwidth}   
            \centering 
            \includegraphics[width=\textwidth,height=3cm]{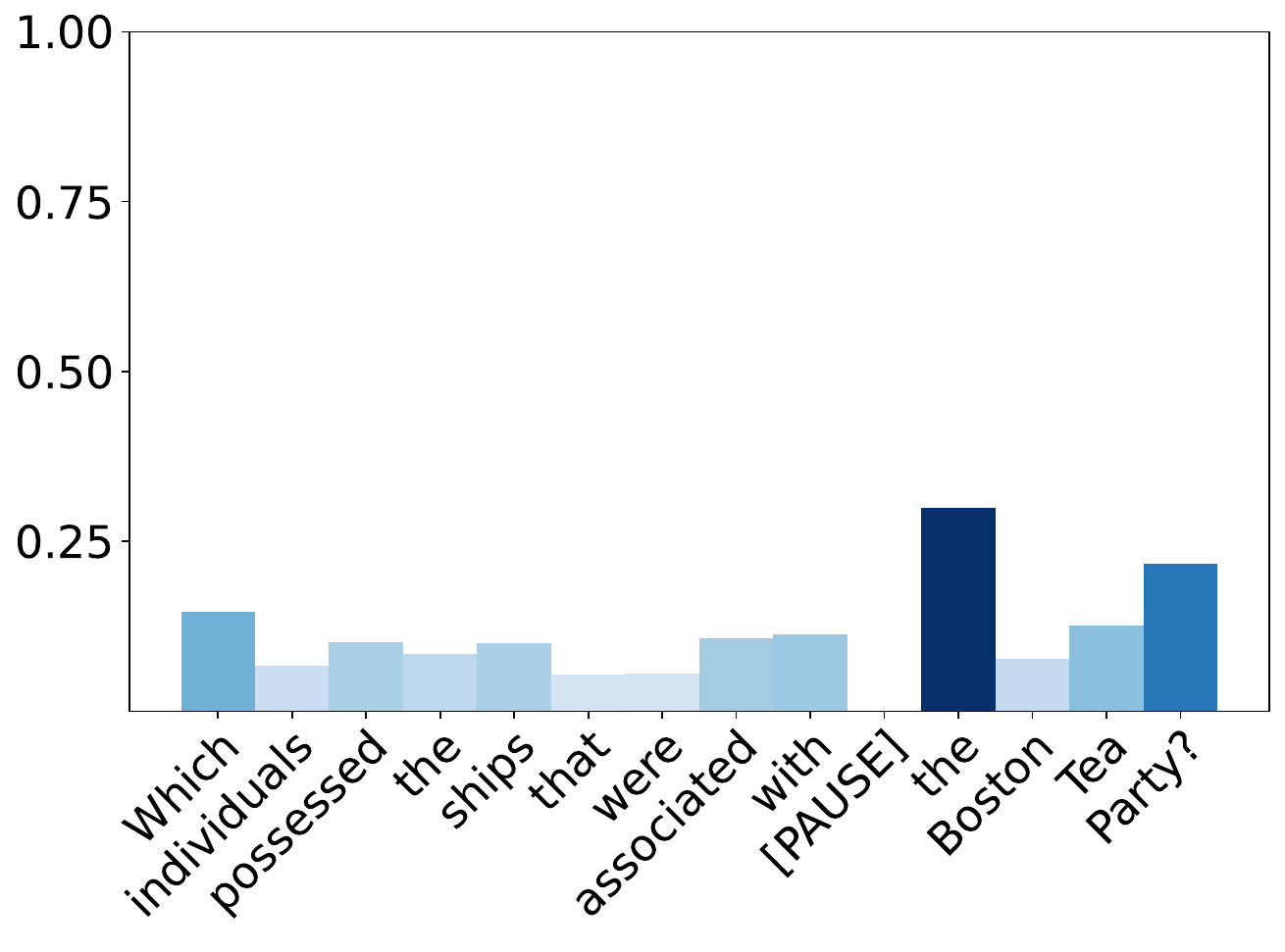}
            \caption[]%
            {{\small After adding \tframed[line width=0.5bp,fill=vred]{\textcolor{white}{\texttt{\textbf{[PAUSE]}}}} tokens} to paraphrase 2.} 
            \label{fig:mean and std of net44}
        \end{subfigure}
        \hfill
        \vskip\baselineskip
        \begin{subfigure}[b]{0.45\textwidth}   
            \centering 
            \includegraphics[width=\textwidth,height=3cm]{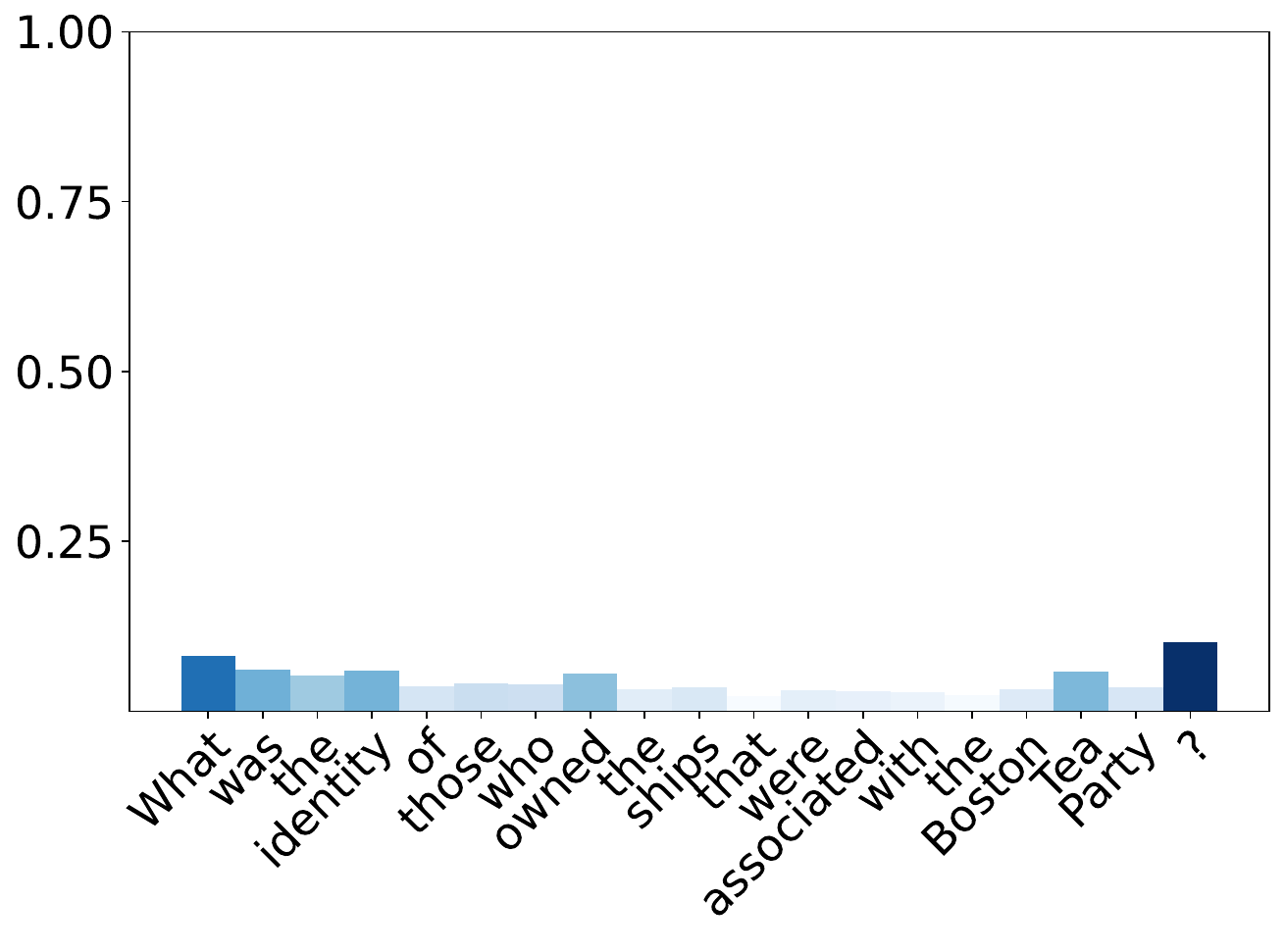}
            \caption[]%
            {{\small Before adding \tframed[line width=0.5bp,fill=vred]{\textcolor{white}{\texttt{\textbf{[PAUSE]}}}} tokens} to paraphrase 3.}
            \label{fig:mean and std of net44}
        \end{subfigure}
        \hfill
        \begin{subfigure}[b]{0.45\textwidth}   
            \centering 
            \includegraphics[width=\textwidth,height=3cm]{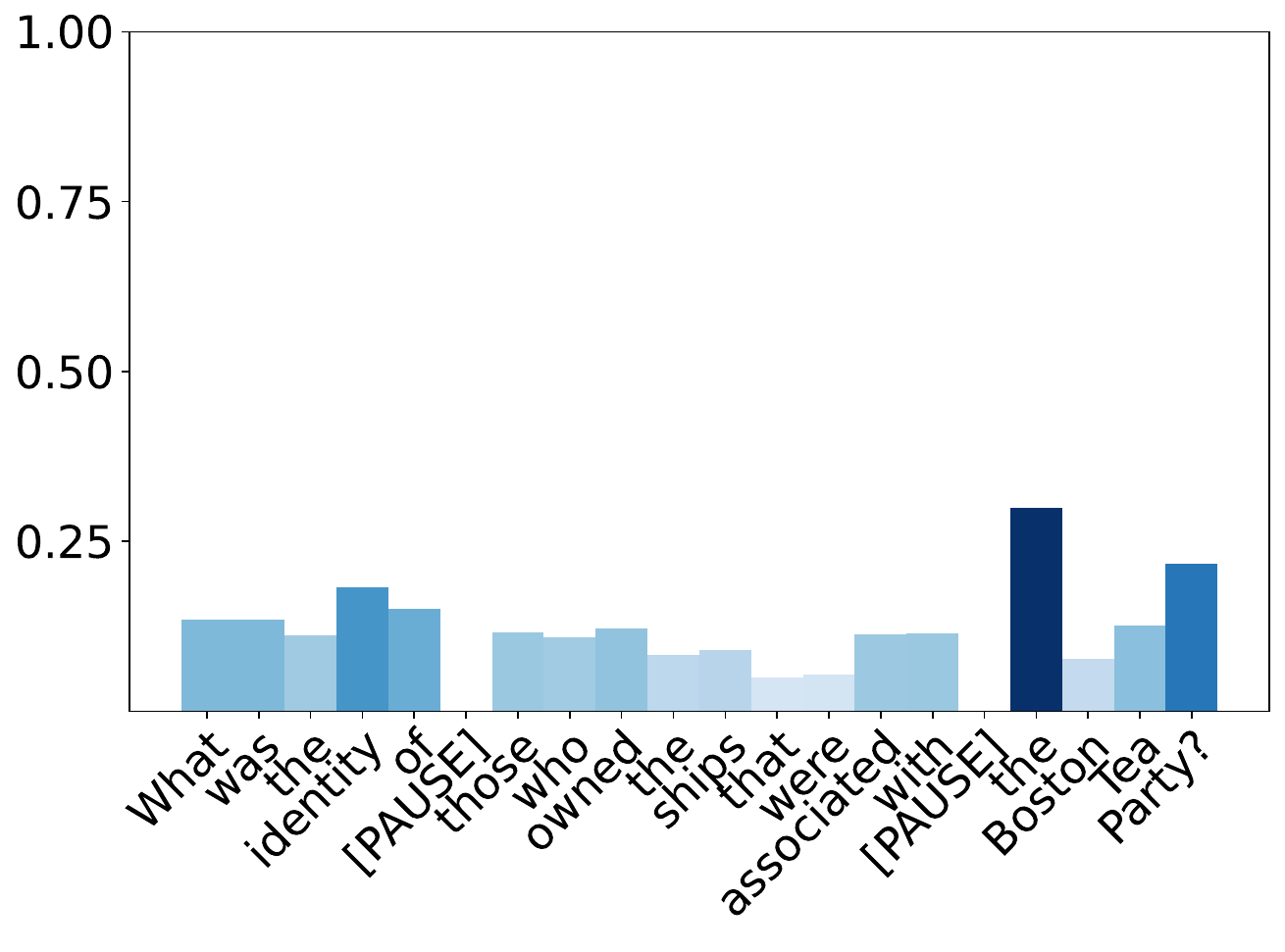}
            \caption[]%
            {{\small After adding \tframed[line width=0.5bp,fill=vred]{\textcolor{white}{\texttt{\textbf{[PAUSE]}}}} tokens} to paraphrase 3.}    
            \label{fig:mean and std of net44}
        \end{subfigure}
        \hfill
        \vskip\baselineskip
        \begin{subfigure}[b]{0.45\textwidth}   
            \centering 
            \includegraphics[width=\textwidth,height=3cm]{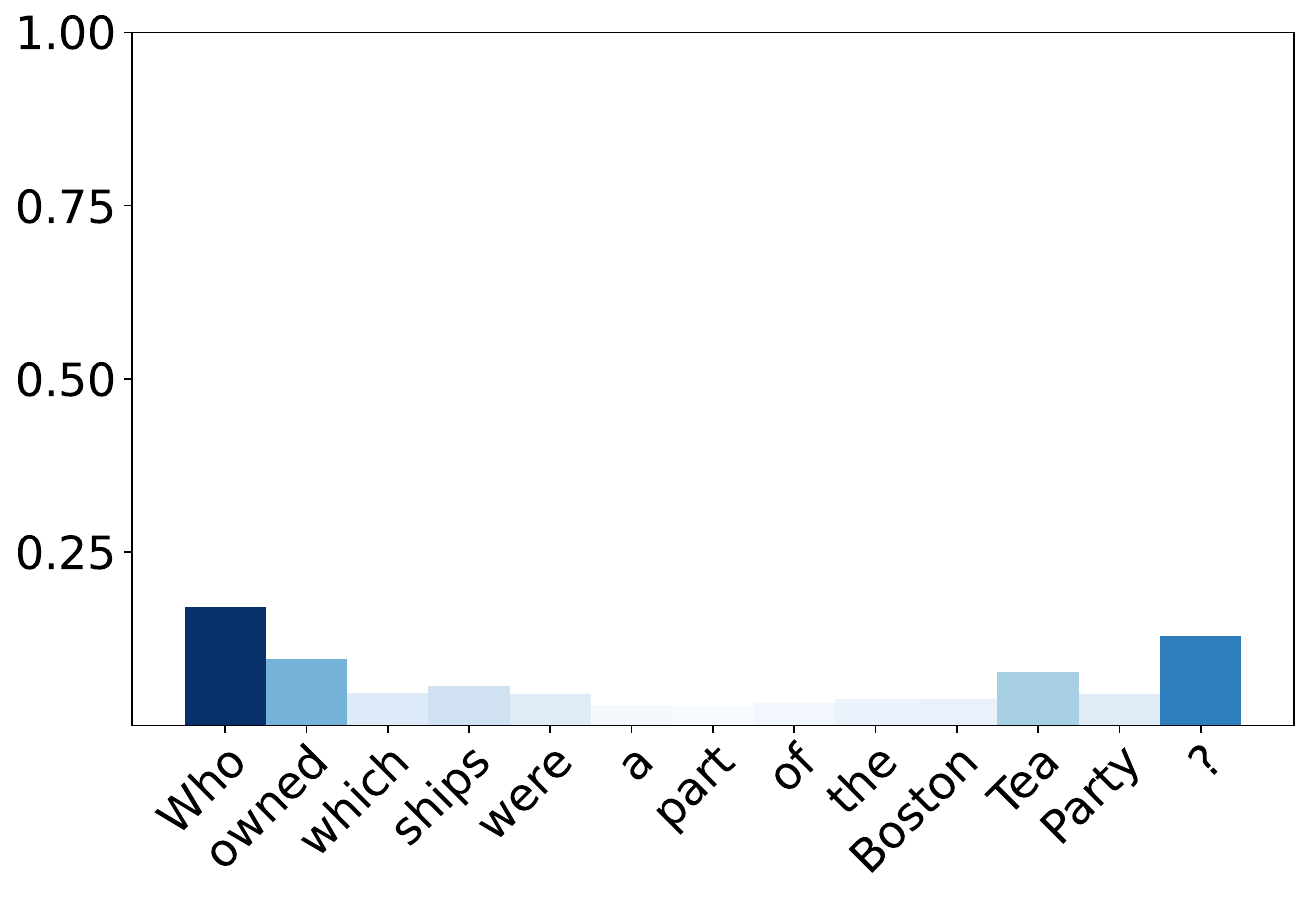}
            \caption[]%
            {{\small Before adding \tframed[line width=0.5bp,fill=vred]{\textcolor{white}{\texttt{\textbf{[PAUSE]}}}} tokens} to paraphrase 4.}    
            \label{fig:mean and std of net44}
        \end{subfigure}
        \hfill
        \begin{subfigure}[b]{0.45\textwidth}   
            \centering 
            \includegraphics[width=\textwidth,height=3cm]{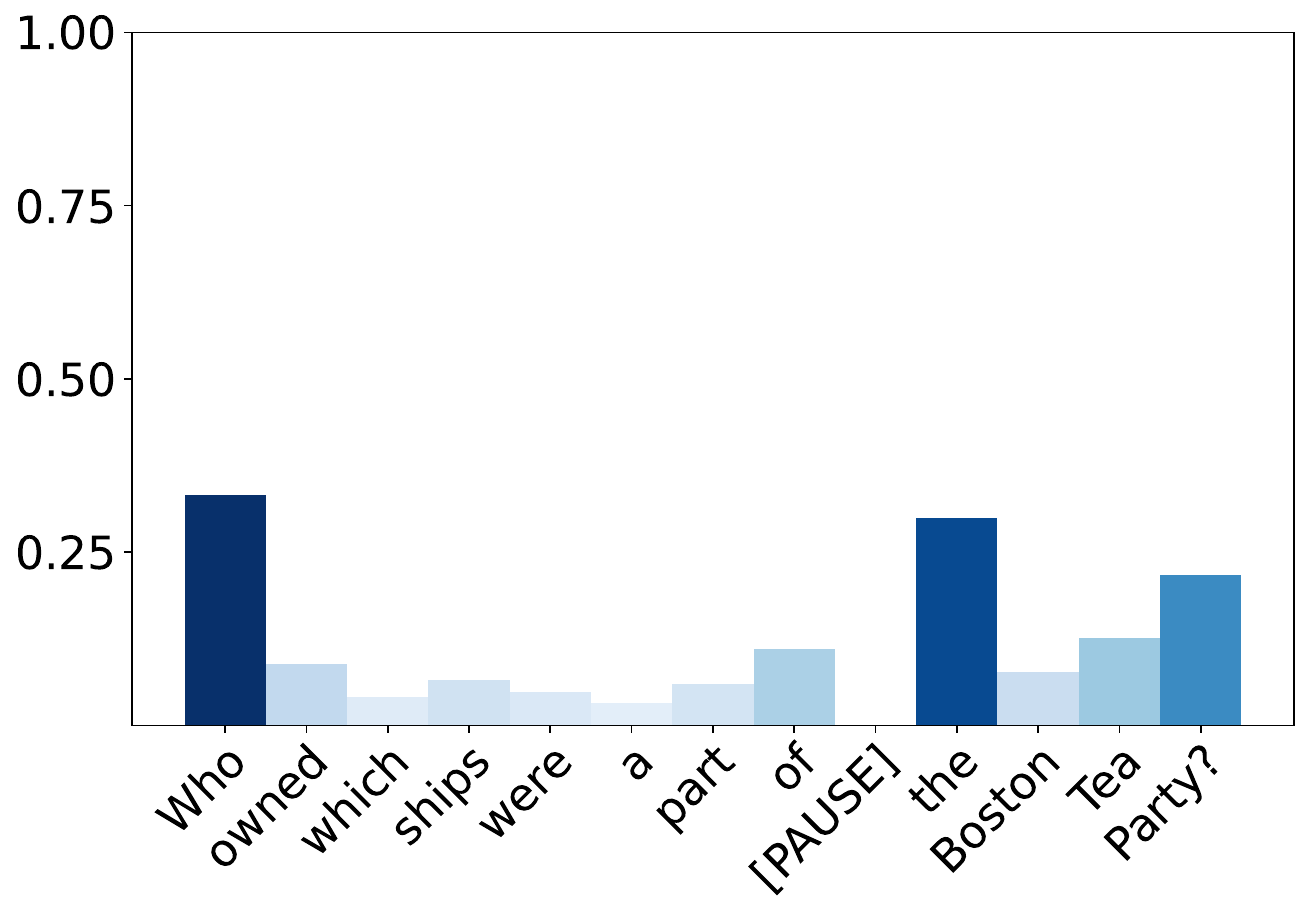}
            \caption[]%
            {{\small After adding \tframed[line width=0.5bp,fill=vred]{\textcolor{white}{\texttt{\textbf{[PAUSE]}}}} tokens} to paraphrase 4.}    
            \label{fig:mean and std of net44}
        \end{subfigure}
        \hfill
        \vskip\baselineskip
        \begin{subfigure}[b]{0.45\textwidth}   
            \centering 
            \includegraphics[width=\textwidth,height=3cm]{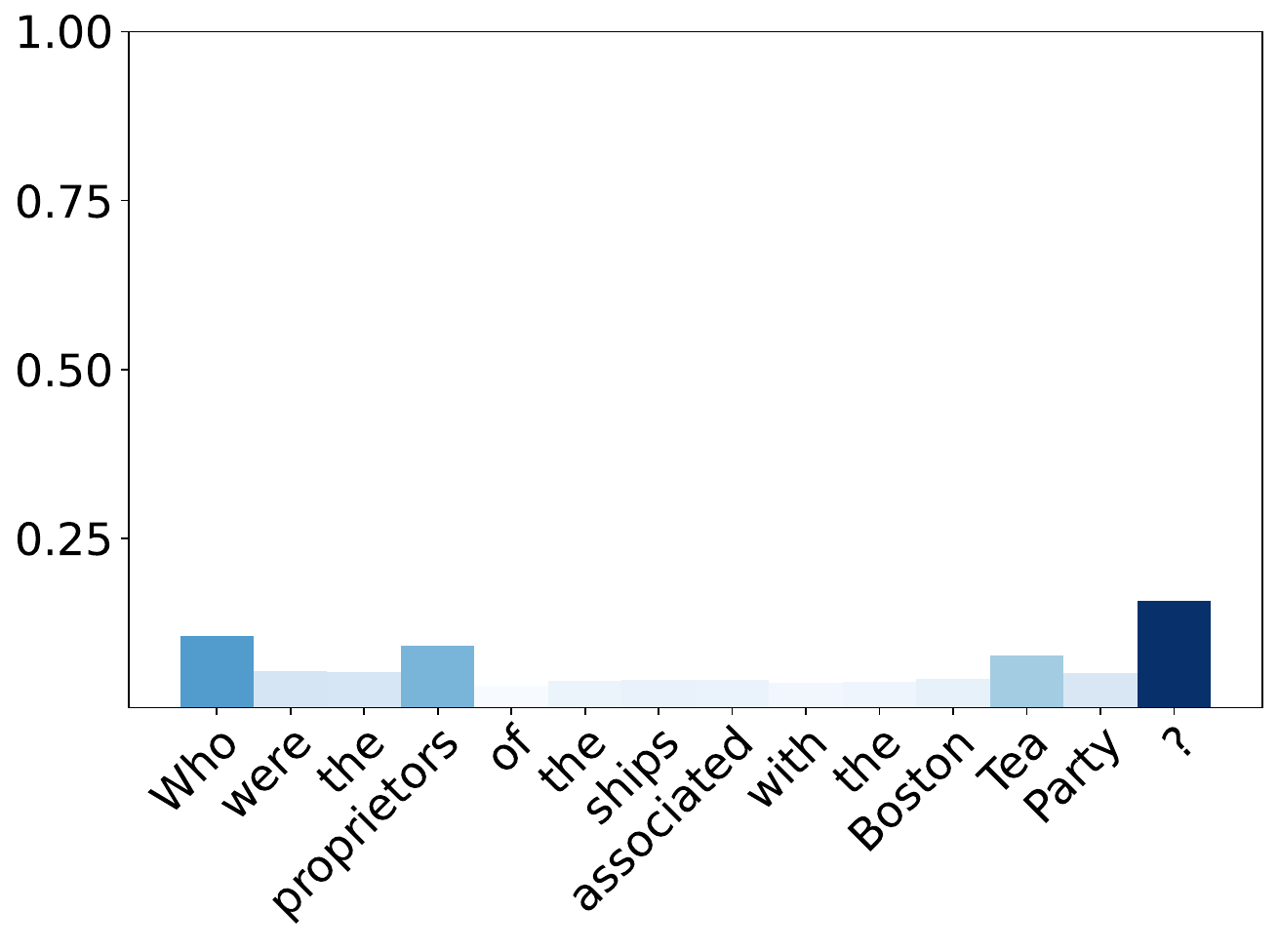}
            \caption[]%
            {{\small Before adding \tframed[line width=0.5bp,fill=vred]{\textcolor{white}{\texttt{\textbf{[PAUSE]}}}} tokens} to paraphrase 5.}    
            \label{fig:mean and std of net44}
        \end{subfigure}
        \hfill
        \begin{subfigure}[b]{0.45\textwidth}   
            \centering 
            \includegraphics[width=\textwidth,height=3cm]{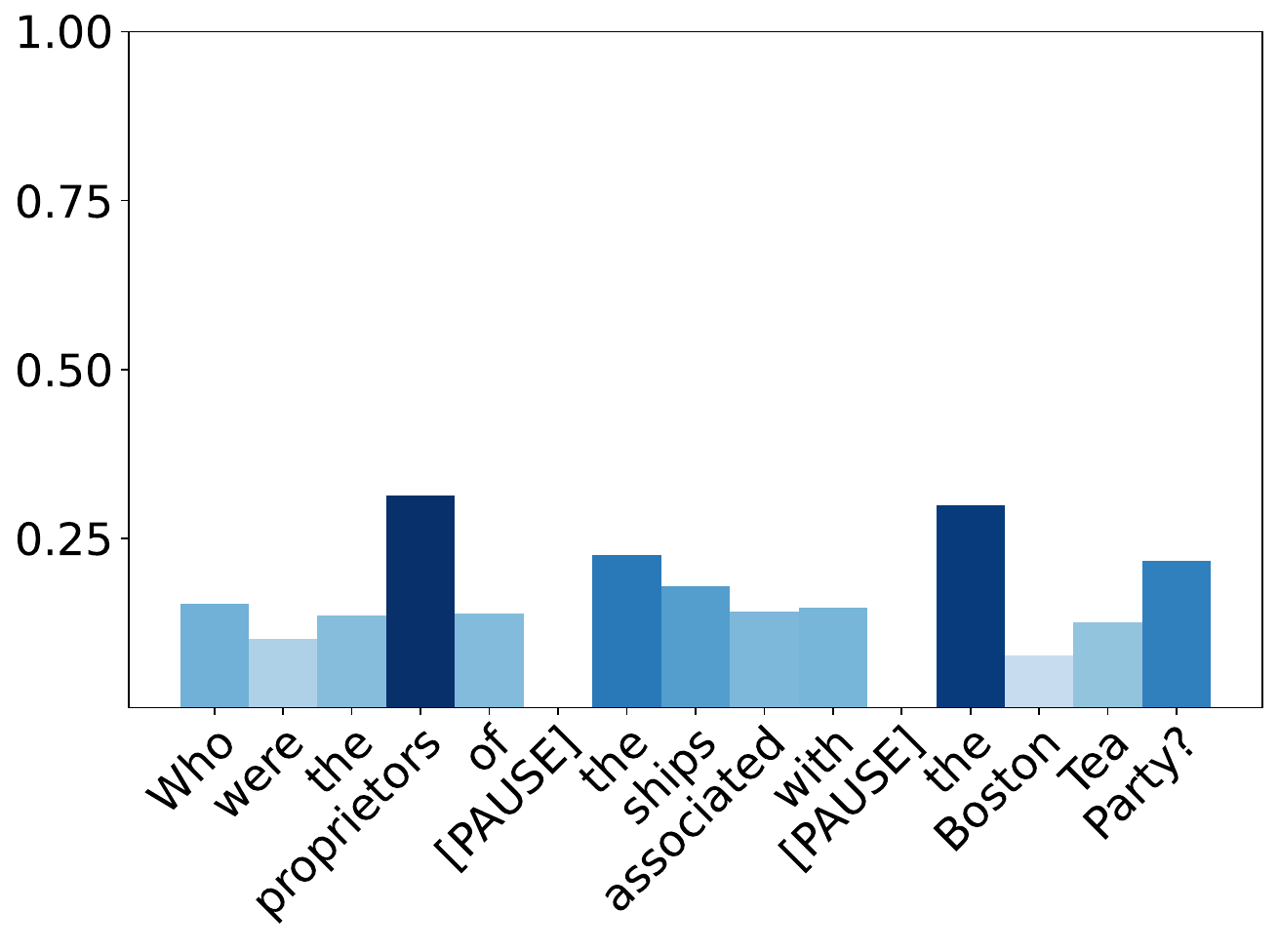}
            \caption[]%
            {{\small After adding \tframed[line width=0.5bp,fill=vred]{\textcolor{white}{\texttt{\textbf{[PAUSE]}}}} tokens} to paraphrase 5.}    
            \label{fig:mean and std of net44}
        \end{subfigure}
        \caption[]%
            {{\small The phrase \textbf{Boston Tea} gets more importance score after adding \tframed[line width=0.5bp,fill=vred]{\textcolor{white}{\texttt{\textbf{[PAUSE]}}}} token for Zephyr.}}   
        \label{fig:Zephyr}
\end{figure*}

In the \cref{fig:alpaca,fig:bloomz,fig:dolly,fig:Falcon,fig:FLAN,fig:GPT Neo,fig:Llama2,fig:OPT,fig:phi,fig:Vicuna,fig:Zephyr} below, we show the impact of adding \tframed[line width=0.5bp,fill=vred]{\textcolor{white}{\texttt{\textbf{[PAUSE]}}}} token to understand the longer prompts.

\end{document}